\documentclass[11pt]{article}

\usepackage[final]{acl}
\usepackage{times}
\usepackage{latexsym}
\usepackage{amsfonts}
\usepackage[T1]{fontenc}
\usepackage[utf8]{inputenc}
\usepackage{microtype}
\usepackage{amsmath}
\usepackage{inconsolata}
\usepackage{graphicx}
\usepackage{natbib}
\usepackage{url} 
\usepackage{booktabs}
\usepackage{multirow}

\title{Evaluating Counterfactual Strategic Reasoning in Large Language Models}

\author{
Dimitrios Georgousis, Maria Lymperaiou, Angeliki Dimitriou,\\
\textbf{Giorgos Filandrianos}, \textbf{Giorgos Stamou} \\
National Technical University of Athens \\
\texttt{\href{mailto:dimjimitris@gmail.com}{dimjimitris@gmail.com}}, \\
\texttt{\{\href{mailto:marialymp@islab.ntua.gr}{marialymp},
\href{mailto:angelikidim@islab.ntua.gr}{angelikidim},
\href{mailto:geofila@islab.ntua.gr}{geofila}\}@ails.ece.ntua.gr}, \\
\texttt{\href{mailto:gstam@cs.ntua.gr}{gstam@cs.ntua.gr}}
}

\begin{document}
\maketitle
\begin{abstract}
We evaluate whether LLMs adapt their strategic behavior when familiar games are counterfactually modified. We introduce a repeated-game evaluation framework covering Prisoner’s Dilemma and Rock–Paper–Scissors under default, label-perturbed, payoff-perturbed, and joint counterfactual variants. This design separates surface robustness to renamed actions from deeper sensitivity to changed incentives. Across multiple frontier LLMs, we find that label perturbations usually cause moderate degradation, whereas payoff perturbations expose stronger failures: LLMs often preserve canonical strategies even when the equilibrium structure changes. In RPS, several LLMs remain close to uniform play despite a payoff-counterfactual equilibrium requiring a biased mixed strategy. Behavioral and efficiency metrics further show that stronger or reasoning-enabled LLMs are not uniformly more strategic: some deliberate more without adapting faster. Overall, counterfactual repeated games provide a compact diagnostic for distinguishing robust incentive-sensitive behavior from brittle template-based strategic execution.
\end{abstract}

\section{Introduction}
Strategic reasoning using Large Language Models (LLMs) forms an upcoming field on the intersection of reasoning and agentic synergy, driven by the rapidly advancing capabilities of state-of-the-art (SoTA) LLMs. Communication between LLMs allows cooperation and competition, i.e. the basic ingredients for allowing \textit{game-playing}
\cite{gandhi2023strategic, zhang2024llm}. Notably, the search for human-level strategic interactions spreads before the LLM-era, demonstrating a long-standing need for autonomous, rational agents  \cite{Silver2016MasteringTG, Berner2019Dota2W, Bakhtin2022HumanlevelPI}.

Interestingly, prompting LLMs as strategic players reveals several behavior traits for the models themselves, enabling the usage of strategic and competitive environments as suitable testbeds for LLM evaluation \cite{zhang2024llm, costarelli2024gamebench, jia2025llm, wang2025tmgbenchsystematicgamebenchmark, GTBENCH}. LLMs are found to often demonstrate a gap between
verbalizing versus executing optimal or equilibrium strategies, demonstrating brittle behavior under competition. Inductive biases, over-reliance to short term outcomes and sensitivity to prompts arise as major issues that influence LLM capability in strategic reasoning, paving the way for more evaluation environments to come.

\begin{figure}[t!]
%\vskip -0.18in
    \centering
    \includegraphics[width=0.75\linewidth]{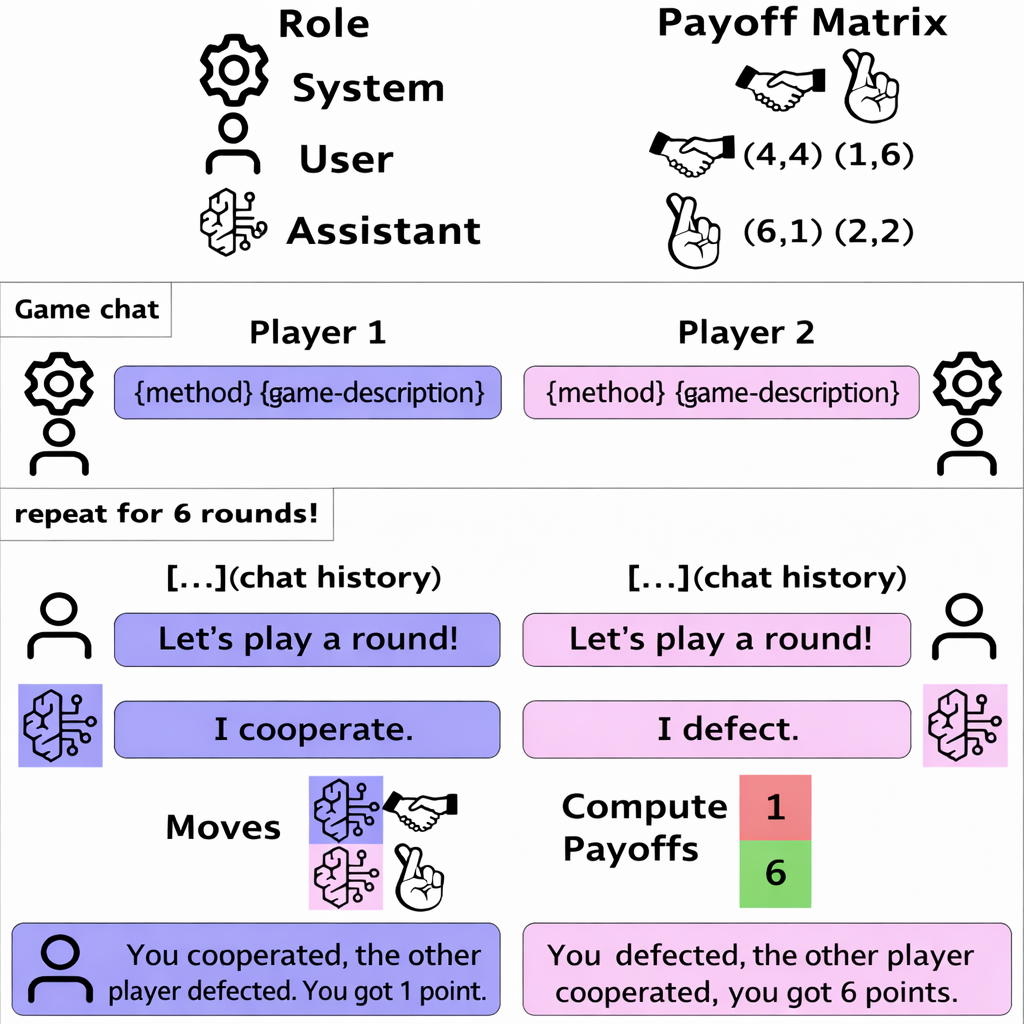}
    \vskip -0.04in
    \caption{LLM interaction for Prisoner's Dilemma}
    \label{fig:teaser-pd}
\end{figure}
%\vskip -0.01in

%In this paper, we evaluate LLMs on games rather than open-ended strategic environments which curtail complex evaluation pitfalls, in the interest of demonstrating LLM abilities on closed-world, well-defined situations. The clear evaluation setup allows for the integration of \textit{counterfactual scenarios}, designed to probe LLM response on alternative setups in place of the well-known default ones. Possible performance deviations in the counterfactual scenario would reveal some reasoning rigidity: LLMs may know how to act strategically in games, thanks to their vast pre-training, but struggle with redefined game parameters that request cognitive flexibility and reasoning out-of-the-box. To this end, we propose a simple, adaptable evaluation setting that considers counterfactual game pairs, showcasing reasoning discrepancies that signal potential memorization over true thinking.
We opt for evaluating LLMs in structured games rather than open-ended strategic environments, enabling controlled and well-defined assessment of their behaviors and strategic capabilities. This setup allows us to introduce counterfactual scenarios that modify standard game parameters to test cognitive flexibility. Performance deviations under such altered conditions reveal potential reasoning rigidity: LLMs may rely on canonical game-template behavior from pre-training yet struggle when familiar structures are redefined. We thus propose a simple and adaptable evaluation framework based on counterfactual game pairs to diagnose behavioral evidence of incentive-sensitive adaptation rather than canonical game-template behavior.

%In order to capture different failure modes, we select two diverging strategic properties: (i) LLM response to stochasticity, where any pattern or bias can be exploited to form a strategy against the other players, and (ii) LLM reaction to situations promoting social wellfare, in which synergetic or egoistic behaviors give rise to varying individual and collective outcomes. These properties are instantiated via the Rock-Paper-Scissors (RPS) and Prisoner's Dilemma (PD) respectively. These problems have been favored in recent LLM game-playing literature \cite{rps, akata2025repeated, fontana2024pd}, setting an ideal starting point to further delve into LLM internal mechanics.
%To capture distinct failure modes, we focus on two strategic properties: (i) LLM responses to stochasticity, where exploitable patterns may emerge, and (ii) LLM behavior in socially sensitive settings, where cooperative and egoistic incentives conflict. These are instantiated through Rock–Paper–Scissors (RPS) and the Prisoner’s Dilemma (PD), respectively. Both games are widely used in recent LLM game-playing literature \cite{rps, akata2025repeated, fontana2024pd}, providing a controlled foundation for probing internal reasoning mechanisms.

To capture distinct failure modes, we focus on two complementary game settings: Rock--Paper--Scissors (RPS), which probes stochasticity and exploitable patterns, and the Prisoner’s Dilemma (PD), which probes cooperation under conflicting incentives. In our setup, RPS tests whether LLMs avoid predictable biases in adversarial play, while PD tests whether they adapt between defection and cooperation under repeated interaction.

In RPS, players choose among three cyclically dominant actions (\textit{Rock beats Scissors, Scissors beats Paper, and Paper beats Rock}), so no move is inherently superior as it depends on other player's actions, but stable patterns become exploitable (e.g. if we always play \textit{Rock} in even rounds and \textit{Paper} in odd rounds). 
In PD, players simultaneously choose between \textit{Cooperate} and \textit{Defect} without knowledge of the opponent’s decision. Mutual cooperation yields shared benefits, while unilateral defection provides a higher immediate payoff. Thus, no action is universally optimal: cooperation risks exploitation, whereas defection sacrifices collective gain. Although defection dominates in the single-round setting, serving as the safest option, repeated interaction allows for adaptation, with persistent defection inviting retaliation, while consistent cooperation  sustaining mutually beneficial outcomes.

Our results reveal three main findings. First, payoff-based counterfactuals are more diagnostic than label counterfactuals: many LLMs tolerate renamed actions but fail when the incentive structure changes. Second, strategic competence is highly opponent-dependent: models often exploit simple algorithmic opponents but become unstable in LLM–LLM interactions, especially in RPS. Third, reasoning-enabled variants do not always improve strategic execution; in several cases, additional deliberation increases token usage without improving adaptation. Overall, our contributions are\footnote{Code is available at \url{https://github.com/dimjimitris/llm_gm_thesis}.}:
\begin{itemize}
    \item We introduce a counterfactual repeated-game framework that separates label robustness from payoff-sensitive strategic adaptation.
    \item We propose behavioral metrics for strategic execution, including opponent comprehension, cooperation/action-distribution behavior, and token-normalized efficiency.
    \item We provide an empirical analysis showing that LLMs often behave strategically in canonical games but remain brittle under payoff and joint counterfactuals.
\end{itemize}

\section{Related work}
%\paragraph{Games for LLM evaluation} A new frontier in LLM evaluation involves probing their behavior in game settings, and examine whether they align with game-theoretic predictions. LLMs have been employed in games that evaluate rational decision-making, including Matching Pennies, Rock--Paper--Scissors, the Ring-Network Game, and the Dictator Game, showcasing that they are often able to articulate equilibrium concepts or optimal strategies, yet exhibit systematic deviations during actual play, particularly in environments requiring randomization or consistent policy execution \cite{matching-pennies,rps}. Such observations highlight a gap between explicit strategic knowledge and realized behavior,  motivating LLM evaluation through interaction-based decision tasks and simulation of human dynamics \cite{aher2023turing,horton2023}. Beyond these, LLMs are evaluated in social dilemma setups including the Prisoner’s Dilemma, Stag Hunt, and other $2 \times 2$ normal-form games which highly evaluate cooperation patterns. In repeated interactions, LLMs often display cooperative or retaliatory patterns, but their behavior remains highly sensitive to prompt framing and opponent specification \cite{akata2025repeated, fontana2024pd}. Moreover, weaker models frequently fail to abstract away from narrative context and instead rely on superficial cues rather than the underlying incentive game structure \cite{loreheydari2024}.
\paragraph{Games for LLM evaluation} constitute a new research frontier, examining whether probed behavior aligns with game-theoretic predictions.
So far, LLMs have demonstrated their ability to approach optimal strategies and equilibrium concepts when employed as rational players in games such as Matching Pennies, RPS, the Ring-Network Game, and the Dictator Game \cite{matching-pennies,rps,kader2025emergencestrategicreasoninglarge}, although they pose systematic deviations in the course of strategic execution, particularly in randomized setups. Cooperative abilities and coordination of LLMs have been probed in social dilemma games, including PD; LLMs display collaborative and retaliation behaviors, even though they remain subject to prompt sensitivity and opponent specification \cite{akata2025repeated,fontana2024pd,mozikov24}.
Moreover, weaker models often fail to abstract away from narrative context and instead rely on superficial cues rather than the actual incentive game structure  \cite{loreheydari2024}.
These works underscore a persistent gap between explicit strategic knowledge and realized decision-making, supporting the use of interactive game environments as diagnostic LLM tools  \cite{aher2023turing,horton2023}. Finally, using LLMs as meta-evaluators of strategic LLM interactions confirms previously observed pitfalls in capturing dominant strategies and handling payoffs, especially in non-canonical game formulations \cite{collins2025evaluating}.

\paragraph{Evaluation of LLMs in Counterfactual Setups}
as opposed to pattern memorization poses a seminal challenge, often operationalized through intervened inputs that explicitly contradict real-world knowledge.
Prior work shows that although models can follow hypothetical premises, their performance deteriorates when these assumptions conflict with well-established facts, suggesting reliance on memorized associations rather than robust integration of altered conditions \cite{li-etal-2023-counterfactual,yamin2025llms,wu-etal-2024-reasoning}.
Subsequent benchmarks formalize this challenge by systematically perturbing task inputs and measuring counterfactual sensitivity, showing that LLMs frequently fail to update their reasoning in response to minimal but semantically meaningful changes \cite{chen2025counterbenchbenchmarkcounterfactualsreasoning,frohberg-binder-2022-crass,akter2026causalconsistencyregularizationtraining}.
Moreover,  apparent improvements from chain-of-thought prompting or instruction tuning may mask underlying brittleness, as LLMs produce fluent yet inconsistent outputs across counterfactual variants \cite{huyuk2025reasoning}, while scaling up LLM size is not a viable cure, since larger models may collapse even more dramatically under reformulated realities \cite{stringli-etal-2025-pitfalls}.
Despite these advances, existing evaluations largely focus on static setups, leaving interactive environments comparatively underexplored.
Particularly, the evaluation of LLMs in \emph{counterfactual game-theoretic settings} remains limited, motivating the use of repeated games as diagnostic environments for strategic reasoning.

\section{Background in Game Theory}
\paragraph{Strategic Games.}
A (finite) \textit{strategic game} in normal form is defined by a tuple:
\begin{equation}
G = \bigl(\{1,2\}, A_1, A_2, u_1, u_2\bigr)
\end{equation}
where $A_i$ is the finite action set of player $i$, and
$u_i : A_1 \times A_2 \to \mathbb{R}, i \in \{1, 2\}$ is player $i$'s payoff function that indicates the quality of an action.
We explicitly focus on \textit{two-player}, \textit{simultaneous} games, i.e. the two players take actions at the same time, without any information regarding the other player's planned action.
A game is \emph{symmetric} if:
\begin{equation}
\begin{aligned}
A_1 = A_2 = A
 \text{ and } 
u_1(a,b) = u_2(b,a) \\
\ \ \forall a,b \in A.
\end{aligned}
\end{equation}
\paragraph{Strategies.}
A \textit{pure strategy} for a player is a single action $a \in A$.
A \textit{mixed strategy} is a probability distribution over actions, denoted by
$\sigma \in \Delta(A)$, where $\Delta(A)$ denotes the set of all valid probability distributions over $A$.
For a mixed-strategy profile $\sigma=(\sigma_1,\sigma_2)$, the expected payoff to player $i$ is:
\begin{equation}
\mathbb{E}_{a \sim \sigma}\bigl[u_i(a_1,a_2)\bigr]
\end{equation}
Here, $\sigma_i$ denotes the mixed strategy of player $i$, $a \sim \sigma$ indicates that actions are sampled according to the players’ mixed strategies, and $u_i$ is the payoff function of player $i$.

\paragraph{Nash equilibrium.}
The central solution concept we employ is the \emph{Nash equilibrium}, which formalizes a notion of strategic stability.
Intuitively, a Nash equilibrium corresponds to a situation in which each player selects a strategy that is optimal given the strategy of the other player, so that no player has an incentive to unilaterally deviate.

A mixed-strategy profile $\sigma^*=(\sigma_1^*,\sigma_2^*)$ denotes a \emph{candidate equilibrium}, where $\sigma_i^*$ is the strategy assigned to player $i$ at equilibrium.
This profile is a Nash equilibrium if, for each player $i$:
\begin{equation}
\mathbb{E}\bigl[u_i(\sigma^*)\bigr]
\;\ge\;
\mathbb{E}\bigl[u_i(\sigma_i,\sigma_{-i}^*)\bigr]
\quad \forall \sigma_i \in \Delta(A).
\end{equation}
Here, $\sigma_{-i}^*$ denotes the \textit{equilibrium} strategy of the opposing player, while $\sigma_i$ represents any alternative (possibly randomized) strategy available to player $i$.
The inequality states that, assuming the other player follows their equilibrium strategy, player $i$ cannot increase their expected payoff by switching unilaterally from $\sigma_i^*$ to any other strategy.

\paragraph{Prisoner's Dilemma (PD).}
Let  $A=\{C,D\}$ denote two distinct actions namely \emph{Cooperate} and \emph{Defect}. The PD payoff structure satisfies:
\begin{equation}
T > R > P > S
\end{equation}
where $R$: \textit{reward} for mutual cooperation $(C,C)$, $T$: \textit{temptation} payoff for unilateral cooperation $(D,C)$, $P$: \textit{punishment} for mutual defection $(D,D)$, $S$: \textit{sucker} payoff for unilateral cooperation $(C,D)$.
Based on the above, cooperation is socially beneficial ($R>P$), defection is individually rational ($T>R$), and unilateral cooperation is strictly worse than mutual defection ($S<P$).

A canonical payoff matrix is:
\begin{equation}
\begin{array}{c|cc} 
 & C & D \\
\hline
C & (R,R) & (S,T) \\
D & (T,S) & (P,P)
\end{array}
\end{equation}
In the one-shot PD, $D$ strictly dominates $C$ and the unique Nash equilibrium is $(D,D)$. PD is a general-sum game, in which players' payoffs need not sum to zero and mutually beneficial outcomes may exist, such as the joint action $(C,C)$.

\paragraph{Repeated games}
are formed by playing a fixed \emph{stage game} multiple times in sequence.
At each round $t$, players simultaneously choose actions and observe the outcome before proceeding to the next round.
Let $a^t$ denote the joint action (combination of actions chosen by all players $a^t=(a^t_1,a^t_2)$) at round $t$. The history $h_t$ available at this round is:
\begin{equation}
h_t = (a^1,\dots,a^{t-1}), \quad t \in \{1,\dots,T\}
\end{equation}
A (behavioral) strategy specifies a distribution over actions conditioned on history:
\begin{equation}
\pi(\cdot \mid h_t) \in \Delta(A)
\end{equation}
The payoff to a player in a repeated game is typically defined as a cumulative sum over rounds:
\begin{equation}
U_i = \sum_{t=1}^{T} u_i(a^t)
\end{equation}
or, in infinite-horizon settings, as a discounted sum with discount factor $\gamma \in (0,1)$.

\paragraph{Rock--Paper--Scissors (RPS).}
Let an action set $A=\{R,P,S\}$ denote \textit{Rock}, \textit{Paper}, \textit{Scissors}. RPS is a two-player \emph{zero-sum} game, i.e. the players' payoffs always sum to zero. In zero-sum games, one player’s gain corresponds exactly to the other's loss, excluding mutual benefits.
Thus, optimal behavior is purely \textit{adversarial}, with a payoff function:
\begin{equation}
u(a,b) =
\begin{cases}
\ \ 1 & \text{if $a$ beats $b$},\\
\ \ 0 & \text{if $a=b$},\\
-1 & \text{if $b$ beats $a$}.
\end{cases}
\end{equation}
RPS has no pure-strategy Nash equilibrium; its unique Nash equilibrium is the mixed strategy:
\begin{equation}
\sigma^*(R)=\sigma^*(P)=\sigma^*(S)=\frac{1}{3}\end{equation}
\section{Method}

\subsection{Counterfactual Games}
One of our main points is evaluating LLMs on counterfactual games, i.e. alternative formulations of  action labels or payoff values. 
%Stable behavior under counterfactual variants suggests abstraction over the underlying payoff structure, whereas large deviations suggest reliance on familiar game templates or surface cues.
An almost invariant counterfactual outcome indicates LLM generalizability and adaptability in novel compositions, or else an overly strict adherence on specific instances rather than reasoning over game dynamics. 
\paragraph{Prisoner's Dilemma Counterfactuals} \mbox{}\\
\textbf{Stag Hunt (SH)} stands as a counterfactual counterpart of PD, challenging the balance between social cooperation and individual safety: two hunters must decide separately whether to hunt a stag ($S$) or a hare ($H$). One hunter on their own can catch a hare, but the payoff is less than catching a stag. However, this requires mutual cooperation, as no hunter on their own is able to capture a stag. The SH payoff satisfies the inequality:
\begin{equation}
R > T \geq P > S
\end{equation}
where $R$: reward for mutual $S$, $T$: hunting $H$ while the other hunts $S$,  $P$: payoff for mutual $H$,  $S$: payoff for hunting $S$ alone.
The game displays two Nash equilibria: the $(S, S)$ is payoff dominant but risky, while the $(H, H)$ is risk-dominant but safe.

The altered SH payoff suggests a change of strategic incentives, shifting the emphasis from defection to coordination; thus, SH is primarily a \textbf{payoff-driven} PD counterfactual. To test LLM adherence to action labels, we substitute PD labels with SH ones ($C \rightarrow S$ and $D \rightarrow H$), without altering the payoff values%; inversely, we substitute the $SH \rightarrow PD$ labels
.
An instance of the employed counterfactual formulation is detailed in App. \ref{sec:pd}.

\paragraph{Rock-Paper-Scissors Counterfactuals} \mbox{}\\ In RPS, we implement both a \textbf{payoff-based} counterfactual as well as a \textbf{label-based} one. While in the default game all wins and losses are of equal magnitude, leading to a unique mixed-strategy equilibrium characterized by uniform randomization, in the payoff-based counterfactual we alter the incentive structure by assigning higher payoff to specific outcomes: the \textit{Rock}-\textit{Paper} combination receives $\times 3$ win-loss value, while the rest of the combinations remain invariant (App. \ref{sec:rps}). This modification breaks the game symmetry and forces rational players to adopt a biased mixed strategy rather than uniform randomization.
Moreover, in the label-based counterfactual game we permute labels to inverse their dominance, so that \textit{Scissors} beats \textit{Rock}, \textit{Rock} beats \textit{Paper} and \textit{Paper} beats \textit{Scissors}. This intervention decouples label memorization from reasoning over the game structure.
Finally, we jointly interfere on payoffs and labels to further stress test LLM players.

For the payoff-based RPS counterfactual, the canonical uniform equilibrium no longer applies. Under the modified payoff matrix, where the Rock–Paper outcome has magnitude 3 while all other win/loss outcomes have magnitude 1, the unique mixed Nash equilibrium is:
\begin{equation}
\pi^*(R)=0.2, \pi^*(P)=0.2, \pi^*(S)=0.6
\end{equation}
Therefore, models that remain close to uniform play (1/3,1/3,1/3) are not merely randomizing; they are failing to adapt to the altered payoff structure.

\section{Experimental Setup}
\subsection{Player Types}
Evaluating an LLM's  strategic thinking is only meaningful in the context of other players' game-playing dynamics. For this reason, we consider the following LLM players, differentiated by their  system prompt: a) \textbf{ZS} (default):  initialized with \textbf{zero-shot} (ZS) system prompt; b) \textbf{CoT}: initialized with  \textbf{Chain of Thought} (CoT) system prompt; c) \textbf{SPP}: initialized with \textbf{Solo Performance} prompt (SPP) \cite{spp}, a collaboration-enhancing approach that combines multiple views from LLM personas for advanced problem-solving; d) \textbf{SC}: employing \textbf{self-consistency} (SC) \cite{wang2023selfconsistency} to augument any of the aforementioned players.

We also contrast LLM players with algorithmic ones, which are based on deterministic strategies, constituting a predictable and well-defined evaluation framework for opposition or collaboration. A potent LLM player is able to recognize the patterns of an algorithmic one, and upon this, decide their optimal strategy or present exploitation tendencies. The following algorithmic players are employed: a) \textbf{SREP}: selects the single-round equilibrium action (pure or mixed depending on the game); in PD this corresponds to always defecting; b) \textbf{PP}: a \textbf{pattern player} follows a cyclic pattern of actions. The next sets of players are instantiated differently for PD and RPS: $c_1$) \textbf{MF}: exploiting the \textbf{most frequent} move of the opponent to maximize their reward for PD; $d_1$) \textbf{TFT}: a \textbf{tit-for-tat}\footnote{In tit-for-tat, a player chooses C in the $1^{st}$ round and in  following rounds it copies the opponent’s previous action.} variant in which the player chooses the move that when paired with the most
recent opponent's move  gives the best reward for PD. Finally, for RPS: $c_2$) \textbf{AP}: an \textbf{adaptive player} that acts so as to counter the opponent’s most frequent move; $d_2$) \textbf{TFT}: based on tit-for-tat, this player chooses the move that counters the opponent’s most recent move.

\section{Experiments}
\subsection{Experimental design} 
\paragraph{Number of rounds} To probe LLMs' comprehension of game dynamics over time, we allow consequent rounds to play per game; specifically, PD is played for 16 rounds, while RPS is repeated 24 times. After preliminary experiments, we conclude that these are adequate repetitions for an LLM to understand and adapt to the opponent considering the number of available variables to select in each round ($C$ or $D$ for PD, and $R$, $P$, or $S$ for RPS.)

\paragraph{Game repetition} Each game is played by the \textbf{non-SC} players 5 different times, and 2 times by \textbf{SC} ones (it is expected that their outcomes remain similar due to the self-consistency mechanism).

\paragraph{Self-Consistency parameters} %Self-consistency is implemented by sampling multiple candidate answers per round and selecting the final move by majority frequency. 
We use 3 samples for PD, where two actions are available, and 5 samples for RPS, where three actions require a larger pool for stable marginalization. Ties at the highest frequency are broken randomly. These settings were chosen as a practical compromise between consistency and computational cost.

\subsection{Models} LLMs instantiated as players include  Claude Sonnet 3.5 v2, Claude Sonnet 3.7 (with vs without thinking),  Claude Sonnet 4 (with vs without thinking), Deepseek R1, Llama 3.3 70B and Mistral Large (24.07). Each LLM plays against a different instance of the same model, or  algorithmic agents.

\subsection{Evaluation Metrics} 
\paragraph{Total points} denote per-player cumulative reward over all rounds, averaged across repetitions. Higher values indicate stronger adaptation to the payoff structure and opponent behavior.

\paragraph{Opponent Comprehension} is a more dynamic metric that evaluates how late an LLM unlocks its opponent's behavior and takes advantage of it. Formally, given a repeated game of $N$ rounds between an LLM agent $A$ and an opponent $B$, we define the \emph{round of opponent comprehension} $m$ as the earliest round after which $A$’s actions yield payoffs that are at least as high as $B$’s in a large majority of subsequent rounds.
To allow for occasional deviations or noise, we introduce a target percentage $tp$, requiring that from round $m$ to round $N$, $A$ achieves payoffs no worse than $B$ in at least $tp$ percent of the rounds.
We default $tp = 90\%$.
Lower values of $m$ indicate earlier comprehension and faster adaptation, while a value of $m$ exceeding the game horizon indicates that the LLM failed to consistently exploit the opponent’s behavior.

%\paragraph{Cooperation Rate} gauges the percentage of rounds in which the LLM cooperates with its opponent (selecting $C$ in PD or $S$ in SH). This metric evaluates social aspects of LLMs, with higher values denoting advanced cooperative over defensive behavior. However, it does not necessarily imply optimal play, since a highly cooperative LLM can be a result of exploitation, which can harm performance in cases of adversarial and deceitful agents.

\paragraph{Cooperation Rate}
Cooperation rate is the proportion of rounds in which the LLM selects the cooperative action (e.g., $C$ in PD or $S$ in SH). It captures social tendencies, though high cooperation does not necessarily imply optimal play.

\paragraph{Efficiency}
is proposed to account for the cost-performance trade-off:
\begin{equation}
    \textit{efficiency} = \frac{\textit{points}}{\textit{tokens}}\times \textit{c}
\end{equation}
The scaling factor \textit{c} has a default value of 1000 for numerical readability.

%\paragraph{Failure Rate} We strictly monitor whether the LLMs follow instructions properly, eliminating any invalid actions; that includes formatting violations or predicting actions outside the allowed action space of each problem. Explicit validity checks are adopted from \citet{liao2024efficacy}.
\paragraph{Failure Rate}
 measures the proportion of invalid actions, including formatting violations and outputs outside the allowed action space.

\subsection{Payoff instantiations}
\label{sec:payoff}

We instantiate PD with payoffs
$(C,C)=(4,4)$, $(C,D)=(1,6)$, $(D,C)=(6,1)$, and $(D,D)=(2,2)$.
Each PD game is repeated for $N=16$ rounds, so per-player cumulative scores satisfy:
\begin{equation}
16 \leq \textit{total points} \leq 96.
\end{equation}
These bounds correspond to always cooperating against a defecting opponent and always defecting against a cooperating opponent, respectively.

For RPS, we use the standard zero-sum payoff instantiation, where a win, loss, and tie yield $w=1$, $\ell=-1$, $t=0$, respectively. Each game is repeated for $N=24$ rounds, with cumulative scores within:
\begin{equation}
-24 \leq \textit{total points} \leq 24.
\end{equation}
Under the single-round mixed Nash equilibrium, the expected payoff per round is 0, yielding an expected cumulative score of 0 over 24 rounds. We therefore interpret RPS results primarily through deviations from this zero baseline.

Since these deterministic bounds are not always informative in uncertain settings (e.g. LLM--LLM interactions), we also use expected-payoff baselines under stationary mixed strategies; the corresponding derivations  are provided in App.~\ref{app:expected_outcomes}.

\paragraph{Result aggregation.}
To support claims beyond isolated examples, we compare default and counterfactual variants over the same repeated runs and prompting variants. We report full per-model means and variability in Appendix~\ref{sec:full-results} and use these values to examine whether degradation is consistent across games, opponents, and metrics. Our analysis is primarily diagnostic: we focus on the direction and recurrence of behavioral changes under label, payoff, and joint counterfactuals.

\section{Results and Analysis}
\label{sec:results}
Full numerical results are shown in App. \ref{sec:full-results}.
\subsection{Prisoner's Dilemma}
%PD primarily reveals whether models can shift between exploitation, defection, and coordination as opponent behavior and payoff structure change.
\paragraph{Total points} widely vary based on the model choice, the opponent category (algorithmic vs LLM) and the prompting technique used. Commencing from algorithmic opponents, the LLM manages to accumulate higher total points, exploiting their predictable algorithmic strategies, although performance fluctuates per opponent. In particular,  \textbf{SREP}  always defects, making defection the optimal response and yielding 32 points over 16 rounds through sustained $(D,D)$ play. Most LLMs cluster around this baseline ($\sim$30 points), with modest prompt-induced variation. A notable exception is Mistral Large, which underperforms substantially and remains unstable across prompting types (from $18.6 \pm 10.6$ under SPP to $29.8 \pm 2.2$ under CoT), suggesting delayed or inconsistent recognition of the dominant response. Against the cyclical \textbf{PP}, LLMs obtain substantially higher scores, typically in the mid-50s to low-60s, by exploiting the repeated \textit{D,C} pattern once recognized.

A different regime emerges in LLM--LLM interactions, where most models converge to mutual cooperation and receive 4 points per round. Claude 3.5/3.7 Sonnet (with and without thinking) and Llama 3.3-70B consistently achieve the maximum 64 points, corresponding to perfect cooperation across all 16 rounds, with virtually no variability across prompting types. In contrast, Claude 4 Sonnet variants and DeepSeek R1 obtain substantially lower totals ($31.4 \pm 0.0$ to $49.4 \pm 15.5$), reflecting a stronger tendency toward defection or unstable strategic exploration. These results suggest that increased reasoning depth does not necessarily improve strategic performance, and may in fact hinder convergence to the mutually beneficial outcome.
\paragraph{Opponent Comprehension} is generally fast against algorithmic opponents, yielding low \textit{m}. This is clearest for \textbf{SREP}, where most models adapt within the first few rounds ($m \leq 4$), and Claude 4 as well as DeepSeek R1 often respond from round 1. These early values align with total points close to the optimal defection baseline. The main exception is Mistral Large, whose comprehension is highly unstable and in some cases never stabilizes within the game horizon ($m>16$), matching its lower cumulative returns. Other algorithmic opponents, especially \textbf{PP}, are harder to infer and therefore produce later comprehension.

In LLM--LLM interactions, comprehension reflects inference of cooperative intent rather than recognition of a fixed policy. For most models and prompting types, this happens very early, which explains the stable mutual cooperation reported in total points. Claude 4 (especially w/o  thinking) and DeepSeek R1 show more variability, indicating delayed or unstable inference of the opponent’s behavior; this is consistent with their lower scores.

%\paragraph{Cooperation Rate} against algorithmic opponents closely reflect the optimal response to each opponent’s strategy. When facing the defecting SREP opponent, cooperation rates cluster tightly $\sim$0.1 for all LLM players, showing almost non-existent cooperation. This residual .1 cooperation is largely confined to early exploratory rounds and is consistent with the opponent comprehension results, where most agents identify persistent defection almost immediately. Against  PP, which alternates between defection and cooperation, cooperation rates are higher but remain well below 0.5, typically ranging between 0.3-0.5 depending on the model and prompting technique. This pattern reflects a brief exploratory phase followed by consistent defection once the cyclic structure is recognized. LLMs that exhibit lower cooperation against PP (Claude 4 w \& w/o thinking, DeepSeek) are the same ones that accumulated fewer total points as a result of defection. For MF and TFT opponents, cooperation rates are again low (mostly 0.1–0.2). Since both opponents converge quickly to defection in PD, these values indicate that LLMs appropriately adapt by minimizing cooperation once retaliatory or exploitative behavior is detected.

\paragraph{Cooperation Rate} against algorithmic opponents largely reflects the optimal response to each strategy. When facing the always-defecting \textbf{SREP}, cooperation rates cluster around $\sim$0.1 across LLMs, indicating almost no cooperation beyond a few exploratory rounds. This aligns with the opponent comprehension results, where most models recognize persistent defection very early. Against \textbf{PP}, which alternates between defection and cooperation, cooperation rises but remains well below 0.5, typically ranging between 0.3 and 0.5 depending on the model and prompt. These values suggest a short exploratory phase followed by sustained defection once the cyclic pattern is understood. For \textbf{MF} and \textbf{TFT}, cooperation is again low (mostly 0.1--0.2), consistent with the fact that both opponents converge quickly to defection in PD.

A starkly different pattern emerges in LLM–LLM interactions, with cooperation rates consistently approaching 1.0 (perfect cooperation) in many cases. Notable exceptions are constituted once again by Claude 4 (particularly thinking) and DeepSeek which persistently defect, pushing cooperation rates $\leq$0.1, confirming the previously reported non-cooperative tendencies.

\paragraph{Efficiency} against algorithmic opponents is generally high, especially when interacting with SREP, as optimal play requires minimal reasoning once persistent defection is identified. The early stabilization (confirmed by lower opponent comprehension scores) minimizes token usage while maintaining (near-)optimal cumulative payoffs. A similar pattern holds for PP; although identifying the cyclic pattern may require a short exploratory phase, once the structure is understood, LLMs consistently defect without further reasoning overhead. LLMs with delayed comprehension incur additional token costs during exploration, which lowers efficiency despite achieving moderately high total points.

\paragraph{Failure Rate} is negligible across models and mostly formatting-related; invalid actions are too rare to affect the main  behavioral conclusions. %Only DeepSeek R1 and Mistral Large fall slightly below the 100\% validity rate level (99.1\% and 99.4\%, respectively). These failures are largely formatting-related rather than strategic, with Mistral Large in particular often violating output-format instructions through markdown-style responses. Overall, invalid actions are too rare to affect the main behavioral conclusions. 

\subsubsection{Counterfactual analysis}

Label-only changes test whether models rely on memorized associations between action names and strategy. Comparing \textbf{PD} with its label-based counterfactual (where PD labels are replaced with SH labels while keeping PD payoffs) shows that stronger LLMs remain largely stable, whereas weaker ones—most notably Mistral Large—become more inconsistent. This degradation is not only reflected in total points, but also in delayed opponent comprehension and increased variability across prompting types. In other words, label changes alone can slow down strategic adaptation even when the underlying incentives are unchanged, indicating partial dependence on surface-level game templates rather than full abstraction over payoffs.

Payoff changes are more revealing, since they alter the strategic structure of the game itself. Moving from PD to SH incentives alleviates strict dominance and turns the interaction into a coordination problem. Many models adapt by shifting away from persistent defection and toward the higher-payoff cooperative equilibrium, often reaching totals close to the SH ceiling. However, this adaptation is uneven across metrics: intermediate total scores suggest partial rather than immediate coordination, delayed comprehension rounds show that several models initially favor the safer equilibrium before switching, and cooperation rates reveal whether higher totals are driven by sustained coordination or by late-stage recovery. Claude 4 variants and DeepSeek R1 remain more cautious across these metrics, often failing to fully exploit the new payoff structure. Overall, payoff-based counterfactuals expose \textbf{stronger reasoning differences} than label-only changes, since successful play requires recomputing incentives rather than merely recognizing renamed actions.

\subsection{Rock-Paper-Scissors}
%RPS primarily reveals whether LLMs can avoid canonical uniform play when exploitable patterns or payoff asymmetries require biased action distributions.
\paragraph{Total points} in RPS are centered around a zero baseline due to the zero-sum nature of the game, contrary to PD. As a result, performance is better interpreted through deviations from zero: positive scores indicate successful exploitation of opponent patterns, while negative scores reflect being exploited. Against algorithmic opponents, LLMs achieve consistently positive scores when exploitable patterns exist. In particular, against \textbf{PP}, stronger models (Claude 3.5/3.7 and Claude 4 variants) obtain the highest gains, reflecting successful identification and exploitation of cyclic behavior, while Llama 3.3 follows with moderate gains and Mistral Large exhibits lower and more variable performance. Against the \textbf{AP} player, total points are closer to zero across all models, indicating limited exploitation due to the opponent’s adaptive behavior. \textbf{SREP}, which follows the equilibrium distribution, also yields near-zero scores, as expected, since no systematic advantage can be obtained.

In LLM--LLM interactions, total points cluster tightly around zero, reflecting mutual adaptation and the absence of stable exploitation. However, variance across runs reveals differences in stability: stronger LLMs maintain scores close to equilibrium, while weaker or more unstable ones occasionally deviate, resulting in subtle gains or losses. Overall, unlike PD, total points in RPS primarily capture the ability to detect and exploit structure, rather than convergence to a single optimum.

\paragraph{Opponent Comprehension} in RPS is weaker and more variable than in PD, reflecting the greater difficulty of reasoning over three actions without a dominant response. Against algorithmic opponents, \textbf{PP} is the most informative benchmark: LLMs obtaining good total points against PP are precisely those that comprehend the cyclic behavior fairly early and adjust their strategy accordingly. Among these, advanced Claude models are the most consistent, while Llama 3.3 acts as a weaker but still competitive runner-up. DeepSeek R1 shows average performance, whereas Mistral Large often adapts late or inconsistently. Against \textbf{AP}, comprehension is generally delayed across models, which matches the more moderate total-point gains observed there and suggests that LLMs need several rounds of interaction before inferring the opponent’s play-style. 

In LLM--LLM interactions, comprehension is less stable than in PD. Since RPS does not admit cooperative convergence, models do not settle into a shared regime as quickly, and $m$ values remain comparatively high. These values are often fairly close to 25(>24), indicating that LLMs do not easily counter other LLMs’ strategies in the default RPS setting. For example, Claude 3.5 Sonnet v2 under zero-shot yields $m=10.6 \pm 13.1$ against its ZS counterpart, but $21.4 \pm 4.6$ and $19.6 \pm 5.6$ against SPP and CoT counterparts, respectively. Overall, unlike PD, RPS does not produce early and stable mutual understanding; instead, opponent comprehension remains delayed and variable even among stronger models. 

\paragraph{Cooperation Rate} is not defined in the same way for RPS as in PD, since the game is strictly adversarial. Instead, this metric captures deviations from uniform play, where a balanced strategy corresponds to selecting each action with probability $\sim \frac{1}{3}$. Against algorithmic opponents, deviations from this baseline indicate exploitation: when facing the \textbf{PP} player, stronger models (Claude 3.7/Claude Sonnet 4 variants) show clear biases toward the countering actions, reflecting successful identification of cyclic patterns. In contrast, against \textbf{AP} and \textbf{SREP}, action frequencies remain closer to uniform, suggesting limited exploitable structure and alignment with near-equilibrium play.

In LLM--LLM interactions, action distributions remain largely balanced across models, with most players approximating uniform randomization. This indicates the absence of stable exploitable patterns and is consistent with total points clustering around zero. However, weaker or less stable models (Mistral Large, DeepSeek R1) occasionally exhibit slight biases, which can lead to small but consistent gains or losses over repeated rounds.

\paragraph{Efficiency} in RPS is generally lower and more variable than in PD, reflecting the increased difficulty of reasoning over three actions without a dominant response. Against algorithmic opponents, efficiency is highest when exploitable structure exists, particularly against the \textbf{PP} player, where stronger models (Claude 3.7 and Claude 4 variants) convert early pattern recognition into higher returns with relatively stable token usage. In contrast, against \textbf{AP} and \textbf{SREP}, efficiency decreases across all models, as limited exploitable structure leads to near-equilibrium play and continued exploration. Models with delayed opponent comprehension, such as Mistral Large and to a lesser extent DeepSeek R1, incur higher token costs without proportional gains in total points, resulting in lower efficiency overall.

\paragraph{Failure Rate} remains negligible and mostly formatting-related, mirroring the PD findings.

%\subsubsection{Counterfactual analysis} RPS counterfactuals are particularly informative because they test both memorized label associations and sensitivity to changed incentives. In the strategy-only counterfactual (\textbf{eq1-alt}), several models, especially weaker ones, lose performance relative to the default setting, indicating difficulty in recomputing the altered dominance relation once familiar labels no longer match the canonical game. In the payoff counterfactuals (\textbf{ba3}, \textbf{ba3-alt}), successful play requires abandoning the uniform RPS heuristic and biasing action frequencies toward the higher-reward outcome; advanced Claude models adapt best, while other models often remain closer to the canonical policy. The joint counterfactual (\textbf{ba3-alt}) is the most demanding and reveals the clearest reasoning rigidity, including cases where enabling thinking appears to hurt rather than help. Overall, the RPS counterfactuals provide strong evidence that strategic competence in the default game does not always transfer to altered payoff or label structures.
\begin{figure*}[h!]
    \centering
    \includegraphics[width=0.9\linewidth, height=4cm]{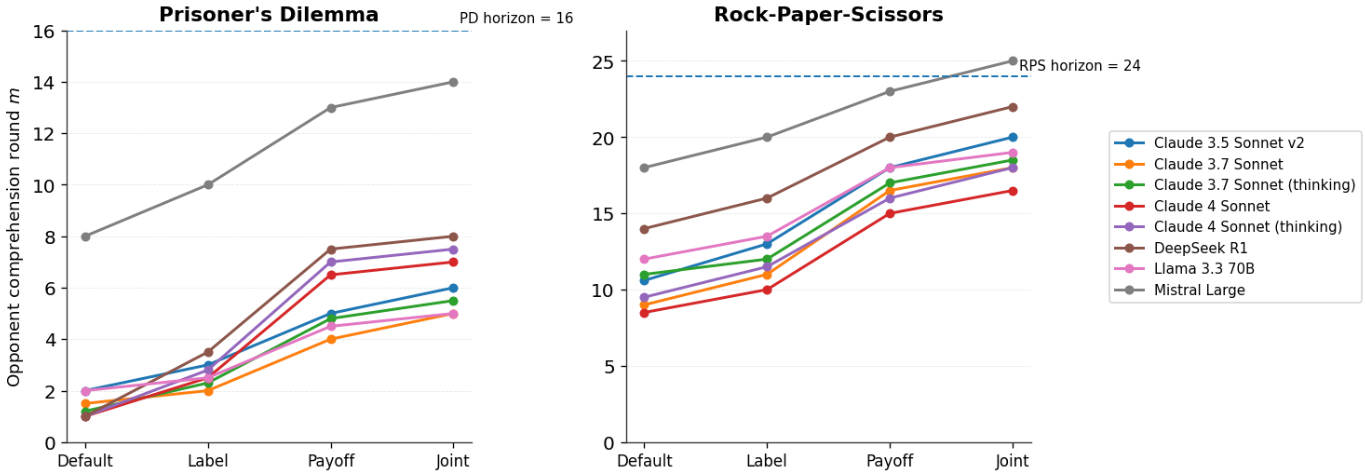}
    \caption{Counterfactual impact on opponent comprehension $m$ across models (zero-shot, LLM-vs-LLM).
    %For each model, we plot the reported opponent-comprehension value under the default, label-based, payoff-based, and joint counterfactual variants.
    Lower values indicate earlier adaptation, while values closer to the game horizon indicate delayed opponent modeling.}
    \label{fig:counterfactual_comprehension}
\end{figure*}
\subsubsection{Counterfactual analysis}

Label-only changes test whether models rely on memorized dominance relations between action names. Comparing the default RPS game with its label-based counterfactual (where action labels are permuted while keeping payoffs unchanged) reveals moderate but consistent degradation for several models. Stronger LLMs, particularly Claude 3.7 and Claude 4 variants, remain relatively stable, maintaining near-equilibrium total points and comparable opponent comprehension. In contrast, weaker or less stable models—most notably Mistral Large and, to a lesser extent, DeepSeek R1—show increased variability, delayed comprehension, and occasional deviations from uniform play. This indicates that even without changing incentives, renaming actions can disrupt learned associations, leading to slower or less consistent adaptation.

Payoff changes are more challenging, as they break game symmetry and require abandoning the canonical uniform mixed strategy. Under payoff-based counterfactuals, successful play requires shifting action frequencies toward the biased mixed equilibrium induced by the asymmetric payoff structure. Stronger LLMs again adapt more effectively, showing improved total points against exploitable opponents and earlier comprehension of advantageous actions. %However, many models remain close to uniform play, resulting in near-zero total points despite altered incentives. 
However, many models remain close to the canonical uniform distribution \((1/3,1/3,1/3)\), even though the payoff-counterfactual equilibrium is \((0.2,0.2,0.6)\). Thus, near-uniform play is no longer evidence of equilibrium behavior, but rather a form of equilibrium rigidity under altered incentives.
This is reflected in higher and more variable comprehension values and lower efficiency, as additional reasoning does not consistently translate into better outcomes. The joint counterfactual is the most demanding, with several models—especially Mistral Large and some thinking-enabled variants—showing near-horizon comprehension and unstable performance. %Overall, as in PD, payoff-based counterfactuals expose \textbf{stronger reasoning limitations} than label-only changes, since they require readjusting optimal strategies.
This motivates a cross-game view of whether delayed comprehension appears systematically as counterfactual pressure increases.

\paragraph{Cross-Game Counterfactual Summary.}

%Figure~\ref{fig:counterfactual_comprehension} summarizes the default-to-counterfactual shift in opponent comprehension $m$ across games and LLM opponents. A consistent behavior is revealed across LLMs tested: counterfactual games challenge model adaptability in unseen game setups, especially when both label and payoff perturbations are considered.
Figure~\ref{fig:counterfactual_comprehension} presents a stress trajectory from familiar to increasingly challenging game formulations. Across PD and RPS, several curves move upward from default to payoff-based and joint variants, showing that adaptation becomes delayed when surface recognition is no longer enough and incentive recomputation is required. Thus, counterfactual games reveal where strategic behavior stops transferring across altered settings.
\paragraph{Failure Modes.}
%Beyond score degradation, these patterns indicate distinct mechanisms by which strategic execution fails under counterfactual games.
Across games, counterfactual variants reveal four recurring failure modes. First, \textit{template persistence} occurs when a model preserves the canonical policy despite changed incentives; this is clearest in payoff-based RPS, where robust play should move away from uniform randomization toward \(\sigma^*=(0.2,0.2,0.6)\). Second, \textit{label anchoring} occurs when renamed actions alter behavior even though payoffs remain unchanged, suggesting partial reliance on surface associations. Third, \textit{delayed adaptation} occurs when models eventually exploit an opponent but only after many rounds, reflected in near-horizon opponent-comprehension values. Fourth, \textit{reasoning-overhead mismatch} occurs when reasoning-enabled variants use more tokens without proportional gains in payoff or adaptation speed. These failures are not random: label perturbations expose semantic anchoring, payoff perturbations expose incentive rigidity, and repeated interaction exposes unstable opponent modeling.

\section{Conclusion}
We introduced a counterfactual repeated-game framework for evaluating LLMs' strategic adaptation under modified labels and incentives. Our results show that canonical game performance can overstate strategic robustness: many models behave competently in default PD or RPS but degrade when payoffs or dominance relations are altered. Payoff-based counterfactuals are especially diagnostic because they require recomputing incentives rather than recognizing familiar action templates. The RPS payoff-counterfactual further shows that models can remain close to canonical uniform play even when the equilibrium shifts to a biased mixed strategy. Across settings, opponent comprehension and efficiency reveal that stronger or more deliberative models are not always better strategic agents. 
%These findings suggest that counterfactual games offer a compact and interpretable testbed for evaluating incentive sensitivity, opponent modeling, and strategic adaptation in LLM agents.
Thus, counterfactual games offer not only a benchmark for strategic robustness, but also a diagnostic lens for separating label anchoring, incentive rigidity, delayed opponent modeling, and inefficient deliberation.

\section*{Limitations}
Our study focuses on controlled, two-player repeated games, which, while enabling precise evaluation, may not fully capture the complexity of real-world strategic interactions. The use of fixed opponents and predefined payoff structures limits ecological validity, and results may vary under more diverse or multi-agent settings. Additionally, our evaluation relies on behavioral metrics derived from observed actions, which may not fully reflect the internal reasoning processes of LLMs. Finally, the set of models, prompting strategies, and counterfactual interventions considered is not exhaustive, and different configurations may yield different conclusions.

%Overall, our conclusions are behavioral rather than mechanistic: observed actions cannot directly reveal the internal reasoning process of the model, and opponent comprehension should therefore be interpreted as behavioral adaptation rather than literal mental-state inference. We therefore interpret counterfactual degradation as evidence of limited strategic robustness rather than definitive proof of memorization. Future work could combine behavioral evaluation with verbalized rationales, log-probability analysis, or controlled policy probes to better separate reasoning, retrieval, and execution failures.

Because our evaluation relies on observed actions, our conclusions are \textbf{behavioral} rather than mechanistic, and reported metrics should be interpreted as behavioral adaptation rather than literal mental-state inference.

%\section*{Acknowledgments}

\bibliography{acl_latex}

@article{Silver2016MasteringTG,
  title={Mastering the game of Go with deep neural networks and tree search},
  author={David Silver and Aja Huang and Chris J. Maddison and Arthur Guez and L. Sifre and George van den Driessche and Julian Schrittwieser and Ioannis Antonoglou and Vedavyas Panneershelvam and Marc Lanctot and Sander Dieleman and Dominik Grewe and John Nham and Nal Kalchbrenner and Ilya Sutskever and Timothy P. Lillicrap and Madeleine Leach and Koray Kavukcuoglu and Thore Graepel and Demis Hassabis},
  journal={Nature},
  year={2016},
  volume={529},
  pages={484-489},
  url={https://api.semanticscholar.org/CorpusID:515925}
}

@article{Bakhtin2022HumanlevelPI,
  title={Human-level play in the game of Diplomacy by combining language models with strategic reasoning},
  author={Anton Bakhtin and Noam Brown and Emily Dinan and Gabriele Farina and Colin Flaherty and Daniel Fried and Andrew Goff and Jonathan Gray and Hengyuan Hu and Athul Paul Jacob and Mo-jtaba Komeili and Karthik Konath and Minae Kwon and Adam Lerer and Mike Lewis and Alexander H. Miller and Sandra Mitts and Adithya Renduchintala and Stephen Roller and Dirk Rowe and Weiyan Shi and Joe Spisak and Alexander Wei and David J. Wu and Hugh Zhang and Markus Zijlstra},
  journal={Science},
  year={2022},
  volume={378},
  pages={1067 - 1074},
  url={https://api.semanticscholar.org/CorpusID:253759631}
}

@article{Berner2019Dota2W,
  title={Dota 2 with Large Scale Deep Reinforcement Learning},
  author={Christopher Berner and Greg Brockman and Brooke Chan and Vicki Cheung and Przemyslaw Debiak and Christy Dennison and David Farhi and Quirin Fischer and Shariq Hashme and Christopher Hesse and Rafal J{\'o}zefowicz and Scott Gray and Catherine Olsson and Jakub W. Pachocki and Michael Petrov and Henrique Pond{\'e} de Oliveira Pinto and Jonathan Raiman and Tim Salimans and Jeremy Schlatter and Jonas Schneider and Szymon Sidor and Ilya Sutskever and Jie Tang and Filip Wolski and Susan Zhang},
  journal={ArXiv},
  year={2019},
  volume={abs/1912.06680},
  url={https://api.semanticscholar.org/CorpusID:209376771}
}

@inproceedings{
gandhi2023strategic,
title={Strategic Reasoning with Language Models},
author={Kanishk Gandhi and Dorsa Sadigh and Noah Goodman},
booktitle={NeurIPS 2023 Foundation Models for Decision Making Workshop},
year={2023},
url={https://openreview.net/forum?id=MUtbsFRZwI}
}

@inproceedings{
zhang2024llm,
title={{LLM} as a Mastermind: A Survey of Strategic Reasoning with Large Language Models},
author={Yadong Zhang and Shaoguang Mao and Tao Ge and Xun Wang and Yan Xia and Wenshan Wu and Ting Song and Man Lan and Furu Wei},
booktitle={First Conference on Language Modeling},
year={2024},
url={https://openreview.net/forum?id=iMqJsQ4evS}
}

@inproceedings{
costarelli2024gamebench,
title={GameBench: Evaluating Strategic Reasoning Abilities of {LLM} Agents},
author={Anthony Costarelli and Mat Allen and Roman Hauksson and Grace Sodunke and Suhas Hariharan and Carlson Cheng and Wenjie Li and Joshua M Clymer and Arjun Yadav},
booktitle={Language Gamification - NeurIPS 2024 Workshop},
year={2024},
url={https://openreview.net/forum?id=qrzKE533Jr}
}

@inproceedings{
jia2025llm,
title={{LLM} Strategic Reasoning: Agentic Study through Behavioral Game Theory},
author={Jingru Jia and Zehua Yuan and Junhao Pan and Paul E McNamara and Deming Chen},
booktitle={The Thirty-ninth Annual Conference on Neural Information Processing Systems},
year={2025},
url={https://openreview.net/forum?id=XQrGTggLvT}
}

@misc{wang2025tmgbenchsystematicgamebenchmark,
      title={TMGBench: A Systematic Game Benchmark for Evaluating Strategic Reasoning Abilities of LLMs}, 
      author={Haochuan Wang and Xiachong Feng and Lei Li and Yu Guo and Zhanyue Qin and Dianbo Sui and Lingpeng Kong},
      year={2025},
      eprint={2410.10479},
      archivePrefix={arXiv},
      primaryClass={cs.AI},
      url={https://arxiv.org/abs/2410.10479}, 
}

@inproceedings{GTBENCH,
author = {Duan, Jinhao and Zhang, Renming and Diffenderfer, James and Kailkhura, Bhavya and Sun, Lichao and Stengel-Eskin, Elias and Bansal, Mohit and Chen, Tianlong and Xu, Kaidi},
title = {GTBENCH: uncovering the strategic reasoning limitations of LLMs via game-theoretic evaluations},
year = {2024},
isbn = {9798331314385},
publisher = {Curran Associates Inc.},
address = {Red Hook, NY, USA},
booktitle = {Proceedings of the 38th International Conference on Neural Information Processing Systems},
articleno = {885},
numpages = {35},
location = {Vancouver, BC, Canada},
series = {NIPS '24}
}

@misc{kader2025emergencestrategicreasoninglarge,
      title={The Emergence of Strategic Reasoning of Large Language Models}, 
      author={Gavin Kader and Dongwoo Lee},
      year={2025},
      eprint={2412.13013},
      archivePrefix={arXiv},
      primaryClass={econ.GN},
      url={https://arxiv.org/abs/2412.13013}, 
}

@inproceedings{matching-pennies,
author = {Silva, Alonso},
title = {Large Language Models Playing Mixed Strategy Nash Equilibrium Games},
year = {2024},
isbn = {978-3-031-78599-3},
publisher = {Springer-Verlag},
address = {Berlin, Heidelberg},
url = {https://doi.org/10.1007/978-3-031-78600-6_13},
doi = {10.1007/978-3-031-78600-6_13},
booktitle = {Network Games, Artificial Intelligence, Control and Optimization: 11th International Conference, NETGCOOP 2024, Lille, France, October 9–11, 2024, Proceedings},
pages = {142–152},
numpages = {11},
location = {Lille, France}
}

@inproceedings{rps,
author = {Fan, Caoyun and Chen, Jindou and Jin, Yaohui and He, Hao},
title = {Can large language models serve as rational players in game theory? a systematic analysis},
year = {2024},
isbn = {978-1-57735-887-9},
publisher = {AAAI Press},
url = {https://doi.org/10.1609/aaai.v38i16.29751},
doi = {10.1609/aaai.v38i16.29751},
booktitle = {Proceedings of the Thirty-Eighth AAAI Conference on Artificial Intelligence and Thirty-Sixth Conference on Innovative Applications of Artificial Intelligence and Fourteenth Symposium on Educational Advances in Artificial Intelligence},
articleno = {2003},
numpages = {8},
series = {AAAI'24/IAAI'24/EAAI'24}
}

@article{akata2025repeated,
  author = {Akata, Elif and Schulz, Eric and others},
  title = {Playing Repeated Games with Large Language Models},
  journal = {Nature Human Behaviour},
  year = {2025}
}

@article{loreheydari2024,
  author  = {Lor{\`e}, Nunzio and Heydari, Babak},
  title   = {Strategic Behavior of Large Language Models and the Role of Game Structure versus Contextual Framing},
  journal = {Scientific Reports},
  volume  = {14},
  number  = {1},
  pages   = {18490},
  year    = {2024},
  doi     = {10.1038/s41598-024-69032-z},
  url     = {https://doi.org/10.1038/s41598-024-69032-z},
  issn    = {2045-2322}
}

@inproceedings{aher2023turing,
author = {Aher, Gati and Arriaga, Rosa I. and Kalai, Adam Tauman},
title = {Using large language models to simulate multiple humans and replicate human subject studies},
year = {2023},
publisher = {JMLR.org},
booktitle = {Proceedings of the 40th International Conference on Machine Learning},
articleno = {17},
numpages = {35},
location = {Honolulu, Hawaii, USA},
series = {ICML'23}
}

@inproceedings{horton2023,
author = {Filippas, Apostolos and Horton, John J. and Manning, Benjamin S.},
title = {Large Language Models as Simulated Economic Agents: What Can We Learn from Homo Silicus?},
year = {2024},
isbn = {9798400707049},
publisher = {Association for Computing Machinery},
address = {New York, NY, USA},
url = {https://doi.org/10.1145/3670865.3673513},
doi = {10.1145/3670865.3673513},
booktitle = {Proceedings of the 25th ACM Conference on Economics and Computation},
pages = {614–615},
numpages = {2},
location = {New Haven, CT, USA},
series = {EC '24}
}

@misc{fontana2024pd,
      title={Nicer Than Humans: How do Large Language Models Behave in the Prisoner's Dilemma?}, 
      author={Nicoló Fontana and Francesco Pierri and Luca Maria Aiello},
      year={2024},
      eprint={2406.13605},
      archivePrefix={arXiv},
      primaryClass={cs.CY},
      url={https://arxiv.org/abs/2406.13605}, 
}

@inproceedings{mozikov24,
author = {Mozikov, Mikhail and Severin, Nikita and Bodishtianu, Valeria and Glushanina, Maria and Nasonov, Ivan and Orekhov, Daniil and Pekhotin, Vladislav and Makovetskiy, Ivan and Baklashkin, Mikhail and Lavrentyev, Vasily and Tsvigun, Akim and Turdakov, Denis and Shavrina, Tatiana and Savchenko, Andrey and Makarov, Ilya},
title = {EAI: emotional decision-making of LLMs in strategic games and ethical dilemmas},
year = {2024},
isbn = {9798331314385},
publisher = {Curran Associates Inc.},
address = {Red Hook, NY, USA},
booktitle = {Proceedings of the 38th International Conference on Neural Information Processing Systems},
articleno = {1709},
numpages = {34},
location = {Vancouver, BC, Canada},
series = {NIPS '24}
}

@inproceedings{
collins2025evaluating,
title={Evaluating Language Models' Evaluations of Games},
author={Katherine M. Collins and Cedegao E. Zhang and Graham Todd and Lance Ying and Mauricio Barba da Costa and Ryan Liu and Adrian Weller and Ionatan Kuperwajs and Lionel Wong and Joshua B. Tenenbaum and Thomas L. Griffiths},
booktitle={NeurIPS 2025 Workshop on Evaluating the Evolving LLM Lifecycle: Benchmarks, Emergent Abilities, and Scaling},
year={2025},
url={https://openreview.net/forum?id=A216uPJtY4}
}

@inproceedings{li-etal-2023-counterfactual,
    title = "Counterfactual reasoning: Testing language models' understanding of hypothetical scenarios",
    author = "Li, Jiaxuan  and
      Yu, Lang  and
      Ettinger, Allyson",
    editor = "Rogers, Anna  and
      Boyd-Graber, Jordan  and
      Okazaki, Naoaki",
    booktitle = "Proceedings of the 61st Annual Meeting of the Association for Computational Linguistics (Volume 2: Short Papers)",
    month = jul,
    year = "2023",
    address = "Toronto, Canada",
    publisher = "Association for Computational Linguistics",
    url = "https://aclanthology.org/2023.acl-short.70/",
    doi = "10.18653/v1/2023.acl-short.70",
    pages = "804--815"
}

@inproceedings{
yamin2025llms,
title={{LLM}s Struggle to Perform Counterfactual Reasoning with Parametric Knowledge},
author={Khurram Yamin and Gaurav Rohit Ghosal and Bryan Wilder},
booktitle={ICML 2025 Workshop on Scaling Up Intervention Models},
year={2025},
url={https://openreview.net/forum?id=EU9qCcKMv1}
}

@misc{chen2025counterbenchbenchmarkcounterfactualsreasoning,
      title={CounterBench: A Benchmark for Counterfactuals Reasoning in Large Language Models}, 
      author={Yuefei Chen and Vivek K. Singh and Jing Ma and Ruxiang Tang},
      year={2025},
      eprint={2502.11008},
      archivePrefix={arXiv},
      primaryClass={cs.CL},
      url={https://arxiv.org/abs/2502.11008}, 
}

@misc{akter2026causalconsistencyregularizationtraining,
      title={Causal Consistency Regularization: Training Verifiably Sensitive Reasoning in Large Language Models}, 
      author={Sanjeda Akter and Ibne Farabi Shihab and Anuj Sharma},
      year={2026},
      eprint={2509.01544},
      archivePrefix={arXiv},
      primaryClass={cs.AI},
      url={https://arxiv.org/abs/2509.01544}, 
}

@inproceedings{stringli-etal-2025-pitfalls,
    title = "Pitfalls of Scale: Investigating the Inverse Task of Redefinition in Large Language Models",
    author = "Stringli, Elena  and
      Lymperaiou, Maria  and
      Filandrianos, Giorgos  and
      Voulodimos, Athanasios  and
      Stamou, Giorgos",
    editor = "Che, Wanxiang  and
      Nabende, Joyce  and
      Shutova, Ekaterina  and
      Pilehvar, Mohammad Taher",
    booktitle = "Findings of the Association for Computational Linguistics: ACL 2025",
    month = jul,
    year = "2025",
    address = "Vienna, Austria",
    publisher = "Association for Computational Linguistics",
    url = "https://aclanthology.org/2025.findings-acl.492/",
    doi = "10.18653/v1/2025.findings-acl.492",
    pages = "9445--9469",
    ISBN = "979-8-89176-256-5"
}

@inproceedings{
huyuk2025reasoning,
title={Reasoning Elicitation in Language Models via Counterfactual Feedback},
author={Alihan H{\"u}y{\"u}k and Xinnuo Xu and Jacqueline R. M. A. Maasch and Aditya V. Nori and Javier Gonzalez},
booktitle={The Thirteenth International Conference on Learning Representations},
year={2025},
url={https://openreview.net/forum?id=VVixJ9QavY}
}

@inproceedings{wu-etal-2024-reasoning,
    title = "Reasoning or Reciting? Exploring the Capabilities and Limitations of Language Models Through Counterfactual Tasks",
    author = {Wu, Zhaofeng  and
      Qiu, Linlu  and
      Ross, Alexis  and
      Aky{\"u}rek, Ekin  and
      Chen, Boyuan  and
      Wang, Bailin  and
      Kim, Najoung  and
      Andreas, Jacob  and
      Kim, Yoon},
    editor = "Duh, Kevin  and
      Gomez, Helena  and
      Bethard, Steven",
    booktitle = "Proceedings of the 2024 Conference of the North American Chapter of the Association for Computational Linguistics: Human Language Technologies (Volume 1: Long Papers)",
    month = jun,
    year = "2024",
    address = "Mexico City, Mexico",
    publisher = "Association for Computational Linguistics",
    url = "https://aclanthology.org/2024.naacl-long.102/",
    doi = "10.18653/v1/2024.naacl-long.102",
    pages = "1819--1862"
}

@inproceedings{frohberg-binder-2022-crass,
    title = "{CRASS}: A Novel Data Set and Benchmark to Test Counterfactual Reasoning of Large Language Models",
    author = {Frohberg, J{\"o}rg  and
      Binder, Frank},
    editor = "Calzolari, Nicoletta  and
      B{\'e}chet, Fr{\'e}d{\'e}ric  and
      Blache, Philippe  and
      Choukri, Khalid  and
      Cieri, Christopher  and
      Declerck, Thierry  and
      Goggi, Sara  and
      Isahara, Hitoshi  and
      Maegaard, Bente  and
      Mariani, Joseph  and
      Mazo, H{\'e}l{\`e}ne  and
      Odijk, Jan  and
      Piperidis, Stelios",
    booktitle = "Proceedings of the Thirteenth Language Resources and Evaluation Conference",
    month = jun,
    year = "2022",
    address = "Marseille, France",
    publisher = "European Language Resources Association",
    url = "https://aclanthology.org/2022.lrec-1.229/",
    pages = "2126--2140"
}

@inproceedings{spp,
    title = "Unleashing the Emergent Cognitive Synergy in Large Language Models: A Task-Solving Agent through Multi-Persona Self-Collaboration",
    author = "Wang, Zhenhailong  and
      Mao, Shaoguang  and
      Wu, Wenshan  and
      Ge, Tao  and
      Wei, Furu  and
      Ji, Heng",
    editor = "Duh, Kevin  and
      Gomez, Helena  and
      Bethard, Steven",
    booktitle = "Proceedings of the 2024 Conference of the North American Chapter of the Association for Computational Linguistics: Human Language Technologies (Volume 1: Long Papers)",
    month = jun,
    year = "2024",
    address = "Mexico City, Mexico",
    publisher = "Association for Computational Linguistics",
    url = "https://aclanthology.org/2024.naacl-long.15/",
    doi = "10.18653/v1/2024.naacl-long.15",
    pages = "257--279"
}

@inproceedings{
wang2023selfconsistency,
title={Self-Consistency Improves Chain of Thought Reasoning in Language Models},
author={Xuezhi Wang and Jason Wei and Dale Schuurmans and Quoc V Le and Ed H. Chi and Sharan Narang and Aakanksha Chowdhery and Denny Zhou},
booktitle={The Eleventh International Conference on Learning Representations },
year={2023},
url={https://openreview.net/forum?id=1PL1NIMMrw}
}

\appendix

\section{Counterfactual game instances}
\label{sec:counterfactual}
\subsection{Prisoner's dilemma}
\label{sec:pd}
In the \textbf{payoff-based} counterfactual of the Prisoner’s Dilemma (PD), we replace the canonical PD payoff structure with that of the Stag Hunt (SH). While the PD satisfies the ordering
\begin{equation*}
 T>R>P>S   
\end{equation*}
which makes defection the dominant strategy in the one-shot game, the Stag Hunt satisfies
\begin{equation*}
 R>T\geq P>S   
\end{equation*}
thereby removing dominance and introducing a coordination structure.

Under SH, two pure Nash equilibria exist: the payoff-dominant equilibrium (S, S) that yields the highest joint reward but requires mutual commitment, and the risk-dominant equilibrium (H, H), which provides a safer but lower payoff. Unlike PD, unilateral deviation from cooperation does not strictly dominate, and strategic incentives shift from individual exploitation to equilibrium selection.
This modification alters the fundamental game dynamics: In PD, rational play in the one-shot setting favors defection; in SH, strategic reasoning requires balancing risk and coordination. Consequently, agents must adjust their behavior from exploitation-oriented reasoning (as in PD) to equilibrium selection reasoning (as in SH).

This counterfactual therefore tests whether LLMs recompute incentives when the dominance structure changes. If a model continues to favor defection-like behavior under SH—despite the absence of strict dominance—this suggests over-reliance on the canonical PD template. Conversely, increased coordination on the payoff-dominant equilibrium indicates sensitivity to the modified incentive structure.

\subsection{Rock-Paper-Scissors}
\label{sec:rps}
In the case of \textbf{payoff-based} RPS counterfactuals, the game is no longer symmetric and the mixed-Nash equilibrium strategy is not uniform. Consequently, players are required to adopt a biased strategy rather than the canonical uniform one:
\begin{equation}
\begin{split}
 \sigma^*(R)=0.2,\\ 
 \sigma^*(P)=0.2,\\ 
 \sigma^*(S)=0.6.   
\end{split}
\end{equation}
In equilibrium, each action played with positive probability must yield equal expected payoff. Because the modified reward structure changes these expectations asymmetrically, the resulting strategy distribution becomes biased toward actions associated with higher expected returns.

This counterfactual formulation serves as a diagnostic for reasoning flexibility. If a model continues to approximate the canonical uniform distribution despite asymmetric incentives, this suggests reliance on memorized RPS heuristics rather than recomputation under the altered payoff structure. Conversely, observable shifts in action frequencies toward the higher-reward actions indicate sensitivity to incentive changes and behavioral adaptation to the modified game dynamics.

\section{Proof of expected outcomes}
\label{app:expected_outcomes}

In this section, we derive the expected per-round payoff under stationary mixed strategies for both Prisoner's Dilemma (PD) and Rock--Paper--Scissors (RPS), including their counterfactual variants.

\subsection{Prisoner's Dilemma}
\label{app:pd_expected_payoff}

Let \(p = \Pr(C)\) denote the probability that player \(i\) cooperates, and let \(q = \Pr(C)\) denote the corresponding probability for the opponent. Then the joint-action probabilities are:
\begin{equation}
 \begin{aligned}
\Pr(C,C) &= pq, \\
\Pr(C,D) &= p(1-q), \\
\Pr(D,C) &= (1-p)q, \\
\Pr(D,D) &= (1-p)(1-q).
\end{aligned}
\end{equation}
Using the payoff instantiation
\begin{equation}
 \begin{aligned}
(C,C)=(4,4), \quad (C,D)=(1,6), \\
\quad (D,C)=(6,1), \quad (D,D)=(2,2)
\end{aligned}   
\end{equation}
the expected payoff for player \(i\) is:
\begin{equation}
\begin{split}
\mathbb{E}[u_i]&= 4pq \\
&\quad+ 1 \cdot p(1-q) \\
&\quad+ 6 \cdot (1-p)q \\
&\quad+ 2 \cdot (1-p)(1-q).
\end{split}   
\end{equation}
Expanding and simplifying gives:
\begin{equation}
\begin{aligned}
\mathbb{E}[u_i]
&= 4pq + p - pq + 6q - 6pq \\
&\quad + 2 - 2p - 2q + 2pq \\
&= 2 - p + 4q - pq.
\end{aligned}   
\end{equation}
This expression provides the expected per-round payoff as a function of the players' mixed strategies.

\subsection{Rock--Paper--Scissors}
\label{app:rps_expected_payoff}

Let \(\sigma_i = (p_R, p_P, p_S)\) and \(\sigma_j = (q_R, q_P, q_S)\) denote the mixed strategies of player \(i\) and the opponent, respectively, where each component corresponds to the probability of selecting Rock, Paper, or Scissors. Thus,
\begin{equation}
\begin{split}
    p_R+p_P+p_S=1 \\ 
   q_R+q_P+q_S=1    
\end{split}
\end{equation}
Under the standard payoff structure, where a win yields \(1\), a loss yields \(-1\), and a tie yields \(0\), the expected payoff for player \(i\) is:
\begin{equation}
   \begin{aligned}
\mathbb{E}[u_i]
&= (p_R q_S + p_P q_R + p_S q_P) \\
&\quad- (p_R q_P + p_P q_S + p_S q_R).
\end{aligned} 
\end{equation}
The first term corresponds to winning outcomes, while the second term corresponds to losing outcomes; ties contribute zero.

At the mixed Nash equilibrium of standard RPS, both players randomize uniformly:
\begin{equation}
    \begin{split}
p_R=p_P=p_S=\frac{1}{3} \\ q_R=q_P=q_S=\frac{1}{3}        
    \end{split}
\end{equation}
Substituting these values into the expected payoff expression gives:
\begin{equation}
\begin{aligned}
\mathbb{E}[u_i]
&=
\left(
\frac{1}{3}\frac{1}{3}
+
\frac{1}{3}\frac{1}{3}
+
\frac{1}{3}\frac{1}{3}
\right) \\
&\quad-
\left(
\frac{1}{3}\frac{1}{3}
+
\frac{1}{3}\frac{1}{3}
+
\frac{1}{3}\frac{1}{3}
\right) \\
&= 0.
\end{aligned}    
\end{equation}
This establishes the zero expected-payoff baseline used in the main analysis. More generally, deviations from the uniform strategy can yield positive or negative expected payoff depending on the opponent's distribution.

\subsection{Counterfactual payoff instantiations}
\label{app:counterfactual_payoff_instantiations}

\subsubsection{Prisoner's Dilemma Counterfactuals}
\label{app:pd_counterfactual_expected_payoff}

We consider three counterfactual instantiations of the PD family: a label-based counterfactual, a payoff-based counterfactual, and a joint counterfactual that alters both labels and payoffs.

\paragraph{Label-based counterfactual.}
In the label-based counterfactual, only the action names are changed, while payoffs remain identical to the default game. We replace \(C \rightarrow S\) and \(D \rightarrow H\), yielding:
\[
\begin{array}{c|cc}
 & S & H \\
\hline
S & (4,4) & (1,6) \\
H & (6,1) & (2,2)
\end{array}
\]

\paragraph{Payoff-based counterfactual.}
In the payoff-based counterfactual, the labels remain \(C\) and \(D\), but the payoff matrix is replaced by the Stag Hunt one:
\[
\begin{array}{c|cc}
 & C & D \\
\hline
C & (6,6) & (1,4) \\
D & (4,1) & (2,2)
\end{array}
\]

\paragraph{Joint counterfactual.}
Finally, the joint counterfactual alters both labels and payoffs, giving the standard Stag Hunt formulation:
\[
\begin{array}{c|cc}
 & S & H \\
\hline
S & (6,6) & (1,4) \\
H & (4,1) & (2,2)
\end{array}
\]

\paragraph{Expected payoff under stationary mixed strategies.}
Let \(p\) denote the probability that player \(i\) selects the first action of the game, and let \(q\) denote the corresponding probability for the opponent. In the default and payoff-based settings, the first action is \(C\); in the label-based and joint settings, the first action is \(S\). The joint-action probabilities are:
\begin{equation}
  \begin{aligned}
\Pr(C,C) &= pq, \\
\Pr(C,D) &= p(1-q), \\
\Pr(D,C) &= (1-p)q, \\
\Pr(D,D) &= (1-p)(1-q),
\end{aligned}  
\end{equation}
with the same notation applying to \((S,H)\) in the label-based settings.

For the default and label-based counterfactual instantiations, the expected per-round payoff for player \(i\) is:
\begin{equation}
\begin{aligned}
\mathbb{E}[u_i]
&= 4pq \\
&\quad+ 1 \cdot p(1-q) \\
&\quad+ 6 \cdot (1-p)q \\
&\quad+ 2 \cdot (1-p)(1-q).
\end{aligned}  
\end{equation}
Expanding and simplifying:
\begin{equation}
  \begin{aligned}
\mathbb{E}[u_i]
&= 4pq + p - pq + 6q -  \\
&\quad 6pq + 2 - 2p - 2q + 2pq \\
&= 2 - p + 4q - pq.
\end{aligned}  
\end{equation}
For the payoff-based and joint counterfactual instantiations, the expected per-round payoff becomes:
\begin{equation}
 \begin{aligned}
\mathbb{E}[u_i]
&= 6pq \\
&\quad + 1 \cdot p(1-q) \\
&\quad + 4 \cdot (1-p)q \\
&\quad + 2 \cdot (1-p)(1-q).
\end{aligned}   
\end{equation}
Expanding and simplifying:
\begin{equation}
 \begin{aligned}
\mathbb{E}[u_i]
&= 6pq + p - pq + 4q - \\
&\quad 4pq + 2 - 2p - 2q + 2pq \\
&= 2 - p + 2q + 3pq.
\end{aligned}   
\end{equation}
These expressions make explicit how the payoff-based counterfactual changes the incentive structure relative to the default PD formulation.

\subsubsection{Rock--Paper--Scissors Counterfactuals}
\label{app:rps_counterfactual_expected_payoff}

We consider three counterfactual instantiations of Rock--Paper--Scissors: a label-based counterfactual, a payoff-based counterfactual, and a joint counterfactual that alters both labels and payoffs.

\paragraph{Label-based counterfactual.}
In the label-based counterfactual, only the dominance induced by the labels is changed, while payoff magnitudes remain identical to the default game. We permute the action names so that the game is presented as:
\[
\begin{array}{c|ccc}
 & P & R & S \\
\hline
P & (0,0) & (-1,1) & (1,-1) \\
R & (1,-1) & (0,0) & (-1,1) \\
S & (-1,1) & (1,-1) & (0,0)
\end{array}
\]

\paragraph{Payoff-based counterfactual.}
In the payoff-based counterfactual, the action labels remain \(R\), \(P\), and \(S\), but one win--loss interaction is amplified. Specifically, the Rock--Paper outcome receives magnitude \(3\) instead of \(1\):
\[
\begin{array}{c|ccc}
 & R & P & S \\
\hline
R & (0,0) & (-3,3) & (1,-1) \\
P & (3,-3) & (0,0) & (-1,1) \\
S & (-1,1) & (1,-1) & (0,0)
\end{array}
\]

\paragraph{Joint counterfactual.}
Finally, the joint counterfactual alters both labels and payoffs, yielding:
\[
\begin{array}{c|ccc}
 & P & R & S \\
\hline
P & (0,0) & (-3,3) & (1,-1) \\
R & (3,-3) & (0,0) & (-1,1) \\
S & (-1,1) & (1,-1) & (0,0)
\end{array}
\]

\paragraph{Expected payoff under stationary mixed strategies.}
Let player \(i\) choose actions with probabilities
\[
p_R = \Pr(R), \qquad p_P = \Pr(P), \qquad p_S = \Pr(S),
\]
with \(p_R+p_P+p_S=1\), and let the opponent choose actions with probabilities
\[
q_R = \Pr(R), \qquad q_P = \Pr(P), \qquad q_S = \Pr(S),
\]
with \(q_R+q_P+q_S=1\).

For the default and label-based counterfactual instantiations, the expected per-round payoff for player \(i\) is:
\begin{equation}
 \begin{aligned}
\mathbb{E}[u_i]
&= (p_R q_S + p_P q_R + p_S q_P) \\
&\quad - (p_R q_P + p_P q_S + p_S q_R).
\end{aligned}   
\end{equation}
The first term collects winning outcomes and the second term collects losing outcomes; ties contribute zero.

For the payoff-based and joint counterfactual instantiations, the Rock--Paper interaction is weighted by \(3\), so the expected payoff becomes:
\begin{equation}
\begin{aligned}
\mathbb{E}[u_i]
&= 3p_P q_R + p_R q_S + p_S q_P \\
&\quad - \left(3p_R q_P + p_P q_S + p_S q_R\right).
\end{aligned}    
\end{equation}
Equivalently, the expected payoffs of the pure actions \(R\), \(P\), and \(S\) against opponent distribution \(q=(q_R,q_P,q_S)\) are:
\begin{equation}
\begin{aligned}
u(R) &= -3q_P + q_S, \\
u(P) &= 3q_R - q_S, \\
u(S) &= -q_R + q_P.
\end{aligned}  
\end{equation}
At a fully mixed Nash equilibrium, all pure actions in support must yield equal expected payoff. Since this payoff matrix is zero-sum and skew-symmetric, the value of the game is \(0\). Therefore:
\begin{equation}
\begin{aligned}
-3q_P + q_S &= 0, \\
3q_R - q_S &= 0, \\
-q_R + q_P &= 0.
\end{aligned}  
\end{equation}
From the third equation, \(q_R=q_P\). From the second equation, \(q_S=3q_R\). Combining these with the normalization constraint,
\[
q_R+q_P+q_S=1,
\]
we obtain:
\[
q_R+q_R+3q_R=1,
\]
and therefore:
\[
q_R=q_P=\frac{1}{5}, \qquad q_S=\frac{3}{5}.
\]
Thus, the unique mixed Nash equilibrium of the payoff-based RPS counterfactual is:
\[
\sigma^*(R)=\frac{1}{5}, \qquad
\sigma^*(P)=\frac{1}{5}, \qquad
\sigma^*(S)=\frac{3}{5}.
\]

Equivalently,
\[
\sigma^* = (0.2, 0.2, 0.6)
\]
over \((R,P,S)\). This shows that the payoff-based counterfactual breaks the canonical uniform equilibrium of standard RPS and requires a biased mixed strategy rather than uniform randomization.

\section{Implementation details}
\label{ref:implementation}
\subsection{System prompts}
A \textbf{Game Description} is the part of a system prompt that effectively describes the game being played.

For Prisoner's Dilemma that Game Description is the following.

\par\noindent\rule{\linewidth}{0.4pt}
\small
You are playing a game against other player(s). The game is based on strategic moves.\\

Game Description:\\
There are two possible moves: \{a\} and \{b\}.\\
- if both players choose \{a\}, both get \{aa\} points.\\
- if both players choose \{b\}, both get \{bb\} points.\\
- if you choose \{a\} and the other player chooses \{b\}, you get \{ab\} points and the other player gets \{ba\} points.\\
- if you choose \{b\} and the other player chooses \{a\}, you get \{ba\} points and the other player gets \{ab\} points.\\

In each round of the game you play against one other player. You and the other player each make a move privately. The moves are compared and you and the other player get your reward/penalty based on the moves made.\\

Message Formats:\\
Move messages are sent when you want to make a move.\\
Move messages are formatted like this:\\
\texttt{[}move\texttt{]} (Optional explanation here) Your move here\\

Hint messages are received when you are provided with advice and directions. These messages are important and you should pay attention to them.\\
Hint messages are formatted like this:\\
\texttt{[}hint\texttt{]} Hint message here\\

Objective:\\
Maximize your points and/or minimize your penalties.\\

Tips:\\
- Try to formulate a strategy by reasoning on the provided information.\\
- Pay attention to the payoff matrix of the game moves.\\
- You should only send move messages. Hint messages will only be provided to you.\\
- Pay attention to hints when provided.\\
- Do not use markdown formatting in your messages (e.g., bold text, italic text, etc.).
\par\noindent\rule{\linewidth}{0.4pt}
\normalsize 

For Rock-Paper-Scissors that Game Description is the following.

\par\noindent\rule{\linewidth}{0.4pt}
\small
You are playing a game against other player(s). The game is based on making moves.\\

Game Description:\\
There are three possible moves: \{a\}, \{b\}, and \{c\}.\\
- \{a\} beats \{c\} giving the winner \{ac\} points and the loser -\{ac\} points.\\
- \{b\} beats \{a\} giving the winner \{ba\} points and the loser -\{ba\} points.\\
- \{c\} beats \{b\} giving the winner \{cb\} points and the loser -\{cb\} points.\\
- If both players make the same move, the game is a tie and no points are awarded.\\

In each round of the game you play against one other player. You and the other player each make a move privately. The moves are compared and you and the other player get your reward/penalty based on the moves made.\\

Message Formats:\\
Move messages are sent when you want to make a move.\\
Move messages are formatted like this:\\
\texttt{[}move\texttt{]} (Optional explanation here) Your move here\\

Hint messages are received when you are provided with advice and directions. These messages are important and you should pay attention to them.\\
Hint messages are formatted like this:\\
\texttt{[}hint\texttt{]} Hint message here\\

Objective:\\
Maximize your points and/or minimize your penalties.\\

Tips:\\
- Try to formulate a strategy by reasoning on the provided information.\\
- Pay attention to the payoff matrix of the game moves.\\
- You should only send move messages. Hint messages will only be provided to you.\\
- Pay attention to hints when provided.\\
- Do not use markdown formatting in your messages (e.g., bold text, italic text, etc.).
\par\noindent\rule{\linewidth}{0.4pt}
\normalsize

\subsection{Prompting techniques}

We employ three prompting strategies to instantiate LLM players: zero-shot (ZS), chain-of-thought (CoT), and solo performance prompting (SPP), optionally combined with self-consistency (SC).

\paragraph{Zero-shot (ZS).}
The model receives a direct instruction describing the game, available actions, and history, and is asked to output a valid action. No explicit reasoning guidance is provided.

\textbf{Template:}
\begin{quote}
You are playing a repeated game. At each round, choose one of the available actions. The history of previous rounds is provided below. Based on this, select your next move. Output only your final action.

History: [previous rounds]

Available actions: [actions]
\end{quote}

\paragraph{Chain-of-Thought (CoT).}
We augment the zero-shot prompt with explicit reasoning instructions, encouraging step-by-step analysis before selecting an action.

\textbf{Template:}
\begin{quote}
You are playing a repeated game. Analyze the opponent's behavior and think step by step before deciding your move. Then output your final action.

History: [previous rounds]

Available actions: [actions]
\end{quote}

\paragraph{Solo Performance Prompting (SPP).}
Following \citet{spp}, we prompt the model to internally consider multiple reasoning perspectives before producing a final decision, aiming to improve strategic consistency.

\textbf{Template:}
\begin{quote}
You are playing a repeated game. Consider multiple possible strategies and perspectives before deciding your next move. Reflect briefly on the opponent's behavior and select the best action.

History: [previous rounds]

Available actions: [actions]
\end{quote}

\paragraph{Self-Consistency (SC).}
Self-consistency is applied on top of the above prompting strategies by sampling multiple candidate responses per round and selecting the final action via majority vote. This reduces variance and improves robustness of the selected strategy.

\paragraph{Decoding parameters}
All models are queried with temperature $=0.7$, top-$p=1.0$, and a maximum of 50 output tokens. %We use greedy decoding for final action extraction after reasoning.
Action extraction is deterministic: after generation, we select the final valid action according to the post-processing rules below.

\paragraph{Action extraction}
Model outputs are post-processed using string matching to extract valid actions (e.g., C/D or R/P/S). If multiple actions are present, the last valid token is selected.  In this case the LLM resubmits its answer until there is a unique valid action present. Invalid outputs are counted toward the failure rate.

\paragraph{Interaction protocol}
At each round, the model is provided with the full history of previous actions and corresponding payoffs. The opponent’s last action is explicitly revealed. Models do not have access to future rounds.

\paragraph{Model access}
All models are accessed via their respective APIs via Bedrock AWS. Claude models are queried through Anthropic API (versions 3.5, 3.7, and 4), DeepSeek R1 via official API, and Llama 3.3-70B via hosted inference endpoints.

\section{Full results}
\label{sec:full-results}
\begin{table*}[h!]
\small
\centering
\begin{tabular}{lllllllll}
\toprule
 &  & \multicolumn{7}{c}{\textbf{PD}} \\
 &  & zs & spp & cot & srep & pp & mf & tft \\
model & prompt &  &  &  &  &  &  &  \\
\midrule
\multirow[t]{6}{*}{C3.5Sv2} & zs & \textbf{64.0 $\pm$ 0.0} & \textbf{64.0 $\pm$ 0.0} & \textbf{64.0 $\pm$ 0.0} & 29.8 $\pm$ 0.4 & 54.2 $\pm$ 4.5 & 30.2 $\pm$ 1.1 & 30.6 $\pm$ 0.9 \\
 & cot & \textbf{64.0 $\pm$ 0.0} & \textbf{64.0 $\pm$ 0.0} & \textbf{64.0 $\pm$ 0.0} & 30.0 $\pm$ 0.0 & 53.6 $\pm$ 1.5 & 31.0 $\pm$ 1.0 & 30.4 $\pm$ 0.9 \\
 & spp & \textbf{64.0 $\pm$ 0.0} & \textbf{64.0 $\pm$ 0.0} & \textbf{64.0 $\pm$ 0.0} & 30.4 $\pm$ 0.5 & 58.4 $\pm$ 4.3 & 30.6 $\pm$ 0.9 & 32.0 $\pm$ 0.0 \\
 & sc-zs & \textbf{64.0 $\pm$ 0.0} & \textbf{64.0 $\pm$ 0.0} & \textbf{64.0 $\pm$ 0.0} & 30.5 $\pm$ 0.7 & 52.0 $\pm$ 0.0 & 32.0 $\pm$ 0.0 & 30.0 $\pm$ 1.4 \\
 & sc-cot & \textbf{64.0 $\pm$ 0.0} & \textbf{64.0 $\pm$ 0.0} & \textbf{64.0 $\pm$ 0.0} & 30.0 $\pm$ 0.0 & 53.5 $\pm$ 2.1 & 32.0 $\pm$ 0.0 & 30.0 $\pm$ 0.0 \\
 & sc-spp & \textbf{64.0 $\pm$ 0.0} & \textbf{64.0 $\pm$ 0.0} & \textbf{64.0 $\pm$ 0.0} & 30.0 $\pm$ 0.0 & 62.0 $\pm$ 0.0 & 31.0 $\pm$ 1.4 & 31.0 $\pm$ 1.4 \\
\cline{1-9}
\multirow[t]{6}{*}{C3.7S} & zs & \textbf{64.0 $\pm$ 0.0} & \textbf{64.0 $\pm$ 0.0} & \textbf{64.0 $\pm$ 0.0} & 30.0 $\pm$ 0.0 & 54.8 $\pm$ 4.1 & 29.8 $\pm$ 0.8 & 30.4 $\pm$ 0.5 \\
 & cot & \textbf{64.0 $\pm$ 0.0} & \textbf{64.0 $\pm$ 0.0} & \textbf{64.0 $\pm$ 0.0} & 30.0 $\pm$ 0.0 & 61.2 $\pm$ 1.1 & 31.0 $\pm$ 0.7 & 31.2 $\pm$ 0.4 \\
 & spp & \textbf{64.0 $\pm$ 0.0} & \textbf{64.0 $\pm$ 0.0} & \textbf{64.0 $\pm$ 0.0} & 30.0 $\pm$ 0.0 & 56.8 $\pm$ 4.4 & 31.0 $\pm$ 1.0 & 30.2 $\pm$ 0.4 \\
 & sc-zs & \textbf{64.0 $\pm$ 0.0} & \textbf{64.0 $\pm$ 0.0} & \textbf{64.0 $\pm$ 0.0} & 30.0 $\pm$ 0.0 & 52.5 $\pm$ 0.7 & 31.0 $\pm$ 1.4 & 30.0 $\pm$ 0.0 \\
 & sc-cot & \textbf{64.0 $\pm$ 0.0} & \textbf{64.0 $\pm$ 0.0} & \textbf{64.0 $\pm$ 0.0} & 30.5 $\pm$ 0.7 & 61.0 $\pm$ 1.4 & 30.5 $\pm$ 0.7 & 31.0 $\pm$ 0.0 \\
 & sc-spp & \textbf{64.0 $\pm$ 0.0} & \textbf{64.0 $\pm$ 0.0} & \textbf{64.0 $\pm$ 0.0} & 30.0 $\pm$ 0.0 & 57.5 $\pm$ 6.4 & 31.0 $\pm$ 1.4 & 31.5 $\pm$ 0.7 \\
\cline{1-9}
\multirow[t]{6}{*}{C3.7S(T)} & zs & \textbf{64.0 $\pm$ 0.0} & \textbf{64.0 $\pm$ 0.0} & \textbf{64.0 $\pm$ 0.0} & 29.8 $\pm$ 0.4 & 52.8 $\pm$ 0.4 & 31.0 $\pm$ 0.7 & 31.6 $\pm$ 0.9 \\
 & cot & \textbf{64.0 $\pm$ 0.0} & \textbf{64.0 $\pm$ 0.0} & \textbf{64.0 $\pm$ 0.0} & 30.6 $\pm$ 0.5 & 57.2 $\pm$ 3.7 & 30.8 $\pm$ 0.8 & 30.2 $\pm$ 0.4 \\
 & spp & \textbf{64.0 $\pm$ 0.0} & \textbf{64.0 $\pm$ 0.0} & \textbf{64.0 $\pm$ 0.0} & 30.2 $\pm$ 0.4 & 58.0 $\pm$ 4.7 & 31.0 $\pm$ 1.0 & 31.0 $\pm$ 1.0 \\
 & sc-zs & \textbf{64.0 $\pm$ 0.0} & \textbf{64.0 $\pm$ 0.0} & \textbf{64.0 $\pm$ 0.0} & 30.0 $\pm$ 0.0 & 52.0 $\pm$ 0.0 & 32.0 $\pm$ 0.0 & 30.0 $\pm$ 0.0 \\
 & sc-cot & \textbf{64.0 $\pm$ 0.0} & \textbf{64.0 $\pm$ 0.0} & \textbf{64.0 $\pm$ 0.0} & 30.5 $\pm$ 0.7 & 53.0 $\pm$ 0.0 & 31.0 $\pm$ 1.4 & 30.5 $\pm$ 0.7 \\
 & sc-spp & \textbf{64.0 $\pm$ 0.0} & \textbf{64.0 $\pm$ 0.0} & \textbf{64.0 $\pm$ 0.0} & 30.5 $\pm$ 0.7 & 57.5 $\pm$ 2.1 & 32.0 $\pm$ 0.0 & 31.0 $\pm$ 1.4 \\
\cline{1-9}
\multirow[t]{6}{*}{C4S} & zs & 49.4 $\pm$ 15.5 & 41.0 $\pm$ 10.7 & 42.0 $\pm$ 10.7 & 31.4 $\pm$ 0.5 & \textbf{55.8 $\pm$ 4.4} & 32.8 $\pm$ 1.6 & 32.2 $\pm$ 1.6 \\
 & cot & 35.2 $\pm$ 3.0 & 39.6 $\pm$ 2.4 & 39.8 $\pm$ 7.4 & 31.4 $\pm$ 0.5 & \textbf{60.2 $\pm$ 4.5} & 32.2 $\pm$ 1.6 & 34.2 $\pm$ 1.9 \\
 & spp & 48.4 $\pm$ 13.0 & 46.2 $\pm$ 13.6 & 36.4 $\pm$ 4.2 & 31.6 $\pm$ 0.5 & \textbf{63.0 $\pm$ 1.0} & 34.4 $\pm$ 2.1 & 31.8 $\pm$ 2.4 \\
 & sc-zs & 34.5 $\pm$ 4.9 & 34.0 $\pm$ 2.8 & 40.0 $\pm$ 4.2 &32.0 $\pm$ 0.0 & \textbf{58.0 $\pm$ 8.5} & 32.0 $\pm$ 0.0 & 32.0 $\pm$ 0.0 \\
 & sc-cot & 39.0 $\pm$ 4.2 & 35.5 $\pm$ 4.9 & 34.0 $\pm$ 2.8 & 32.0 $\pm$ 0.0 & \textbf{64.0 $\pm$ 0.0} & 35.0 $\pm$ 0.0 & 33.5 $\pm$ 2.1 \\
 & sc-spp & 32.0 $\pm$ 1.4 & 38.0 $\pm$ 1.4 & 35.5 $\pm$ 4.9 & 30.5 $\pm$ 0.7 & \textbf{63.5 $\pm$ 0.7} & 35.5 $\pm$ 0.7 & 31.5 $\pm$ 0.7 \\
\cline{1-9}
\multirow[t]{6}{*}{C4S(T)} & zs & 32.0 $\pm$ 0.0 & 34.8 $\pm$ 2.7 & 32.0 $\pm$ 0.0 & 32.0 $\pm$ 0.0 & \textbf{63.4 $\pm$ 0.9} & 35.2 $\pm$ 1.8 & 35.8 $\pm$ 0.4 \\
 & cot & 32.0 $\pm$ 0.0 & 40.8 $\pm$ 8.0 & 34.6 $\pm$ 3.6 & 31.6 $\pm$ 0.5 & \textbf{64.0 $\pm$ 0.0} & 33.8 $\pm$ 1.8 & 34.4 $\pm$ 2.2 \\
 & spp & 31.4 $\pm$ 0.9 & 50.6 $\pm$ 15.4 & 41.8 $\pm$ 12.4 & 31.4 $\pm$ 0.5 & \textbf{59.6 $\pm$ 5.2} & 34.8 $\pm$ 1.8 & 34.8 $\pm$ 2.2 \\
 & sc-zs & 37.5 $\pm$ 7.8 & 34.0 $\pm$ 2.8 & 32.0 $\pm$ 0.0 & 32.0 $\pm$ 0.0 & \textbf{64.0 $\pm$ 0.0} & 32.0 $\pm$ 0.0 & 33.5 $\pm$ 2.1 \\
 & sc-cot & 32.0 $\pm$ 0.0 & 37.0 $\pm$ 1.4 & 32.0 $\pm$ 0.0 & 32.0 $\pm$ 0.0 & \textbf{64.0 $\pm$ 0.0} & 36.0 $\pm$ 0.0 & 32.0 $\pm$ 0.0 \\
 & sc-spp & 47.0 $\pm$ 21.2 & 36.0 $\pm$ 0.0 & 32.0 $\pm$ 0.0 & 31.5 $\pm$ 0.7 & \textbf{62.0 $\pm$ 2.8} & 35.5 $\pm$ 0.7 & 33.5 $\pm$ 2.1 \\
\cline{1-9}
\multirow[t]{6}{*}{DS-R1} & zs & 36.8 $\pm$ 4.1 & 39.0 $\pm$ 2.5 & 30.2 $\pm$ 13.0 & 31.6 $\pm$ 0.5 & \textbf{62.6 $\pm$ 1.5} & 32.8 $\pm$ 2.0 & 32.6 $\pm$ 2.2 \\
 & cot & 33.2 $\pm$ 2.6 & 32.2 $\pm$ 2.2 & 33.0 $\pm$ 6.6 & 28.6 $\pm$ 7.1 & \textbf{64.0 $\pm$ 0.0} & 32.2 $\pm$ 1.8 & 23.6 $\pm$ 8.7 \\
 & spp & 36.4 $\pm$ 4.2 & 32.0 $\pm$ 6.4 & 31.2 $\pm$ 6.3 & 31.2 $\pm$ 0.4 & \textbf{61.4 $\pm$ 3.7} & 33.2 $\pm$ 2.6 & 33.2 $\pm$ 1.6 \\
 & sc-zs & 32.5 $\pm$ 2.1 & 33.0 $\pm$ 2.8 & 33.5 $\pm$ 3.5 & 31.5 $\pm$ 0.7 & \textbf{64.0 $\pm$ 0.0} & 35.0 $\pm$ 0.0 & 35.5 $\pm$ 0.7 \\
 & sc-cot & 34.5 $\pm$ 4.9 & 34.0 $\pm$ 2.8 & 27.0 $\pm$ 7.1 & 32.0 $\pm$ 0.0 & \textbf{64.0 $\pm$ 0.0} & 31.5 $\pm$ 0.7 & 31.5 $\pm$ 0.7 \\
 & sc-spp & 35.0 $\pm$ 4.2 & 36.0 $\pm$ 5.7 & 34.0 $\pm$ 2.8 & 31.0 $\pm$ 0.0 & \textbf{61.0 $\pm$ 4.2} & 33.5 $\pm$ 2.1 & 32.0 $\pm$ 0.0 \\
\cline{1-9}
\multirow[t]{6}{*}{L3.3-70B} & zs & \textbf{64.0 $\pm$ 0.0} & \textbf{64.0 $\pm$ 0.0} & \textbf{64.0 $\pm$ 0.0} & 29.6 $\pm$ 0.5 & 52.0 $\pm$ 0.0 & 31.6 $\pm$ 0.9 & 31.6 $\pm$ 0.9 \\
 & cot & \textbf{64.0 $\pm$ 0.0} & \textbf{64.0 $\pm$ 0.0} & \textbf{64.0 $\pm$ 0.0} & 29.8 $\pm$ 0.4 & 55.6 $\pm$ 4.6 & 30.8 $\pm$ 1.1 & 30.4 $\pm$ 0.9 \\
 & spp & \textbf{64.0 $\pm$ 0.0} & \textbf{64.0 $\pm$ 0.0} & \textbf{64.0 $\pm$ 0.0} & 29.8 $\pm$ 0.4 & 53.8 $\pm$ 2.9 & 30.8 $\pm$ 1.1 & 30.4 $\pm$ 0.9 \\
 & sc-zs & \textbf{64.0 $\pm$ 0.0} & \textbf{64.0 $\pm$ 0.0} & \textbf{64.0 $\pm$ 0.0} & 29.0 $\pm$ 0.0 & 52.0 $\pm$ 0.0 & 31.0 $\pm$ 1.4 & 32.0 $\pm$ 0.0 \\
 & sc-cot & \textbf{64.0 $\pm$ 0.0} & \textbf{64.0 $\pm$ 0.0} & \textbf{64.0 $\pm$ 0.0} & 30.0 $\pm$ 0.0 & 52.0 $\pm$ 0.0 & 32.0 $\pm$ 0.0 & 31.0 $\pm$ 1.4 \\
 & sc-spp & \textbf{64.0 $\pm$ 0.0} & \textbf{64.0 $\pm$ 0.0} & \textbf{64.0 $\pm$ 0.0} & 29.5 $\pm$ 0.7 & 55.5 $\pm$ 4.9 & 30.0 $\pm$ 0.0 & 32.0 $\pm$ 0.0 \\
\cline{1-9}
\multirow[t]{6}{*}{Mistral} & zs & \textbf{61.4 $\pm$ 3.2} & 60.0 $\pm$ 4.4 & 59.6 $\pm$ 9.8 & 24.2 $\pm$ 0.4 & 53.8 $\pm$ 2.0 & 26.0 $\pm$ 1.6 & 25.8 $\pm$ 0.8 \\
 & cot & \textbf{64.0 $\pm$ 0.0} & 60.2 $\pm$ 3.6 & 62.0 $\pm$ 2.8 & 29.8 $\pm$ 2.2 & 57.8 $\pm$ 3.0 & 29.0 $\pm$ 1.2 & 26.0 $\pm$ 1.2 \\
 & spp & \textbf{64.0 $\pm$ 0.0} & 62.2 $\pm$ 4.0 & 60.8 $\pm$ 7.2 & 18.6 $\pm$ 10.6 & 53.2 $\pm$ 2.2 & 26.4 $\pm$ 3.2 & 24.4 $\pm$ 1.1 \\
 & sc-zs & \textbf{64.0 $\pm$ 0.0} & \textbf{64.0 $\pm$ 0.0} & \textbf{64.0 $\pm$ 0.0} & 24.0 $\pm$ 0.0 & 54.0 $\pm$ 2.8 & 25.0 $\pm$ 1.4 & 25.0 $\pm$ 1.4 \\
 & sc-cot & 62.0 $\pm$ 2.8 & \textbf{64.0 $\pm$ 0.0} & 60.5 $\pm$ 4.9 & 27.5 $\pm$ 4.9 & 60.0 $\pm$ 5.7 & 33.5 $\pm$ 3.5 & 29.0 $\pm$ 4.2 \\
 & sc-spp & \textbf{64.0 $\pm$ 0.0} & \textbf{64.0 $\pm$ 0.0} & \textbf{64.0 $\pm$ 0.0} & 24.0 $\pm$ 0.0 & 56.0 $\pm$ 0.0 & 22.5 $\pm$ 4.9 & 22.5 $\pm$ 2.1 \\
\bottomrule
\end{tabular}
\caption{Total Points Averaged Over All Iterations for the default PD. \textbf{Bold} indicates the highest total points per row.}
\label{tab:pd_total_points_avg_heatmap_pd}
\end{table*}

Analysis in Section \ref{sec:results} is backed by numerical results fully presented in the following tables. These tables exhibit LLM behavior across prompting strategies and opponents for all evaluation metrics used in this paper. Due to space constraints, we use shorthand names in tables:

C3.5Sv2 = Claude Sonnet 3.5 v2,

C3.7S = Claude Sonnet 3.7,

C3.7S(T) = Claude Sonnet 3.7 (thinking),

C4S = Claude Sonnet 4,

C4S(T) = Claude Sonnet 4 (thinking),

DS-R1 = DeepSeek R1,

L3.3-70B = Llama 3.3 70B,

Mistral = Mistral Large (24.07).
\subsection{PD Total points}
\begin{table*}[h!]
\small
\centering
\begin{tabular}{p{1cm}llllllll}
\toprule
 &  & \multicolumn{7}{c}{\textbf{PD label-based counterfactual}} \\
 &  & zs & spp & cot & srep & pp & mf & tft \\
model & prompt &  &  &  &  &  &  &  \\
\midrule
\multirow[t]{6}{*}{C3.5Sv2} & zs & \textbf{96.0 $\pm$ 0.0} & \textbf{96.0 $\pm$ 0.0} & \textbf{96.0 $\pm$ 0.0} & 44.0 $\pm$ 5.3 & 52.0 $\pm$ 2.1 & 89.0 $\pm$ 0.0 & 60.6 $\pm$ 32.3 \\
 & cot & \textbf{96.0 $\pm$ 0.0} & \textbf{96.0 $\pm$ 0.0} & \textbf{96.0 $\pm$ 0.0} & 47.8 $\pm$ 3.8 & 49.6 $\pm$ 2.1 & 91.8 $\pm$ 3.8 & 60.0 $\pm$ 32.9 \\
 & spp & \textbf{96.0 $\pm$ 0.0} & \textbf{96.0 $\pm$ 0.0} & \textbf{96.0 $\pm$ 0.0} & 44.8 $\pm$ 9.8 & 47.6 $\pm$ 1.7 & 90.4 $\pm$ 3.1 & 72.0 $\pm$ 32.9 \\
 & sc-zs & \textbf{96.0 $\pm$ 0.0} & \textbf{96.0 $\pm$ 0.0} & \textbf{96.0 $\pm$ 0.0} & 40.0 $\pm$ 7.1 & 54.0 $\pm$ 1.4 & 89.0 $\pm$ 0.0 & \textbf{96.0 $\pm$ 0.0} \\
 & sc-cot & \textbf{96.0 $\pm$ 0.0} & \textbf{96.0 $\pm$ 0.0} & \textbf{96.0 $\pm$ 0.0} & 52.5 $\pm$ 4.9 & 52.0 $\pm$ 0.0 & 89.0 $\pm$ 0.0 & 66.0 $\pm$ 42.4 \\
 & sc-spp & \textbf{96.0 $\pm$ 0.0} & \textbf{96.0 $\pm$ 0.0} & \textbf{96.0 $\pm$ 0.0} & 52.5 $\pm$ 13.4 & 52.0 $\pm$ 8.5 & 92.5 $\pm$ 4.9 & 34.5 $\pm$ 0.7 \\
\cline{1-9}
\multirow[t]{6}{*}{C3.7S} & zs & \textbf{96.0 $\pm$ 0.0} & \textbf{96.0 $\pm$ 0.0} & \textbf{96.0 $\pm$ 0.0} & 43.6 $\pm$ 8.1 & 49.0 $\pm$ 3.0 & 91.8 $\pm$ 3.8 & 72.6 $\pm$ 32.1 \\
 & cot & \textbf{96.0 $\pm$ 0.0} & \textbf{96.0 $\pm$ 0.0} & \textbf{96.0 $\pm$ 0.0} & 45.0 $\pm$ 5.0 & 52.2 $\pm$ 3.3 & 93.2 $\pm$ 3.8 & 60.2 $\pm$ 32.7 \\
 & spp & \textbf{96.0 $\pm$ 0.0} & \textbf{96.0 $\pm$ 0.0} & \textbf{96.0 $\pm$ 0.0} & 43.2 $\pm$ 6.7 & 48.2 $\pm$ 4.4 & 94.6 $\pm$ 3.1 & 48.6 $\pm$ 26.6 \\
 & sc-zs & \textbf{96.0 $\pm$ 0.0} & \textbf{96.0 $\pm$ 0.0} & \textbf{96.0 $\pm$ 0.0} & 43.0 $\pm$ 14.1 & 46.5 $\pm$ 0.7 & 92.5 $\pm$ 4.9 & 37.5 $\pm$ 0.7 \\
 & sc-cot & \textbf{96.0 $\pm$ 0.0} & \textbf{96.0 $\pm$ 0.0} & \textbf{96.0 $\pm$ 0.0} & 40.0 $\pm$ 9.9 & 52.0 $\pm$ 8.5 & 92.5 $\pm$ 4.9 & 66.5 $\pm$ 41.7 \\
 & sc-spp & \textbf{96.0 $\pm$ 0.0} & \textbf{96.0 $\pm$ 0.0} & \textbf{96.0 $\pm$ 0.0} & 42.0 $\pm$ 5.7 & 49.0 $\pm$ 4.2 & 92.5 $\pm$ 4.9 & 36.5 $\pm$ 0.7 \\
\cline{1-9}
\multirow[t]{6}{*}{C3.7S(T)} & zs & \textbf{96.0 $\pm$ 0.0} & \textbf{96.0 $\pm$ 0.0} & \textbf{96.0 $\pm$ 0.0} & 44.6 $\pm$ 8.1 & 50.0 $\pm$ 2.0 & 67.2 $\pm$ 29.9 & 48.2 $\pm$ 26.7 \\
 & cot & \textbf{96.0 $\pm$ 0.0} & 90.8 $\pm$ 11.6 & \textbf{96.0 $\pm$ 0.0} & 47.8 $\pm$ 7.3 & 49.8 $\pm$ 4.8 & \textbf{96.0 $\pm$ 0.0} & 72.2 $\pm$ 32.6 \\
 & spp & \textbf{96.0 $\pm$ 0.0} & \textbf{96.0 $\pm$ 0.0} & \textbf{96.0 $\pm$ 0.0} & 37.8 $\pm$ 4.6 & 49.6 $\pm$ 7.2 & 91.8 $\pm$ 3.8 & 72.2 $\pm$ 32.6 \\
 & sc-zs & \textbf{96.0 $\pm$ 0.0} & \textbf{96.0 $\pm$ 0.0} & \textbf{96.0 $\pm$ 0.0} & 49.5 $\pm$ 20.5 & 48.0 $\pm$ 1.4 & 61.5 $\pm$ 38.9 & \textbf{96.0 $\pm$ 0.0} \\
 & sc-cot & \textbf{96.0 $\pm$ 0.0} & \textbf{96.0 $\pm$ 0.0} & \textbf{96.0 $\pm$ 0.0} & 45.0 $\pm$ 1.4 & \textbf{55.5 $\pm$ 3.5} & 89.0 $\pm$ 0.0 & 35.5 $\pm$ 0.7 \\
 & sc-spp & \textbf{96.0 $\pm$ 0.0} & \textbf{96.0 $\pm$ 0.0} & \textbf{96.0 $\pm$ 0.0} & 41.0 $\pm$ 8.5 & 48.5 $\pm$ 2.1 & 92.5 $\pm$ 4.9 & 65.5 $\pm$ 43.1 \\
\cline{1-9}
\multirow[t]{6}{*}{C4S} & zs & 71.8 $\pm$ 33.1 & \textbf{94.6 $\pm$ 3.1} & 80.4 $\pm$ 25.2 & 43.0 $\pm$ 5.7 & 47.0 $\pm$ 1.2 & 66.2 $\pm$ 31.2 & 47.4 $\pm$ 27.2 \\
 & cot & \textbf{92.2 $\pm$ 3.5} & 66.8 $\pm$ 29.1 & 55.2 $\pm$ 27.3 & 38.2 $\pm$ 3.5 & 50.0 $\pm$ 5.2 & 45.0 $\pm$ 28.5 & 47.0 $\pm$ 27.4 \\
 & spp & \textbf{83.0 $\pm$ 25.3} & 46.2 $\pm$ 27.9 & 46.8 $\pm$ 27.5 & 39.6 $\pm$ 3.0 & 49.2 $\pm$ 5.3 & 69.2 $\pm$ 33.6 & 45.4 $\pm$ 24.4 \\
 & sc-zs & \textbf{96.0 $\pm$ 0.0} & 93.0 $\pm$ 4.2 & 88.0 $\pm$ 1.4 & 38.5 $\pm$ 3.5 & 46.0 $\pm$ 0.0 & 89.0 $\pm$ 0.0 & 36.0 $\pm$ 2.8 \\
 & sc-cot & 63.0 $\pm$ 41.0 & 32.5 $\pm$ 0.7 & 34.5 $\pm$ 0.7 & 41.5 $\pm$ 0.7 & 45.5 $\pm$ 0.7 & \textbf{63.5 $\pm$ 46.0} & 36.0 $\pm$ 1.4 \\
 & sc-spp & 94.0 $\pm$ 2.8 & 93.0 $\pm$ 4.2 & 57.5 $\pm$ 38.9 & 42.0 $\pm$ 4.2 & 47.0 $\pm$ 1.4 & \textbf{96.0 $\pm$ 0.0} & 54.0 $\pm$ 26.9 \\
\cline{1-9}
\multirow[t]{6}{*}{C4S(T)} & zs & \textbf{96.0 $\pm$ 0.0} & 84.0 $\pm$ 26.8 & 83.0 $\pm$ 29.1 & 41.6 $\pm$ 5.7 & 47.4 $\pm$ 1.1 & 79.0 $\pm$ 26.4 & 59.6 $\pm$ 33.2 \\
 & cot & 59.0 $\pm$ 33.8 & 71.6 $\pm$ 33.4 & 70.0 $\pm$ 35.6 & 45.2 $\pm$ 6.1 & 47.2 $\pm$ 1.3 & 55.4 $\pm$ 30.7 & \textbf{71.8 $\pm$ 33.1} \\
 & spp & 82.2 $\pm$ 26.1 & 68.0 $\pm$ 31.0 & 81.8 $\pm$ 27.0 & 39.6 $\pm$ 5.2 & 50.2 $\pm$ 5.5 & \textbf{91.0 $\pm$ 3.1} & 83.6 $\pm$ 27.7 \\
 & sc-zs & \textbf{96.0 $\pm$ 0.0} & \textbf{96.0 $\pm$ 0.0} & 91.5 $\pm$ 6.4 & 57.0 $\pm$ 2.8 & 50.5 $\pm$ 2.1 & 33.5 $\pm$ 2.1 & 36.0 $\pm$ 1.4 \\
 & sc-cot & 65.5 $\pm$ 43.1 & 63.5 $\pm$ 46.0 & \textbf{96.0 $\pm$ 0.0} & 43.0 $\pm$ 4.2 & 47.5 $\pm$ 0.7 & 64.0 $\pm$ 45.3 & 65.5 $\pm$ 43.1 \\
 & sc-spp & \textbf{92.5 $\pm$ 4.9} & 60.5 $\pm$ 41.7 & 65.0 $\pm$ 43.8 & 42.5 $\pm$ 2.1 & 47.0 $\pm$ 1.4 & 60.5 $\pm$ 40.3 & 33.0 $\pm$ 1.4 \\
\cline{1-9}
\multirow[t]{6}{*}{DS-R1} & zs & 46.6 $\pm$ 27.6 & 33.4 $\pm$ 2.4 & 30.6 $\pm$ 4.8 & 44.4 $\pm$ 2.9 & 48.0 $\pm$ 1.9 & \textbf{56.4 $\pm$ 33.1} & 46.8 $\pm$ 27.5 \\
 & cot & 32.0 $\pm$ 1.9 & 32.6 $\pm$ 1.5 & 45.4 $\pm$ 28.3 & 38.6 $\pm$ 4.3 & \textbf{47.8 $\pm$ 0.8} & 32.2 $\pm$ 1.5 & 45.4 $\pm$ 28.3 \\
 & spp & 31.6 $\pm$ 2.2 & 37.0 $\pm$ 11.8 & \textbf{53.4 $\pm$ 27.9} & 42.0 $\pm$ 2.2 & 47.8 $\pm$ 2.4 & 45.0 $\pm$ 28.5 & 51.6 $\pm$ 27.3 \\
 & sc-zs & 33.0 $\pm$ 1.4 & 35.0 $\pm$ 5.7 & 35.0 $\pm$ 1.4 & 39.0 $\pm$ 1.4 & \textbf{48.0 $\pm$ 1.4} & 32.0 $\pm$ 1.4 & 33.5 $\pm$ 2.1 \\
 & sc-cot & 63.5 $\pm$ 46.0 & 34.0 $\pm$ 2.8 & 34.0 $\pm$ 0.0 & 46.5 $\pm$ 9.2 & 46.5 $\pm$ 0.7 & 65.0 $\pm$ 43.8 & \textbf{65.5 $\pm$ 43.1} \\
 & sc-spp & 28.5 $\pm$ 9.2 & 33.0 $\pm$ 1.4 & 35.0 $\pm$ 1.4 & 46.0 $\pm$ 2.8 & \textbf{47.0 $\pm$ 1.4} & 31.0 $\pm$ 0.0 & 34.0 $\pm$ 0.0 \\
\cline{1-9}
\multirow[t]{6}{*}{L3.3-70B} & zs & \textbf{96.0 $\pm$ 0.0} & \textbf{96.0 $\pm$ 0.0} & \textbf{96.0 $\pm$ 0.0} & 41.4 $\pm$ 4.3 & 52.0 $\pm$ 0.0 & 70.8 $\pm$ 34.5 & 47.0 $\pm$ 27.4 \\
 & cot & \textbf{96.0 $\pm$ 0.0} & \textbf{96.0 $\pm$ 0.0} & \textbf{96.0 $\pm$ 0.0} & 45.0 $\pm$ 2.9 & 49.6 $\pm$ 3.3 & 81.0 $\pm$ 26.0 & 46.8 $\pm$ 27.5 \\
 & spp & \textbf{96.0 $\pm$ 0.0} & \textbf{96.0 $\pm$ 0.0} & \textbf{96.0 $\pm$ 0.0} & 41.4 $\pm$ 10.5 & 48.0 $\pm$ 1.4 & 91.8 $\pm$ 3.8 & 83.8 $\pm$ 27.3 \\
 & sc-zs & \textbf{96.0 $\pm$ 0.0} & \textbf{96.0 $\pm$ 0.0} & \textbf{96.0 $\pm$ 0.0} & 35.5 $\pm$ 6.4 & 52.0 $\pm$ 0.0 & \textbf{96.0 $\pm$ 0.0} & \textbf{96.0 $\pm$ 0.0} \\
 & sc-cot & \textbf{96.0 $\pm$ 0.0} & \textbf{96.0 $\pm$ 0.0} & \textbf{96.0 $\pm$ 0.0} & 39.5 $\pm$ 0.7 & 46.0 $\pm$ 0.0 & \textbf{96.0 $\pm$ 0.0} & 65.0 $\pm$ 43.8 \\
 & sc-spp & \textbf{96.0 $\pm$ 0.0} & \textbf{96.0 $\pm$ 0.0} & \textbf{96.0 $\pm$ 0.0} & 44.0 $\pm$ 0.0 & 52.0 $\pm$ 0.0 & 65.5 $\pm$ 43.1 & \textbf{96.0 $\pm$ 0.0} \\
\cline{1-9}
\multirow[t]{6}{*}{Mistral} & zs & \textbf{96.0 $\pm$ 0.0} & 79.0 $\pm$ 32.2 & \textbf{96.0 $\pm$ 0.0} & 46.2 $\pm$ 6.6 & 42.4 $\pm$ 3.9 & 79.8 $\pm$ 28.6 & 83.4 $\pm$ 25.5 \\
 & cot & 90.4 $\pm$ 11.4 & 84.8 $\pm$ 25.0 & \textbf{96.0 $\pm$ 0.0} & 43.2 $\pm$ 3.8 & 43.4 $\pm$ 4.8 & 79.4 $\pm$ 29.5 & 72.2 $\pm$ 30.4 \\
 & spp & \textbf{96.0 $\pm$ 0.0} & 62.6 $\pm$ 42.1 & 92.6 $\pm$ 5.0 & 43.2 $\pm$ 9.9 & 41.2 $\pm$ 2.7 & 82.0 $\pm$ 24.8 & 59.4 $\pm$ 28.9 \\
 & sc-zs & \textbf{96.0 $\pm$ 0.0} & \textbf{96.0 $\pm$ 0.0} & 83.5 $\pm$ 17.7 & 42.5 $\pm$ 4.9 & 49.0 $\pm$ 0.0 & 92.5 $\pm$ 4.9 & \textbf{96.0 $\pm$ 0.0} \\
 & sc-cot & \textbf{96.0 $\pm$ 0.0} & \textbf{96.0 $\pm$ 0.0} & \textbf{96.0 $\pm$ 0.0} & 37.5 $\pm$ 2.1 & 44.5 $\pm$ 2.1 & 92.5 $\pm$ 4.9 & 68.0 $\pm$ 39.6 \\
 & sc-spp & \textbf{96.0 $\pm$ 0.0} & 60.5 $\pm$ 50.2 & \textbf{96.0 $\pm$ 0.0} & 37.0 $\pm$ 2.8 & 40.0 $\pm$ 0.0 & 89.0 $\pm$ 0.0 & 58.5 $\pm$ 26.2 \\
\bottomrule
\end{tabular}
\caption{Total Points Averaged Over All Iterations for the label-based PD counterfactual.}
\label{tab:pd_total_points_avg_heatmap_pd-alt}
\end{table*}

Table \ref{tab:pd_total_points_avg_heatmap_pd} showcases total points averaged for the number of iterations default PD was played (16). Bold entries denote the maximum total points achieved per LLM in PD. Higher total points indicate better strategic performance, which may arise from either cooperation (e.g., in LLM--LLM interactions) or successful exploitation (e.g., against algorithmic opponents).

\begin{table*}[h!]
\small
\centering
\begin{tabular}{p{1cm}p{0.9cm}lllllll}
\toprule
 &  & \multicolumn{7}{c}{\textbf{SH (payoff-based counterfactual of PD)}} \\
 &  & zs & spp & cot & srep & pp & mf & tft \\
model & prompt &  &  &  &  &  &  &  \\
\midrule
\multirow[t]{6}{*}{C3.5Sv2} & zs & \textbf{96.0 $\pm$ 0.0} & \textbf{96.0 $\pm$ 0.0} & \textbf{96.0 $\pm$ 0.0} & 43.8 $\pm$ 3.7 & 51.2 $\pm$ 5.1 & 91.8 $\pm$ 3.8 & 59.2 $\pm$ 33.6 \\
 & cot & \textbf{96.0 $\pm$ 0.0} & \textbf{96.0 $\pm$ 0.0} & \textbf{96.0 $\pm$ 0.0} & 43.0 $\pm$ 3.1 & 48.2 $\pm$ 0.8 & 59.4 $\pm$ 33.4 & 46.8 $\pm$ 27.5 \\
 & spp & \textbf{96.0 $\pm$ 0.0} & \textbf{96.0 $\pm$ 0.0} & \textbf{96.0 $\pm$ 0.0} & 36.8 $\pm$ 3.8 & 46.8 $\pm$ 1.3 & 79.8 $\pm$ 24.7 & 83.6 $\pm$ 27.7 \\
 & sc-zs & \textbf{96.0 $\pm$ 0.0} & \textbf{96.0 $\pm$ 0.0} & \textbf{96.0 $\pm$ 0.0} & 44.5 $\pm$ 6.4 & 52.0 $\pm$ 8.5 & 92.5 $\pm$ 4.9 & \textbf{96.0 $\pm$ 0.0} \\
 & sc-cot & \textbf{96.0 $\pm$ 0.0} & \textbf{96.0 $\pm$ 0.0} & \textbf{96.0 $\pm$ 0.0} & 40.5 $\pm$ 2.1 & 53.0 $\pm$ 7.1 & 34.0 $\pm$ 0.0 & 65.5 $\pm$ 43.1 \\
 & sc-spp & \textbf{96.0 $\pm$ 0.0} & \textbf{96.0 $\pm$ 0.0} & \textbf{96.0 $\pm$ 0.0} & \textbf{49.0 $\pm$ 2.8} & 46.0 $\pm$ 0.0 & 66.0 $\pm$ 42.4 & 65.0 $\pm$ 43.8 \\
\cline{1-9}
\multirow[t]{6}{*}{C3.7S} & zs & \textbf{96.0 $\pm$ 0.0} & \textbf{96.0 $\pm$ 0.0} & \textbf{96.0 $\pm$ 0.0} & 46.8 $\pm$ 5.4 & 52.0 $\pm$ 4.7 & 93.2 $\pm$ 3.8 & 72.8 $\pm$ 31.8 \\
 & cot & \textbf{96.0 $\pm$ 0.0} & \textbf{96.0 $\pm$ 0.0} & \textbf{96.0 $\pm$ 0.0} & 42.6 $\pm$ 6.6 & 51.2 $\pm$ 3.7 & 93.2 $\pm$ 3.8 & 71.4 $\pm$ 33.7 \\
 & spp & \textbf{96.0 $\pm$ 0.0} & \textbf{96.0 $\pm$ 0.0} & \textbf{96.0 $\pm$ 0.0} & 40.4 $\pm$ 2.3 & 53.2 $\pm$ 4.5 & 93.2 $\pm$ 3.8 & 60.6 $\pm$ 32.3 \\
 & sc-zs & \textbf{96.0 $\pm$ 0.0} & \textbf{96.0 $\pm$ 0.0} & \textbf{96.0 $\pm$ 0.0} & 39.5 $\pm$ 0.7 & 48.0 $\pm$ 1.4 & 89.0 $\pm$ 0.0 & 65.5 $\pm$ 43.1 \\
 & sc-cot & \textbf{96.0 $\pm$ 0.0} & \textbf{96.0 $\pm$ 0.0} & \textbf{96.0 $\pm$ 0.0} & 45.0 $\pm$ 1.4 & 56.5 $\pm$ 2.1 & 92.5 $\pm$ 4.9 & 35.0 $\pm$ 1.4 \\
 & sc-spp & \textbf{96.0 $\pm$ 0.0} & \textbf{96.0 $\pm$ 0.0} & \textbf{96.0 $\pm$ 0.0} & 42.5 $\pm$ 6.4 & 52.0 $\pm$ 8.5 & 92.5 $\pm$ 4.9 & \textbf{96.0 $\pm$ 0.0} \\
\cline{1-9}
\multirow[t]{6}{*}{C3.7S(T)} & zs & \textbf{96.0 $\pm$ 0.0} & \textbf{96.0 $\pm$ 0.0} & \textbf{96.0 $\pm$ 0.0} & 47.2 $\pm$ 5.8 & 50.2 $\pm$ 3.4 & 81.4 $\pm$ 26.1 & 59.6 $\pm$ 33.2 \\
 & cot & \textbf{96.0 $\pm$ 0.0} & \textbf{96.0 $\pm$ 0.0} & \textbf{96.0 $\pm$ 0.0} & 40.4 $\pm$ 4.8 & 52.2 $\pm$ 4.9 & 94.6 $\pm$ 3.1 & 59.0 $\pm$ 33.8 \\
 & spp & \textbf{96.0 $\pm$ 0.0} & \textbf{96.0 $\pm$ 0.0} & \textbf{96.0 $\pm$ 0.0} & 39.8 $\pm$ 4.8 & 54.0 $\pm$ 2.8 & 91.8 $\pm$ 3.8 & 59.6 $\pm$ 33.2 \\
 & sc-zs & \textbf{96.0 $\pm$ 0.0} & \textbf{96.0 $\pm$ 0.0} & \textbf{96.0 $\pm$ 0.0} & 41.5 $\pm$ 3.5 & 48.0 $\pm$ 1.4 & 92.5 $\pm$ 4.9 & 35.0 $\pm$ 1.4 \\
 & sc-cot & \textbf{96.0 $\pm$ 0.0} & \textbf{96.0 $\pm$ 0.0} & \textbf{96.0 $\pm$ 0.0} & 39.0 $\pm$ 1.4 & 55.5 $\pm$ 0.7 & \textbf{96.0 $\pm$ 0.0} & 65.5 $\pm$ 43.1 \\
 & sc-spp & \textbf{96.0 $\pm$ 0.0} & \textbf{96.0 $\pm$ 0.0} & \textbf{96.0 $\pm$ 0.0} & 41.5 $\pm$ 10.6 & 50.5 $\pm$ 6.4 & \textbf{96.0 $\pm$ 0.0} & 35.0 $\pm$ 0.0 \\
\cline{1-9}
\multirow[t]{6}{*}{C4S} & zs & \textbf{84.4 $\pm$ 25.9} & 58.8 $\pm$ 31.8 & 70.4 $\pm$ 29.0 & 39.6 $\pm$ 4.9 & 52.0 $\pm$ 6.0 & 68.8 $\pm$ 34.2 & 52.4 $\pm$ 23.4 \\
 & cot & 45.2 $\pm$ 25.1 & 47.0 $\pm$ 25.2 & \textbf{57.2 $\pm$ 30.4} & 45.4 $\pm$ 6.5 & 50.4 $\pm$ 5.9 & 32.0 $\pm$ 1.2 & 37.4 $\pm$ 4.9 \\
 & spp & 46.2 $\pm$ 25.6 & 36.0 $\pm$ 1.9 & 67.0 $\pm$ 22.8 & 41.6 $\pm$ 4.0 & 51.2 $\pm$ 5.4 & 44.6 $\pm$ 26.0 & \textbf{88.6 $\pm$ 5.7} \\
 & sc-zs & \textbf{96.0 $\pm$ 0.0} & \textbf{96.0 $\pm$ 0.0} & 62.0 $\pm$ 38.2 & 41.0 $\pm$ 12.7 & 54.5 $\pm$ 4.9 & 92.5 $\pm$ 4.9 & 65.5 $\pm$ 43.1 \\
 & sc-cot & 37.0 $\pm$ 4.2 & \textbf{59.5 $\pm$ 36.1} & 54.5 $\pm$ 27.6 & 42.0 $\pm$ 2.8 & 51.0 $\pm$ 1.4 & 33.0 $\pm$ 0.0 & 36.5 $\pm$ 2.1 \\
 & sc-spp & \textbf{66.0 $\pm$ 42.4} & 35.0 $\pm$ 0.0 & 63.0 $\pm$ 41.0 & 43.5 $\pm$ 14.8 & 47.5 $\pm$ 2.1 & 61.0 $\pm$ 39.6 & 62.5 $\pm$ 37.5 \\
\cline{1-9}
\multirow[t]{6}{*}{C4S(T)} & zs & 68.0 $\pm$ 32.6 & \textbf{77.6 $\pm$ 25.6} & 68.2 $\pm$ 30.9 & 38.6 $\pm$ 5.2 & 48.8 $\pm$ 3.6 & 56.6 $\pm$ 32.9 & 47.6 $\pm$ 27.1 \\
 & cot & 71.2 $\pm$ 33.1 & \textbf{77.0 $\pm$ 25.6} & 42.8 $\pm$ 23.6 & 48.6 $\pm$ 10.1 & 48.0 $\pm$ 1.2 & 49.2 $\pm$ 24.1 & 50.4 $\pm$ 20.8 \\
 & spp & 83.8 $\pm$ 27.3 & \textbf{88.0 $\pm$ 6.4} & 68.8 $\pm$ 35.5 & 46.4 $\pm$ 4.7 & 50.2 $\pm$ 5.0 & 81.8 $\pm$ 28.0 & 49.0 $\pm$ 22.2 \\
 & sc-zs & 62.5 $\pm$ 34.6 & 57.0 $\pm$ 29.7 & 92.5 $\pm$ 4.9 & 42.0 $\pm$ 1.4 & \textbf{58.0 $\pm$ 4.2} & \textbf{96.0 $\pm$ 0.0} & 35.0 $\pm$ 0.0 \\
 & sc-cot & 32.5 $\pm$ 0.7 & \textbf{94.0 $\pm$ 2.8} & 66.0 $\pm$ 42.4 & 43.5 $\pm$ 2.1 & 48.5 $\pm$ 0.7 & 61.0 $\pm$ 39.6 & 65.5 $\pm$ 43.1 \\
 & sc-spp & \textbf{89.0 $\pm$ 0.0} & 84.5 $\pm$ 10.6 & 63.0 $\pm$ 41.0 & 42.0 $\pm$ 1.4 & 49.5 $\pm$ 12.0 & 32.5 $\pm$ 0.7 & 36.0 $\pm$ 2.8 \\
\cline{1-9}
\multirow[t]{6}{*}{DS-R1} & zs & 43.8 $\pm$ 22.0 & \textbf{51.0 $\pm$ 27.8} & 32.0 $\pm$ 1.9 & 45.4 $\pm$ 5.0 & 47.8 $\pm$ 1.3 & 45.8 $\pm$ 28.1 & 34.2 $\pm$ 1.1 \\
 & cot & 44.8 $\pm$ 28.7 & 33.4 $\pm$ 0.5 & 33.8 $\pm$ 0.8 & 44.0 $\pm$ 7.6 & \textbf{49.0 $\pm$ 2.1} & 33.4 $\pm$ 1.8 & 34.4 $\pm$ 0.9 \\
 & spp & 44.6 $\pm$ 22.0 & 33.4 $\pm$ 1.3 & \textbf{58.0 $\pm$ 32.9} & 38.6 $\pm$ 1.9 & 48.0 $\pm$ 1.6 & 43.4 $\pm$ 25.5 & 33.4 $\pm$ 1.1 \\
 & sc-zs & 33.5 $\pm$ 0.7 & \textbf{58.0 $\pm$ 36.8} & 33.0 $\pm$ 2.8 & 44.5 $\pm$ 2.1 & 48.0 $\pm$ 1.4 & 31.0 $\pm$ 0.0 & 33.0 $\pm$ 1.4 \\
 & sc-cot & 34.0 $\pm$ 0.0 & 35.0 $\pm$ 2.8 & 34.5 $\pm$ 3.5 & 40.5 $\pm$ 0.7 & \textbf{48.0 $\pm$ 0.0} & 34.5 $\pm$ 2.1 & 33.5 $\pm$ 0.7 \\
 & sc-spp & 32.5 $\pm$ 0.7 & 34.0 $\pm$ 1.4 & 32.5 $\pm$ 0.7 & 41.0 $\pm$ 1.4 & \textbf{48.0 $\pm$ 0.0} & 32.5 $\pm$ 2.1 & 33.5 $\pm$ 0.7 \\
\cline{1-9}
\multirow[t]{6}{*}{L3.3-70B} & zs & \textbf{96.0 $\pm$ 0.0} & \textbf{96.0 $\pm$ 0.0} & \textbf{96.0 $\pm$ 0.0} & 47.0 $\pm$ 11.5 & 51.4 $\pm$ 1.3 & 82.0 $\pm$ 27.6 & 49.8 $\pm$ 25.9 \\
 & cot & 86.8 $\pm$ 20.6 & \textbf{96.0 $\pm$ 0.0} & \textbf{96.0 $\pm$ 0.0} & 45.0 $\pm$ 5.7 & 48.8 $\pm$ 2.0 & 94.6 $\pm$ 3.1 & 83.8 $\pm$ 27.3 \\
 & spp & \textbf{96.0 $\pm$ 0.0} & \textbf{96.0 $\pm$ 0.0} & \textbf{96.0 $\pm$ 0.0} & 47.2 $\pm$ 4.1 & 50.2 $\pm$ 1.6 & 68.6 $\pm$ 31.3 & 71.6 $\pm$ 33.4 \\
 & sc-zs & \textbf{96.0 $\pm$ 0.0} & \textbf{96.0 $\pm$ 0.0} & \textbf{96.0 $\pm$ 0.0} & 43.5 $\pm$ 3.5 & 52.0 $\pm$ 0.0 & \textbf{96.0 $\pm$ 0.0} & \textbf{96.0 $\pm$ 0.0} \\
 & sc-cot & \textbf{96.0 $\pm$ 0.0} & \textbf{96.0 $\pm$ 0.0} & \textbf{96.0 $\pm$ 0.0} & 41.5 $\pm$ 10.6 & 47.0 $\pm$ 1.4 & 65.0 $\pm$ 43.8 & 65.0 $\pm$ 43.8 \\
 & sc-spp & \textbf{96.0 $\pm$ 0.0} & \textbf{96.0 $\pm$ 0.0} & \textbf{96.0 $\pm$ 0.0} & 43.5 $\pm$ 6.4 & 50.5 $\pm$ 2.1 & \textbf{96.0 $\pm$ 0.0} & \textbf{96.0 $\pm$ 0.0} \\
\cline{1-9}
\multirow[t]{6}{*}{Mistral} & zs & 66.2 $\pm$ 20.0 & 83.2 $\pm$ 20.1 & 52.2 $\pm$ 26.1 & 36.4 $\pm$ 9.7 & 44.2 $\pm$ 7.8 & \textbf{88.8 $\pm$ 5.5} & 74.6 $\pm$ 29.5 \\
 & cot & 72.6 $\pm$ 24.4 & \textbf{77.2 $\pm$ 17.0} & 62.6 $\pm$ 26.7 & 43.4 $\pm$ 5.7 & 50.4 $\pm$ 8.0 & 43.4 $\pm$ 25.8 & 49.2 $\pm$ 26.2 \\
 & spp & 70.4 $\pm$ 25.9 & 68.4 $\pm$ 25.3 & 57.6 $\pm$ 27.6 & 38.8 $\pm$ 8.3 & 45.0 $\pm$ 4.5 & \textbf{78.8 $\pm$ 27.3} & 59.0 $\pm$ 27.1 \\
 & sc-zs & \textbf{96.0 $\pm$ 0.0} & 74.5 $\pm$ 30.4 & 76.5 $\pm$ 27.6 & 41.5 $\pm$ 2.1 & 40.0 $\pm$ 0.0 & 92.5 $\pm$ 4.9 & 39.0 $\pm$ 1.4 \\
 & sc-cot & \textbf{76.0 $\pm$ 28.3} & 63.0 $\pm$ 22.6 & 42.0 $\pm$ 14.1 & 39.0 $\pm$ 0.0 & 48.0 $\pm$ 7.1 & 61.5 $\pm$ 48.8 & 37.0 $\pm$ 2.8 \\
 & sc-spp & 48.0 $\pm$ 0.0 & 68.5 $\pm$ 31.8 & 46.5 $\pm$ 12.0 & 44.5 $\pm$ 7.8 & 40.0 $\pm$ 0.0 & \textbf{96.0 $\pm$ 0.0} & 64.5 $\pm$ 37.5 \\
\bottomrule
\end{tabular}
\caption{Total Points Averaged Over All Iterations for Stag Hunt (payoff-based PD counterfactual)}
\label{tab:pd_total_points_avg_heatmap_sh}
\end{table*}

In Table \ref{tab:pd_total_points_avg_heatmap_pd-alt} we present results for the label-based PD counterfactual in which we substitute labels as 
$C \rightarrow S (\textit{Stag})$ and $D \rightarrow H (\textit{Hare})$, without modifying the payoffs. In this setting, stronger models such as Claude 3.5/3.7 and Llama 3.3 remain largely invariant, consistently achieving the maximum cumulative reward (96 points) across prompting strategies and LLM--LLM interactions. This indicates robust abstraction over action labels and reliance on underlying incentives rather than memorized action semantics. In contrast, Claude 4 variants and DeepSeek R1 exhibit substantial degradation, with lower total points and significantly higher variance, particularly under reasoning-heavy prompting, suggesting increased sensitivity to label changes and unstable adaptation. Mistral Large remains the most inconsistent model, showing large variability across prompts and occasional recovery only under self-consistency. Across algorithmic opponents, performance remains relatively high—especially against MF—indicating that exploitation of simple patterns is preserved even under relabeling, while interactions with TFT reveal substantial instability. Overall, the label-based counterfactual highlights that while some models generalize well across superficial changes, others exhibit strong dependence on action naming, leading to degraded and inconsistent strategic behavior.

\begin{table*}[h!]
\small
\centering
\begin{tabular}{lllllllll}
\toprule
 &  & \multicolumn{7}{c}{\textbf{Joint counterfactual of PD}} \\
 &  & zs & spp & cot & srep & pp & mf & tft \\
model & prompt &  &  &  &  &  &  &  \\
\midrule
\multirow[t]{6}{*}{C3.5Sv2} & zs & \textbf{64.0 $\pm$ 0.0} & \textbf{64.0 $\pm$ 0.0} & \textbf{64.0 $\pm$ 0.0} & 30.2 $\pm$ 0.4 & 58.8 $\pm$ 5.0 & 31.2 $\pm$ 1.1 & 31.0 $\pm$ 1.0 \\
 & cot & \textbf{64.0 $\pm$ 0.0} & \textbf{64.0 $\pm$ 0.0} & \textbf{64.0 $\pm$ 0.0} & 30.8 $\pm$ 0.4 & 58.4 $\pm$ 1.8 & 31.4 $\pm$ 0.5 & 31.6 $\pm$ 0.9 \\
 & spp & \textbf{64.0 $\pm$ 0.0} & \textbf{64.0 $\pm$ 0.0} & \textbf{64.0 $\pm$ 0.0} & 30.4 $\pm$ 0.5 & 61.6 $\pm$ 0.9 & 31.6 $\pm$ 0.9 & 31.6 $\pm$ 1.8 \\
 & sc-zs & \textbf{64.0 $\pm$ 0.0} & \textbf{64.0 $\pm$ 0.0} & \textbf{64.0 $\pm$ 0.0} & 30.5 $\pm$ 0.7 & 59.5 $\pm$ 0.7 & 31.0 $\pm$ 1.4 & 32.0 $\pm$ 0.0 \\
 & sc-cot & \textbf{64.0 $\pm$ 0.0} & \textbf{64.0 $\pm$ 0.0} & \textbf{64.0 $\pm$ 0.0} & 30.5 $\pm$ 0.7 & 58.5 $\pm$ 2.1 & 31.5 $\pm$ 0.7 & 32.0 $\pm$ 0.0 \\
 & sc-spp & \textbf{64.0 $\pm$ 0.0} & \textbf{64.0 $\pm$ 0.0} & \textbf{64.0 $\pm$ 0.0} & 30.5 $\pm$ 0.7 & 59.5 $\pm$ 3.5 & 31.0 $\pm$ 1.4 & 31.5 $\pm$ 0.7 \\
\cline{1-9}
\multirow[t]{6}{*}{C3.7S} & zs & \textbf{64.0 $\pm$ 0.0} & \textbf{64.0 $\pm$ 0.0} & \textbf{64.0 $\pm$ 0.0} & 30.0 $\pm$ 0.0 & 55.6 $\pm$ 4.1 & 31.2 $\pm$ 0.8 & 30.8 $\pm$ 0.8 \\
 & cot & \textbf{64.0 $\pm$ 0.0} & \textbf{64.0 $\pm$ 0.0} & \textbf{64.0 $\pm$ 0.0} & 30.6 $\pm$ 0.5 & 55.4 $\pm$ 4.2 & 31.0 $\pm$ 1.0 & 31.2 $\pm$ 0.4 \\
 & spp & \textbf{64.0 $\pm$ 0.0} & 58.4 $\pm$ 10.3 & 63.6 $\pm$ 0.9 & 30.2 $\pm$ 0.4 & 60.4 $\pm$ 2.6 & 32.8 $\pm$ 1.6 & 31.4 $\pm$ 1.5 \\
 & sc-zs & \textbf{64.0 $\pm$ 0.0} & \textbf{64.0 $\pm$ 0.0} & 62.5 $\pm$ 2.1 & 30.0 $\pm$ 0.0 & 57.5 $\pm$ 6.4 & 31.0 $\pm$ 1.4 & 31.5 $\pm$ 0.7 \\
 & sc-cot & \textbf{64.0 $\pm$ 0.0} & 52.5 $\pm$ 16.3 & \textbf{64.0 $\pm$ 0.0} & 30.0 $\pm$ 0.0 & 55.5 $\pm$ 9.2 & 30.5 $\pm$ 0.7 & 32.0 $\pm$ 0.0 \\
 & sc-spp & \textbf{64.0 $\pm$ 0.0} & \textbf{64.0 $\pm$ 0.0} & 50.5 $\pm$ 19.1 & 30.5 $\pm$ 0.7 & 60.5 $\pm$ 2.1 & 30.5 $\pm$ 0.7 & 32.0 $\pm$ 0.0 \\
\cline{1-9}
\multirow[t]{6}{*}{C3.7S(T)} & zs & \textbf{64.0 $\pm$ 0.0} & 59.6 $\pm$ 9.8 & \textbf{64.0 $\pm$ 0.0} & 30.4 $\pm$ 0.5 & 53.6 $\pm$ 2.2 & 30.8 $\pm$ 1.1 & 31.0 $\pm$ 1.4 \\
 & cot & 55.0 $\pm$ 12.3 & \textbf{64.0 $\pm$ 0.0} & \textbf{64.0 $\pm$ 0.0} & 30.8 $\pm$ 0.4 & 58.6 $\pm$ 4.3 & 31.8 $\pm$ 0.4 & 31.2 $\pm$ 0.8 \\
 & spp & \textbf{63.4 $\pm$ 1.3} & 57.8 $\pm$ 13.9 & \textbf{63.4 $\pm$ 1.3} & 30.2 $\pm$ 0.4 & 59.0 $\pm$ 3.5 & 31.0 $\pm$ 1.2 & 31.6 $\pm$ 0.9 \\
 & sc-zs & \textbf{64.0 $\pm$ 0.0} & \textbf{64.0 $\pm$ 0.0} & \textbf{64.0 $\pm$ 0.0} & 30.0 $\pm$ 0.0 & 57.0 $\pm$ 7.1 & 32.0 $\pm$ 0.0 & 30.5 $\pm$ 0.7 \\
 & sc-cot & \textbf{64.0 $\pm$ 0.0} & \textbf{64.0 $\pm$ 0.0} & \textbf{64.0 $\pm$ 0.0} & 30.5 $\pm$ 0.7 & 61.5 $\pm$ 2.1 & 31.0 $\pm$ 0.0 & 31.5 $\pm$ 0.7 \\
 & sc-spp & 63.0 $\pm$ 1.4 & \textbf{64.0 $\pm$ 0.0} & \textbf{64.0 $\pm$ 0.0} & 30.5 $\pm$ 0.7 & 56.0 $\pm$ 5.7 & 32.0 $\pm$ 0.0 & 32.0 $\pm$ 0.0 \\
\cline{1-9}
\multirow[t]{6}{*}{C4S} & zs & 40.8 $\pm$ 13.7 & 38.6 $\pm$ 2.2 & 44.0 $\pm$ 13.1 & 31.2 $\pm$ 0.8 & \textbf{58.0 $\pm$ 4.5} & 32.2 $\pm$ 2.3 & 32.4 $\pm$ 2.3 \\
 & cot & 35.8 $\pm$ 4.3 & 40.0 $\pm$ 12.5 & 38.0 $\pm$ 5.3 & 30.6 $\pm$ 0.5 & \textbf{60.4 $\pm$ 2.3} & 33.6 $\pm$ 1.7 & 32.6 $\pm$ 2.1 \\
 & spp & 34.2 $\pm$ 3.5 & 36.4 $\pm$ 13.8 & 38.6 $\pm$ 9.4 & 31.4 $\pm$ 0.5 & \textbf{57.4 $\pm$ 4.5} & 32.4 $\pm$ 1.8 & 34.2 $\pm$ 2.2 \\
 & sc-zs & 37.0 $\pm$ 1.4 & 35.0 $\pm$ 5.7 & 31.5 $\pm$ 0.7 & 31.5 $\pm$ 0.7 & \textbf{64.0 $\pm$ 0.0} & 33.5 $\pm$ 3.5 & 35.0 $\pm$ 1.4 \\
 & sc-cot & 31.5 $\pm$ 0.7 & 32.0 $\pm$ 0.0 & 34.5 $\pm$ 4.9 & 32.0 $\pm$ 0.0 & \textbf{64.0 $\pm$ 0.0} & 32.0 $\pm$ 0.0 & 33.0 $\pm$ 1.4 \\
 & sc-spp & 34.5 $\pm$ 3.5 & 33.5 $\pm$ 3.5 & 34.5 $\pm$ 4.9 & 32.0 $\pm$ 0.0 & \textbf{60.5 $\pm$ 2.1} & 32.0 $\pm$ 1.4 & 32.5 $\pm$ 0.7 \\
\cline{1-9}
\multirow[t]{6}{*}{C4S(T)} & zs & 33.0 $\pm$ 2.0 & 42.0 $\pm$ 11.5 & 32.0 $\pm$ 1.9 & 31.4 $\pm$ 0.5 & \textbf{59.4 $\pm$ 5.9} & 32.0 $\pm$ 1.9 & 34.0 $\pm$ 2.3 \\
 & cot & 33.2 $\pm$ 3.9 & 34.0 $\pm$ 2.7 & 31.8 $\pm$ 0.4 & 32.0 $\pm$ 0.0 & \textbf{62.4 $\pm$ 3.6} & 34.2 $\pm$ 2.5 & 33.6 $\pm$ 2.2 \\
 & spp & 34.2 $\pm$ 5.4 & 39.8 $\pm$ 12.7 & 31.4 $\pm$ 0.5 & 31.2 $\pm$ 0.4 & \textbf{60.2 $\pm$ 3.1} & 32.4 $\pm$ 2.1 & 31.4 $\pm$ 0.5 \\
 & sc-zs & 36.5 $\pm$ 3.5 & 40.0 $\pm$ 0.0 & 47.5 $\pm$ 23.3 & 31.0 $\pm$ 0.0 & \textbf{64.0 $\pm$ 0.0} & 33.5 $\pm$ 3.5 & 36.0 $\pm$ 0.0 \\
 & sc-cot & 40.0 $\pm$ 5.7 & 35.5 $\pm$ 6.4 & 32.0 $\pm$ 0.0 & 32.0 $\pm$ 0.0 & \textbf{64.0 $\pm$ 0.0} & 32.0 $\pm$ 0.0 & 36.0 $\pm$ 0.0 \\
 & sc-spp & 32.0 $\pm$ 0.0 & 38.0 $\pm$ 2.8 & 41.5 $\pm$ 13.4 & 32.0 $\pm$ 0.0 & \textbf{58.0 $\pm$ 8.5} & 35.5 $\pm$ 0.7 & 33.0 $\pm$ 1.4 \\
\cline{1-9}
\multirow[t]{6}{*}{DS-R1} & zs & 32.8 $\pm$ 2.5 & 34.8 $\pm$ 0.8 & 32.0 $\pm$ 1.2 & 31.0 $\pm$ 0.7 & \textbf{62.2 $\pm$ 2.0} & 31.8 $\pm$ 1.8 & 33.6 $\pm$ 2.4 \\
 & cot & 34.4 $\pm$ 1.9 & 34.0 $\pm$ 3.7 & 32.0 $\pm$ 1.7 & 31.4 $\pm$ 0.5 & \textbf{62.2 $\pm$ 2.2} & 33.6 $\pm$ 2.4 & 34.6 $\pm$ 2.1 \\
 & spp & 36.2 $\pm$ 1.8 & 34.8 $\pm$ 2.6 & 33.6 $\pm$ 3.2 & 31.0 $\pm$ 0.0 & \textbf{63.6 $\pm$ 0.9} & 31.4 $\pm$ 0.5 & 32.6 $\pm$ 2.7 \\
 & sc-zs & 33.5 $\pm$ 2.1 & 33.5 $\pm$ 3.5 & 31.5 $\pm$ 0.7 & 31.5 $\pm$ 0.7 & \textbf{64.0 $\pm$ 0.0} & 35.5 $\pm$ 0.7 & 34.0 $\pm$ 2.8 \\
 & sc-cot & 35.5 $\pm$ 0.7 & 33.5 $\pm$ 3.5 & 35.5 $\pm$ 0.7 & 31.5 $\pm$ 0.7 & \textbf{64.0 $\pm$ 0.0} & 36.0 $\pm$ 0.0 & 33.5 $\pm$ 3.5 \\
 & sc-spp & 34.0 $\pm$ 2.8 & 38.0 $\pm$ 2.8 & 36.0 $\pm$ 0.0 & 32.0 $\pm$ 0.0 & \textbf{64.0 $\pm$ 0.0} & 35.5 $\pm$ 0.7 & 35.5 $\pm$ 0.7 \\
\cline{1-9}
\multirow[t]{6}{*}{L3.3-70B} & zs & \textbf{64.0 $\pm$ 0.0} & \textbf{64.0 $\pm$ 0.0} & \textbf{64.0 $\pm$ 0.0} & 28.0 $\pm$ 1.2 & 54.8 $\pm$ 3.6 & 30.0 $\pm$ 1.4 & 30.8 $\pm$ 0.8 \\
 & cot & \textbf{64.0 $\pm$ 0.0} & \textbf{64.0 $\pm$ 0.0} & \textbf{64.0 $\pm$ 0.0} & 29.2 $\pm$ 1.1 & 53.4 $\pm$ 3.1 & 30.8 $\pm$ 1.1 & 31.2 $\pm$ 1.1 \\
 & spp & \textbf{64.0 $\pm$ 0.0} & \textbf{64.0 $\pm$ 0.0} & \textbf{64.0 $\pm$ 0.0} & 29.4 $\pm$ 0.5 & 54.0 $\pm$ 2.8 & 31.0 $\pm$ 1.0 & 31.0 $\pm$ 1.0 \\
 & sc-zs & \textbf{64.0 $\pm$ 0.0} & \textbf{64.0 $\pm$ 0.0} & \textbf{64.0 $\pm$ 0.0} & 28.5 $\pm$ 0.7 & 52.5 $\pm$ 0.7 & 31.0 $\pm$ 1.4 & 29.5 $\pm$ 3.5 \\
 & sc-cot & \textbf{64.0 $\pm$ 0.0} & \textbf{64.0 $\pm$ 0.0} & \textbf{64.0 $\pm$ 0.0} & 29.0 $\pm$ 0.0 & 56.0 $\pm$ 5.7 & 30.0 $\pm$ 0.0 & 31.0 $\pm$ 1.4 \\
 & sc-spp & \textbf{64.0 $\pm$ 0.0} & \textbf{64.0 $\pm$ 0.0} & \textbf{64.0 $\pm$ 0.0} & 29.0 $\pm$ 1.4 & 56.0 $\pm$ 4.2 & 31.0 $\pm$ 1.4 & 29.5 $\pm$ 0.7 \\
\cline{1-9}
\multirow[t]{6}{*}{Mistral} & zs & 52.6 $\pm$ 9.8 & 50.0 $\pm$ 9.9 & 39.2 $\pm$ 9.4 & 23.8 $\pm$ 0.4 & \textbf{53.8 $\pm$ 3.5} & 25.6 $\pm$ 1.7 & 26.0 $\pm$ 1.4 \\
 & cot & 58.8 $\pm$ 3.7 & \textbf{59.6 $\pm$ 4.4} & 55.0 $\pm$ 13.1 & 26.8 $\pm$ 2.5 & 56.6 $\pm$ 6.4 & 28.8 $\pm$ 2.0 & 28.2 $\pm$ 3.3 \\
 & spp & \textbf{57.2 $\pm$ 4.9} & 56.8 $\pm$ 7.3 & 51.8 $\pm$ 11.7 & 24.4 $\pm$ 4.1 & 54.0 $\pm$ 1.2 & 27.2 $\pm$ 1.6 & 24.8 $\pm$ 2.2 \\
 & sc-zs & 54.0 $\pm$ 1.4 & \textbf{58.0 $\pm$ 5.7} & 54.0 $\pm$ 14.1 & 24.0 $\pm$ 0.0 & 56.0 $\pm$ 0.0 & 25.0 $\pm$ 1.4 & 26.0 $\pm$ 2.8 \\
 & sc-cot & 58.5 $\pm$ 3.5 & \textbf{63.5 $\pm$ 9.2} & 43.5 $\pm$ 3.5 & 28.0 $\pm$ 5.7 & 56.5 $\pm$ 0.7 & 31.5 $\pm$ 2.1 & 30.0 $\pm$ 0.0 \\
 & sc-spp & 59.0 $\pm$ 0.0 & \textbf{61.5 $\pm$ 3.5} & 32.5 $\pm$ 12.0 & 26.0 $\pm$ 4.2 & 51.0 $\pm$ 5.7 & 24.0 $\pm$ 0.0 & 27.0 $\pm$ 1.4 \\
\bottomrule
\end{tabular}
\caption{Total Points Averaged Over All Iterations for the joint PD counterfactual (label-based and payoff-based).}
\label{tab:pd_total_points_avg_heatmap_sh-alt}
\end{table*}
In Table \ref{tab:pd_total_points_avg_heatmap_sh} we demonstrate total point results for Stag Hunt, as a payoff-based counterfactual of PD. The clearest result is that the strongest models—Claude 3.5, Claude 3.7, Claude 3.7 thinking, and Llama 3.3—very often reach the maximum 96 points in LLM–LLM interactions, indicating that they successfully shift from the default PD defection logic to the payoff-dominant cooperative equilibrium of Stag Hunt. This is important because, unlike the label-based counterfactual, the payoff structure here actually changes; consistently reaching 96 therefore suggests genuine sensitivity to altered incentives rather than mere invariance to relabeling.

A second important pattern is that Claude 4 and DeepSeek R1 remain much less stable under the Stag Hunt payoff change. Claude 4 improves relative to its weaker PD cooperation behavior in some settings, but still exhibits large variance and frequent drops well below the optimum, especially under CoT- or SC-based prompting. DeepSeek R1 performs even worse overall, mostly staying in the 30–50 range and rarely approaching the coordination optimum. This supports the claim that some models fail not only under label changes but also under genuine incentive shifts, often defaulting to cautious or inconsistent play instead of converging to the payoff-dominant equilibrium.

Third, several algorithmic-opponent columns, especially MF and sometimes TFT, produce very high scores, often in the 90s for the stronger models. This suggests that once the altered incentives are understood, these opponents become highly exploitable. By contrast, the SREP and PP columns remain much lower, typically around the 40s–50s, indicating that not all opponents benefit equally from the SH payoff shift and that coordination/exploitation depends strongly on the opponent’s dynamics.

Finally, Mistral remains the most unstable model, but the table also shows occasional recovery under self-consistency, including some 96-point outcomes. So while Mistral is still brittle, the Stag Hunt results suggest that SC can sometimes help it coordinate under changed incentives, even if this behavior is not reliable. Overall, the table provides strong evidence that payoff-based counterfactuals are more revealing than label-only ones: the best models adapt to the new equilibrium structure, whereas weaker or more unstable models show large variance and frequent failure to coordinate on the optimal SH outcome.

Finally, in Table \ref{tab:pd_total_points_avg_heatmap_sh-alt} results regarding the joint PD counterfactual game are presented. Once both labels and incentives are altered together, robustness becomes much more model-dependent than in either single intervention. The strongest evidence comes from Claude 3.5, Claude 3.7, Claude 3.7 thinking, and Llama 3.3, which still preserve near-ceiling performance in LLM--LLM interactions, usually reaching $64.0 \pm 0.0$, indicating that these models can often maintain coordination even when both the semantics and the strategic structure are perturbed. However, this robustness is weaker than in the isolated label-only or payoff-only cases, as some prompting configurations now show mild degradation or increased variance. By contrast, Claude 4 variants and DeepSeek R1 remain substantially below the ceiling in most LLM--LLM settings, typically clustering in the 30s and 40s, which suggests that combining label and payoff changes amplifies their brittleness. Mistral is again the least stable overall: it improves in some prompting configurations, but remains far from ceiling performance and highly variable. Another useful observation is that the \textbf{PP} column remains relatively high across most models, meaning that cyclic algorithmic structure is still exploitable even in the joint counterfactual, whereas the other algorithmic-opponent columns stay much closer to default-PD ranges. Overall, the table supports a strong conclusion: when both surface form and incentives change simultaneously, only the most robust models preserve strong strategic behavior, while weaker or less stable models fail to recompute the new game reliably.

\subsection{PD Opponent comprehension}
\begin{table*}[h!]
\small
\centering
\begin{tabular}{lllllllll}
\toprule
 &  & \multicolumn{7}{c}{\textbf{PD}} \\
 &  & zs & spp & cot & srep & pp & mf & tft \\
model & prompt &  &  &  &  &  &  &  \\
\midrule
\multirow[t]{6}{*}{C3.5Sv2} & zs & \textbf{1.0 $\pm$ 0.0} & \textbf{1.0 $\pm$ 0.0} & \textbf{1.0 $\pm$ 0.0} & 2.2 $\pm$ 0.4 & 8.4 $\pm$ 7.0 & 5.6 $\pm$ 6.4 & 2.8 $\pm$ 1.3 \\
 & cot & \textbf{1.0 $\pm$ 0.0} & \textbf{1.0 $\pm$ 0.0} & \textbf{1.0 $\pm$ 0.0} & 2.0 $\pm$ 0.0 & 14.0 $\pm$ 2.0 & 3.0 $\pm$ 1.2 & 2.2 $\pm$ 0.4 \\
 & spp & \textbf{1.0 $\pm$ 0.0} & \textbf{1.0 $\pm$ 0.0} & \textbf{1.0 $\pm$ 0.0} & 1.6 $\pm$ 0.5 & 7.6 $\pm$ 6.1 & 2.0 $\pm$ 0.7 & 3.0 $\pm$ 0.0 \\
 & sc-zs & \textbf{1.0 $\pm$ 0.0} & \textbf{1.0 $\pm$ 0.0} & \textbf{1.0 $\pm$ 0.0} & 1.5 $\pm$ 0.7 & 12.0 $\pm$ 0.0 & 3.0 $\pm$ 0.0 & 4.5 $\pm$ 0.7 \\
 & sc-cot & \textbf{1.0 $\pm$ 0.0} & \textbf{1.0 $\pm$ 0.0} & \textbf{1.0 $\pm$ 0.0} & 2.0 $\pm$ 0.0 & 9.0 $\pm$ 4.2 & 3.0 $\pm$ 0.0 & 2.0 $\pm$ 0.0 \\
 & sc-spp & \textbf{1.0 $\pm$ 0.0} & \textbf{1.0 $\pm$ 0.0} & \textbf{1.0 $\pm$ 0.0} & 2.0 $\pm$ 0.0 & 2.0 $\pm$ 0.0 & 2.5 $\pm$ 0.7 & 2.5 $\pm$ 0.7 \\
\cline{1-9}
\multirow[t]{6}{*}{C3.7S} & zs & \textbf{1.0 $\pm$ 0.0} & \textbf{1.0 $\pm$ 0.0} & \textbf{1.0 $\pm$ 0.0} & 2.0 $\pm$ 0.0 & 11.6 $\pm$ 5.5 & 3.6 $\pm$ 1.5 & 3.2 $\pm$ 1.6 \\
 & cot & \textbf{1.0 $\pm$ 0.0} & \textbf{1.0 $\pm$ 0.0} & \textbf{1.0 $\pm$ 0.0} & 2.0 $\pm$ 0.0 & 2.0 $\pm$ 0.0 & 3.2 $\pm$ 1.8 & 3.2 $\pm$ 2.3 \\
 & spp & \textbf{1.0 $\pm$ 0.0} & \textbf{1.0 $\pm$ 0.0} & \textbf{1.0 $\pm$ 0.0} & 2.0 $\pm$ 0.0 & 9.6 $\pm$ 6.2 & 3.0 $\pm$ 1.2 & 3.6 $\pm$ 2.3 \\
 & sc-zs & \textbf{1.0 $\pm$ 0.0} & \textbf{1.0 $\pm$ 0.0} & \textbf{1.0 $\pm$ 0.0} & 2.0 $\pm$ 0.0 & 13.0 $\pm$ 1.4 & 2.5 $\pm$ 0.7 & 2.0 $\pm$ 0.0 \\
 & sc-cot & \textbf{1.0 $\pm$ 0.0} & \textbf{1.0 $\pm$ 0.0} & \textbf{1.0 $\pm$ 0.0} & 1.5 $\pm$ 0.7 & 2.0 $\pm$ 0.0 & 3.5 $\pm$ 2.1 & 5.0 $\pm$ 0.0 \\
 & sc-spp & \textbf{1.0 $\pm$ 0.0} & \textbf{1.0 $\pm$ 0.0} & \textbf{1.0 $\pm$ 0.0} & 2.0 $\pm$ 0.0 & 8.0 $\pm$ 8.5 & 2.5 $\pm$ 0.7 & 4.0 $\pm$ 1.4 \\
\cline{1-9}
\multirow[t]{6}{*}{C3.7S(T)} & zs & \textbf{1.0 $\pm$ 0.0} & \textbf{1.0 $\pm$ 0.0} & \textbf{1.0 $\pm$ 0.0} & 2.6 $\pm$ 1.3 & 12.8 $\pm$ 4.1 & 4.0 $\pm$ 1.4 & 2.8 $\pm$ 0.4 \\
 & cot & \textbf{1.0 $\pm$ 0.0} & \textbf{1.0 $\pm$ 0.0} & \textbf{1.0 $\pm$ 0.0} & 1.4 $\pm$ 0.5 & 10.4 $\pm$ 6.1 & 2.6 $\pm$ 1.5 & 1.8 $\pm$ 0.4 \\
 & spp & \textbf{1.0 $\pm$ 0.0} & \textbf{1.0 $\pm$ 0.0} & \textbf{1.0 $\pm$ 0.0} & 1.8 $\pm$ 0.4 & 6.8 $\pm$ 6.7 & 3.0 $\pm$ 1.2 & 3.0 $\pm$ 1.2 \\
 & sc-zs & \textbf{1.0 $\pm$ 0.0} & \textbf{1.0 $\pm$ 0.0} & \textbf{1.0 $\pm$ 0.0} & 2.0 $\pm$ 0.0 & 13.0 $\pm$ 1.4 & 3.0 $\pm$ 0.0 & 2.0 $\pm$ 0.0 \\
 & sc-cot & \textbf{1.0 $\pm$ 0.0} & \textbf{1.0 $\pm$ 0.0} & \textbf{1.0 $\pm$ 0.0} & 1.5 $\pm$ 0.7 & 15.0 $\pm$ 1.4 & 2.5 $\pm$ 0.7 & 1.5 $\pm$ 0.7 \\
 & sc-spp & \textbf{1.0 $\pm$ 0.0} & \textbf{1.0 $\pm$ 0.0} & \textbf{1.0 $\pm$ 0.0} & 1.5 $\pm$ 0.7 & 9.0 $\pm$ 7.1 & 3.0 $\pm$ 0.0 & 2.5 $\pm$ 0.7 \\
\cline{1-9}
\multirow[t]{6}{*}{C4S} & zs & 1.2 $\pm$ 0.4 & 1.4 $\pm$ 0.9 & 3.0 $\pm$ 4.5 & \textbf{1.0 $\pm$ 0.0} & 12.0 $\pm$ 5.5 & 1.8 $\pm$ 1.1 & 2.2 $\pm$ 1.6 \\
 & cot & 3.0 $\pm$ 2.8 & 10.8 $\pm$ 4.1 & 3.2 $\pm$ 2.3 & \textbf{1.0 $\pm$ 0.0} & 6.0 $\pm$ 6.1 & 2.2 $\pm$ 1.6 & 1.6 $\pm$ 1.3 \\
 & spp & 5.2 $\pm$ 6.9 & 1.6 $\pm$ 1.3 & 4.6 $\pm$ 2.1 & \textbf{1.0 $\pm$ 0.0} & 1.8 $\pm$ 1.3 & 4.2 $\pm$ 7.2 & 2.2 $\pm$ 1.8 \\
 & sc-zs & 2.5 $\pm$ 2.1 & \textbf{1.0 $\pm$ 0.0} & 7.0 $\pm$ 8.5 & \textbf{1.0 $\pm$ 0.0} & 6.5 $\pm$ 7.8 & \textbf{1.0 $\pm$ 0.0} & 2.0 $\pm$ 1.4 \\
 & sc-cot & 3.5 $\pm$ 3.5 & \textbf{1.0 $\pm$ 0.0} & \textbf{1.0 $\pm$ 0.0} & \textbf{1.0 $\pm$ 0.0} & \textbf{1.0 $\pm$ 0.0} & \textbf{1.0 $\pm$ 0.0} & \textbf{1.0 $\pm$ 0.0} \\
 & sc-spp & 4.0 $\pm$ 4.2 & 3.0 $\pm$ 2.8 & \textbf{1.0 $\pm$ 0.0} & 2.0 $\pm$ 1.4 & \textbf{1.0 $\pm$ 0.0} & \textbf{1.0 $\pm$ 0.0} & \textbf{1.0 $\pm$ 0.0} \\
\cline{1-9}
\multirow[t]{6}{*}{C4S(T)} & zs & \textbf{1.0 $\pm$ 0.0} & 2.0 $\pm$ 2.2 & \textbf{1.0 $\pm$ 0.0} & \textbf{1.0 $\pm$ 0.0} & 1.2 $\pm$ 0.4 & \textbf{1.0 $\pm$ 0.0} & \textbf{1.0 $\pm$ 0.0} \\
 & cot & \textbf{1.0 $\pm$ 0.0} & 4.2 $\pm$ 7.2 & 4.0 $\pm$ 6.7 & \textbf{1.0 $\pm$ 0.0} & \textbf{1.0 $\pm$ 0.0} & 1.6 $\pm$ 1.3 & \textbf{1.0 $\pm$ 0.0} \\
 & spp & 1.4 $\pm$ 0.9 & 2.2 $\pm$ 2.2 & 2.4 $\pm$ 2.2 & \textbf{1.0 $\pm$ 0.0} & 5.2 $\pm$ 6.4 & 1.6 $\pm$ 1.3 & \textbf{1.0 $\pm$ 0.0} \\
 & sc-zs & \textbf{1.0 $\pm$ 0.0} & \textbf{1.0 $\pm$ 0.0} & \textbf{1.0 $\pm$ 0.0} & \textbf{1.0 $\pm$ 0.0} & \textbf{1.0 $\pm$ 0.0} & \textbf{1.0 $\pm$ 0.0} & \textbf{1.0 $\pm$ 0.0} \\
 & sc-cot & \textbf{1.0 $\pm$ 0.0} & 3.0 $\pm$ 2.8 & \textbf{1.0 $\pm$ 0.0} & \textbf{1.0 $\pm$ 0.0} & \textbf{1.0 $\pm$ 0.0} & \textbf{1.0 $\pm$ 0.0} & \textbf{1.0 $\pm$ 0.0} \\
 & sc-spp & \textbf{1.0 $\pm$ 0.0} & \textbf{1.0 $\pm$ 0.0} & \textbf{1.0 $\pm$ 0.0} & \textbf{1.0 $\pm$ 0.0} & 2.5 $\pm$ 2.1 & \textbf{1.0 $\pm$ 0.0} & 2.0 $\pm$ 1.4 \\
\cline{1-9}
\multirow[t]{6}{*}{DS-R1} & zs & 3.2 $\pm$ 2.2 & 2.0 $\pm$ 2.2 & 8.0 $\pm$ 8.3 & \textbf{1.0 $\pm$ 0.0} & 4.0 $\pm$ 6.7 & \textbf{1.0 $\pm$ 0.0} & \textbf{1.0 $\pm$ 0.0} \\
 & cot & \textbf{1.0 $\pm$ 0.0} & \textbf{1.0 $\pm$ 0.0} & \textbf{1.0 $\pm$ 0.0} & \textbf{1.0 $\pm$ 0.0} & \textbf{1.0 $\pm$ 0.0} & 2.2 $\pm$ 2.7 & 10.6 $\pm$ 8.8 \\
 & spp & 5.2 $\pm$ 6.6 & 7.4 $\pm$ 7.6 & 5.2 $\pm$ 6.9 & \textbf{1.0 $\pm$ 0.0} & 4.8 $\pm$ 5.4 & \textbf{1.0 $\pm$ 0.0} & \textbf{1.0 $\pm$ 0.0} \\
 & sc-zs & 6.5 $\pm$ 7.8 & \textbf{1.0 $\pm$ 0.0} & \textbf{1.0 $\pm$ 0.0} & \textbf{1.0 $\pm$ 0.0} & \textbf{1.0 $\pm$ 0.0} & \textbf{1.0 $\pm$ 0.0} & \textbf{1.0 $\pm$ 0.0} \\
 & sc-cot & 3.5 $\pm$ 3.5 & \textbf{1.0 $\pm$ 0.0} & 9.0 $\pm$ 11.3 & \textbf{1.0 $\pm$ 0.0} & \textbf{1.0 $\pm$ 0.0} & \textbf{1.0 $\pm$ 0.0} & \textbf{1.0 $\pm$ 0.0} \\
 & sc-spp & 2.5 $\pm$ 2.1 & \textbf{1.0 $\pm$ 0.0} & \textbf{1.0 $\pm$ 0.0} & \textbf{1.0 $\pm$ 0.0} & 7.5 $\pm$ 9.2 & \textbf{1.0 $\pm$ 0.0} & \textbf{1.0 $\pm$ 0.0} \\
\cline{1-9}
\multirow[t]{6}{*}{L3.3-70B} & zs & \textbf{1.0 $\pm$ 0.0} & \textbf{1.0 $\pm$ 0.0} & \textbf{1.0 $\pm$ 0.0} & 2.8 $\pm$ 1.1 & 14.0 $\pm$ 0.0 & 2.8 $\pm$ 0.4 & 2.8 $\pm$ 0.4 \\
 & cot & \textbf{1.0 $\pm$ 0.0} & \textbf{1.0 $\pm$ 0.0} & \textbf{1.0 $\pm$ 0.0} & 2.4 $\pm$ 0.9 & 10.0 $\pm$ 6.5 & 2.4 $\pm$ 0.5 & 2.2 $\pm$ 0.4 \\
 & spp & \textbf{1.0 $\pm$ 0.0} & \textbf{1.0 $\pm$ 0.0} & \textbf{1.0 $\pm$ 0.0} & 2.4 $\pm$ 0.9 & 12.8 $\pm$ 5.0 & 2.4 $\pm$ 0.5 & 2.2 $\pm$ 0.4 \\
 & sc-zs & \textbf{1.0 $\pm$ 0.0} & \textbf{1.0 $\pm$ 0.0} & \textbf{1.0 $\pm$ 0.0} & 4.0 $\pm$ 0.0 & 14.0 $\pm$ 0.0 & 2.5 $\pm$ 0.7 & 3.0 $\pm$ 0.0 \\
 & sc-cot & \textbf{1.0 $\pm$ 0.0} & \textbf{1.0 $\pm$ 0.0} & \textbf{1.0 $\pm$ 0.0} & 2.0 $\pm$ 0.0 & 14.0 $\pm$ 0.0 & 3.0 $\pm$ 0.0 & 2.5 $\pm$ 0.7 \\
 & sc-spp & \textbf{1.0 $\pm$ 0.0} & \textbf{1.0 $\pm$ 0.0} & \textbf{1.0 $\pm$ 0.0} & 3.0 $\pm$ 1.4 & 9.0 $\pm$ 7.1 & 2.0 $\pm$ 0.0 & 3.0 $\pm$ 0.0 \\
\cline{1-9}
\multirow[t]{6}{*}{Mistral} & zs & 10.2 $\pm$ 8.4 & 6.8 $\pm$ 6.8 & \textbf{3.8 $\pm$ 6.3} & 16.4 $\pm$ 0.5 & 15.2 $\pm$ 1.1 & 16.2 $\pm$ 0.4 & 16.0 $\pm$ 0.7 \\
 & cot & \textbf{1.0 $\pm$ 0.0} & 7.0 $\pm$ 8.2 & 4.2 $\pm$ 7.2 & 4.4 $\pm$ 7.1 & 12.6 $\pm$ 6.5 & 8.6 $\pm$ 7.3 & 16.2 $\pm$ 0.8 \\
 & spp & \textbf{1.0 $\pm$ 0.0} & 3.2 $\pm$ 4.9 & 4.2 $\pm$ 7.2 & 13.4 $\pm$ 6.9 & 15.6 $\pm$ 0.9 & 17.0 $\pm$ 0.0 & 16.4 $\pm$ 0.9 \\
 & sc-zs & \textbf{1.0 $\pm$ 0.0} & \textbf{1.0 $\pm$ 0.0} & \textbf{1.0 $\pm$ 0.0} & 16.0 $\pm$ 0.0 & 15.0 $\pm$ 1.4 & 16.5 $\pm$ 0.7 & 16.5 $\pm$ 0.7 \\
 & sc-cot & 8.5 $\pm$ 10.6 & \textbf{1.0 $\pm$ 0.0} & 8.5 $\pm$ 10.6 & 9.0 $\pm$ 11.3 & 8.5 $\pm$ 10.6 & \textbf{1.0 $\pm$ 0.0} & 10.0 $\pm$ 9.9 \\
 & sc-spp & \textbf{1.0 $\pm$ 0.0} & \textbf{1.0 $\pm$ 0.0} & \textbf{1.0 $\pm$ 0.0} & 16.5 $\pm$ 0.7 & 16.0 $\pm$ 0.0 & 17.0 $\pm$ 0.0 & 16.0 $\pm$ 0.0 \\
\bottomrule
\end{tabular}
\caption{Round $m$\# where the LLM player understood the opponent's Strategy in default PD. Bold for best results (lower number of rounds $m$ at which opponent comprehension was realized).}
\label{tab:pd_round_heatmap_pd}
\end{table*}

Opponent comprehension for the default PD is provided in Table \ref{tab:pd_round_heatmap_pd}, with registered values corresponding to the number of rounds at which the LLM understood the opponent's play.

The main pattern is that PD opponent comprehension is generally very early for most strong models, especially in LLM--LLM interactions, where Claude 3.5, Claude 3.7, Claude 3.7(T), and Llama 3.3 almost always achieve 
$m=1.0\pm 0.0$. This means they establish the relevant strategic regime immediately—consistent with the perfect-cooperation totals reported earlier. In contrast, Claude 4, DeepSeek R1, and especially Mistral are less stable in these same settings, with higher and more variable 
$m$, indicating delayed or inconsistent inference of the opponent’s behavior.

Against algorithmic opponents, the easiest case is clearly SREP, where many models adapt within the first 
1–3 rounds. Claude 4 and Claude 4(T) are particularly strong here, often reaching 
$m=1.0\pm 0.0$, and DeepSeek also performs well on several prompting variants. By contrast, PP is much harder: for the otherwise strong Claude 3.x and Llama models,
$m$ often rises to roughly 8–15, showing that cyclic structure takes longer to exploit than persistent defection. This supports the interpretation that PD comprehension depends strongly on opponent type rather than only model size.

A third notable point is that Mistral is the weakest and most unstable model overall. It often remains near the game horizon for algorithmic opponents—for example 
$m\approx 16$ for SREP, MF, and TFT in several settings—indicating that it frequently fails to reliably infer the opponent within the game. This sharply contrasts with the other models and aligns with its poorer total-point performance.

Finally, self-consistency sometimes stabilizes comprehension but does not uniformly improve it. For example, Claude 4(T) under SC often collapses to 
$m=1$ across many opponent types, while for Claude 3.x and Llama the gains are smaller because they are already near-optimal. Overall, the table shows that default PD is easy for the strongest models in cooperative LLM--LLM settings and against simple defecting opponents, but more demanding for cyclic algorithmic play and substantially harder for weaker models such as Mistral.

\begin{table*}[h!]
\small
\centering
\begin{tabular}{lllllllll}
\toprule
 &  & \multicolumn{7}{c}{\textbf{PD label-based counterfactual}} \\
 &  & zs & spp & cot & srep & pp & mf & tft \\
model & prompt &  &  &  &  &  &  &  \\
\midrule
\multirow[t]{6}{*}{C3.5Sv2} & zs & \textbf{1.0 $\pm$ 0.0} & \textbf{1.0 $\pm$ 0.0} & \textbf{1.0 $\pm$ 0.0} & 14.0 $\pm$ 1.9 & 13.4 $\pm$ 2.2 & 2.0 $\pm$ 0.0 & 10.0 $\pm$ 8.2 \\
 & cot & \textbf{1.0 $\pm$ 0.0} & \textbf{1.0 $\pm$ 0.0} & \textbf{1.0 $\pm$ 0.0} & 16.6 $\pm$ 0.5 & 15.4 $\pm$ 1.7 & 1.6 $\pm$ 0.5 & 9.0 $\pm$ 8.0 \\
 & spp & \textbf{1.0 $\pm$ 0.0} & \textbf{1.0 $\pm$ 0.0} & \textbf{1.0 $\pm$ 0.0} & 15.6 $\pm$ 1.1 & 16.6 $\pm$ 0.9 & 1.8 $\pm$ 0.4 & 6.0 $\pm$ 7.3 \\
 & sc-zs & \textbf{1.0 $\pm$ 0.0} & \textbf{1.0 $\pm$ 0.0} & \textbf{1.0 $\pm$ 0.0} & 16.5 $\pm$ 0.7 & 11.0 $\pm$ 0.0 & 2.0 $\pm$ 0.0 & \textbf{1.0 $\pm$ 0.0} \\
 & sc-cot & \textbf{1.0 $\pm$ 0.0} & \textbf{1.0 $\pm$ 0.0} & \textbf{1.0 $\pm$ 0.0} & 16.0 $\pm$ 1.4 & 15.0 $\pm$ 2.8 & 2.0 $\pm$ 0.0 & 9.0 $\pm$ 11.3 \\
 & sc-spp & \textbf{1.0 $\pm$ 0.0} & \textbf{1.0 $\pm$ 0.0} & \textbf{1.0 $\pm$ 0.0} & 16.0 $\pm$ 1.4 & 10.5 $\pm$ 9.2 & 1.5 $\pm$ 0.7 & 5.5 $\pm$ 2.1 \\
\cline{1-9}
\multirow[t]{6}{*}{C3.7S} & zs & \textbf{1.0 $\pm$ 0.0} & \textbf{1.0 $\pm$ 0.0} & \textbf{1.0 $\pm$ 0.0} & 15.6 $\pm$ 2.1 & 15.4 $\pm$ 2.2 & 1.6 $\pm$ 0.5 & 5.4 $\pm$ 7.0 \\
 & cot & \textbf{1.0 $\pm$ 0.0} & \textbf{1.0 $\pm$ 0.0} & \textbf{1.0 $\pm$ 0.0} & 15.8 $\pm$ 1.1 & 12.2 $\pm$ 3.3 & 1.4 $\pm$ 0.5 & 8.8 $\pm$ 7.4 \\
 & spp & \textbf{1.0 $\pm$ 0.0} & \textbf{1.0 $\pm$ 0.0} & \textbf{1.0 $\pm$ 0.0} & 17.0 $\pm$ 0.0 & 14.6 $\pm$ 5.4 & 1.2 $\pm$ 0.4 & 9.8 $\pm$ 7.0 \\
 & sc-zs & \textbf{1.0 $\pm$ 0.0} & \textbf{1.0 $\pm$ 0.0} & \textbf{1.0 $\pm$ 0.0} & 15.0 $\pm$ 2.8 & 17.0 $\pm$ 0.0 & 1.5 $\pm$ 0.7 & 16.0 $\pm$ 1.4 \\
 & sc-cot & \textbf{1.0 $\pm$ 0.0} & \textbf{1.0 $\pm$ 0.0} & \textbf{1.0 $\pm$ 0.0} & 11.0 $\pm$ 8.5 & 11.5 $\pm$ 7.8 & 1.5 $\pm$ 0.7 & 8.5 $\pm$ 10.6 \\
 & sc-spp & \textbf{1.0 $\pm$ 0.0} & \textbf{1.0 $\pm$ 0.0} & \textbf{1.0 $\pm$ 0.0} & 17.0 $\pm$ 0.0 & 16.0 $\pm$ 1.4 & 1.5 $\pm$ 0.7 & 12.5 $\pm$ 3.5 \\
\cline{1-9}
\multirow[t]{6}{*}{C3.7S(T)} & zs & \textbf{1.0 $\pm$ 0.0} & \textbf{1.0 $\pm$ 0.0} & \textbf{1.0 $\pm$ 0.0} & 15.4 $\pm$ 1.1 & 14.2 $\pm$ 2.7 & 7.4 $\pm$ 7.5 & 11.4 $\pm$ 6.0 \\
 & cot & \textbf{1.0 $\pm$ 0.0} & 1.6 $\pm$ 1.3 & \textbf{1.0 $\pm$ 0.0} & 16.4 $\pm$ 0.9 & 14.0 $\pm$ 5.7 & \textbf{1.0 $\pm$ 0.0} & 5.8 $\pm$ 6.6 \\
 & spp & \textbf{1.0 $\pm$ 0.0} & \textbf{1.0 $\pm$ 0.0} & \textbf{1.0 $\pm$ 0.0} & 13.0 $\pm$ 4.2 & 12.4 $\pm$ 5.7 & 1.6 $\pm$ 0.5 & 5.2 $\pm$ 6.6 \\
 & sc-zs & \textbf{1.0 $\pm$ 0.0} & \textbf{1.0 $\pm$ 0.0} & \textbf{1.0 $\pm$ 0.0} & 16.5 $\pm$ 0.7 & 16.0 $\pm$ 1.4 & 8.0 $\pm$ 8.5 & \textbf{1.0 $\pm$ 0.0} \\
 & sc-cot & \textbf{1.0 $\pm$ 0.0} & \textbf{1.0 $\pm$ 0.0} & \textbf{1.0 $\pm$ 0.0} & 16.5 $\pm$ 0.7 & 7.5 $\pm$ 4.9 & 2.0 $\pm$ 0.0 & 9.5 $\pm$ 3.5 \\
 & sc-spp & \textbf{1.0 $\pm$ 0.0} & \textbf{1.0 $\pm$ 0.0} & \textbf{1.0 $\pm$ 0.0} & 14.0 $\pm$ 0.0 & 16.0 $\pm$ 1.4 & 1.5 $\pm$ 0.7 & 4.0 $\pm$ 4.2 \\
\cline{1-9}
\multirow[t]{6}{*}{C4S} & zs & 2.4 $\pm$ 2.6 & \textbf{1.2 $\pm$ 0.4} & 2.0 $\pm$ 2.2 & 17.0 $\pm$ 0.0 & 16.6 $\pm$ 0.9 & 2.6 $\pm$ 1.9 & 7.2 $\pm$ 4.5 \\
 & cot & \textbf{1.2 $\pm$ 0.4} & 5.8 $\pm$ 6.1 & 8.8 $\pm$ 7.4 & 16.2 $\pm$ 0.8 & 14.0 $\pm$ 5.7 & 3.0 $\pm$ 1.4 & 7.0 $\pm$ 6.0 \\
 & spp & 2.4 $\pm$ 2.6 & 5.8 $\pm$ 4.3 & 4.2 $\pm$ 2.6 & 15.2 $\pm$ 2.5 & 14.8 $\pm$ 4.9 & \textbf{2.2 $\pm$ 2.2} & 7.2 $\pm$ 5.8 \\
 & sc-zs & \textbf{1.0 $\pm$ 0.0} & \textbf{1.0 $\pm$ 0.0} & 1.5 $\pm$ 0.7 & 14.0 $\pm$ 1.4 & 17.0 $\pm$ 0.0 & 2.0 $\pm$ 0.0 & 10.5 $\pm$ 7.8 \\
 & sc-cot & 3.0 $\pm$ 2.8 & 11.5 $\pm$ 2.1 & 10.5 $\pm$ 7.8 & 15.5 $\pm$ 2.1 & 17.0 $\pm$ 0.0 & \textbf{1.0 $\pm$ 0.0} & 10.0 $\pm$ 0.0 \\
 & sc-spp & \textbf{1.0 $\pm$ 0.0} & \textbf{1.0 $\pm$ 0.0} & 2.5 $\pm$ 0.7 & 14.5 $\pm$ 3.5 & 17.0 $\pm$ 0.0 & \textbf{1.0 $\pm$ 0.0} & 9.5 $\pm$ 6.4 \\
\cline{1-9}
\multirow[t]{6}{*}{C4S(T)} & zs & \textbf{1.0 $\pm$ 0.0} & 1.2 $\pm$ 0.4 & 2.0 $\pm$ 2.2 & 15.6 $\pm$ 1.1 & 17.0 $\pm$ 0.0 & 2.4 $\pm$ 1.5 & 4.4 $\pm$ 5.0 \\
 & cot & 3.0 $\pm$ 4.5 & 2.2 $\pm$ 2.2 & 2.0 $\pm$ 1.7 & 14.8 $\pm$ 2.0 & 16.6 $\pm$ 0.9 & \textbf{1.6 $\pm$ 0.5} & 4.6 $\pm$ 5.0 \\
 & spp & 1.2 $\pm$ 0.4 & 1.8 $\pm$ 1.3 & \textbf{1.0 $\pm$ 0.0} & 12.8 $\pm$ 4.8 & 14.2 $\pm$ 6.3 & 4.2 $\pm$ 5.5 & 1.8 $\pm$ 1.8 \\
 & sc-zs & \textbf{1.0 $\pm$ 0.0} & \textbf{1.0 $\pm$ 0.0} & \textbf{1.0 $\pm$ 0.0} & 17.0 $\pm$ 0.0 & 14.0 $\pm$ 1.4 & 3.5 $\pm$ 3.5 & 10.0 $\pm$ 5.7 \\
 & sc-cot & 2.5 $\pm$ 2.1 & \textbf{1.0 $\pm$ 0.0} & \textbf{1.0 $\pm$ 0.0} & 17.0 $\pm$ 0.0 & 17.0 $\pm$ 0.0 & \textbf{1.0 $\pm$ 0.0} & 5.0 $\pm$ 5.7 \\
 & sc-spp & 1.5 $\pm$ 0.7 & \textbf{1.0 $\pm$ 0.0} & \textbf{1.0 $\pm$ 0.0} & 16.5 $\pm$ 0.7 & 17.0 $\pm$ 0.0 & 1.5 $\pm$ 0.7 & 3.0 $\pm$ 2.8 \\
\cline{1-9}
\multirow[t]{6}{*}{DS-R1} & zs & 8.4 $\pm$ 4.9 & 8.2 $\pm$ 5.9 & 11.2 $\pm$ 8.0 & 15.0 $\pm$ 1.6 & 17.0 $\pm$ 0.0 & \textbf{1.6 $\pm$ 0.5} & 8.2 $\pm$ 6.6 \\
 & cot & 4.2 $\pm$ 5.2 & 5.8 $\pm$ 6.6 & 3.6 $\pm$ 5.8 & 13.6 $\pm$ 5.0 & 17.0 $\pm$ 0.0 & \textbf{3.2 $\pm$ 2.7} & 3.8 $\pm$ 4.1 \\
 & spp & 8.0 $\pm$ 5.8 & 4.6 $\pm$ 3.6 & 1.6 $\pm$ 1.3 & 14.8 $\pm$ 1.6 & 16.6 $\pm$ 0.9 & \textbf{1.2 $\pm$ 0.4} & 5.6 $\pm$ 6.6 \\
 & sc-zs & \textbf{1.0 $\pm$ 0.0} & 8.5 $\pm$ 10.6 & 4.0 $\pm$ 4.2 & 15.5 $\pm$ 0.7 & 17.0 $\pm$ 0.0 & 1.5 $\pm$ 0.7 & 8.5 $\pm$ 10.6 \\
 & sc-cot & \textbf{1.0 $\pm$ 0.0} & 7.0 $\pm$ 8.5 & 9.0 $\pm$ 11.3 & 16.0 $\pm$ 0.0 & 17.0 $\pm$ 0.0 & \textbf{1.0 $\pm$ 0.0} & 6.0 $\pm$ 7.1 \\
 & sc-spp & 3.0 $\pm$ 2.8 & \textbf{1.0 $\pm$ 0.0} & 6.0 $\pm$ 1.4 & 16.0 $\pm$ 1.4 & 17.0 $\pm$ 0.0 & \textbf{1.0 $\pm$ 0.0} & 5.0 $\pm$ 5.7 \\
\cline{1-9}
\multirow[t]{6}{*}{L3.3-70B} & zs & \textbf{1.0 $\pm$ 0.0} & \textbf{1.0 $\pm$ 0.0} & \textbf{1.0 $\pm$ 0.0} & 16.4 $\pm$ 0.9 & 15.0 $\pm$ 0.0 & 3.0 $\pm$ 2.7 & 6.8 $\pm$ 3.8 \\
 & cot & \textbf{1.0 $\pm$ 0.0} & \textbf{1.0 $\pm$ 0.0} & \textbf{1.0 $\pm$ 0.0} & 16.4 $\pm$ 0.9 & 15.8 $\pm$ 1.1 & 2.4 $\pm$ 2.1 & 5.2 $\pm$ 3.0 \\
 & spp & \textbf{1.0 $\pm$ 0.0} & \textbf{1.0 $\pm$ 0.0} & \textbf{1.0 $\pm$ 0.0} & 13.8 $\pm$ 4.5 & 17.0 $\pm$ 0.0 & 1.6 $\pm$ 0.5 & 2.2 $\pm$ 2.7 \\
 & sc-zs & \textbf{1.0 $\pm$ 0.0} & \textbf{1.0 $\pm$ 0.0} & \textbf{1.0 $\pm$ 0.0} & 12.0 $\pm$ 7.1 & 15.0 $\pm$ 0.0 & \textbf{1.0 $\pm$ 0.0} & \textbf{1.0 $\pm$ 0.0} \\
 & sc-cot & \textbf{1.0 $\pm$ 0.0} & \textbf{1.0 $\pm$ 0.0} & \textbf{1.0 $\pm$ 0.0} & 12.5 $\pm$ 0.7 & 17.0 $\pm$ 0.0 & \textbf{1.0 $\pm$ 0.0} & 3.0 $\pm$ 2.8 \\
 & sc-spp & \textbf{1.0 $\pm$ 0.0} & \textbf{1.0 $\pm$ 0.0} & \textbf{1.0 $\pm$ 0.0} & 15.5 $\pm$ 2.1 & 15.0 $\pm$ 0.0 & 3.5 $\pm$ 3.5 & \textbf{1.0 $\pm$ 0.0} \\
\cline{1-9}
\multirow[t]{6}{*}{Mistral} & zs & \textbf{1.0 $\pm$ 0.0} & 6.4 $\pm$ 7.6 & \textbf{1.0 $\pm$ 0.0} & 16.6 $\pm$ 0.5 & 17.0 $\pm$ 0.0 & 4.4 $\pm$ 6.5 & 4.0 $\pm$ 6.7 \\
 & cot & 4.2 $\pm$ 7.2 & 4.2 $\pm$ 7.2 & \textbf{1.0 $\pm$ 0.0} & 16.6 $\pm$ 0.5 & 17.0 $\pm$ 0.0 & 4.4 $\pm$ 6.5 & 7.4 $\pm$ 8.8 \\
 & spp & \textbf{1.0 $\pm$ 0.0} & 7.6 $\pm$ 8.6 & 1.2 $\pm$ 0.4 & 16.6 $\pm$ 0.5 & 17.0 $\pm$ 0.0 & 4.4 $\pm$ 7.1 & 9.0 $\pm$ 8.0 \\
 & sc-zs & \textbf{1.0 $\pm$ 0.0} & \textbf{1.0 $\pm$ 0.0} & 3.0 $\pm$ 2.8 & 16.0 $\pm$ 1.4 & 16.0 $\pm$ 1.4 & 1.5 $\pm$ 0.7 & \textbf{1.0 $\pm$ 0.0} \\
 & sc-cot & \textbf{1.0 $\pm$ 0.0} & \textbf{1.0 $\pm$ 0.0} & \textbf{1.0 $\pm$ 0.0} & 16.5 $\pm$ 0.7 & 17.0 $\pm$ 0.0 & 1.5 $\pm$ 0.7 & 9.0 $\pm$ 11.3 \\
 & sc-spp & \textbf{1.0 $\pm$ 0.0} & 9.0 $\pm$ 11.3 & \textbf{1.0 $\pm$ 0.0} & 17.0 $\pm$ 0.0 & 17.0 $\pm$ 0.0 & 2.0 $\pm$ 0.0 & 11.0 $\pm$ 8.5 \\
\bottomrule
\end{tabular}
\caption{Round $m$\# where the LLM player understood the opponent's Strategy in the label-based counterfactual PD setup.}
\label{tab:pd_round_heatmap_pd-alt}
\end{table*}

The label-based counterfactual opponent comprehension results are showcased in Table \ref{tab:pd_round_heatmap_pd-alt}.  This table shows that the label-based counterfactual preserves early comprehension in LLM--LLM play for the strongest models, but it substantially hurts comprehension against algorithmic opponents. Claude 3.5/3.7 and Llama 3.3 still achieve 
$m=1.0 \pm 0.0$ across the LLM-opponent columns, indicating that relabeling alone does not disrupt mutual coordination when the opponent is another LLM. However, for the same models, comprehension against \textbf{SREP} and \textbf{PP} often deteriorates sharply to near-horizon values ($m\approx 14–
17$), showing that changing action names can strongly delay recognition of deterministic strategies even when payoffs remain unchanged.
\begin{table*}[h!]
\small
\centering
\begin{tabular}{lllllllll}
\toprule
 &  & \multicolumn{7}{c}{\textbf{SH (payoff-based counterfactual of PD)}} \\
 &  & zs & spp & cot & srep & pp & mf & tft \\
model & prompt &  &  &  &  &  &  &  \\
\midrule
\multirow[t]{6}{*}{C3.5Sv2} & zs & \textbf{1.0 $\pm$ 0.0} & \textbf{1.0 $\pm$ 0.0} & \textbf{1.0 $\pm$ 0.0} & 15.8 $\pm$ 0.8 & 13.2 $\pm$ 5.8 & 1.6 $\pm$ 0.5 & 4.2 $\pm$ 4.1 \\
 & cot & \textbf{1.0 $\pm$ 0.0} & \textbf{1.0 $\pm$ 0.0} & \textbf{1.0 $\pm$ 0.0} & 16.0 $\pm$ 1.2 & 17.0 $\pm$ 0.0 & 3.4 $\pm$ 2.3 & 6.2 $\pm$ 4.0 \\
 & spp & \textbf{1.0 $\pm$ 0.0} & \textbf{1.0 $\pm$ 0.0} & \textbf{1.0 $\pm$ 0.0} & 11.8 $\pm$ 5.8 & 17.0 $\pm$ 0.0 & 2.4 $\pm$ 1.5 & 1.8 $\pm$ 1.8 \\
 & sc-zs & \textbf{1.0 $\pm$ 0.0} & \textbf{1.0 $\pm$ 0.0} & \textbf{1.0 $\pm$ 0.0} & 16.0 $\pm$ 0.0 & 11.5 $\pm$ 7.8 & 1.5 $\pm$ 0.7 & \textbf{1.0 $\pm$ 0.0} \\
 & sc-cot & \textbf{1.0 $\pm$ 0.0} & \textbf{1.0 $\pm$ 0.0} & \textbf{1.0 $\pm$ 0.0} & 15.0 $\pm$ 2.8 & 11.5 $\pm$ 7.8 & 4.0 $\pm$ 0.0 & 5.5 $\pm$ 6.4 \\
 & sc-spp & \textbf{1.0 $\pm$ 0.0} & \textbf{1.0 $\pm$ 0.0} & \textbf{1.0 $\pm$ 0.0} & 16.5 $\pm$ 0.7 & 17.0 $\pm$ 0.0 & 3.0 $\pm$ 2.8 & 2.5 $\pm$ 2.1 \\
\cline{1-9}
\multirow[t]{6}{*}{C3.7S} & zs & \textbf{1.0 $\pm$ 0.0} & \textbf{1.0 $\pm$ 0.0} & \textbf{1.0 $\pm$ 0.0} & 16.4 $\pm$ 0.9 & 12.0 $\pm$ 5.0 & 1.4 $\pm$ 0.5 & 6.2 $\pm$ 7.4 \\
 & cot & \textbf{1.0 $\pm$ 0.0} & \textbf{1.0 $\pm$ 0.0} & \textbf{1.0 $\pm$ 0.0} & 14.4 $\pm$ 2.6 & 13.2 $\pm$ 3.0 & 1.4 $\pm$ 0.5 & 4.8 $\pm$ 6.9 \\
 & spp & \textbf{1.0 $\pm$ 0.0} & \textbf{1.0 $\pm$ 0.0} & \textbf{1.0 $\pm$ 0.0} & 15.0 $\pm$ 2.3 & 11.6 $\pm$ 4.0 & 1.4 $\pm$ 0.5 & 8.6 $\pm$ 8.0 \\
 & sc-zs & \textbf{1.0 $\pm$ 0.0} & \textbf{1.0 $\pm$ 0.0} & \textbf{1.0 $\pm$ 0.0} & 17.0 $\pm$ 0.0 & 16.0 $\pm$ 1.4 & 2.0 $\pm$ 0.0 & 4.0 $\pm$ 4.2 \\
 & sc-cot & \textbf{1.0 $\pm$ 0.0} & \textbf{1.0 $\pm$ 0.0} & \textbf{1.0 $\pm$ 0.0} & 16.5 $\pm$ 0.7 & 8.5 $\pm$ 3.5 & 1.5 $\pm$ 0.7 & 8.5 $\pm$ 4.9 \\
 & sc-spp & \textbf{1.0 $\pm$ 0.0} & \textbf{1.0 $\pm$ 0.0} & \textbf{1.0 $\pm$ 0.0} & 15.5 $\pm$ 0.7 & 10.5 $\pm$ 9.2 & 1.5 $\pm$ 0.7 & \textbf{1.0 $\pm$ 0.0} \\
\cline{1-9}
\multirow[t]{6}{*}{C3.7S(T)} & zs & \textbf{1.0 $\pm$ 0.0} & \textbf{1.0 $\pm$ 0.0} & \textbf{1.0 $\pm$ 0.0} & 16.4 $\pm$ 0.5 & 15.0 $\pm$ 1.4 & 2.2 $\pm$ 2.2 & 6.6 $\pm$ 6.5 \\
 & cot & \textbf{1.0 $\pm$ 0.0} & \textbf{1.0 $\pm$ 0.0} & \textbf{1.0 $\pm$ 0.0} & 16.0 $\pm$ 1.0 & 11.2 $\pm$ 5.4 & 1.2 $\pm$ 0.4 & 4.4 $\pm$ 4.1 \\
 & spp & \textbf{1.0 $\pm$ 0.0} & \textbf{1.0 $\pm$ 0.0} & \textbf{1.0 $\pm$ 0.0} & 14.6 $\pm$ 2.4 & 10.0 $\pm$ 4.1 & 1.6 $\pm$ 0.5 & 7.0 $\pm$ 6.6 \\
 & sc-zs & \textbf{1.0 $\pm$ 0.0} & \textbf{1.0 $\pm$ 0.0} & \textbf{1.0 $\pm$ 0.0} & 16.0 $\pm$ 1.4 & 16.0 $\pm$ 1.4 & 1.5 $\pm$ 0.7 & 10.0 $\pm$ 7.1 \\
 & sc-cot & \textbf{1.0 $\pm$ 0.0} & \textbf{1.0 $\pm$ 0.0} & \textbf{1.0 $\pm$ 0.0} & 11.5 $\pm$ 3.5 & 7.0 $\pm$ 2.8 & \textbf{1.0 $\pm$ 0.0} & 4.0 $\pm$ 4.2 \\
 & sc-spp & \textbf{1.0 $\pm$ 0.0} & \textbf{1.0 $\pm$ 0.0} & \textbf{1.0 $\pm$ 0.0} & \textbf{11.0 $\pm$ 7.1} & 13.0 $\pm$ 5.7 & \textbf{1.0 $\pm$ 0.0} & 10.0 $\pm$ 1.4 \\
\cline{1-9}
\multirow[t]{6}{*}{C4S} & zs & 3.8 $\pm$ 6.3 & 7.6 $\pm$ 7.2 & 6.4 $\pm$ 7.6 & 16.8 $\pm$ 0.4 & 13.6 $\pm$ 6.5 & \textbf{1.8 $\pm$ 1.3} & 8.8 $\pm$ 5.9 \\
 & cot & \textbf{3.0 $\pm$ 2.1} & 9.6 $\pm$ 7.1 & 4.4 $\pm$ 2.7 & 15.4 $\pm$ 1.9 & 13.4 $\pm$ 6.1 & 10.0 $\pm$ 6.6 & 12.4 $\pm$ 6.6 \\
 & spp & 6.6 $\pm$ 5.9 & 8.4 $\pm$ 5.9 & 6.2 $\pm$ 5.4 & 15.8 $\pm$ 2.7 & 12.0 $\pm$ 6.9 & 4.4 $\pm$ 3.8 & \textbf{2.2 $\pm$ 1.1} \\
 & sc-zs & \textbf{1.0 $\pm$ 0.0} & \textbf{1.0 $\pm$ 0.0} & 8.5 $\pm$ 9.2 & 14.5 $\pm$ 2.1 & 9.5 $\pm$ 7.8 & 1.5 $\pm$ 0.7 & 4.0 $\pm$ 4.2 \\
 & sc-cot & 14.5 $\pm$ 2.1 & \textbf{2.0 $\pm$ 1.4} & 9.5 $\pm$ 6.4 & 14.5 $\pm$ 3.5 & 17.0 $\pm$ 0.0 & 4.0 $\pm$ 2.8 & 14.5 $\pm$ 2.1 \\
 & sc-spp & \textbf{3.5 $\pm$ 3.5} & 8.5 $\pm$ 6.4 & 6.5 $\pm$ 7.8 & 15.5 $\pm$ 2.1 & 17.0 $\pm$ 0.0 & 4.0 $\pm$ 2.8 & 7.5 $\pm$ 7.8 \\
\cline{1-9}
\multirow[t]{6}{*}{C4S(T)} & zs & \textbf{1.0 $\pm$ 0.0} & 4.8 $\pm$ 5.9 & 3.6 $\pm$ 3.4 & 15.0 $\pm$ 2.9 & 15.4 $\pm$ 3.6 & 2.6 $\pm$ 1.8 & 5.4 $\pm$ 3.0 \\
 & cot & \textbf{1.2 $\pm$ 0.4} & 5.0 $\pm$ 6.7 & 2.6 $\pm$ 1.8 & 16.0 $\pm$ 0.0 & 17.0 $\pm$ 0.0 & 6.0 $\pm$ 6.4 & 10.2 $\pm$ 4.8 \\
 & spp & 3.0 $\pm$ 4.5 & 2.2 $\pm$ 1.3 & 2.2 $\pm$ 2.2 & 16.6 $\pm$ 0.5 & 14.4 $\pm$ 5.8 & \textbf{2.0 $\pm$ 1.7} & 7.8 $\pm$ 6.1 \\
 & sc-zs & 3.5 $\pm$ 3.5 & 5.0 $\pm$ 1.4 & 1.5 $\pm$ 0.7 & 15.0 $\pm$ 1.4 & \textbf{6.5 $\pm$ 3.5} & \textbf{1.0 $\pm$ 0.0} & 9.5 $\pm$ 3.5 \\
 & sc-cot & 6.5 $\pm$ 4.9 & \textbf{1.0 $\pm$ 0.0} & 1.5 $\pm$ 0.7 & 15.5 $\pm$ 2.1 & 17.0 $\pm$ 0.0 & 4.0 $\pm$ 2.8 & 2.5 $\pm$ 2.1 \\
 & sc-spp & \textbf{2.0 $\pm$ 0.0} & \textbf{2.0 $\pm$ 1.4} & 6.0 $\pm$ 7.1 & 17.0 $\pm$ 0.0 & 11.5 $\pm$ 7.8 & 5.0 $\pm$ 1.4 & 7.0 $\pm$ 8.5 \\
\cline{1-9}
\multirow[t]{6}{*}{DS-R1} & zs & 4.8 $\pm$ 3.6 & \textbf{1.2 $\pm$ 0.4} & 9.0 $\pm$ 5.5 & 15.8 $\pm$ 0.8 & 17.0 $\pm$ 0.0 & 2.2 $\pm$ 1.6 & 7.2 $\pm$ 4.0 \\
 & cot & \textbf{1.8 $\pm$ 1.3} & \textbf{1.8 $\pm$ 0.8} & 5.4 $\pm$ 4.9 & 14.6 $\pm$ 1.8 & 16.6 $\pm$ 0.9 & 3.0 $\pm$ 1.6 & 7.6 $\pm$ 5.1 \\
 & spp & 3.4 $\pm$ 4.3 & 11.2 $\pm$ 3.0 & 3.6 $\pm$ 5.8 & 12.2 $\pm$ 3.0 & 16.6 $\pm$ 0.9 & \textbf{1.6 $\pm$ 0.5} & 4.2 $\pm$ 3.0 \\
 & sc-zs & 1.5 $\pm$ 0.7 & 4.0 $\pm$ 4.2 & 6.0 $\pm$ 7.1 & 13.5 $\pm$ 3.5 & 17.0 $\pm$ 0.0 & \textbf{1.0 $\pm$ 0.0} & \textbf{1.0 $\pm$ 0.0} \\
 & sc-cot & \textbf{1.0 $\pm$ 0.0} & 9.0 $\pm$ 8.5 & 10.0 $\pm$ 7.1 & 16.5 $\pm$ 0.7 & 17.0 $\pm$ 0.0 & 3.5 $\pm$ 2.1 & 7.0 $\pm$ 5.7 \\
 & sc-spp & 5.0 $\pm$ 5.7 & 11.0 $\pm$ 8.5 & 7.5 $\pm$ 6.4 & 17.0 $\pm$ 0.0 & 17.0 $\pm$ 0.0 & \textbf{1.0 $\pm$ 0.0} & 1.5 $\pm$ 0.7 \\
\cline{1-9}
\multirow[t]{6}{*}{L3.3-70B} & zs & \textbf{1.0 $\pm$ 0.0} & \textbf{1.0 $\pm$ 0.0} & \textbf{1.0 $\pm$ 0.0} & 15.4 $\pm$ 1.8 & 15.4 $\pm$ 0.9 & 2.2 $\pm$ 2.2 & 11.6 $\pm$ 6.8 \\
 & cot & 4.0 $\pm$ 6.7 & \textbf{1.0 $\pm$ 0.0} & \textbf{1.0 $\pm$ 0.0} & 15.6 $\pm$ 1.1 & 16.6 $\pm$ 0.9 & 1.2 $\pm$ 0.4 & 2.2 $\pm$ 2.7 \\
 & spp & \textbf{1.0 $\pm$ 0.0} & \textbf{1.0 $\pm$ 0.0} & \textbf{1.0 $\pm$ 0.0} & 15.0 $\pm$ 1.4 & 16.2 $\pm$ 1.1 & 5.4 $\pm$ 5.2 & 3.8 $\pm$ 3.9 \\
 & sc-zs & \textbf{1.0 $\pm$ 0.0} & \textbf{1.0 $\pm$ 0.0} & \textbf{1.0 $\pm$ 0.0} & 16.0 $\pm$ 1.4 & 15.0 $\pm$ 0.0 & \textbf{1.0 $\pm$ 0.0} & \textbf{1.0 $\pm$ 0.0} \\
 & sc-cot & \textbf{1.0 $\pm$ 0.0} & \textbf{1.0 $\pm$ 0.0} & \textbf{1.0 $\pm$ 0.0} & 11.5 $\pm$ 7.8 & 17.0 $\pm$ 0.0 & 2.5 $\pm$ 2.1 & 3.0 $\pm$ 2.8 \\
 & sc-spp & \textbf{1.0 $\pm$ 0.0} & \textbf{1.0 $\pm$ 0.0} & \textbf{1.0 $\pm$ 0.0} & 16.5 $\pm$ 0.7 & 16.0 $\pm$ 1.4 & \textbf{1.0 $\pm$ 0.0} & \textbf{1.0 $\pm$ 0.0} \\
\cline{1-9}
\multirow[t]{6}{*}{M-L(24.07)} & zs & 13.4 $\pm$ 7.0 & 8.2 $\pm$ 7.6 & 12.6 $\pm$ 6.5 & 15.8 $\pm$ 1.6 & 14.4 $\pm$ 4.8 & \textbf{3.4 $\pm$ 1.8} & 7.4 $\pm$ 8.8 \\
 & cot & 8.0 $\pm$ 7.9 & \textbf{5.0 $\pm$ 6.5} & 10.0 $\pm$ 7.4 & 16.4 $\pm$ 0.5 & 13.4 $\pm$ 7.0 & 12.8 $\pm$ 6.4 & 12.2 $\pm$ 6.9 \\
 & spp & 7.4 $\pm$ 8.3 & 9.8 $\pm$ 8.1 & 11.2 $\pm$ 7.9 & 16.4 $\pm$ 1.3 & 17.0 $\pm$ 0.0 & \textbf{4.8 $\pm$ 6.4} & 11.0 $\pm$ 8.2 \\
 & sc-zs & \textbf{1.0 $\pm$ 0.0} & 8.0 $\pm$ 9.9 & 9.0 $\pm$ 11.3 & 17.0 $\pm$ 0.0 & 17.0 $\pm$ 0.0 & 1.5 $\pm$ 0.7 & 17.0 $\pm$ 0.0 \\
 & sc-cot & \textbf{8.5 $\pm$ 10.6} & 11.5 $\pm$ 7.8 & 16.0 $\pm$ 1.4 & 17.0 $\pm$ 0.0 & 16.0 $\pm$ 1.4 & \textbf{8.5 $\pm$ 10.6} & 12.5 $\pm$ 4.9 \\
 & sc-spp & 16.5 $\pm$ 0.7 & 9.0 $\pm$ 11.3 & 15.0 $\pm$ 1.4 & 15.5 $\pm$ 2.1 & 17.0 $\pm$ 0.0 & \textbf{1.0 $\pm$ 0.0} & 9.0 $\pm$ 11.3 \\
\bottomrule
\end{tabular}
\caption{Round $m$\# where the LLM player understood the opponent's Strategy in the SH counterfactual.}
\label{tab:pd_round_heatmap_sh}
\end{table*}

A second notable pattern is that MF remains comparatively easy for many models under the label shift, often with 
$m$ close to 1-3, while \textbf{TFT} becomes much more variable. This suggests that simple frequency-based exploitation transfers more robustly across renamed actions than response policies that depend on correctly interpreting the opponent’s most recent move. Among the stronger models, Claude 4 variants remain mixed, with good LLM--LLM comprehension but unstable algorithmic performance, while DeepSeek R1 and especially Mistral are the least robust overall, frequently showing delayed comprehension across multiple opponent types. Overall, the table supports the claim that label-only changes mainly impair abstraction over opponent policy rather than cooperation with other LLMs.

\begin{table*}[h!]
\small
\centering
\begin{tabular}{lllllllll}
\toprule
 &  & \multicolumn{7}{c}{\textbf{Joint counterfactual of PD}} \\
 &  & zs & spp & cot & srep & pp & mf & tft \\
model & prompt &  &  &  &  &  &  &  \\
\midrule
\multirow[t]{6}{*}{C3.5Sv2} & zs & \textbf{1.0 $\pm$ 0.0} & \textbf{1.0 $\pm$ 0.0} & \textbf{1.0 $\pm$ 0.0} & 1.8 $\pm$ 0.4 & 2.0 $\pm$ 0.0 & 2.6 $\pm$ 0.5 & 2.2 $\pm$ 0.8 \\
 & cot & \textbf{1.0 $\pm$ 0.0} & \textbf{1.0 $\pm$ 0.0} & \textbf{1.0 $\pm$ 0.0} & 1.2 $\pm$ 0.4 & 6.0 $\pm$ 5.8 & 1.8 $\pm$ 1.1 & 2.8 $\pm$ 0.4 \\
 & spp & \textbf{1.0 $\pm$ 0.0} & \textbf{1.0 $\pm$ 0.0} & \textbf{1.0 $\pm$ 0.0} & 1.6 $\pm$ 0.5 & 2.0 $\pm$ 0.0 & 2.8 $\pm$ 0.4 & 1.8 $\pm$ 0.8 \\
 & sc-zs & \textbf{1.0 $\pm$ 0.0} & \textbf{1.0 $\pm$ 0.0} & \textbf{1.0 $\pm$ 0.0} & 1.5 $\pm$ 0.7 & 3.0 $\pm$ 1.4 & 2.5 $\pm$ 0.7 & 3.0 $\pm$ 0.0 \\
 & sc-cot & \textbf{1.0 $\pm$ 0.0} & \textbf{1.0 $\pm$ 0.0} & \textbf{1.0 $\pm$ 0.0} & 1.5 $\pm$ 0.7 & 4.0 $\pm$ 2.8 & 2.0 $\pm$ 1.4 & 3.0 $\pm$ 0.0 \\
 & sc-spp & \textbf{1.0 $\pm$ 0.0} & \textbf{1.0 $\pm$ 0.0} & \textbf{1.0 $\pm$ 0.0} & 1.5 $\pm$ 0.7 & 4.0 $\pm$ 2.8 & 2.5 $\pm$ 0.7 & 2.0 $\pm$ 1.4 \\
\cline{1-9}
\multirow[t]{6}{*}{C3.7S} & zs & \textbf{1.0 $\pm$ 0.0} & \textbf{1.0 $\pm$ 0.0} & \textbf{1.0 $\pm$ 0.0} & 2.0 $\pm$ 0.0 & 11.6 $\pm$ 5.7 & 3.6 $\pm$ 1.3 & 3.4 $\pm$ 1.5 \\
 & cot & \textbf{1.0 $\pm$ 0.0} & \textbf{1.0 $\pm$ 0.0} & \textbf{1.0 $\pm$ 0.0} & 1.4 $\pm$ 0.5 & 8.8 $\pm$ 4.6 & 2.2 $\pm$ 0.8 & 4.6 $\pm$ 0.9 \\
 & spp & \textbf{1.0 $\pm$ 0.0} & 3.6 $\pm$ 5.8 & \textbf{1.0 $\pm$ 0.0} & 1.8 $\pm$ 0.4 & 4.4 $\pm$ 4.3 & 3.4 $\pm$ 1.1 & 4.0 $\pm$ 1.4 \\
 & sc-zs & \textbf{1.0 $\pm$ 0.0} & \textbf{1.0 $\pm$ 0.0} & \textbf{1.0 $\pm$ 0.0} & 2.0 $\pm$ 0.0 & 8.0 $\pm$ 8.5 & 2.5 $\pm$ 0.7 & 2.0 $\pm$ 1.4 \\
 & sc-cot & \textbf{1.0 $\pm$ 0.0} & 3.0 $\pm$ 2.8 & \textbf{1.0 $\pm$ 0.0} & 2.0 $\pm$ 0.0 & 1.5 $\pm$ 0.7 & 3.5 $\pm$ 2.1 & 3.0 $\pm$ 0.0 \\
 & sc-spp & \textbf{1.0 $\pm$ 0.0} & \textbf{1.0 $\pm$ 0.0} & 3.5 $\pm$ 3.5 & 1.5 $\pm$ 0.7 & 3.0 $\pm$ 1.4 & 1.5 $\pm$ 0.7 & 3.0 $\pm$ 0.0 \\
\cline{1-9}
\multirow[t]{6}{*}{C3.7S(T)} & zs & \textbf{1.0 $\pm$ 0.0} & 1.2 $\pm$ 0.4 & \textbf{1.0 $\pm$ 0.0} & 1.6 $\pm$ 0.5 & 12.8 $\pm$ 3.9 & 5.0 $\pm$ 1.4 & 3.2 $\pm$ 1.1 \\
 & cot & 2.2 $\pm$ 1.8 & \textbf{1.0 $\pm$ 0.0} & \textbf{1.0 $\pm$ 0.0} & 1.2 $\pm$ 0.4 & 3.2 $\pm$ 1.8 & 2.6 $\pm$ 0.9 & 2.8 $\pm$ 1.5 \\
 & spp & \textbf{1.0 $\pm$ 0.0} & 1.6 $\pm$ 1.3 & \textbf{1.0 $\pm$ 0.0} & 1.8 $\pm$ 0.4 & 5.2 $\pm$ 5.0 & 2.0 $\pm$ 1.2 & 2.8 $\pm$ 0.4 \\
 & sc-zs & \textbf{1.0 $\pm$ 0.0} & \textbf{1.0 $\pm$ 0.0} & \textbf{1.0 $\pm$ 0.0} & 2.0 $\pm$ 0.0 & 8.0 $\pm$ 8.5 & 3.0 $\pm$ 0.0 & 3.5 $\pm$ 2.1 \\
 & sc-cot & \textbf{1.0 $\pm$ 0.0} & \textbf{1.0 $\pm$ 0.0} & \textbf{1.0 $\pm$ 0.0} & 1.5 $\pm$ 0.7 & 1.5 $\pm$ 0.7 & 3.0 $\pm$ 2.8 & 2.0 $\pm$ 1.4 \\
 & sc-spp & 1.5 $\pm$ 0.7 & \textbf{1.0 $\pm$ 0.0} & \textbf{1.0 $\pm$ 0.0} & 1.5 $\pm$ 0.7 & 8.0 $\pm$ 5.7 & 3.0 $\pm$ 0.0 & 3.0 $\pm$ 0.0 \\
\cline{1-9}
\multirow[t]{6}{*}{C4S} & zs & 3.6 $\pm$ 4.8 & 2.2 $\pm$ 1.3 & 1.4 $\pm$ 0.9 & \textbf{1.2 $\pm$ 0.4} & 7.4 $\pm$ 6.3 & 2.0 $\pm$ 1.0 & 2.2 $\pm$ 1.3 \\
 & cot & \textbf{1.0 $\pm$ 0.0} & 3.0 $\pm$ 4.5 & 2.0 $\pm$ 2.2 & 1.8 $\pm$ 1.1 & 4.2 $\pm$ 5.6 & 2.2 $\pm$ 1.6 & 1.6 $\pm$ 1.3 \\
 & spp & 2.0 $\pm$ 1.4 & 2.4 $\pm$ 1.3 & \textbf{1.0 $\pm$ 0.0} & \textbf{1.0 $\pm$ 0.0} & 5.6 $\pm$ 0.9 & 3.0 $\pm$ 1.6 & 1.6 $\pm$ 1.3 \\
 & sc-zs & 2.0 $\pm$ 1.4 & \textbf{1.0 $\pm$ 0.0} & \textbf{1.0 $\pm$ 0.0} & \textbf{1.0 $\pm$ 0.0} & \textbf{1.0 $\pm$ 0.0} & \textbf{1.0 $\pm$ 0.0} & 2.0 $\pm$ 1.4 \\
 & sc-cot & \textbf{1.0 $\pm$ 0.0} & \textbf{1.0 $\pm$ 0.0} & 5.5 $\pm$ 6.4 & \textbf{1.0 $\pm$ 0.0} & \textbf{1.0 $\pm$ 0.0} & \textbf{1.0 $\pm$ 0.0} & 3.0 $\pm$ 2.8 \\
 & sc-spp & 3.5 $\pm$ 3.5 & \textbf{1.0 $\pm$ 0.0} & 3.5 $\pm$ 3.5 & \textbf{1.0 $\pm$ 0.0} & 3.5 $\pm$ 3.5 & 3.5 $\pm$ 3.5 & 3.5 $\pm$ 3.5 \\
\cline{1-9}
\multirow[t]{6}{*}{C4S(T)} & zs & 1.2 $\pm$ 0.4 & \textbf{1.0 $\pm$ 0.0} & 1.2 $\pm$ 0.4 & \textbf{1.0 $\pm$ 0.0} & 6.2 $\pm$ 7.3 & 1.6 $\pm$ 0.9 & 1.4 $\pm$ 0.9 \\
 & cot & 1.2 $\pm$ 0.4 & \textbf{1.0 $\pm$ 0.0} & \textbf{1.0 $\pm$ 0.0} & \textbf{1.0 $\pm$ 0.0} & 4.0 $\pm$ 6.7 & \textbf{1.0 $\pm$ 0.0} & \textbf{1.0 $\pm$ 0.0} \\
 & spp & 4.2 $\pm$ 5.2 & 1.6 $\pm$ 1.3 & \textbf{1.0 $\pm$ 0.0} & \textbf{1.0 $\pm$ 0.0} & 5.6 $\pm$ 5.1 & \textbf{1.0 $\pm$ 0.0} & 1.6 $\pm$ 1.3 \\
 & sc-zs & 2.0 $\pm$ 1.4 & \textbf{1.0 $\pm$ 0.0} & \textbf{1.0 $\pm$ 0.0} & \textbf{1.0 $\pm$ 0.0} & \textbf{1.0 $\pm$ 0.0} & \textbf{1.0 $\pm$ 0.0} & \textbf{1.0 $\pm$ 0.0} \\
 & sc-cot & \textbf{1.0 $\pm$ 0.0} & \textbf{1.0 $\pm$ 0.0} & \textbf{1.0 $\pm$ 0.0} & \textbf{1.0 $\pm$ 0.0} & \textbf{1.0 $\pm$ 0.0} & \textbf{1.0 $\pm$ 0.0} & \textbf{1.0 $\pm$ 0.0} \\
 & sc-spp & \textbf{1.0 $\pm$ 0.0} & \textbf{1.0 $\pm$ 0.0} & 3.5 $\pm$ 3.5 & \textbf{1.0 $\pm$ 0.0} & 8.5 $\pm$ 10.6 & \textbf{1.0 $\pm$ 0.0} & 5.0 $\pm$ 5.7 \\
\cline{1-9}
\multirow[t]{6}{*}{DS-R1} & zs & \textbf{1.0 $\pm$ 0.0} & 3.0 $\pm$ 2.8 & 1.4 $\pm$ 0.9 & 1.8 $\pm$ 1.8 & \textbf{1.0 $\pm$ 0.0} & \textbf{1.0 $\pm$ 0.0} & \textbf{1.0 $\pm$ 0.0} \\
 & cot & \textbf{1.0 $\pm$ 0.0} & \textbf{1.0 $\pm$ 0.0} & \textbf{1.0 $\pm$ 0.0} & \textbf{1.0 $\pm$ 0.0} & \textbf{1.0 $\pm$ 0.0} & \textbf{1.0 $\pm$ 0.0} & \textbf{1.0 $\pm$ 0.0} \\
 & spp & 4.2 $\pm$ 7.2 & 2.2 $\pm$ 1.8 & 2.0 $\pm$ 2.2 & \textbf{1.0 $\pm$ 0.0} & 1.6 $\pm$ 1.3 & \textbf{1.0 $\pm$ 0.0} & 4.0 $\pm$ 6.7 \\
 & sc-zs & \textbf{1.0 $\pm$ 0.0} & \textbf{1.0 $\pm$ 0.0} & \textbf{1.0 $\pm$ 0.0} & \textbf{1.0 $\pm$ 0.0} & \textbf{1.0 $\pm$ 0.0} & \textbf{1.0 $\pm$ 0.0} & \textbf{1.0 $\pm$ 0.0} \\
 & sc-cot & \textbf{1.0 $\pm$ 0.0} & \textbf{1.0 $\pm$ 0.0} & \textbf{1.0 $\pm$ 0.0} & \textbf{1.0 $\pm$ 0.0} & \textbf{1.0 $\pm$ 0.0} & \textbf{1.0 $\pm$ 0.0} & \textbf{1.0 $\pm$ 0.0} \\
 & sc-spp & \textbf{1.0 $\pm$ 0.0} & \textbf{1.0 $\pm$ 0.0} & \textbf{1.0 $\pm$ 0.0} & \textbf{1.0 $\pm$ 0.0} & \textbf{1.0 $\pm$ 0.0} & \textbf{1.0 $\pm$ 0.0} & \textbf{1.0 $\pm$ 0.0} \\
\cline{1-9}
\multirow[t]{6}{*}{L3.3-70B} & zs & \textbf{1.0 $\pm$ 0.0} & \textbf{1.0 $\pm$ 0.0} & \textbf{1.0 $\pm$ 0.0} & 6.4 $\pm$ 3.3 & 10.0 $\pm$ 5.7 & 4.2 $\pm$ 1.1 & 3.6 $\pm$ 1.8 \\
 & cot & \textbf{1.0 $\pm$ 0.0} & \textbf{1.0 $\pm$ 0.0} & \textbf{1.0 $\pm$ 0.0} & 3.8 $\pm$ 1.8 & 12.0 $\pm$ 4.5 & 2.4 $\pm$ 0.5 & 2.6 $\pm$ 0.5 \\
 & spp & \textbf{1.0 $\pm$ 0.0} & \textbf{1.0 $\pm$ 0.0} & \textbf{1.0 $\pm$ 0.0} & 3.4 $\pm$ 1.3 & 13.2 $\pm$ 5.2 & 3.0 $\pm$ 1.2 & 3.0 $\pm$ 1.2 \\
 & sc-zs & \textbf{1.0 $\pm$ 0.0} & \textbf{1.0 $\pm$ 0.0} & \textbf{1.0 $\pm$ 0.0} & 5.0 $\pm$ 1.4 & 15.0 $\pm$ 1.4 & 2.5 $\pm$ 0.7 & 9.5 $\pm$ 9.2 \\
 & sc-cot & \textbf{1.0 $\pm$ 0.0} & \textbf{1.0 $\pm$ 0.0} & \textbf{1.0 $\pm$ 0.0} & 4.0 $\pm$ 0.0 & 8.0 $\pm$ 8.5 & 2.0 $\pm$ 0.0 & 2.5 $\pm$ 0.7 \\
 & sc-spp & \textbf{1.0 $\pm$ 0.0} & \textbf{1.0 $\pm$ 0.0} & \textbf{1.0 $\pm$ 0.0} & 4.5 $\pm$ 3.5 & 10.0 $\pm$ 8.5 & 2.5 $\pm$ 0.7 & 3.0 $\pm$ 1.4 \\
\cline{1-9}
\multirow[t]{6}{*}{Mistral} & zs & 15.6 $\pm$ 1.5 & 16.2 $\pm$ 0.8 & \textbf{10.4 $\pm$ 7.7} & 16.4 $\pm$ 0.5 & 13.0 $\pm$ 6.7 & 16.6 $\pm$ 0.5 & 17.0 $\pm$ 0.0 \\
 & cot & 9.6 $\pm$ 7.9 & 7.4 $\pm$ 7.9 & 9.4 $\pm$ 7.3 & 11.4 $\pm$ 7.2 & \textbf{5.2 $\pm$ 5.4} & 13.6 $\pm$ 6.0 & 16.8 $\pm$ 0.4 \\
 & spp & 10.0 $\pm$ 7.4 & \textbf{7.2 $\pm$ 8.5} & \textbf{7.2 $\pm$ 6.5} & 14.4 $\pm$ 4.2 & 16.0 $\pm$ 0.0 & 13.8 $\pm$ 6.1 & 16.6 $\pm$ 0.5 \\
 & sc-zs & 16.0 $\pm$ 1.4 & 9.0 $\pm$ 9.9 & \textbf{8.0 $\pm$ 9.9} & 16.0 $\pm$ 0.0 & 16.0 $\pm$ 0.0 & 16.5 $\pm$ 0.7 & 16.5 $\pm$ 0.7 \\
 & sc-cot & \textbf{8.5 $\pm$ 10.6} & 10.0 $\pm$ 5.7 & 16.5 $\pm$ 0.7 & \textbf{8.5 $\pm$ 10.6} & 16.0 $\pm$ 0.0 & 11.0 $\pm$ 7.1 & 11.5 $\pm$ 6.4 \\
 & sc-spp & 16.5 $\pm$ 0.7 & \textbf{9.0 $\pm$ 11.3} & 16.5 $\pm$ 0.7 & 10.5 $\pm$ 9.2 & 16.5 $\pm$ 0.7 & 16.5 $\pm$ 0.7 & 17.0 $\pm$ 0.0 \\
\bottomrule
\end{tabular}
\caption{Round $m$\# where the LLM player understood the opponent's strategy for the joint PD counterfactual (label-based and payoff-based).}
\label{tab:pd_round_heatmap_sh-alt}
\end{table*}
Opponent comprehension results for the payoff-based PD counterfactual are presented in Table \ref{tab:pd_round_heatmap_sh}. In this case, opponent comprehension remains immediate in LLM--LLM play for the strongest models, with Claude 3.5/3.7 and Llama 3.3 almost uniformly achieving 
$m=1.0 \pm 0.0$ across the LLM-opponent columns. This suggests that when facing another LLM, these models still establish the cooperative SH regime very quickly despite the altered incentives. However, comprehension against algorithmic opponents is substantially delayed, especially for \textbf{SREP} and often \textbf{PP}, where many values move close to the game horizon ($m \approx 14–17$). This indicates that switching from defection-dominant PD incentives to coordination-based SH makes deterministic opponents harder to interpret consistently, even for otherwise strong models.

A second notable trend is that MF remains the easiest algorithmic opponent under SH for most strong models, often with 
$m$ near 1-3, while \textbf{TFT} is much more variable. Claude 4 variants remain mixed: they can still show low $m$ in LLM--LLM settings and occasionally on MF, but are less stable overall than Claude 3.5/3.7. DeepSeek R1 is uneven, with some early comprehension but frequent delays on SREP and PP, while Mistral is again the weakest and most inconsistent model, often showing mid- to late-horizon comprehension across several opponent types. Overall, the table supports the idea that the payoff change affects understanding of deterministic policies more than coordination with other LLMs, and that SH introduces a harder adaptation problem than default PD.

Finally, opponent comprehension for the joint PD counterfactual setup is demonstrated in Table \ref{tab:pd_round_heatmap_sh-alt}. In this case, opponent comprehension is again very early for the strongest models, especially in LLM--LLM interactions: Claude 3.5/3.7, Claude 4 thinking, and DeepSeek frequently achieve $m=1$ across the first three columns, showing that combining payoff and label changes does not disrupt mutual adaptation for all models equally. Among algorithmic opponents, Claude 4 variants and DeepSeek R1 adapt best overall, often keeping $m$ close to 1-2 even for SREP, MF, and TFT. Claude 3.5 is also highly stable, though its comprehension against \textbf{PP} is sometimes slightly later. In contrast, Llama 3.3 shows noticeably delayed comprehension on algorithmic opponents in this joint setting, particularly against SREP and PP, despite remaining perfect in LLM--LLM play.

The clearest outlier is again Mistral, which remains the weakest and most inconsistent model: many values lie near the horizon (
$m \approx 10–
17$), indicating frequent failure to reliably infer the opponent even after most of the game has passed. Overall, the joint counterfactual appears less disruptive than the payoff-only SH setup for the strongest models, but it still preserves a clear separation between robust models (Claude family, often DeepSeek) and brittle ones (especially Mistral).

\subsection{PD cooperation rates}
\begin{table*}[h!]
\small
\centering
\begin{tabular}{lllllllll}
\toprule
 &  & \multicolumn{7}{c}{\textbf{PD}} \\
 &  & zs & spp & cot & srep & pp & mf & tft \\
model & prompt &  &  &  &  &  &  &  \\
\midrule
\multirow[t]{6}{*}{C3.5Sv2} & zs & \textbf{1.0 $\pm$ 0.0} & \textbf{1.0 $\pm$ 0.0} & \textbf{1.0 $\pm$ 0.0} & 0.1 $\pm$ 0.0 & 0.4 $\pm$ 0.2 & 0.2 $\pm$ 0.1 & 0.2 $\pm$ 0.1 \\
 & cot & \textbf{1.0 $\pm$ 0.0} & \textbf{1.0 $\pm$ 0.0} & \textbf{1.0 $\pm$ 0.0} & 0.1 $\pm$ 0.0 & 0.5 $\pm$ 0.1 & 0.2 $\pm$ 0.1 & 0.1 $\pm$ 0.0 \\
 & spp & \textbf{1.0 $\pm$ 0.0} & \textbf{1.0 $\pm$ 0.0} & \textbf{1.0 $\pm$ 0.0} & 0.1 $\pm$ 0.0 & 0.3 $\pm$ 0.2 & 0.1 $\pm$ 0.0 & 0.2 $\pm$ 0.0 \\
 & sc-zs & \textbf{1.0 $\pm$ 0.0} & \textbf{1.0 $\pm$ 0.0} & \textbf{1.0 $\pm$ 0.0} & 0.1 $\pm$ 0.0 & 0.5 $\pm$ 0.0 & 0.2 $\pm$ 0.0 & 0.2 $\pm$ 0.0 \\
 & sc-cot & \textbf{1.0 $\pm$ 0.0} & \textbf{1.0 $\pm$ 0.0} & \textbf{1.0 $\pm$ 0.0} & 0.1 $\pm$ 0.0 & 0.4 $\pm$ 0.1 & 0.2 $\pm$ 0.0 & 0.1 $\pm$ 0.0 \\
 & sc-spp & \textbf{1.0 $\pm$ 0.0} & \textbf{1.0 $\pm$ 0.0} & \textbf{1.0 $\pm$ 0.0} & 0.1 $\pm$ 0.0 & 0.1 $\pm$ 0.0 & 0.2 $\pm$ 0.0 & 0.2 $\pm$ 0.0 \\
\cline{1-9}
\multirow[t]{6}{*}{C3.7S} & zs & \textbf{1.0 $\pm$ 0.0} & \textbf{1.0 $\pm$ 0.0} & \textbf{1.0 $\pm$ 0.0} & 0.1 $\pm$ 0.0 & 0.4 $\pm$ 0.2 & 0.2 $\pm$ 0.1 & 0.2 $\pm$ 0.1 \\
 & cot & \textbf{1.0 $\pm$ 0.0} & \textbf{1.0 $\pm$ 0.0} & \textbf{1.0 $\pm$ 0.0} & 0.1 $\pm$ 0.0 & 0.1 $\pm$ 0.0 & 0.2 $\pm$ 0.1 & 0.2 $\pm$ 0.1 \\
 & spp & \textbf{1.0 $\pm$ 0.0} & \textbf{1.0 $\pm$ 0.0} & \textbf{1.0 $\pm$ 0.0} & 0.1 $\pm$ 0.0 & 0.3 $\pm$ 0.2 & 0.2 $\pm$ 0.1 & 0.2 $\pm$ 0.1 \\
 & sc-zs & \textbf{1.0 $\pm$ 0.0} & \textbf{1.0 $\pm$ 0.0} & \textbf{1.0 $\pm$ 0.0} & 0.1 $\pm$ 0.0 & 0.5 $\pm$ 0.0 & 0.2 $\pm$ 0.0 & 0.1 $\pm$ 0.0 \\
 & sc-cot & \textbf{1.0 $\pm$ 0.0} & \textbf{1.0 $\pm$ 0.0} & \textbf{1.0 $\pm$ 0.0} & 0.1 $\pm$ 0.0 & 0.2 $\pm$ 0.0 & 0.2 $\pm$ 0.1 & 0.2 $\pm$ 0.0 \\
 & sc-spp & \textbf{1.0 $\pm$ 0.0} & \textbf{1.0 $\pm$ 0.0} & \textbf{1.0 $\pm$ 0.0} & 0.1 $\pm$ 0.0 & 0.3 $\pm$ 0.3 & 0.2 $\pm$ 0.0 & 0.2 $\pm$ 0.0 \\
\cline{1-9}
\multirow[t]{6}{*}{C3.7S(T)} & zs & \textbf{1.0 $\pm$ 0.0} & \textbf{1.0 $\pm$ 0.0} & \textbf{1.0 $\pm$ 0.0} & 0.1 $\pm$ 0.0 & 0.5 $\pm$ 0.0 & 0.2 $\pm$ 0.1 & 0.2 $\pm$ 0.0 \\
 & cot & \textbf{1.0 $\pm$ 0.0} & \textbf{1.0 $\pm$ 0.0} & \textbf{1.0 $\pm$ 0.0} & 0.1 $\pm$ 0.0 & 0.3 $\pm$ 0.1 & 0.1 $\pm$ 0.1 & 0.1 $\pm$ 0.0 \\
 & spp & \textbf{1.0 $\pm$ 0.0} & \textbf{1.0 $\pm$ 0.0} & \textbf{1.0 $\pm$ 0.0} & 0.1 $\pm$ 0.0 & 0.3 $\pm$ 0.2 & 0.2 $\pm$ 0.1 & 0.2 $\pm$ 0.1 \\
 & sc-zs & \textbf{1.0 $\pm$ 0.0} & \textbf{1.0 $\pm$ 0.0} & \textbf{1.0 $\pm$ 0.0} & 0.1 $\pm$ 0.0 & 0.5 $\pm$ 0.0 & 0.2 $\pm$ 0.0 & 0.1 $\pm$ 0.0 \\
 & sc-cot & \textbf{1.0 $\pm$ 0.0} & \textbf{1.0 $\pm$ 0.0} & \textbf{1.0 $\pm$ 0.0} & 0.1 $\pm$ 0.0 & 0.5 $\pm$ 0.0 & 0.2 $\pm$ 0.0 & 0.1 $\pm$ 0.0 \\
 & sc-spp & \textbf{1.0 $\pm$ 0.0} & \textbf{1.0 $\pm$ 0.0} & \textbf{1.0 $\pm$ 0.0} & 0.1 $\pm$ 0.0 & 0.3 $\pm$ 0.1 & 0.2 $\pm$ 0.0 & 0.2 $\pm$ 0.0 \\
\cline{1-9}
\multirow[t]{6}{*}{C4S} & zs & \textbf{0.6 $\pm$ 0.4} & 0.2 $\pm$ 0.4 & 0.2 $\pm$ 0.4 & 0.0 $\pm$ 0.0 & 0.4 $\pm$ 0.2 & 0.1 $\pm$ 0.1 & 0.1 $\pm$ 0.0 \\
 & cot & 0.1 $\pm$ 0.1 & 0.2 $\pm$ 0.0 & \textbf{0.2 $\pm$ 0.2} & 0.0 $\pm$ 0.0 & 0.2 $\pm$ 0.2 & 0.1 $\pm$ 0.0 & 0.1 $\pm$ 0.0 \\
 & spp & \textbf{0.5 $\pm$ 0.4} & 0.4 $\pm$ 0.5 & 0.1 $\pm$ 0.0 & 0.0 $\pm$ 0.0 & 0.1 $\pm$ 0.1 & 0.1 $\pm$ 0.1 & 0.1 $\pm$ 0.1 \\
 & sc-zs & 0.1 $\pm$ 0.0 & 0.0 $\pm$ 0.0 & 0.1 $\pm$ 0.1 & 0.0 $\pm$ 0.0 & \textbf{0.2 $\pm$ 0.4} & 0.0 $\pm$ 0.0 & 0.1 $\pm$ 0.1 \\
 & sc-cot & \textbf{0.1 $\pm$ 0.1} & 0.0 $\pm$ 0.0 & 0.0 $\pm$ 0.0 & 0.0 $\pm$ 0.0 & 0.0 $\pm$ 0.0 & \textbf{0.1 $\pm$ 0.0} & 0.0 $\pm$ 0.0 \\
 & sc-spp & \textbf{0.1 $\pm$ 0.1} & \textbf{0.1 $\pm$ 0.1} & 0.0 $\pm$ 0.0 & 0.1 $\pm$ 0.0 & 0.0 $\pm$ 0.0 & 0.0 $\pm$ 0.0 & 0.0 $\pm$ 0.0 \\
\cline{1-9}
\multirow[t]{6}{*}{C4S(T)} & zs & 0.0 $\pm$ 0.0 & 0.0 $\pm$ 0.1 & 0.0 $\pm$ 0.0 & 0.0 $\pm$ 0.0 & \textbf{0.0 $\pm$ 0.1} & 0.0 $\pm$ 0.0 & 0.0 $\pm$ 0.0 \\
 & cot & 0.0 $\pm$ 0.0 & \textbf{0.2 $\pm$ 0.3} & 0.0 $\pm$ 0.1 & 0.0 $\pm$ 0.0 & 0.0 $\pm$ 0.0 & 0.0 $\pm$ 0.1 & 0.0 $\pm$ 0.0 \\
 & spp & 0.0 $\pm$ 0.1 & \textbf{0.6 $\pm$ 0.4} & 0.3 $\pm$ 0.3 & 0.0 $\pm$ 0.0 & 0.2 $\pm$ 0.2 & 0.0 $\pm$ 0.1 & 0.0 $\pm$ 0.0 \\
 & sc-zs & 0.0 $\pm$ 0.0 & 0.0 $\pm$ 0.0 & 0.0 $\pm$ 0.0 & 0.0 $\pm$ 0.0 & 0.0 $\pm$ 0.0 & 0.0 $\pm$ 0.0 & 0.0 $\pm$ 0.0 \\
 & sc-cot & 0.0 $\pm$ 0.0 & \textbf{0.1 $\pm$ 0.1} & 0.0 $\pm$ 0.0 & 0.0 $\pm$ 0.0 & 0.0 $\pm$ 0.0 & 0.0 $\pm$ 0.0 & 0.0 $\pm$ 0.0 \\
 & sc-spp & \textbf{0.5 $\pm$ 0.7} & 0.0 $\pm$ 0.0 & 0.0 $\pm$ 0.0 & 0.0 $\pm$ 0.0 & 0.1 $\pm$ 0.1 & 0.0 $\pm$ 0.0 & 0.1 $\pm$ 0.1 \\
\cline{1-9}
\multirow[t]{6}{*}{DS-R1} & zs & 0.1 $\pm$ 0.0 & 0.1 $\pm$ 0.0 & \textbf{0.1 $\pm$ 0.1} & 0.0 $\pm$ 0.0 & 0.1 $\pm$ 0.1 & 0.1 $\pm$ 0.0 & 0.1 $\pm$ 0.0 \\
 & cot & 0.0 $\pm$ 0.0 & 0.0 $\pm$ 0.0 & 0.0 $\pm$ 0.0 & 0.0 $\pm$ 0.0 & 0.0 $\pm$ 0.0 & \textbf{0.0 $\pm$ 0.1} & 0.0 $\pm$ 0.0 \\
 & spp & 0.1 $\pm$ 0.1 & 0.1 $\pm$ 0.1 & 0.1 $\pm$ 0.1 & 0.1 $\pm$ 0.0 & \textbf{0.1 $\pm$ 0.2} & 0.0 $\pm$ 0.0 & 0.0 $\pm$ 0.0 \\
 & sc-zs & \textbf{0.1 $\pm$ 0.0} & 0.1 $\pm$ 0.0 & 0.0 $\pm$ 0.0 & 0.0 $\pm$ 0.0 & 0.0 $\pm$ 0.0 & 0.1 $\pm$ 0.0 & 0.0 $\pm$ 0.0 \\
 & sc-cot & \textbf{0.1 $\pm$ 0.0} & 0.0 $\pm$ 0.0 & 0.0 $\pm$ 0.0 & 0.0 $\pm$ 0.0 & 0.0 $\pm$ 0.0 & 0.0 $\pm$ 0.0 & 0.0 $\pm$ 0.0 \\
 & sc-spp & 0.1 $\pm$ 0.1 & 0.0 $\pm$ 0.0 & 0.0 $\pm$ 0.0 & 0.1 $\pm$ 0.0 & \textbf{0.2 $\pm$ 0.2} & 0.0 $\pm$ 0.0 & 0.0 $\pm$ 0.0 \\
\cline{1-9}
\multirow[t]{6}{*}{L3.3-70B} & zs & \textbf{1.0 $\pm$ 0.0} & \textbf{1.0 $\pm$ 0.0} & \textbf{1.0 $\pm$ 0.0} & 0.1 $\pm$ 0.0 & 0.5 $\pm$ 0.0 & 0.2 $\pm$ 0.0 & 0.2 $\pm$ 0.0 \\
 & cot & \textbf{1.0 $\pm$ 0.0} & \textbf{1.0 $\pm$ 0.0} & \textbf{1.0 $\pm$ 0.0} & 0.1 $\pm$ 0.0 & 0.4 $\pm$ 0.2 & 0.1 $\pm$ 0.0 & 0.1 $\pm$ 0.0 \\
 & spp & \textbf{1.0 $\pm$ 0.0} & \textbf{1.0 $\pm$ 0.0} & \textbf{1.0 $\pm$ 0.0} & 0.1 $\pm$ 0.0 & 0.5 $\pm$ 0.1 & 0.1 $\pm$ 0.0 & 0.1 $\pm$ 0.0 \\
 & sc-zs & \textbf{1.0 $\pm$ 0.0} & \textbf{1.0 $\pm$ 0.0} & \textbf{1.0 $\pm$ 0.0} & 0.2 $\pm$ 0.0 & 0.5 $\pm$ 0.0 & 0.2 $\pm$ 0.0 & 0.2 $\pm$ 0.0 \\
 & sc-cot & \textbf{1.0 $\pm$ 0.0} & \textbf{1.0 $\pm$ 0.0} & \textbf{1.0 $\pm$ 0.0} & 0.1 $\pm$ 0.0 & 0.5 $\pm$ 0.0 & 0.2 $\pm$ 0.0 & 0.2 $\pm$ 0.0 \\
 & sc-spp & \textbf{1.0 $\pm$ 0.0} & \textbf{1.0 $\pm$ 0.0} & \textbf{1.0 $\pm$ 0.0} & 0.2 $\pm$ 0.0 & 0.4 $\pm$ 0.2 & 0.1 $\pm$ 0.0 & 0.2 $\pm$ 0.0 \\
\cline{1-9}
\multirow[t]{6}{*}{Mistral} & zs & 0.8 $\pm$ 0.2 & 0.8 $\pm$ 0.2 & \textbf{0.9 $\pm$ 0.3} & 0.5 $\pm$ 0.0 & 0.5 $\pm$ 0.0 & 0.5 $\pm$ 0.1 & 0.5 $\pm$ 0.1 \\
 & cot & \textbf{1.0 $\pm$ 0.0} & 0.6 $\pm$ 0.4 & 0.7 $\pm$ 0.4 & 0.1 $\pm$ 0.1 & 0.3 $\pm$ 0.2 & 0.3 $\pm$ 0.2 & 0.5 $\pm$ 0.1 \\
 & spp & \textbf{1.0 $\pm$ 0.0} & 1.0 $\pm$ 0.1 & 0.9 $\pm$ 0.3 & 0.4 $\pm$ 0.3 & 0.5 $\pm$ 0.1 & 0.4 $\pm$ 0.2 & 0.5 $\pm$ 0.1 \\
 & sc-zs & \textbf{1.0 $\pm$ 0.0} & \textbf{1.0 $\pm$ 0.0} & \textbf{1.0 $\pm$ 0.0} & 0.5 $\pm$ 0.0 & 0.5 $\pm$ 0.0 & 0.5 $\pm$ 0.0 & 0.5 $\pm$ 0.0 \\
 & sc-cot & 0.9 $\pm$ 0.2 & \textbf{1.0 $\pm$ 0.0} & 0.7 $\pm$ 0.4 & 0.3 $\pm$ 0.3 & 0.2 $\pm$ 0.4 & 0.0 $\pm$ 0.0 & 0.4 $\pm$ 0.3 \\
 & sc-spp & \textbf{1.0 $\pm$ 0.0} & \textbf{1.0 $\pm$ 0.0} & \textbf{1.0 $\pm$ 0.0} & 0.5 $\pm$ 0.0 & 0.5 $\pm$ 0.0 & 0.8 $\pm$ 0.3 & 0.5 $\pm$ 0.0 \\
\bottomrule
\end{tabular}
\caption{Average Cooperation (Ratio of Cooperative Moves) in the default PD case. Bold denotes higher cooperation rates per LLM-prompt instance. }
\label{tab:pd_cooperation_avg_heatmap_pd}
\end{table*}
Regarding cooperation rates, the results for the default PD setup are presented in Table \ref{tab:pd_cooperation_avg_heatmap_pd}. Cooperation rates in default PD closely mirror the strategic split already seen in total points and comprehension.
In LLM–LLM interactions (first three columns), Claude 3.5, Claude 3.7, Claude 3.7 Thinking, and Llama 3.3 show perfect cooperation across all prompting types (
$\approx 1.0$), confirming robust convergence to mutual cooperation. By contrast, Claude 4, Claude 4 Thinking, and DeepSeek R1 cooperate far less, often near 0-0.2, indicating persistent distrust and over-defection even against other LLMs. Mistral is unstable: some settings remain highly cooperative, but others drop substantially, showing weaker consistency.

Against algorithmic opponents, cooperation is appropriately low for most strong models. On SREP, rates cluster around 
0.1, reflecting rapid adaptation to always defect. Against PP, cooperation rises to roughly 
0.3–0.5 for Claude 3.x and Llama 3.3, suggesting a short exploratory phase before exploiting the pattern. Cooperation against MF and TFT is again low, typically 
0.1–0.2, which is consistent with these opponents quickly inducing defection. The main exception is Mistral, whose cooperation remains unusually high against algorithmic players as well ($\sim 0.4–
0.8$ in several cases), matching its weaker strategic adaptation and slower opponent comprehension. Overall, the table confirms that stronger default-PD performers are those that cooperate reliably with other LLMs but suppress cooperation quickly against exploitative algorithmic opponents.

\begin{table*}[h!]
\small
\centering
\begin{tabular}{lllllllll}
\toprule
 &  & \multicolumn{7}{c}{\textbf{PD label-based counterfactual}} \\
 &  & zs & spp & cot & srep & pp & mf & tft \\
model & prompt &  &  &  &  &  &  &  \\
\midrule
\multirow[t]{6}{*}{C3.5Sv2} & zs & \textbf{1.0 $\pm$ 0.0} & \textbf{1.0 $\pm$ 0.0} & \textbf{1.0 $\pm$ 0.0} & 0.3 $\pm$ 0.2 & 0.5 $\pm$ 0.0 & 0.9 $\pm$ 0.0 & 0.6 $\pm$ 0.4 \\
 & cot & \textbf{1.0 $\pm$ 0.0} & \textbf{1.0 $\pm$ 0.0} & \textbf{1.0 $\pm$ 0.0} & 0.4 $\pm$ 0.1 & 0.4 $\pm$ 0.1 & 1.0 $\pm$ 0.0 & 0.6 $\pm$ 0.4 \\
 & spp & \textbf{1.0 $\pm$ 0.0} & \textbf{1.0 $\pm$ 0.0} & \textbf{1.0 $\pm$ 0.0} & 0.4 $\pm$ 0.2 & 0.3 $\pm$ 0.2 & 0.9 $\pm$ 0.0 & 0.7 $\pm$ 0.4 \\
 & sc-zs & \textbf{1.0 $\pm$ 0.0} & \textbf{1.0 $\pm$ 0.0} & \textbf{1.0 $\pm$ 0.0} & 0.3 $\pm$ 0.2 & 0.5 $\pm$ 0.0 & 0.9 $\pm$ 0.0 & \textbf{1.0 $\pm$ 0.0} \\
 & sc-cot & \textbf{1.0 $\pm$ 0.0} & \textbf{1.0 $\pm$ 0.0} & \textbf{1.0 $\pm$ 0.0} & 0.4 $\pm$ 0.0 & 0.5 $\pm$ 0.0 & 0.9 $\pm$ 0.0 & 0.7 $\pm$ 0.4 \\
 & sc-spp & \textbf{1.0 $\pm$ 0.0} & \textbf{1.0 $\pm$ 0.0} & \textbf{1.0 $\pm$ 0.0} & 0.5 $\pm$ 0.3 & 0.3 $\pm$ 0.3 & 1.0 $\pm$ 0.0 & 0.2 $\pm$ 0.0 \\
\cline{1-9}
\multirow[t]{6}{*}{C3.7S} & zs & \textbf{1.0 $\pm$ 0.0} & \textbf{1.0 $\pm$ 0.0} & \textbf{1.0 $\pm$ 0.0} & 0.4 $\pm$ 0.2 & 0.5 $\pm$ 0.0 & 1.0 $\pm$ 0.0 & 0.7 $\pm$ 0.4 \\
 & cot & \textbf{1.0 $\pm$ 0.0} & \textbf{1.0 $\pm$ 0.0} & \textbf{1.0 $\pm$ 0.0} & 0.5 $\pm$ 0.1 & 0.5 $\pm$ 0.1 & 1.0 $\pm$ 0.0 & 0.6 $\pm$ 0.4 \\
 & spp & \textbf{1.0 $\pm$ 0.0} & \textbf{1.0 $\pm$ 0.0} & \textbf{1.0 $\pm$ 0.0} & 0.4 $\pm$ 0.1 & 0.2 $\pm$ 0.1 & 1.0 $\pm$ 0.0 & 0.5 $\pm$ 0.3 \\
 & sc-zs & \textbf{1.0 $\pm$ 0.0} & \textbf{1.0 $\pm$ 0.0} & \textbf{1.0 $\pm$ 0.0} & 0.4 $\pm$ 0.3 & 0.3 $\pm$ 0.2 & 1.0 $\pm$ 0.0 & 0.3 $\pm$ 0.0 \\
 & sc-cot & \textbf{1.0 $\pm$ 0.0} & \textbf{1.0 $\pm$ 0.0} & \textbf{1.0 $\pm$ 0.0} & 0.4 $\pm$ 0.2 & 0.5 $\pm$ 0.0 & 1.0 $\pm$ 0.0 & 0.7 $\pm$ 0.5 \\
 & sc-spp & \textbf{1.0 $\pm$ 0.0} & \textbf{1.0 $\pm$ 0.0} & \textbf{1.0 $\pm$ 0.0} & 0.5 $\pm$ 0.2 & 0.3 $\pm$ 0.3 & 1.0 $\pm$ 0.0 & 0.3 $\pm$ 0.0 \\
\cline{1-9}
\multirow[t]{6}{*}{C3.7S(T)} & zs & \textbf{1.0 $\pm$ 0.0} & \textbf{1.0 $\pm$ 0.0} & \textbf{1.0 $\pm$ 0.0} & 0.5 $\pm$ 0.2 & 0.4 $\pm$ 0.1 & 0.7 $\pm$ 0.4 & 0.4 $\pm$ 0.3 \\
 & cot & \textbf{1.0 $\pm$ 0.0} & 0.9 $\pm$ 0.1 & \textbf{1.0 $\pm$ 0.0} & 0.5 $\pm$ 0.1 & 0.4 $\pm$ 0.1 & \textbf{1.0 $\pm$ 0.0} & 0.7 $\pm$ 0.4 \\
 & spp & \textbf{1.0 $\pm$ 0.0} & \textbf{1.0 $\pm$ 0.0} & \textbf{1.0 $\pm$ 0.0} & 0.3 $\pm$ 0.1 & 0.4 $\pm$ 0.2 & 1.0 $\pm$ 0.0 & 0.7 $\pm$ 0.4 \\
 & sc-zs & \textbf{1.0 $\pm$ 0.0} & \textbf{1.0 $\pm$ 0.0} & \textbf{1.0 $\pm$ 0.0} & 0.7 $\pm$ 0.1 & 0.4 $\pm$ 0.2 & 0.6 $\pm$ 0.5 & \textbf{1.0 $\pm$ 0.0} \\
 & sc-cot & \textbf{1.0 $\pm$ 0.0} & \textbf{1.0 $\pm$ 0.0} & \textbf{1.0 $\pm$ 0.0} & 0.4 $\pm$ 0.0 & 0.5 $\pm$ 0.0 & 0.9 $\pm$ 0.0 & 0.2 $\pm$ 0.0 \\
 & sc-spp & \textbf{1.0 $\pm$ 0.0} & \textbf{1.0 $\pm$ 0.0} & \textbf{1.0 $\pm$ 0.0} & 0.4 $\pm$ 0.3 & 0.3 $\pm$ 0.1 & 1.0 $\pm$ 0.0 & 0.6 $\pm$ 0.6 \\
\cline{1-9}
\multirow[t]{6}{*}{C4S} & zs & 0.6 $\pm$ 0.5 & \textbf{1.0 $\pm$ 0.0} & 0.8 $\pm$ 0.4 & 0.4 $\pm$ 0.1 & 0.2 $\pm$ 0.2 & 0.6 $\pm$ 0.4 & 0.4 $\pm$ 0.4 \\
 & cot & \textbf{0.9 $\pm$ 0.1} & 0.6 $\pm$ 0.4 & 0.5 $\pm$ 0.4 & 0.2 $\pm$ 0.1 & 0.2 $\pm$ 0.2 & 0.3 $\pm$ 0.4 & 0.3 $\pm$ 0.4 \\
 & spp & \textbf{0.8 $\pm$ 0.4} & 0.3 $\pm$ 0.4 & 0.3 $\pm$ 0.4 & 0.2 $\pm$ 0.1 & 0.3 $\pm$ 0.2 & 0.6 $\pm$ 0.5 & 0.3 $\pm$ 0.4 \\
 & sc-zs & \textbf{1.0 $\pm$ 0.0} & 0.9 $\pm$ 0.1 & 0.9 $\pm$ 0.0 & 0.2 $\pm$ 0.1 & 0.1 $\pm$ 0.0 & 0.9 $\pm$ 0.0 & 0.2 $\pm$ 0.1 \\
 & sc-cot & \textbf{0.5 $\pm$ 0.6} & 0.2 $\pm$ 0.0 & 0.2 $\pm$ 0.0 & 0.3 $\pm$ 0.1 & 0.2 $\pm$ 0.0 & \textbf{0.5 $\pm$ 0.7} & 0.2 $\pm$ 0.0 \\
 & sc-spp & 1.0 $\pm$ 0.0 & 0.9 $\pm$ 0.1 & 0.5 $\pm$ 0.5 & 0.3 $\pm$ 0.1 & 0.2 $\pm$ 0.2 & \textbf{1.0 $\pm$ 0.0} & 0.5 $\pm$ 0.4 \\
\cline{1-9}
\multirow[t]{6}{*}{C4S(T)} & zs & \textbf{1.0 $\pm$ 0.0} & 0.8 $\pm$ 0.4 & 0.8 $\pm$ 0.4 & 0.4 $\pm$ 0.1 & 0.3 $\pm$ 0.1 & 0.8 $\pm$ 0.4 & 0.5 $\pm$ 0.5 \\
 & cot & 0.4 $\pm$ 0.5 & 0.6 $\pm$ 0.5 & 0.7 $\pm$ 0.5 & 0.3 $\pm$ 0.2 & 0.2 $\pm$ 0.1 & 0.4 $\pm$ 0.5 & \textbf{0.7 $\pm$ 0.4} \\
 & spp & 0.8 $\pm$ 0.4 & 0.6 $\pm$ 0.5 & 0.8 $\pm$ 0.4 & 0.3 $\pm$ 0.1 & 0.3 $\pm$ 0.2 & \textbf{0.9 $\pm$ 0.0} & 0.8 $\pm$ 0.4 \\
 & sc-zs & \textbf{1.0 $\pm$ 0.0} & \textbf{1.0 $\pm$ 0.0} & 1.0 $\pm$ 0.0 & 0.7 $\pm$ 0.0 & 0.5 $\pm$ 0.0 & 0.1 $\pm$ 0.1 & 0.2 $\pm$ 0.1 \\
 & sc-cot & 0.5 $\pm$ 0.7 & 0.5 $\pm$ 0.7 & \textbf{1.0 $\pm$ 0.0} & 0.1 $\pm$ 0.1 & 0.0 $\pm$ 0.0 & 0.5 $\pm$ 0.7 & 0.6 $\pm$ 0.6 \\
 & sc-spp & \textbf{1.0 $\pm$ 0.0} & 0.5 $\pm$ 0.6 & 0.5 $\pm$ 0.7 & 0.3 $\pm$ 0.0 & 0.2 $\pm$ 0.0 & 0.5 $\pm$ 0.7 & 0.1 $\pm$ 0.1 \\
\cline{1-9}
\multirow[t]{6}{*}{DS-R1} & zs & 0.3 $\pm$ 0.4 & 0.1 $\pm$ 0.0 & 0.1 $\pm$ 0.1 & 0.3 $\pm$ 0.1 & 0.2 $\pm$ 0.2 & \textbf{0.4 $\pm$ 0.5} & 0.3 $\pm$ 0.4 \\
 & cot & 0.1 $\pm$ 0.1 & 0.1 $\pm$ 0.0 & 0.2 $\pm$ 0.4 & \textbf{0.3 $\pm$ 0.2} & 0.1 $\pm$ 0.1 & 0.1 $\pm$ 0.1 & 0.2 $\pm$ 0.4 \\
 & spp & 0.1 $\pm$ 0.1 & 0.1 $\pm$ 0.2 & 0.3 $\pm$ 0.4 & 0.2 $\pm$ 0.1 & 0.2 $\pm$ 0.1 & 0.2 $\pm$ 0.4 & \textbf{0.3 $\pm$ 0.4} \\
 & sc-zs & 0.0 $\pm$ 0.0 & 0.1 $\pm$ 0.1 & 0.1 $\pm$ 0.1 & 0.0 $\pm$ 0.0 & \textbf{0.2 $\pm$ 0.2} & 0.1 $\pm$ 0.0 & 0.1 $\pm$ 0.1 \\
 & sc-cot & \textbf{0.5 $\pm$ 0.7} & 0.1 $\pm$ 0.1 & 0.1 $\pm$ 0.1 & 0.3 $\pm$ 0.2 & 0.1 $\pm$ 0.0 & 0.5 $\pm$ 0.7 & \textbf{0.5 $\pm$ 0.7} \\
 & sc-spp & 0.0 $\pm$ 0.0 & 0.0 $\pm$ 0.0 & 0.1 $\pm$ 0.0 & \textbf{0.2 $\pm$ 0.3} & 0.1 $\pm$ 0.1 & 0.1 $\pm$ 0.0 & 0.1 $\pm$ 0.1 \\
\cline{1-9}
\multirow[t]{6}{*}{L3.3-70B} & zs & \textbf{1.0 $\pm$ 0.0} & \textbf{1.0 $\pm$ 0.0} & \textbf{1.0 $\pm$ 0.0} & 0.3 $\pm$ 0.2 & 0.5 $\pm$ 0.0 & 0.7 $\pm$ 0.4 & 0.3 $\pm$ 0.4 \\
 & cot & \textbf{1.0 $\pm$ 0.0} & \textbf{1.0 $\pm$ 0.0} & \textbf{1.0 $\pm$ 0.0} & 0.4 $\pm$ 0.1 & 0.3 $\pm$ 0.2 & 0.8 $\pm$ 0.4 & 0.3 $\pm$ 0.4 \\
 & spp & \textbf{1.0 $\pm$ 0.0} & \textbf{1.0 $\pm$ 0.0} & \textbf{1.0 $\pm$ 0.0} & 0.3 $\pm$ 0.2 & 0.4 $\pm$ 0.2 & 1.0 $\pm$ 0.0 & 0.8 $\pm$ 0.4 \\
 & sc-zs & \textbf{1.0 $\pm$ 0.0} & \textbf{1.0 $\pm$ 0.0} & \textbf{1.0 $\pm$ 0.0} & 0.2 $\pm$ 0.1 & 0.5 $\pm$ 0.0 & \textbf{1.0 $\pm$ 0.0} & \textbf{1.0 $\pm$ 0.0} \\
 & sc-cot & \textbf{1.0 $\pm$ 0.0} & \textbf{1.0 $\pm$ 0.0} & \textbf{1.0 $\pm$ 0.0} & 0.3 $\pm$ 0.1 & 0.1 $\pm$ 0.0 & \textbf{1.0 $\pm$ 0.0} & 0.6 $\pm$ 0.6 \\
 & sc-spp & \textbf{1.0 $\pm$ 0.0} & \textbf{1.0 $\pm$ 0.0} & \textbf{1.0 $\pm$ 0.0} & 0.4 $\pm$ 0.0 & 0.5 $\pm$ 0.0 & 0.6 $\pm$ 0.6 & \textbf{1.0 $\pm$ 0.0} \\
\cline{1-9}
\multirow[t]{6}{*}{Mistral} & zs & \textbf{1.0 $\pm$ 0.0} & 0.9 $\pm$ 0.2 & \textbf{1.0 $\pm$ 0.0} & 0.6 $\pm$ 0.1 & 0.5 $\pm$ 0.0 & 0.9 $\pm$ 0.2 & 0.9 $\pm$ 0.3 \\
 & cot & 0.9 $\pm$ 0.1 & 0.9 $\pm$ 0.2 & \textbf{1.0 $\pm$ 0.0} & 0.5 $\pm$ 0.1 & 0.4 $\pm$ 0.2 & 0.8 $\pm$ 0.3 & 0.8 $\pm$ 0.3 \\
 & spp & \textbf{1.0 $\pm$ 0.0} & 0.7 $\pm$ 0.4 & 1.0 $\pm$ 0.0 & 0.6 $\pm$ 0.1 & 0.5 $\pm$ 0.1 & 0.9 $\pm$ 0.2 & 0.6 $\pm$ 0.4 \\
 & sc-zs & \textbf{1.0 $\pm$ 0.0} & \textbf{1.0 $\pm$ 0.0} & 0.9 $\pm$ 0.1 & 0.6 $\pm$ 0.0 & 0.5 $\pm$ 0.0 & 1.0 $\pm$ 0.0 & \textbf{1.0 $\pm$ 0.0} \\
 & sc-cot & \textbf{1.0 $\pm$ 0.0} & \textbf{1.0 $\pm$ 0.0} & \textbf{1.0 $\pm$ 0.0} & 0.4 $\pm$ 0.2 & 0.5 $\pm$ 0.0 & 1.0 $\pm$ 0.0 & 0.8 $\pm$ 0.4 \\
 & sc-spp & \textbf{1.0 $\pm$ 0.0} & 0.8 $\pm$ 0.3 & \textbf{1.0 $\pm$ 0.0} & 0.6 $\pm$ 0.1 & 0.5 $\pm$ 0.0 & 0.9 $\pm$ 0.0 & 0.7 $\pm$ 0.3 \\
\bottomrule
\end{tabular}
\caption{Average Cooperation (Ratio of Cooperative Moves) in the label-based PD counterfactual setup.}
\label{tab:pd_cooperation_avg_heatmap_pd-alt}
\end{table*}

Table \ref{tab:pd_cooperation_avg_heatmap_pd-alt} presents the cooperation rate results for the label-based PD counterfactual. The label-based PD counterfactual produces a clear shift in cooperation patterns relative to default PD. In LLM–LLM interactions, the strongest models—Claude 3.5, Claude 3.7, Claude 3.7 Thinking, and Llama 3.3—remain almost perfectly cooperative, showing that simple relabeling does not substantially disrupt mutual coordination for stronger models. Mistral is also much more cooperative here than in the default setup, often approaching 1.0, which suggests that the renamed actions may actually encourage more socially aligned behavior in weaker models. By contrast, DeepSeek R1 remains the least cooperative, with rates mostly around 0.0–0.3, indicating that label shifts do not resolve its general tendency toward defection or instability.

Against algorithmic opponents, cooperation becomes much higher than in default PD for many models. In particular, cooperation against MF often rises to 0.9–1.0 for Claude 3.x, Llama 3.3, and Mistral, while TFT also elicits moderate-to-high cooperation in many settings. Even against SREP and PP, rates are frequently around 0.3–0.5, much higher than the near-zero cooperation observed in the default game. This suggests that the label substitution weakens immediate recognition of the underlying incentive structure: models often preserve cooperative behavior under renamed actions instead of quickly switching to the more exploitative default-PD response. Overall, the table supports the claim that label-only counterfactuals do not strongly harm strong LLM–LLM cooperation, but they do alter adaptation to algorithmic opponents by making models more cooperative and therefore less tightly aligned with the original PD incentives.

\begin{table*}[h!]
\small
\centering
\begin{tabular}{lllllllll}
\toprule
 &  & \multicolumn{7}{c}{\textbf{SH (payoff-based counterfactual of PD)}} \\
 &  & zs & spp & cot & srep & pp & mf & tft \\
model & prompt &  &  &  &  &  &  &  \\
\midrule
\multirow[t]{6}{*}{C3.5Sv2} & zs & \textbf{1.0 $\pm$ 0.0} & \textbf{1.0 $\pm$ 0.0} & \textbf{1.0 $\pm$ 0.0} & 0.3 $\pm$ 0.1 & 0.3 $\pm$ 0.2 & 1.0 $\pm$ 0.0 & 0.5 $\pm$ 0.5 \\
 & cot & \textbf{1.0 $\pm$ 0.0} & \textbf{1.0 $\pm$ 0.0} & \textbf{1.0 $\pm$ 0.0} & 0.2 $\pm$ 0.1 & 0.3 $\pm$ 0.1 & 0.5 $\pm$ 0.5 & 0.3 $\pm$ 0.4 \\
 & spp & \textbf{1.0 $\pm$ 0.0} & \textbf{1.0 $\pm$ 0.0} & \textbf{1.0 $\pm$ 0.0} & 0.1 $\pm$ 0.0 & 0.2 $\pm$ 0.1 & 0.8 $\pm$ 0.4 & 0.8 $\pm$ 0.4 \\
 & sc-zs & \textbf{1.0 $\pm$ 0.0} & \textbf{1.0 $\pm$ 0.0} & \textbf{1.0 $\pm$ 0.0} & 0.1 $\pm$ 0.0 & 0.3 $\pm$ 0.3 & 1.0 $\pm$ 0.0 & \textbf{1.0 $\pm$ 0.0} \\
 & sc-cot & \textbf{1.0 $\pm$ 0.0} & \textbf{1.0 $\pm$ 0.0} & \textbf{1.0 $\pm$ 0.0} & 0.2 $\pm$ 0.1 & 0.3 $\pm$ 0.2 & 0.1 $\pm$ 0.0 & 0.6 $\pm$ 0.6 \\
 & sc-spp & \textbf{1.0 $\pm$ 0.0} & \textbf{1.0 $\pm$ 0.0} & \textbf{1.0 $\pm$ 0.0} & 0.3 $\pm$ 0.0 & 0.1 $\pm$ 0.0 & 0.6 $\pm$ 0.6 & 0.6 $\pm$ 0.6 \\
\cline{1-9}
\multirow[t]{6}{*}{C3.7S} & zs & \textbf{1.0 $\pm$ 0.0} & \textbf{1.0 $\pm$ 0.0} & \textbf{1.0 $\pm$ 0.0} & 0.5 $\pm$ 0.2 & 0.5 $\pm$ 0.1 & 1.0 $\pm$ 0.0 & 0.7 $\pm$ 0.4 \\
 & cot & \textbf{1.0 $\pm$ 0.0} & \textbf{1.0 $\pm$ 0.0} & \textbf{1.0 $\pm$ 0.0} & 0.4 $\pm$ 0.2 & 0.5 $\pm$ 0.0 & 1.0 $\pm$ 0.0 & 0.7 $\pm$ 0.4 \\
 & spp & \textbf{1.0 $\pm$ 0.0} & \textbf{1.0 $\pm$ 0.0} & \textbf{1.0 $\pm$ 0.0} & 0.3 $\pm$ 0.1 & 0.4 $\pm$ 0.2 & 1.0 $\pm$ 0.0 & 0.6 $\pm$ 0.4 \\
 & sc-zs & \textbf{1.0 $\pm$ 0.0} & \textbf{1.0 $\pm$ 0.0} & \textbf{1.0 $\pm$ 0.0} & 0.4 $\pm$ 0.0 & 0.4 $\pm$ 0.2 & 0.9 $\pm$ 0.0 & 0.6 $\pm$ 0.6 \\
 & sc-cot & \textbf{1.0 $\pm$ 0.0} & \textbf{1.0 $\pm$ 0.0} & \textbf{1.0 $\pm$ 0.0} & 0.4 $\pm$ 0.0 & 0.5 $\pm$ 0.0 & 1.0 $\pm$ 0.0 & 0.2 $\pm$ 0.1 \\
 & sc-spp & \textbf{1.0 $\pm$ 0.0} & \textbf{1.0 $\pm$ 0.0} & \textbf{1.0 $\pm$ 0.0} & 0.3 $\pm$ 0.0 & 0.3 $\pm$ 0.3 & 1.0 $\pm$ 0.0 & \textbf{1.0 $\pm$ 0.0} \\
\cline{1-9}
\multirow[t]{6}{*}{C3.7S(T)} & zs & \textbf{1.0 $\pm$ 0.0} & \textbf{1.0 $\pm$ 0.0} & \textbf{1.0 $\pm$ 0.0} & 0.5 $\pm$ 0.1 & 0.4 $\pm$ 0.2 & 0.8 $\pm$ 0.4 & 0.5 $\pm$ 0.4 \\
 & cot & \textbf{1.0 $\pm$ 0.0} & \textbf{1.0 $\pm$ 0.0} & \textbf{1.0 $\pm$ 0.0} & 0.3 $\pm$ 0.1 & 0.4 $\pm$ 0.2 & 1.0 $\pm$ 0.0 & 0.5 $\pm$ 0.5 \\
 & spp & \textbf{1.0 $\pm$ 0.0} & \textbf{1.0 $\pm$ 0.0} & \textbf{1.0 $\pm$ 0.0} & 0.2 $\pm$ 0.1 & 0.5 $\pm$ 0.0 & 1.0 $\pm$ 0.0 & 0.6 $\pm$ 0.4 \\
 & sc-zs & \textbf{1.0 $\pm$ 0.0} & \textbf{1.0 $\pm$ 0.0} & \textbf{1.0 $\pm$ 0.0} & 0.3 $\pm$ 0.2 & 0.4 $\pm$ 0.2 & 1.0 $\pm$ 0.0 & 0.2 $\pm$ 0.1 \\
 & sc-cot & \textbf{1.0 $\pm$ 0.0} & \textbf{1.0 $\pm$ 0.0} & \textbf{1.0 $\pm$ 0.0} & 0.2 $\pm$ 0.1 & 0.5 $\pm$ 0.0 & \textbf{1.0 $\pm$ 0.0} & 0.6 $\pm$ 0.6 \\
 & sc-spp & \textbf{1.0 $\pm$ 0.0} & \textbf{1.0 $\pm$ 0.0} & \textbf{1.0 $\pm$ 0.0} & 0.3 $\pm$ 0.3 & 0.3 $\pm$ 0.3 & \textbf{1.0 $\pm$ 0.0} & 0.2 $\pm$ 0.0 \\
\cline{1-9}
\multirow[t]{6}{*}{C4S} & zs & \textbf{0.8 $\pm$ 0.3} & 0.6 $\pm$ 0.4 & 0.7 $\pm$ 0.4 & 0.2 $\pm$ 0.1 & 0.3 $\pm$ 0.2 & 0.6 $\pm$ 0.5 & 0.5 $\pm$ 0.4 \\
 & cot & 0.2 $\pm$ 0.4 & 0.3 $\pm$ 0.4 & \textbf{0.4 $\pm$ 0.5} & 0.3 $\pm$ 0.1 & 0.3 $\pm$ 0.2 & 0.2 $\pm$ 0.1 & 0.2 $\pm$ 0.2 \\
 & spp & 0.3 $\pm$ 0.4 & 0.1 $\pm$ 0.1 & 0.6 $\pm$ 0.3 & 0.3 $\pm$ 0.1 & 0.4 $\pm$ 0.1 & 0.3 $\pm$ 0.4 & \textbf{0.9 $\pm$ 0.0} \\
 & sc-zs & \textbf{1.0 $\pm$ 0.0} & \textbf{1.0 $\pm$ 0.0} & 0.6 $\pm$ 0.5 & 0.2 $\pm$ 0.0 & 0.5 $\pm$ 0.0 & 1.0 $\pm$ 0.0 & 0.6 $\pm$ 0.6 \\
 & sc-cot & 0.2 $\pm$ 0.2 & 0.4 $\pm$ 0.6 & \textbf{0.5 $\pm$ 0.4} & 0.0 $\pm$ 0.0 & \textbf{0.5 $\pm$ 0.0} & 0.1 $\pm$ 0.1 & 0.2 $\pm$ 0.0 \\
 & sc-spp & 0.6 $\pm$ 0.6 & 0.1 $\pm$ 0.1 & \textbf{0.6 $\pm$ 0.5} & 0.4 $\pm$ 0.2 & 0.2 $\pm$ 0.1 & 0.6 $\pm$ 0.5 & \textbf{0.6 $\pm$ 0.5} \\
\cline{1-9}
\multirow[t]{6}{*}{C4S(T)} & zs & 0.6 $\pm$ 0.5 & \textbf{0.8 $\pm$ 0.3} & 0.6 $\pm$ 0.4 & 0.2 $\pm$ 0.2 & 0.3 $\pm$ 0.2 & 0.4 $\pm$ 0.5 & 0.3 $\pm$ 0.4 \\
 & cot & 0.6 $\pm$ 0.5 & \textbf{0.8 $\pm$ 0.3} & 0.2 $\pm$ 0.4 & 0.4 $\pm$ 0.2 & 0.2 $\pm$ 0.1 & 0.3 $\pm$ 0.4 & 0.5 $\pm$ 0.3 \\
 & spp & 0.8 $\pm$ 0.4 & \textbf{0.9 $\pm$ 0.1} & 0.7 $\pm$ 0.4 & 0.3 $\pm$ 0.1 & 0.3 $\pm$ 0.2 & 0.8 $\pm$ 0.4 & 0.4 $\pm$ 0.4 \\
 & sc-zs & 0.5 $\pm$ 0.6 & 0.5 $\pm$ 0.5 & 1.0 $\pm$ 0.0 & 0.2 $\pm$ 0.1 & 0.5 $\pm$ 0.0 & \textbf{1.0 $\pm$ 0.0} & 0.2 $\pm$ 0.0 \\
 & sc-cot & 0.2 $\pm$ 0.0 & \textbf{1.0 $\pm$ 0.0} & 0.5 $\pm$ 0.7 & 0.2 $\pm$ 0.2 & 0.3 $\pm$ 0.0 & 0.6 $\pm$ 0.5 & 0.5 $\pm$ 0.7 \\
 & sc-spp & \textbf{0.9 $\pm$ 0.0} & 0.8 $\pm$ 0.1 & 0.6 $\pm$ 0.5 & 0.3 $\pm$ 0.0 & 0.5 $\pm$ 0.0 & 0.2 $\pm$ 0.0 & 0.1 $\pm$ 0.2 \\
\cline{1-9}
\multirow[t]{6}{*}{DS-R1} & zs & 0.2 $\pm$ 0.4 & \textbf{0.3 $\pm$ 0.4} & 0.1 $\pm$ 0.0 & 0.2 $\pm$ 0.2 & 0.1 $\pm$ 0.1 & 0.3 $\pm$ 0.4 & 0.1 $\pm$ 0.1 \\
 & cot & \textbf{0.2 $\pm$ 0.4} & 0.0 $\pm$ 0.0 & 0.0 $\pm$ 0.0 & 0.1 $\pm$ 0.1 & 0.2 $\pm$ 0.1 & 0.1 $\pm$ 0.0 & 0.1 $\pm$ 0.1 \\
 & spp & 0.2 $\pm$ 0.4 & 0.1 $\pm$ 0.0 & \textbf{0.4 $\pm$ 0.5} & 0.1 $\pm$ 0.1 & 0.1 $\pm$ 0.1 & 0.2 $\pm$ 0.4 & 0.1 $\pm$ 0.0 \\
 & sc-zs & 0.0 $\pm$ 0.0 & \textbf{0.5 $\pm$ 0.5} & 0.1 $\pm$ 0.0 & 0.2 $\pm$ 0.1 & 0.1 $\pm$ 0.0 & 0.1 $\pm$ 0.0 & 0.0 $\pm$ 0.0 \\
 & sc-cot & 0.0 $\pm$ 0.0 & 0.1 $\pm$ 0.0 & \textbf{0.1 $\pm$ 0.0} & 0.1 $\pm$ 0.1 & \textbf{0.1 $\pm$ 0.1} & \textbf{0.1 $\pm$ 0.0} & \textbf{0.1 $\pm$ 0.0} \\
 & sc-spp & 0.0 $\pm$ 0.0 & 0.1 $\pm$ 0.0 & 0.1 $\pm$ 0.0 & \textbf{0.1 $\pm$ 0.0} & 0.0 $\pm$ 0.0 & 0.0 $\pm$ 0.0 & 0.0 $\pm$ 0.0 \\
\cline{1-9}
\multirow[t]{6}{*}{L3.3-70B} & zs & \textbf{1.0 $\pm$ 0.0} & \textbf{1.0 $\pm$ 0.0} & \textbf{1.0 $\pm$ 0.0} & 0.5 $\pm$ 0.1 & 0.5 $\pm$ 0.0 & 0.8 $\pm$ 0.4 & 0.5 $\pm$ 0.3 \\
 & cot & 0.9 $\pm$ 0.2 & \textbf{1.0 $\pm$ 0.0} & \textbf{1.0 $\pm$ 0.0} & 0.4 $\pm$ 0.2 & 0.3 $\pm$ 0.1 & 1.0 $\pm$ 0.0 & 0.8 $\pm$ 0.4 \\
 & spp & \textbf{1.0 $\pm$ 0.0} & \textbf{1.0 $\pm$ 0.0} & \textbf{1.0 $\pm$ 0.0} & 0.4 $\pm$ 0.1 & 0.5 $\pm$ 0.0 & 0.7 $\pm$ 0.4 & 0.7 $\pm$ 0.4 \\
 & sc-zs & \textbf{1.0 $\pm$ 0.0} & \textbf{1.0 $\pm$ 0.0} & \textbf{1.0 $\pm$ 0.0} & 0.4 $\pm$ 0.1 & 0.5 $\pm$ 0.0 & \textbf{1.0 $\pm$ 0.0} & \textbf{1.0 $\pm$ 0.0} \\
 & sc-cot & \textbf{1.0 $\pm$ 0.0} & \textbf{1.0 $\pm$ 0.0} & \textbf{1.0 $\pm$ 0.0} & 0.3 $\pm$ 0.2 & 0.2 $\pm$ 0.0 & 0.6 $\pm$ 0.6 & 0.6 $\pm$ 0.6 \\
 & sc-spp & \textbf{1.0 $\pm$ 0.0} & \textbf{1.0 $\pm$ 0.0} & \textbf{1.0 $\pm$ 0.0} & 0.4 $\pm$ 0.4 & 0.5 $\pm$ 0.0 & \textbf{1.0 $\pm$ 0.0} & \textbf{1.0 $\pm$ 0.0} \\
\cline{1-9}
\multirow[t]{6}{*}{Mistral} & zs & 0.7 $\pm$ 0.2 & 0.9 $\pm$ 0.2 & 0.7 $\pm$ 0.2 & 0.6 $\pm$ 0.1 & 0.5 $\pm$ 0.0 & \textbf{0.9 $\pm$ 0.1} & 0.8 $\pm$ 0.3 \\
 & cot & 0.6 $\pm$ 0.4 & \textbf{0.8 $\pm$ 0.3} & 0.5 $\pm$ 0.4 & 0.5 $\pm$ 0.2 & 0.5 $\pm$ 0.0 & 0.4 $\pm$ 0.3 & 0.5 $\pm$ 0.3 \\
 & spp & 0.7 $\pm$ 0.3 & 0.7 $\pm$ 0.4 & 0.6 $\pm$ 0.3 & 0.5 $\pm$ 0.1 & 0.5 $\pm$ 0.1 & \textbf{0.8 $\pm$ 0.2} & 0.7 $\pm$ 0.3 \\
 & sc-zs & \textbf{1.0 $\pm$ 0.0} & 0.8 $\pm$ 0.3 & 0.8 $\pm$ 0.2 & 0.6 $\pm$ 0.0 & 0.5 $\pm$ 0.0 & 1.0 $\pm$ 0.0 & 0.4 $\pm$ 0.1 \\
 & sc-cot & 0.7 $\pm$ 0.4 & 0.4 $\pm$ 0.5 & 0.2 $\pm$ 0.3 & 0.4 $\pm$ 0.3 & 0.5 $\pm$ 0.0 & \textbf{0.7 $\pm$ 0.4} & 0.2 $\pm$ 0.2 \\
 & sc-spp & 0.2 $\pm$ 0.4 & 0.8 $\pm$ 0.3 & 0.5 $\pm$ 0.0 & 0.6 $\pm$ 0.1 & 0.5 $\pm$ 0.0 & \textbf{1.0 $\pm$ 0.0} & 0.7 $\pm$ 0.4 \\
\bottomrule
\end{tabular}
\caption{Average Cooperation (Ratio of Cooperative Moves) in the Stag Hunt counterfactual.}
\label{tab:pd_cooperation_avg_heatmap_sh}
\end{table*}
Table \ref{tab:pd_cooperation_avg_heatmap_sh} presents the cooperation rate results for the payoff-based PD counterfactual. The payoff-based counterfactual successfully shifts many models toward more cooperative behavior, but this shift is highly model-dependent.

The clearest pattern is that Claude 3.5, Claude 3.7, Claude 3.7(T), and Llama 3.3 remain strongly cooperative in LLM–LLM interactions, almost always staying at or near 1.0 cooperation across ZS, CoT, SPP, and SC. This suggests that these models adapt well to the SH coordination incentives and consistently select the socially efficient action when paired with other cooperative LLMs. They also often show very high cooperation against MF and sometimes against TFT, indicating that once the opponent appears coordination-friendly, these models sustain cooperative play.

By contrast, Claude 4 variants are much more mixed. Their cooperation is often moderate rather than maximal, and strongly depends on the prompting variant. Self-consistency sometimes helps substantially—for example, pushing some LLM–LLM cases or MF cases close to full cooperation—but overall these models appear less stable and more cautious than the Claude 3.x models. DeepSeek R1 is the least cooperative among the stronger models: cooperation is mostly low across opponent types, with only occasional moderate increases under specific prompts, suggesting persistent reluctance to move toward SH-style coordination. Mistral is notable in a different way: unlike in default PD, it often displays moderate to high cooperation in SH, especially in LLM–LLM settings and against MF, but remains highly variable across prompts and opponents.

Across algorithmic opponents, cooperation is no longer uniformly low as in default PD. Against SREP and PP, many models now show intermediate cooperation levels (roughly 0.2–0.5, sometimes higher), which is consistent with SH no longer rewarding universal defection. Against MF, several strong models frequently reach near-perfect cooperation, while TFT produces more variable outcomes, reflecting the difficulty of sustaining reciprocal coordination over time. Overall, the table supports the main claim that the SH payoff change induces a real behavioral shift from defection toward coordination, with the strongest evidence coming from Claude 3.x and Llama 3.3, while Claude 4, DeepSeek, and Mistral remain more unstable in how fully they exploit the new incentive structure.

\begin{table*}[h!]
\small
\centering
\begin{tabular}{lllllllll}
\toprule
 &  & \multicolumn{7}{c}{\textbf{Joint counterfactual of PD}} \\
 &  & zs & spp & cot & srep & pp & mf & tft \\
model & prompt &  &  &  &  &  &  &  \\
\midrule
\multirow[t]{6}{*}{C3.5Sv2} & zs & \textbf{1.0 $\pm$ 0.0} & \textbf{1.0 $\pm$ 0.0} & \textbf{1.0 $\pm$ 0.0} & 0.1 $\pm$ 0.0 & 0.2 $\pm$ 0.2 & 0.2 $\pm$ 0.0 & 0.1 $\pm$ 0.1 \\
 & cot & \textbf{1.0 $\pm$ 0.0} & \textbf{1.0 $\pm$ 0.0} & \textbf{1.0 $\pm$ 0.0} & 0.1 $\pm$ 0.0 & 0.3 $\pm$ 0.1 & 0.1 $\pm$ 0.1 & 0.2 $\pm$ 0.0 \\
 & spp & \textbf{1.0 $\pm$ 0.0} & \textbf{1.0 $\pm$ 0.0} & \textbf{1.0 $\pm$ 0.0} & 0.1 $\pm$ 0.0 & 0.1 $\pm$ 0.0 & 0.2 $\pm$ 0.0 & 0.1 $\pm$ 0.0 \\
 & sc-zs & \textbf{1.0 $\pm$ 0.0} & \textbf{1.0 $\pm$ 0.0} & \textbf{1.0 $\pm$ 0.0} & 0.1 $\pm$ 0.0 & 0.2 $\pm$ 0.0 & 0.2 $\pm$ 0.0 & 0.2 $\pm$ 0.0 \\
 & sc-cot & \textbf{1.0 $\pm$ 0.0} & \textbf{1.0 $\pm$ 0.0} & \textbf{1.0 $\pm$ 0.0} & 0.1 $\pm$ 0.0 & 0.2 $\pm$ 0.1 & 0.1 $\pm$ 0.1 & 0.2 $\pm$ 0.0 \\
 & sc-spp & \textbf{1.0 $\pm$ 0.0} & \textbf{1.0 $\pm$ 0.0} & \textbf{1.0 $\pm$ 0.0} & 0.1 $\pm$ 0.0 & 0.2 $\pm$ 0.1 & 0.2 $\pm$ 0.0 & 0.1 $\pm$ 0.1 \\
 \cline{1-9}
\multirow[t]{6}{*}{C3.7S} & zs & \textbf{1.0 $\pm$ 0.0} & \textbf{1.0 $\pm$ 0.0} & \textbf{1.0 $\pm$ 0.0} & 0.1 $\pm$ 0.0 & 0.4 $\pm$ 0.2 & 0.2 $\pm$ 0.1 & 0.2 $\pm$ 0.1 \\
 & cot & \textbf{1.0 $\pm$ 0.0} & \textbf{1.0 $\pm$ 0.0} & \textbf{1.0 $\pm$ 0.0} & 0.1 $\pm$ 0.0 & 0.4 $\pm$ 0.2 & 0.1 $\pm$ 0.1 & 0.2 $\pm$ 0.0 \\
 & spp & \textbf{1.0 $\pm$ 0.0} & 0.8 $\pm$ 0.3 & 1.0 $\pm$ 0.0 & 0.1 $\pm$ 0.0 & 0.2 $\pm$ 0.1 & 0.1 $\pm$ 0.0 & 0.2 $\pm$ 0.1 \\
 & sc-zs & \textbf{1.0 $\pm$ 0.0} & \textbf{1.0 $\pm$ 0.0} & 0.9 $\pm$ 0.1 & 0.1 $\pm$ 0.0 & 0.3 $\pm$ 0.3 & 0.2 $\pm$ 0.0 & 0.1 $\pm$ 0.1 \\
 & sc-cot & \textbf{1.0 $\pm$ 0.0} & 0.6 $\pm$ 0.6 & \textbf{1.0 $\pm$ 0.0} & 0.1 $\pm$ 0.0 & 0.3 $\pm$ 0.3 & 0.2 $\pm$ 0.1 & 0.2 $\pm$ 0.0 \\
 & sc-spp & \textbf{1.0 $\pm$ 0.0} & \textbf{1.0 $\pm$ 0.0} & 0.6 $\pm$ 0.6 & 0.1 $\pm$ 0.0 & 0.2 $\pm$ 0.1 & 0.1 $\pm$ 0.0 & 0.2 $\pm$ 0.0 \\
\cline{1-9}
\multirow[t]{6}{*}{C3.7S(T)} & zs & \textbf{1.0 $\pm$ 0.0} & 0.8 $\pm$ 0.4 & \textbf{1.0 $\pm$ 0.0} & 0.1 $\pm$ 0.0 & 0.5 $\pm$ 0.1 & 0.2 $\pm$ 0.0 & 0.2 $\pm$ 0.0 \\
 & cot & 0.7 $\pm$ 0.5 & \textbf{1.0 $\pm$ 0.0} & \textbf{1.0 $\pm$ 0.0} & 0.1 $\pm$ 0.0 & 0.2 $\pm$ 0.1 & 0.2 $\pm$ 0.1 & 0.2 $\pm$ 0.1 \\
 & spp & \textbf{1.0 $\pm$ 0.1} & 0.8 $\pm$ 0.4 & \textbf{1.0 $\pm$ 0.1} & 0.1 $\pm$ 0.0 & 0.2 $\pm$ 0.1 & 0.1 $\pm$ 0.1 & 0.2 $\pm$ 0.0 \\
 & sc-zs & \textbf{1.0 $\pm$ 0.0} & \textbf{1.0 $\pm$ 0.0} & \textbf{1.0 $\pm$ 0.0} & 0.1 $\pm$ 0.0 & 0.3 $\pm$ 0.3 & 0.2 $\pm$ 0.0 & 0.2 $\pm$ 0.1 \\
 & sc-cot & \textbf{1.0 $\pm$ 0.0} & \textbf{1.0 $\pm$ 0.0} & \textbf{1.0 $\pm$ 0.0} & 0.1 $\pm$ 0.0 & 0.1 $\pm$ 0.1 & 0.2 $\pm$ 0.1 & 0.1 $\pm$ 0.1 \\
 & sc-spp & 0.9 $\pm$ 0.1 & \textbf{1.0 $\pm$ 0.0} & \textbf{1.0 $\pm$ 0.0} & 0.1 $\pm$ 0.0 & 0.3 $\pm$ 0.2 & 0.2 $\pm$ 0.0 & 0.2 $\pm$ 0.0 \\
\cline{1-9}
\multirow[t]{6}{*}{C4S} & zs & 0.3 $\pm$ 0.4 & 0.1 $\pm$ 0.1 & \textbf{0.3 $\pm$ 0.4} & 0.1 $\pm$ 0.1 & 0.3 $\pm$ 0.2 & 0.1 $\pm$ 0.1 & 0.1 $\pm$ 0.1 \\
 & cot & 0.0 $\pm$ 0.0 & \textbf{0.3 $\pm$ 0.4} & 0.0 $\pm$ 0.1 & 0.1 $\pm$ 0.0 & 0.2 $\pm$ 0.1 & 0.1 $\pm$ 0.1 & 0.1 $\pm$ 0.1 \\
 & spp & 0.1 $\pm$ 0.1 & 0.3 $\pm$ 0.3 & 0.2 $\pm$ 0.4 & 0.0 $\pm$ 0.0 & \textbf{0.3 $\pm$ 0.1} & 0.1 $\pm$ 0.0 & 0.1 $\pm$ 0.1 \\
 & sc-zs & \textbf{0.1 $\pm$ 0.1} & \textbf{0.1 $\pm$ 0.0} & 0.0 $\pm$ 0.0 & 0.0 $\pm$ 0.0 & 0.0 $\pm$ 0.0 & 0.0 $\pm$ 0.0 & \textbf{0.1 $\pm$ 0.1} \\
 & sc-cot & 0.0 $\pm$ 0.0 & 0.0 $\pm$ 0.0 & \textbf{0.1 $\pm$ 0.0} & 0.0 $\pm$ 0.0 & 0.0 $\pm$ 0.0 & 0.0 $\pm$ 0.0 & 0.1 $\pm$ 0.1 \\
 & sc-spp & 0.1 $\pm$ 0.1 & 0.0 $\pm$ 0.0 & 0.1 $\pm$ 0.0 & 0.0 $\pm$ 0.0 & \textbf{0.2 $\pm$ 0.1} & 0.1 $\pm$ 0.1 & 0.1 $\pm$ 0.1 \\
\cline{1-9}
\multirow[t]{6}{*}{C4S(T)} & zs & 0.0 $\pm$ 0.1 & \textbf{0.2 $\pm$ 0.4} & 0.1 $\pm$ 0.1 & 0.0 $\pm$ 0.0 & 0.2 $\pm$ 0.2 & 0.1 $\pm$ 0.1 & 0.1 $\pm$ 0.1 \\
 & cot & 0.0 $\pm$ 0.1 & 0.0 $\pm$ 0.0 & 0.0 $\pm$ 0.0 & 0.0 $\pm$ 0.0 & \textbf{0.1 $\pm$ 0.2} & 0.0 $\pm$ 0.0 & 0.0 $\pm$ 0.0 \\
 & spp & 0.1 $\pm$ 0.1 & \textbf{0.2 $\pm$ 0.4} & 0.0 $\pm$ 0.0 & 0.1 $\pm$ 0.0 & 0.2 $\pm$ 0.2 & 0.0 $\pm$ 0.0 & 0.1 $\pm$ 0.1 \\
 & sc-zs & 0.1 $\pm$ 0.0 & 0.0 $\pm$ 0.0 & \textbf{0.5 $\pm$ 0.7} & 0.1 $\pm$ 0.0 & 0.0 $\pm$ 0.0 & 0.0 $\pm$ 0.0 & 0.0 $\pm$ 0.0 \\
 & sc-cot & \textbf{0.1 $\pm$ 0.1} & 0.0 $\pm$ 0.0 & 0.0 $\pm$ 0.0 & 0.0 $\pm$ 0.0 & 0.0 $\pm$ 0.0 & 0.0 $\pm$ 0.0 & 0.0 $\pm$ 0.0 \\
 & sc-spp & 0.0 $\pm$ 0.0 & 0.0 $\pm$ 0.0 & 0.1 $\pm$ 0.2 & 0.0 $\pm$ 0.0 & \textbf{0.2 $\pm$ 0.4} & 0.0 $\pm$ 0.0 & 0.1 $\pm$ 0.1 \\
\cline{1-9}
\multirow[t]{6}{*}{DS-R1} & zs & 0.1 $\pm$ 0.0 & \textbf{0.1 $\pm$ 0.1} & 0.1 $\pm$ 0.1 & 0.1 $\pm$ 0.0 & 0.1 $\pm$ 0.1 & 0.1 $\pm$ 0.0 & 0.1 $\pm$ 0.0 \\
 & cot & 0.1 $\pm$ 0.0 & 0.0 $\pm$ 0.0 & 0.1 $\pm$ 0.0 & 0.0 $\pm$ 0.0 & \textbf{0.1 $\pm$ 0.1} & 0.1 $\pm$ 0.0 & 0.0 $\pm$ 0.0 \\
 & spp & 0.0 $\pm$ 0.1 & \textbf{0.1 $\pm$ 0.1} & 0.1 $\pm$ 0.1 & 0.1 $\pm$ 0.0 & 0.0 $\pm$ 0.1 & 0.0 $\pm$ 0.0 & 0.1 $\pm$ 0.0 \\
 & sc-zs & 0.0 $\pm$ 0.0& 0.0 $\pm$ 0.0 & 0.0 $\pm$ 0.0 & 0.0 $\pm$ 0.0 & 0.0 $\pm$ 0.0 & 0.0 $\pm$ 0.0 & 0.0 $\pm$ 0.0 \\
 & sc-cot & 0.0 $\pm$ 0.0 & 0.0 $\pm$ 0.0 & 0.0 $\pm$ 0.0 & 0.0 $\pm$ 0.0 & 0.0 $\pm$ 0.0 & 0.0 $\pm$ 0.0 & 0.0 $\pm$ 0.0 \\
 & sc-spp & 0.0 $\pm$ 0.0 & 0.0 $\pm$ 0.0 & 0.0 $\pm$ 0.0 & 0.0 $\pm$ 0.0 & 0.0 $\pm$ 0.0 & 0.0 $\pm$ 0.0 & 0.0 $\pm$ 0.0 \\
\cline{1-9}
\multirow[t]{6}{*}{L3.3-70B} & zs & \textbf{1.0 $\pm$ 0.0} & \textbf{1.0 $\pm$ 0.0} & \textbf{1.0 $\pm$ 0.0} & 0.2 $\pm$ 0.1 & 0.4 $\pm$ 0.1 & 0.2 $\pm$ 0.0 & 0.2 $\pm$ 0.1 \\
 & cot & \textbf{1.0 $\pm$ 0.0} & \textbf{1.0 $\pm$ 0.0} & \textbf{1.0 $\pm$ 0.0} & 0.2 $\pm$ 0.1 & 0.5 $\pm$ 0.1 & 0.1 $\pm$ 0.0 & 0.2 $\pm$ 0.0 \\
 & spp & \textbf{1.0 $\pm$ 0.0} & \textbf{1.0 $\pm$ 0.0} & \textbf{1.0 $\pm$ 0.0} & 0.2 $\pm$ 0.0 & 0.5 $\pm$ 0.1 & 0.2 $\pm$ 0.1 & 0.2 $\pm$ 0.1 \\
 & sc-zs & \textbf{1.0 $\pm$ 0.0} & \textbf{1.0 $\pm$ 0.0} & \textbf{1.0 $\pm$ 0.0} & 0.2 $\pm$ 0.0 & 0.5 $\pm$ 0.0 & 0.2 $\pm$ 0.0 & 0.3 $\pm$ 0.2 \\
 & sc-cot & \textbf{1.0 $\pm$ 0.0} & \textbf{1.0 $\pm$ 0.0} & \textbf{1.0 $\pm$ 0.0} & 0.2 $\pm$ 0.0 & 0.3 $\pm$ 0.2 & 0.1 $\pm$ 0.0 & 0.2 $\pm$ 0.0 \\
 & sc-spp & \textbf{1.0 $\pm$ 0.0} & \textbf{1.0 $\pm$ 0.0} & \textbf{1.0 $\pm$ 0.0} & 0.2 $\pm$ 0.1 & 0.4 $\pm$ 0.2 & 0.2 $\pm$ 0.0 & 0.2 $\pm$ 0.0 \\
\cline{1-9}
\multirow[t]{6}{*}{Mistral} & zs & \textbf{0.7 $\pm$ 0.2} & 0.6 $\pm$ 0.2 & 0.5 $\pm$ 0.0 & 0.5 $\pm$ 0.0 & 0.5 $\pm$ 0.0 & 0.5 $\pm$ 0.0 & 0.5 $\pm$ 0.0 \\
 & cot & \textbf{0.5 $\pm$ 0.2} & 0.4 $\pm$ 0.2 & 0.4 $\pm$ 0.3 & 0.3 $\pm$ 0.2 & 0.3 $\pm$ 0.2 & 0.4 $\pm$ 0.1 & 0.3 $\pm$ 0.2 \\
 & spp & 0.7 $\pm$ 0.1 & 0.6 $\pm$ 0.3 & \textbf{0.7 $\pm$ 0.2} & 0.5 $\pm$ 0.3 & 0.6 $\pm$ 0.0 & 0.5 $\pm$ 0.2 & 0.5 $\pm$ 0.0 \\
 & sc-zs & 0.7 $\pm$ 0.1 & \textbf{0.9 $\pm$ 0.0} & \textbf{0.9 $\pm$ 0.2} & 0.5 $\pm$ 0.0 & 0.5 $\pm$ 0.0 & 0.5 $\pm$ 0.0 & 0.5 $\pm$ 0.0 \\
 & sc-cot & 0.4 $\pm$ 0.1 & 0.5 $\pm$ 0.0 & \textbf{0.6 $\pm$ 0.0} & 0.2 $\pm$ 0.4 & 0.4 $\pm$ 0.1 & 0.3 $\pm$ 0.1 & 0.3 $\pm$ 0.0 \\
 & sc-spp & 0.6 $\pm$ 0.1 & \textbf{0.7 $\pm$ 0.3} & 0.6 $\pm$ 0.1 & 0.4 $\pm$ 0.3 & 0.5 $\pm$ 0.0 & 0.5 $\pm$ 0.0 & 0.4 $\pm$ 0.0 \\
\bottomrule
\end{tabular}
\caption{Average Cooperation (Ratio of Cooperative Moves) in the joint counterfactual setup.}
\label{tab:pd_cooperation_avg_heatmap_sh-alt}
\end{table*}
Finally, in Table \ref{tab:pd_cooperation_avg_heatmap_sh} we demonstrate the cooperation rate results for the joint PD counterfactual. The joint PD counterfactual setup behaves much more like the default PD regime than like SH: for most strong models, cooperation remains maximal in LLM–LLM play but stays low against algorithmic opponents, indicating that the label+payoff intervention did not induce the broad coordination shift observed in the pure SH payoff counterfactual.

The clearest pattern is that Claude 3.5, Claude 3.7, Claude 3.7(T), and Llama 3.3 still cooperate almost perfectly ($\approx 1.0$) against LLM counterparts across nearly all prompting variants. This suggests that mutual cooperation remains easy to sustain in same-model interactions even under the joint reformulation. However, against SREP, PP, MF, and TFT, these same models mostly drop back to low cooperation rates (typically around 0.1–0.3, with only occasional values around 0.4–0.5), which is much closer to the default PD pattern than to the SH payoff-based counterfactual. In other words, once both labels and payoffs are altered together, these models do not generalize the cooperative SH behavior as strongly as they do when only payoffs are changed.

\subsection{Efficiency in PD}
\begin{table*}[h!]
\centering \small
\begin{tabular}{llllll}
\toprule
 &  & \textbf{PD} & \textbf{Label-based} & \textbf{Payoff-based} & \textbf{Joint} \\
model & prompt &  &  &  &  \\
\midrule
\multirow[t]{6}{*}{C3.5Sv2} & zs & 18.35 $\pm$ 10.24 & \textbf{28.04 $\pm$ 12.12} & 21.84 $\pm$ 12.24 & 15.56 $\pm$ 9.72 \\
 & cot & 9.00 $\pm$ 4.05 & \textbf{15.88 $\pm$ 8.80} & 11.25 $\pm$ 5.37 & 7.85 $\pm$ 2.92 \\
 & spp & 7.54 $\pm$ 2.98 & \textbf{11.99 $\pm$ 4.73} & 9.60 $\pm$ 4.14 & 5.93 $\pm$ 2.41 \\
 & sc-zs & 6.37 $\pm$ 3.42 & \textbf{9.88 $\pm$ 4.56} & 7.63 $\pm$ 3.60 & 4.59 $\pm$ 2.71 \\
 & sc-cot & 2.96 $\pm$ 1.46 & \textbf{6.06 $\pm$ 4.01} & 5.50 $\pm$ 5.26 & 2.58 $\pm$ 1.00 \\
 & sc-spp & 2.54 $\pm$ 1.09 & 3.81 $\pm$ 1.63 & \textbf{4.70 $\pm$ 4.25} & 1.96 $\pm$ 0.76 \\
\cline{1-6}
\multirow[t]{6}{*}{C3.7S} & zs & 14.03 $\pm$ 7.88 & \textbf{24.35 $\pm$ 12.64} & 24.04 $\pm$ 12.10 & 13.16 $\pm$ 7.68 \\
 & cot & 6.39 $\pm$ 2.64 & 10.14 $\pm$ 4.40 & \textbf{10.68 $\pm$ 4.72} & 6.17 $\pm$ 2.67 \\
 & spp & 5.48 $\pm$ 3.66 & \textbf{7.89 $\pm$ 5.63} & 7.22 $\pm$ 3.47 & 4.57 $\pm$ 3.10 \\
 & sc-zs & 4.88 $\pm$ 2.49 & 7.02 $\pm$ 4.64 & \textbf{7.70 $\pm$ 4.79} & 4.37 $\pm$ 2.50 \\
 & sc-cot & 2.34 $\pm$ 1.32 & \textbf{3.63 $\pm$ 1.78} & 3.15 $\pm$ 1.60 & 2.01 $\pm$ 1.07 \\
 & sc-spp & 2.36 $\pm$ 1.37 & \textbf{3.21 $\pm$ 2.72} & 2.67 $\pm$ 1.01 & 1.51 $\pm$ 0.69 \\
\cline{1-6}
\multirow[t]{6}{*}{C3.7S(T)} & zs & 15.67 $\pm$ 6.23 & 22.93 $\pm$ 10.25 & \textbf{30.01 $\pm$ 15.17} & 19.17 $\pm$ 9.07 \\
 & cot & 8.44 $\pm$ 3.39 & \textbf{13.73 $\pm$ 5.86} & 13.68 $\pm$ 6.63 & 7.86 $\pm$ 3.05 \\
 & spp & 6.82 $\pm$ 2.85 & \textbf{10.99 $\pm$ 4.98} & 9.87 $\pm$ 4.58 & 6.10 $\pm$ 2.53 \\
 & sc-zs & 5.35 $\pm$ 2.15 & 9.53 $\pm$ 4.24 & \textbf{9.88 $\pm$ 5.87} & 6.90 $\pm$ 2.89 \\
 & sc-cot & 2.67 $\pm$ 1.19 & 4.16 $\pm$ 2.16 & \textbf{4.45 $\pm$ 2.13} & 2.71 $\pm$ 1.05 \\
 & sc-spp & 2.21 $\pm$ 0.86 & 3.45 $\pm$ 1.66 & \textbf{3.49 $\pm$ 1.84} & 2.08 $\pm$ 0.91 \\
\cline{1-6}
\multirow[t]{6}{*}{C4S} & zs & 7.45 $\pm$ 3.99 & 13.85 $\pm$ 8.83 & \textbf{15.12 $\pm$ 12.56} & 7.92 $\pm$ 4.45 \\
 & cot & 4.15 $\pm$ 0.81 & \textbf{6.34 $\pm$ 3.79} & 4.53 $\pm$ 2.42 & 4.19 $\pm$ 0.82 \\
 & spp & 3.85 $\pm$ 1.26 & \textbf{4.49 $\pm$ 2.59} & 4.46 $\pm$ 2.23 & 3.45 $\pm$ 0.81 \\
 & sc-zs & 2.15 $\pm$ 0.48 & 4.86 $\pm$ 3.10 & \textbf{5.45 $\pm$ 3.33} & 1.99 $\pm$ 0.41 \\
 & sc-cot & 1.36 $\pm$ 0.18 & 1.50 $\pm$ 0.94 & \textbf{1.51 $\pm$ 0.47} & 1.41 $\pm$ 0.25 \\
 & sc-spp & 1.13 $\pm$ 0.35 & \textbf{2.01 $\pm$ 0.99} & 1.36 $\pm$ 0.81 & 1.04 $\pm$ 0.31 \\
\cline{1-6}
\multirow[t]{6}{*}{C4S(T)} & zs & 11.59 $\pm$ 3.05 & \textbf{24.90 $\pm$ 12.96} & 19.88 $\pm$ 11.98 & 11.66 $\pm$ 3.63 \\
 & cot & 5.84 $\pm$ 1.51 & \textbf{9.81 $\pm$ 6.49} & 8.43 $\pm$ 5.02 & 5.37 $\pm$ 1.18 \\
 & spp & 3.99 $\pm$ 1.62 & \textbf{6.68 $\pm$ 2.97} & 6.04 $\pm$ 2.96 & 3.50 $\pm$ 0.87 \\
 & sc-zs & 3.99 $\pm$ 1.00 & \textbf{7.91 $\pm$ 3.90} & 7.51 $\pm$ 3.45 & 4.19 $\pm$ 1.89 \\
 & sc-cot & 1.88 $\pm$ 0.38 & \textbf{3.43 $\pm$ 2.19} & 3.13 $\pm$ 2.20 & 1.94 $\pm$ 0.54 \\
 & sc-spp & 1.32 $\pm$ 0.36 & \textbf{1.80 $\pm$ 0.90} & 1.68 $\pm$ 0.87 & 1.07 $\pm$ 0.28 \\
\cline{1-6}
\multirow[t]{6}{*}{DS-R1} & zs & \textbf{13.86 $\pm$ 5.08} & 12.59 $\pm$ 5.53 & 13.47 $\pm$ 6.02 & 12.38 $\pm$ 4.01 \\
 & cot & 8.48 $\pm$ 5.19 & 9.71 $\pm$ 4.00 & 11.77 $\pm$ 3.26 & \textbf{12.48 $\pm$ 3.23} \\
 & spp & 8.22 $\pm$ 3.06 & \textbf{8.79 $\pm$ 5.23} & 8.51 $\pm$ 4.02 & 8.68 $\pm$ 2.39 \\
 & sc-zs & 4.37 $\pm$ 1.43 & 3.66 $\pm$ 0.95 & \textbf{4.72 $\pm$ 2.32} & 4.51 $\pm$ 1.36 \\
 & sc-cot & 2.74 $\pm$ 1.31 & \textbf{4.70 $\pm$ 2.15} & 4.41 $\pm$ 0.99 & 4.39 $\pm$ 1.34 \\
 & sc-spp & 2.69 $\pm$ 0.86 & 2.29 $\pm$ 0.92 & 2.40 $\pm$ 0.37 & \textbf{3.02 $\pm$ 1.29} \\
\cline{1-6}
\multirow[t]{6}{*}{L3.3-70B} & zs & 25.03 $\pm$ 9.53 & 38.08 $\pm$ 17.58 & \textbf{40.90 $\pm$ 17.03} & 25.09 $\pm$ 9.76 \\
 & cot & 18.62 $\pm$ 7.48 & 28.85 $\pm$ 13.32 & \textbf{30.11 $\pm$ 12.13} & 17.57 $\pm$ 6.97 \\
 & spp & 16.46 $\pm$ 6.41 & \textbf{27.37 $\pm$ 9.98} & 25.53 $\pm$ 10.79 & 16.63 $\pm$ 6.44 \\
 & sc-zs & 8.37 $\pm$ 3.35 & \textbf{15.04 $\pm$ 5.46} & 14.98 $\pm$ 5.11 & 8.50 $\pm$ 3.24 \\
 & sc-cot & 6.22 $\pm$ 2.55 & \textbf{10.28 $\pm$ 4.55} & 9.55 $\pm$ 4.55 & 5.68 $\pm$ 2.40 \\
 & sc-spp & 5.50 $\pm$ 2.14 & 8.98 $\pm$ 3.58 & \textbf{9.78 $\pm$ 3.30} & 5.62 $\pm$ 2.20 \\
\cline{1-6}
\multirow[t]{6}{*}{Mistral} & zs & 27.04 $\pm$ 11.01 & \textbf{47.46 $\pm$ 18.77} & 39.23 $\pm$ 17.11 & 23.17 $\pm$ 8.92 \\
 & cot & 6.07 $\pm$ 3.11 & \textbf{9.60 $\pm$ 4.27} & 7.18 $\pm$ 3.49 & 5.57 $\pm$ 2.45 \\
 & spp & 4.51 $\pm$ 2.62 & \textbf{7.15 $\pm$ 4.36} & 6.01 $\pm$ 4.00 & 4.45 $\pm$ 2.33 \\
 & sc-zs & 9.65 $\pm$ 4.33 & \textbf{16.86 $\pm$ 5.59} & 13.38 $\pm$ 5.80 & 8.68 $\pm$ 3.49 \\
 & sc-cot & 2.19 $\pm$ 0.84 & \textbf{3.52 $\pm$ 1.21} & 2.14 $\pm$ 1.12 & 1.72 $\pm$ 0.57 \\
 & sc-spp & 1.84 $\pm$ 1.08 & \textbf{2.71 $\pm$ 1.55} & 2.10 $\pm$ 1.08 & 1.61 $\pm$ 1.01 \\
\bottomrule
\end{tabular}
\caption{Average Efficiency (Points per kilo-token) for all PD setups.}
\label{tab:pd_efficiency_avg_heatmap}
\end{table*}

Table \ref{tab:pd_efficiency_avg_heatmap} exhibits the efficiency metric over the default and the three counterfactual PD settings. A concise takeaway is that efficiency does not perfectly track raw capability. Some of the strongest models in total points are not the most efficient, while some lighter or less reliable models achieve high points-per-token because they use fewer tokens or behave more tersely.

The most notable result is that Llama 3.3 70B is the strongest overall efficiency performer among the robust models. It posts the highest or near-highest efficiency values across many settings, especially in pd-alt and sh, suggesting a very favorable trade-off between strategic performance and token cost. This supports the broader claim that Llama is not always the strongest on absolute strategic sophistication, but it is often one of the best cost-performance choices.

Among the Claude family, Claude 3.5 Sonnet v2, Claude 3.7 Sonnet, and Claude 3.7 Sonnet (Thinking) remain reasonably efficient, but efficiency drops substantially once one moves from ZS to CoT/SPP, and drops again under SC. This shows that extra reasoning and majority-vote aggregation often add token cost faster than they improve payoff. In particular, the thinking variants do not yield proportional efficiency gains: even when they improve robustness or points in some settings, their extra deliberation often lowers points-per-token.

Claude Sonnet 4 and Claude Sonnet 4 Thinking are clearly less efficient overall than Claude 3.5/3.7 and Llama. Their values are consistently lower, especially under SC, which fits the earlier observation that these models often over-reason or behave cautiously without enough performance gain to justify the added token cost. So although Claude 4 can be competitive in some raw outcomes, it appears less attractive from an efficiency standpoint.

DeepSeek R1 is interesting because it is fairly competitive in efficiency despite mixed strategic quality. Its values are often moderate and stable, and in a few cases it attains the best score in a row. This suggests that DeepSeek’s weaker strategic consistency is partly offset by lower token usage. Still, its efficiency advantage should be interpreted carefully, since it does not necessarily reflect stronger reasoning.

Finally, Mistral shows the clearest “efficiency can be misleading” pattern. It achieves some of the highest efficiency scores in the table, especially in pd-alt and under some SC settings, but earlier results showed that it is also one of the most brittle models overall. This likely means that Mistral is efficient because it is relatively cheap in tokens, not because it is consistently the best strategic reasoner. In that sense, Mistral looks like a high-variance, high-efficiency option, whereas Llama is the more convincing balanced winner.

Overall, the table reinforces three points: (i) zero-shot is usually the most efficient prompting strategy, (ii) CoT/SPP and especially SC often reduce efficiency unless they deliver a substantial behavioral gain, and (iii) the best trade-off between capability and efficiency appears to come from Llama 3.3 70B, while Claude 4 variants are the clearest examples of stronger models whose extra reasoning is not matched by proportional efficiency gains.

\subsection{PD validity rate}
\begin{table}[h!]
\centering \small
\begin{tabular}{ll}
\toprule
model & avg validity rate\\
\midrule
C3.5Sv2 & 100.0 $\pm$ 0.0 \\
C3.7S & 100.0 $\pm$ 0.0 \\
C3.7S(T) & 100.0 $\pm$ 0.0 \\
C4S & 100.0 $\pm$ 0.0 \\
C4S(T) & 100.0 $\pm$ 0.0 \\
DS-R1 & 99.1 $\pm$ 6.3 \\
L3.3-70B & 100.0 $\pm$ 0.0 \\
Mistral  & 99.4 $\pm$ 6.5 \\
\bottomrule
\end{tabular}
\caption{Average Valid Rate (\% of Valid Outcomes)}
\label{tab:pd_valid_rates}
\end{table}

We present validity rates for PD in Table \ref{tab:pd_valid_rates}. Validity in PD is effectively perfect for nearly all models, indicating that formatting compliance and action-space adherence do not meaningfully affect the behavioral results. Claude 3.5 Sonnet v2, Claude 3.7, Claude 3.7(T), Claude 4, Claude 4(T), and Llama 3.3 all achieve a strict 
$100.0\%\pm 0.0$ validity rate, showing completely stable instruction following in this simpler two-action setting. The only minor deviations come from DeepSeek R1 ($99.1\%\pm 6.3$) and Mistral ($
99.4\%\pm 6.5$), whose averages remain very high but whose larger variance suggests occasional formatting or parsing failures. Overall, invalid outputs are too rare to explain any of the observed differences in total points, opponent comprehension, cooperation, or efficiency. Instead, the PD results should be interpreted as reflecting genuine strategic differences across models, with Mistral and DeepSeek showing slightly weaker output reliability in addition to their broader behavioral instability.

\subsection{RPS total points}
\begin{table*}[h!]
\small
\centering
\begin{tabular}{p{1cm}p{0.9cm}p{1.6cm}p{1.6cm}p{1.6cm}p{1.4cm}p{1.5cm}p{1.5cm}p{1.5cm}}
\toprule
 &  & \multicolumn{7}{c}{\textbf{RPS}} \\
 &  & zs & spp & cot & srep & pp & ap & tft \\
model & prompt &  &  &  &  &  &  &  \\
\midrule
\multirow[t]{6}{*}{C3.5Sv2} & zs & 6.6 $\pm$ 13.5 & -1.2 $\pm$ 9.2 & -1.8 $\pm$ 6.9 & -0.4 $\pm$ 3.3 & 6.0 $\pm$ 10.1 & 11.2 $\pm$ 6.4 & \textbf{16.6 $\pm$ 6.9} \\
 & cot & 0.8 $\pm$ 5.4 & 2.0 $\pm$ 5.7 & 5.4 $\pm$ 4.8 & 0.2 $\pm$ 5.3 & 3.8 $\pm$ 11.9 & 6.0 $\pm$ 3.7 & \textbf{13.0 $\pm$ 6.8} \\
 & spp & 8.0 $\pm$ 9.9 & 7.2 $\pm$ 11.6 & 1.6 $\pm$ 6.9 & -1.4 $\pm$ 2.2 & 9.4 $\pm$ 11.2 & 6.8 $\pm$ 1.5 & \textbf{12.8 $\pm$ 7.6} \\
 & sc-zs & -12.5 $\pm$ 12.0 & -0.5 $\pm$ 21.9 & 0.5 $\pm$ 4.9 & -0.5 $\pm$ 9.2 & 1.5 $\pm$ 2.1 & 9.0 $\pm$ 1.4 & \textbf{21.5 $\pm$ 0.7} \\
 & sc-cot & 9.0 $\pm$ 0.0 & -1.0 $\pm$ 2.8 & 0.0 $\pm$ 7.1 & 1.0 $\pm$ 4.2 & 10.5 $\pm$ 14.8 & 14.0 $\pm$ 12.7 & \textbf{16.5 $\pm$ 2.1} \\
 & sc-spp & -2.5 $\pm$ 21.9 & -9.0 $\pm$ 17.0 & 0.0 $\pm$ 9.9 & 7.0 $\pm$ 1.4 & -1.5 $\pm$ 3.5 & 5.0 $\pm$ 5.7 & \textbf{12.5 $\pm$ 14.8} \\
\cline{1-9}
\multirow[t]{6}{*}{C3.7S} & zs & 1.4 $\pm$ 9.6 & -4.6 $\pm$ 11.3 & -3.0 $\pm$ 11.9 & -2.0 $\pm$ 5.0 & \textbf{13.6 $\pm$ 12.5} & 8.8 $\pm$ 2.6 & 4.8 $\pm$ 7.1 \\
 & cot & 9.6 $\pm$ 5.2 & 2.6 $\pm$ 9.8 & 0.4 $\pm$ 5.5 & 1.4 $\pm$ 2.1 & 19.6 $\pm$ 1.3 & 8.2 $\pm$ 1.5 & \textbf{20.2 $\pm$ 3.6} \\
 & spp & 3.6 $\pm$ 9.6 & 4.8 $\pm$ 6.0 & -5.2 $\pm$ 8.6 & -2.4 $\pm$ 2.3 & \textbf{19.2 $\pm$ 1.3} & 14.2 $\pm$ 7.2 & 19.0 $\pm$ 4.8 \\
 & sc-zs & -2.0 $\pm$ 28.3 & 0.0 $\pm$ 7.1 & -15.5 $\pm$ 9.2 & -4.0 $\pm$ 2.8 & \textbf{14.0 $\pm$ 14.1} & 6.5 $\pm$ 0.7 & 0.0 $\pm$ 0.0 \\
 & sc-cot & 20.5 $\pm$ 4.9 & -4.5 $\pm$ 4.9 & 1.5 $\pm$ 2.1 & -3.5 $\pm$ 6.4 & 21.0 $\pm$ 0.0 & 9.5 $\pm$ 0.7 & \textbf{22.0 $\pm$ 2.8} \\
 & sc-spp & 9.0 $\pm$ 12.7 & -0.5 $\pm$ 0.7 & -1.0 $\pm$ 5.7 & 2.5 $\pm$ 10.6 & 19.5 $\pm$ 2.1 & 12.0 $\pm$ 1.4 & \textbf{21.0 $\pm$ 2.8} \\
\cline{1-9}
\multirow[t]{6}{*}{C3.7S(T)} & zs & 0.2 $\pm$ 3.6 & 0.0 $\pm$ 5.3 & -4.2 $\pm$ 7.3 & -0.6 $\pm$ 3.6 & \textbf{19.6 $\pm$ 1.9} & 7.6 $\pm$ 2.8 & 17.6 $\pm$ 4.0 \\
 & cot & -2.0 $\pm$ 6.0 & 0.2 $\pm$ 3.2 & -3.8 $\pm$ 5.1 & -0.4 $\pm$ 3.2 & \textbf{21.0 $\pm$ 0.0} & 9.8 $\pm$ 4.4 & 15.6 $\pm$ 6.7 \\
 & spp & -3.2 $\pm$ 4.9 & 2.2 $\pm$ 10.9 & -1.6 $\pm$ 2.9 & 5.2 $\pm$ 4.1 & \textbf{20.8 $\pm$ 0.4} & 8.4 $\pm$ 1.7 & 15.8 $\pm$ 9.3 \\
 & sc-zs & -5.0 $\pm$ 2.8 & -7.5 $\pm$ 0.7 & 7.5 $\pm$ 3.5 & -3.5 $\pm$ 4.9 & \textbf{21.0 $\pm$ 0.0} & \textbf{15.5 $\pm$ 9.2} & 1.0 $\pm$ 1.4 \\
 & sc-cot & -2.0 $\pm$ 12.7 & -8.0 $\pm$ 9.9 & -9.5 $\pm$ 6.4 & 3.5 $\pm$ 4.9 & 18.5 $\pm$ 0.7 & 14.0 $\pm$ 8.5 & \textbf{22.0 $\pm$ 0.0} \\
 & sc-spp & -2.0 $\pm$ 11.3 & -9.5 $\pm$ 12.0 & -16.5 $\pm$ 4.9 & -4.0 $\pm$ 4.2 & 21.0 $\pm$ 0.0 & 9.0 $\pm$ 2.8 & \textbf{22.5 $\pm$ 2.1} \\
\cline{1-9}
\multirow[t]{6}{*}{C4S} & zs & 0.2 $\pm$ 4.8 & -5.4 $\pm$ 5.3 & -4.2 $\pm$ 8.1 & 2.4 $\pm$ 1.9 & \textbf{18.2 $\pm$ 8.0} & 10.6 $\pm$ 1.7 & 9.2 $\pm$ 8.4 \\
 & cot & 3.4 $\pm$ 17.3 & 3.2 $\pm$ 4.6 & 4.4 $\pm$ 12.5 & 1.2 $\pm$ 3.8 & \textbf{19.6 $\pm$ 2.1} & 12.4 $\pm$ 6.1 & 12.2 $\pm$ 7.3 \\
 & spp & 5.0 $\pm$ 8.5 & -7.6 $\pm$ 16.9 & 4.6 $\pm$ 9.4 & 1.0 $\pm$ 6.4 & \textbf{18.0 $\pm$ 3.7} & 12.4 $\pm$ 3.5 & 10.4 $\pm$ 7.3 \\
 & sc-zs & 12.5 $\pm$ 9.2 & -19.5 $\pm$ 3.5 & -11.5 $\pm$ 12.0 & 0.0 $\pm$ 0.0 & \textbf{24.0 $\pm$ 0.0} & 10.5 $\pm$ 2.1 & 18.0 $\pm$ 8.5 \\
 & sc-cot & -5.5 $\pm$ 6.4 & 1.0 $\pm$ 18.4 & 1.5 $\pm$ 14.8 & 3.5 $\pm$ 4.9 & \textbf{21.0 $\pm$ 0.0} & 9.5 $\pm$ 0.7 & 18.0 $\pm$ 8.5 \\
 & sc-spp & -0.5 $\pm$ 2.1 & 6.5 $\pm$ 9.2 & -6.0 $\pm$ 4.2 & 1.0 $\pm$ 2.8 & \textbf{21.0 $\pm$ 0.0} & 15.5 $\pm$ 9.2 & 20.0 $\pm$ 1.4 \\
\cline{1-9}
\multirow[t]{6}{*}{C4S(T)} & zs & -0.2 $\pm$ 2.3 & -2.8 $\pm$ 7.3 & -0.4 $\pm$ 15.9 & 1.0 $\pm$ 6.2 & \textbf{19.2 $\pm$ 4.0} & 11.8 $\pm$ 0.8 & 15.0 $\pm$ 10.6 \\
 & cot & 7.6 $\pm$ 10.4 & 0.0 $\pm$ 3.4 & -2.6 $\pm$ 11.5 & 2.0 $\pm$ 4.6 & \textbf{19.8 $\pm$ 1.6} & 10.0 $\pm$ 7.2 & 14.6 $\pm$ 8.1 \\
 & spp & -2.2 $\pm$ 5.4 & -1.8 $\pm$ 10.6 & -1.2 $\pm$ 5.7 & 0.4 $\pm$ 5.6 & \textbf{20.8 $\pm$ 0.4} & 8.4 $\pm$ 2.4 & 15.2 $\pm$ 8.3 \\
 & sc-zs & -5.0 $\pm$ 7.1 & 0.0 $\pm$ 2.8 & 0.5 $\pm$ 9.2 & 2.5 $\pm$ 0.7 & 21.0 $\pm$ 0.0 & 7.0 $\pm$ 5.7 & \textbf{22.0 $\pm$ 2.8} \\
 & sc-cot & 11.5 $\pm$ 6.4 & 8.5 $\pm$ 2.1 & 12.0 $\pm$ 2.8 & -0.5 $\pm$ 3.5 & 21.0 $\pm$ 0.0 & 11.0 $\pm$ 8.5 & \textbf{23.0 $\pm$ 1.4} \\
 & sc-spp & 5.0 $\pm$ 1.4 & -1.0 $\pm$ 0.0 & 1.0 $\pm$ 9.9 & 3.0 $\pm$ 1.4 & 21.0 $\pm$ 0.0 & 10.5 $\pm$ 2.1 & \textbf{23.5 $\pm$ 0.7} \\
\cline{1-9}
\multirow[t]{6}{*}{DS-R1} & zs & 1.2 $\pm$ 7.8 & -1.0 $\pm$ 1.6 & -3.2 $\pm$ 9.7 & 0.8 $\pm$ 3.0 & 5.8 $\pm$ 2.9 & 6.8 $\pm$ 5.0 & \textbf{14.6 $\pm$ 3.8} \\
 & cot & 4.0 $\pm$ 8.2 & 5.0 $\pm$ 4.3 & -4.2 $\pm$ 4.5 & 4.4 $\pm$ 4.2 & 12.4 $\pm$ 4.2 & 9.6 $\pm$ 3.0 & \textbf{17.8 $\pm$ 3.4} \\
 & spp & 0.8 $\pm$ 6.3 & -2.4 $\pm$ 5.9 & -0.4 $\pm$ 6.3 & -1.4 $\pm$ 3.5 & 6.2 $\pm$ 4.8 & 10.6 $\pm$ 2.2 & \textbf{13.4 $\pm$ 5.4} \\
 & sc-zs & 5.0 $\pm$ 9.9 & -3.5 $\pm$ 4.9 & -0.5 $\pm$ 6.4 & -3.5 $\pm$ 0.7 & 9.0 $\pm$ 8.5 & 6.5 $\pm$ 3.5 & \textbf{14.5 $\pm$ 0.7} \\
 & sc-cot & -0.5 $\pm$ 13.4 & -3.5 $\pm$ 4.9 & -3.5 $\pm$ 9.2 & 1.0 $\pm$ 1.4 & 15.0 $\pm$ 1.4 & 15.5 $\pm$ 3.5 & \textbf{20.5 $\pm$ 2.1} \\
 & sc-spp & -6.0 $\pm$ 5.7 & -2.5 $\pm$ 2.1 & -6.0 $\pm$ 1.4 & 0.0 $\pm$ 1.4 & 13.0 $\pm$ 2.8 & 7.0 $\pm$ 2.8 & \textbf{20.0 $\pm$ 2.8} \\
\cline{1-9}
\multirow[t]{6}{*}{L3.3-70B} & zs & -0.8 $\pm$ 1.8 & -0.4 $\pm$ 7.1 & 1.8 $\pm$ 5.4 & -2.0 $\pm$ 4.7 & \textbf{14.8 $\pm$ 12.6} & 6.6 $\pm$ 2.1 & 1.0 $\pm$ 1.0 \\
 & cot & -0.6 $\pm$ 1.3 & -1.6 $\pm$ 8.8 & 0.0 $\pm$ 2.4 & 0.6 $\pm$ 3.8 & \textbf{19.0 $\pm$ 8.7} & 8.2 $\pm$ 3.1 & 10.0 $\pm$ 8.9 \\
 & spp & 1.0 $\pm$ 1.7 & -1.6 $\pm$ 2.4 & -1.4 $\pm$ 1.8 & -0.4 $\pm$ 2.9 & 6.2 $\pm$ 7.8 & \textbf{9.8 $\pm$ 6.0} & 4.4 $\pm$ 4.0 \\
 & sc-zs & 0.0 $\pm$ 0.0 & -8.0 $\pm$ 11.3 & 2.0 $\pm$ 0.0 & -1.0 $\pm$ 2.8 & \textbf{12.5 $\pm$ 16.3} & 9.0 $\pm$ 2.8 & 0.0 $\pm$ 0.0 \\
 & sc-cot & -0.5 $\pm$ 0.7 & 2.0 $\pm$ 2.8 & 1.5 $\pm$ 2.1 & -0.5 $\pm$ 2.1 & 6.0 $\pm$ 2.8 & 7.0 $\pm$ 1.4 & \textbf{16.5 $\pm$ 9.2} \\
 & sc-spp & 0.0 $\pm$ 0.0 & -10.5 $\pm$ 16.3 & -0.5 $\pm$ 2.1 & 5.5 $\pm$ 0.7 & \textbf{19.0 $\pm$ 7.1} & 12.5 $\pm$ 2.1 & 2.0 $\pm$ 2.8 \\
\cline{1-9}
\multirow[t]{6}{*}{Mistral} & zs & 0.0 $\pm$ 1.4 & -10.6 $\pm$ 7.7 & -3.8 $\pm$ 2.8 & 0.8 $\pm$ 5.2 & \textbf{20.8 $\pm$ 7.2} & 5.4 $\pm$ 3.6 & 0.8 $\pm$ 0.8 \\
 & cot & 8.8 $\pm$ 5.4 & 0.0 $\pm$ 2.6 & -4.2 $\pm$ 10.7 & 0.4 $\pm$ 4.2 & -1.0 $\pm$ 12.9 & 1.6 $\pm$ 4.2 & \textbf{16.0 $\pm$ 9.5} \\
 & spp & 9.2 $\pm$ 5.4 & 7.0 $\pm$ 13.2 & 0.6 $\pm$ 8.4 & 0.8 $\pm$ 1.5 & 6.8 $\pm$ 8.2 & 2.2 $\pm$ 4.1 & \textbf{12.6 $\pm$ 6.6} \\
 & sc-zs & 0.5 $\pm$ 0.7 & -4.0 $\pm$ 2.8 & -13.5 $\pm$ 0.7 & 1.0 $\pm$ 1.4 & \textbf{24.0 $\pm$ 0.0} & 0.5 $\pm$ 0.7 & 0.0 $\pm$ 0.0 \\
 & sc-cot & 11.5 $\pm$ 10.6 & 5.5 $\pm$ 3.5 & -2.5 $\pm$ 30.4 & -0.5 $\pm$ 0.7 & 11.5 $\pm$ 17.7 & -0.5 $\pm$ 7.8 & \textbf{17.5 $\pm$ 7.8} \\
 & sc-spp & -0.5 $\pm$ 0.7 & -0.5 $\pm$ 0.7 & 2.5 $\pm$ 17.7 & 0.0 $\pm$ 4.2 & 6.0 $\pm$ 7.1 & -4.0 $\pm$ 2.8 & \textbf{6.5 $\pm$ 21.9} \\
\bottomrule
\end{tabular}
\caption{Total Points Averaged Over All Iterations for the default RPS setup. Bold denotes best results (most total points).}
\label{tab:rps_total_points_avg_heatmap_eq1}
\end{table*}

Considering the full repetition experiment of 24 RPS rounds, the following Tables are going to demonstrate the averaged point count in all RPS setups considered (default and three counterfactuals).

Commencing from the default RPS setup, results are presented in Table \ref{tab:rps_total_points_avg_heatmap_eq1}. A concise interpretation of the default RPS total points table is that performance is highly opponent-dependent, with clear separation between algorithmic opponents (especially PP, AP, TFT) and LLM–LLM play, where scores tend to remain close to zero.

First, across almost all models and prompting strategies, LLM–LLM settings (zs, spp, cot) produce scores close to 0, often with high variance. This is consistent with the mixed Nash equilibrium of RPS: when both agents behave unpredictably, neither side gains a consistent advantage. The large variance indicates that models do not perfectly randomize, but deviations are inconsistent rather than systematically exploitable.

In contrast, algorithmic opponents are consistently exploitable, and this is where most positive scores arise. In particular, Patterned Player (PP) and Tit-for-Tat (TFT) often yield the highest gains, with many models reaching 15–24 points, especially under self-consistency (SC) variants. This suggests that models are capable of detecting and exploiting regularities or reactive patterns, even if they struggle to maintain optimal mixed strategies in symmetric play.

Among models, Claude 3.7 (both standard and thinking) and Claude 4 variants are the strongest overall in RPS. They achieve the most consistent high scores against algorithmic opponents and benefit notably from SC prompting, which often boosts results toward the upper range ($\approx 20–24$). However, their performance in LLM–LLM settings still centers around zero, indicating that gains come primarily from exploitation rather than equilibrium play.

DeepSeek R1 and Claude 3.5 show moderate performance: they can exploit structured opponents (especially TFT and PP) but with less consistency and lower ceilings. Llama 3.3 is relatively stable but less aggressive, often achieving smaller gains, suggesting more conservative or less adaptive strategies.

Finally, Mistral exhibits the most unstable behavior: while it sometimes achieves high scores (e.g., against PP or under SC), it also shows large negative values and high variance, especially under CoT and SC. This reinforces its earlier profile as brittle and inconsistent, even when occasional exploitation succeeds.

Overall, the results highlight that in default RPS, success is driven more by opponent exploitation than by equilibrium play, and that prompting strategies like SC amplify exploitation ability rather than improving adherence to optimal mixed strategies.

\begin{table*}[h!]
\small
\centering
\begin{tabular}{p{1cm}p{0.9cm}lllllll}
\toprule
 &  & \multicolumn{7}{c}{\textbf{Label-based RPS counterfactual }} \\
 &  & zs & spp & cot & srep & pp & ap & tft \\
model & prompt &  &  &  &  &  &  &  \\
\midrule
\multirow[t]{6}{*}{C3.5Sv2} & zs & -1.6 $\pm$ 3.5 & -3.4 $\pm$ 2.3 & -0.8 $\pm$ 3.2 & -0.4 $\pm$ 3.4 & -4.4 $\pm$ 2.5 & \textbf{9.0 $\pm$ 2.6} & 5.2 $\pm$ 7.1 \\
 & cot & 0.8 $\pm$ 8.0 & -2.8 $\pm$ 6.7 & -0.4 $\pm$ 1.7 & -5.0 $\pm$ 4.7 & -9.0 $\pm$ 10.9 & \textbf{4.4 $\pm$ 3.9} & -0.8 $\pm$ 3.4 \\
 & spp & 1.6 $\pm$ 3.7 & -2.6 $\pm$ 1.5 & 1.8 $\pm$ 4.6 & 0.2 $\pm$ 5.0 & -8.0 $\pm$ 5.0 & \textbf{6.6 $\pm$ 3.9} & 2.0 $\pm$ 4.4 \\
 & sc-zs & 3.0 $\pm$ 5.7 & -2.0 $\pm$ 4.2 & 3.5 $\pm$ 2.1 & 3.5 $\pm$ 2.1 & -12.0 $\pm$ 2.8 & 7.0 $\pm$ 1.4 & \textbf{12.0 $\pm$ 1.4} \\
 & sc-cot & -2.5 $\pm$ 0.7 & -2.5 $\pm$ 0.7 & -4.5 $\pm$ 12.0 & 0.5 $\pm$ 2.1 & -20.5 $\pm$ 0.7 & 4.0 $\pm$ 0.0 & \textbf{5.5 $\pm$ 4.9} \\
 & sc-spp & -9.0 $\pm$ 15.6 & 5.0 $\pm$ 11.3 & 0.5 $\pm$ 6.4 & 4.5 $\pm$ 2.1 & -17.0 $\pm$ 5.7 & \textbf{9.0 $\pm$ 1.4} & 3.0 $\pm$ 0.0 \\
\cline{1-9}
\multirow[t]{6}{*}{C3.7S} & zs & -4.8 $\pm$ 5.5 & -2.4 $\pm$ 9.2 & 0.4 $\pm$ 8.2 & -0.8 $\pm$ 5.5 & -12.4 $\pm$ 11.9 & 10.2 $\pm$ 1.3 & \textbf{12.0 $\pm$ 6.0} \\
 & cot & 4.2 $\pm$ 6.4 & 1.4 $\pm$ 2.5 & -3.4 $\pm$ 5.9 & -0.2 $\pm$ 3.5 & \textbf{18.4 $\pm$ 7.0} & 13.8 $\pm$ 5.0 & 12.0 $\pm$ 12.4 \\
 & spp & 0.6 $\pm$ 5.9 & 5.6 $\pm$ 5.5 & -2.2 $\pm$ 10.0 & -2.4 $\pm$ 5.8 & \textbf{18.8 $\pm$ 2.3} & 10.6 $\pm$ 1.3 & 17.8 $\pm$ 3.3 \\
 & sc-zs & -6.0 $\pm$ 1.4 & -2.5 $\pm$ 9.2 & 0.0 $\pm$ 5.7 & 3.0 $\pm$ 1.4 & -8.0 $\pm$ 22.6 & 7.0 $\pm$ 2.8 & \textbf{7.5 $\pm$ 2.1} \\
 & sc-cot & -1.0 $\pm$ 1.4 & 2.5 $\pm$ 0.7 & 7.0 $\pm$ 4.2 & 3.0 $\pm$ 2.8 & \textbf{18.5 $\pm$ 7.8} & 10.0 $\pm$ 2.8 & 13.5 $\pm$ 12.0 \\
 & sc-spp & 6.0 $\pm$ 7.1 & 5.0 $\pm$ 21.2 & 4.0 $\pm$ 0.0 & -3.5 $\pm$ 9.2 & 15.0 $\pm$ 0.0 & 11.5 $\pm$ 0.7 & \textbf{16.5 $\pm$ 2.1} \\
\cline{1-9}
\multirow[t]{6}{*}{C3.7S(T)} & zs & -1.6 $\pm$ 5.5 & -2.6 $\pm$ 8.0 & -0.4 $\pm$ 3.5 & 1.0 $\pm$ 2.3 & 2.2 $\pm$ 14.8 & \textbf{6.4 $\pm$ 1.9} & 4.6 $\pm$ 3.4 \\
 & cot & 0.8 $\pm$ 3.6 & 0.0 $\pm$ 4.7 & -4.0 $\pm$ 2.2 & -1.6 $\pm$ 4.6 & 8.8 $\pm$ 17.3 & 9.2 $\pm$ 2.2 & \textbf{13.2 $\pm$ 6.1} \\
 & spp & 4.2 $\pm$ 8.5 & -0.2 $\pm$ 7.7 & -0.8 $\pm$ 6.9 & 0.6 $\pm$ 3.4 & \textbf{18.8 $\pm$ 6.1} & 9.4 $\pm$ 1.8 & 17.4 $\pm$ 5.7 \\
 & sc-zs & -8.5 $\pm$ 7.8 & -11.5 $\pm$ 2.1 & 1.5 $\pm$ 13.4 & 0.5 $\pm$ 3.5 & 18.0 $\pm$ 8.5 & 10.0 $\pm$ 1.4 & \textbf{20.5 $\pm$ 2.1} \\
 & sc-cot & 1.5 $\pm$ 2.1 & -3.0 $\pm$ 4.2 & 1.5 $\pm$ 3.5 & 3.5 $\pm$ 0.7 & 14.0 $\pm$ 14.1 & 7.5 $\pm$ 2.1 & \textbf{24.0 $\pm$ 0.0} \\
 & sc-spp & 12.5 $\pm$ 3.5 & 7.0 $\pm$ 2.8 & -4.5 $\pm$ 9.2 & -4.0 $\pm$ 1.4 & \textbf{21.5 $\pm$ 3.5} & 9.0 $\pm$ 1.4 & 10.0 $\pm$ 12.7 \\
\cline{1-9}
\multirow[t]{6}{*}{C4S} & zs & -2.6 $\pm$ 5.6 & -1.0 $\pm$ 2.1 & -2.8 $\pm$ 4.8 & -0.6 $\pm$ 4.2 & -2.6 $\pm$ 17.9 & \textbf{7.4 $\pm$ 1.9} & 3.8 $\pm$ 9.1 \\
 & cot & 1.2 $\pm$ 2.7 & -2.6 $\pm$ 5.9 & -4.6 $\pm$ 5.9 & -1.8 $\pm$ 4.6 & \textbf{15.8 $\pm$ 5.4} & 9.0 $\pm$ 0.7 & 8.6 $\pm$ 4.2 \\
 & spp & 1.0 $\pm$ 11.1 & 4.4 $\pm$ 8.4 & 2.0 $\pm$ 5.8 & -2.4 $\pm$ 4.4 & \textbf{16.4 $\pm$ 2.6} & 1.2 $\pm$ 12.8 & -1.4 $\pm$ 6.8 \\
 & sc-zs & -0.5 $\pm$ 3.5 & -7.5 $\pm$ 0.7 & -0.5 $\pm$ 2.1 & 0.0 $\pm$ 2.8 & \textbf{17.0 $\pm$ 1.4} & 7.5 $\pm$ 0.7 & -2.5 $\pm$ 4.9 \\
 & sc-cot & 1.0 $\pm$ 1.4 & 3.5 $\pm$ 7.8 & -3.0 $\pm$ 2.8 & 1.0 $\pm$ 11.3 & \textbf{20.0 $\pm$ 4.2} & 10.0 $\pm$ 0.0 & 3.5 $\pm$ 3.5 \\
 & sc-spp & 1.0 $\pm$ 5.7 & -2.0 $\pm$ 0.0 & -2.5 $\pm$ 2.1 & 1.0 $\pm$ 2.8 & \textbf{17.5 $\pm$ 3.5} & 9.0 $\pm$ 1.4 & 2.5 $\pm$ 0.7 \\
\cline{1-9}
\multirow[t]{6}{*}{C4S(T)} & zs & -3.2 $\pm$ 1.3 & 0.6 $\pm$ 2.9 & -1.2 $\pm$ 3.1 & -1.6 $\pm$ 3.4 & -7.2 $\pm$ 14.5 & \textbf{5.6 $\pm$ 3.1} & 3.2 $\pm$ 3.3 \\
 & cot & 0.0 $\pm$ 3.2 & 0.8 $\pm$ 3.3 & 2.4 $\pm$ 4.3 & 1.6 $\pm$ 4.3 & \textbf{12.8 $\pm$ 8.2} & 9.4 $\pm$ 1.9 & 4.6 $\pm$ 9.8 \\
 & spp & 0.0 $\pm$ 7.5 & 0.8 $\pm$ 6.4 & -1.2 $\pm$ 2.4 & -5.6 $\pm$ 4.7 & \textbf{13.0 $\pm$ 8.0} & 8.4 $\pm$ 2.7 & 10.0 $\pm$ 13.5 \\
 & sc-zs & 1.0 $\pm$ 2.8 & -6.0 $\pm$ 1.4 & -9.0 $\pm$ 8.5 & -2.0 $\pm$ 2.8 & -11.5 $\pm$ 6.4 & \textbf{6.5 $\pm$ 3.5} & -4.0 $\pm$ 0.0 \\
 & sc-cot & 6.5 $\pm$ 3.5 & -3.0 $\pm$ 4.2 & 0.0 $\pm$ 2.8 & -2.0 $\pm$ 2.8 & \textbf{22.5 $\pm$ 2.1} & 11.0 $\pm$ 0.0 & 11.0 $\pm$ 9.9 \\
 & sc-spp & 4.5 $\pm$ 3.5 & -4.0 $\pm$ 8.5 & 4.0 $\pm$ 7.1 & 1.5 $\pm$ 6.4 & \textbf{13.5 $\pm$ 4.9} & 9.0 $\pm$ 1.4 & 6.5 $\pm$ 20.5 \\
\cline{1-9}
\multirow[t]{6}{*}{DS-R1} & zs & 2.0 $\pm$ 3.5 & 0.4 $\pm$ 2.7 & 2.6 $\pm$ 3.2 & 1.2 $\pm$ 2.2 & -7.4 $\pm$ 2.5 & \textbf{6.2 $\pm$ 2.0} & 4.8 $\pm$ 2.3 \\
 & cot & 0.8 $\pm$ 3.8 & 2.8 $\pm$ 5.1 & 0.2 $\pm$ 4.8 & 1.0 $\pm$ 4.8 & -4.0 $\pm$ 4.6 & \textbf{6.4 $\pm$ 2.1} & 5.0 $\pm$ 4.6 \\
 & spp & -3.0 $\pm$ 2.5 & 0.2 $\pm$ 1.9 & -2.4 $\pm$ 5.3 & 0.0 $\pm$ 2.8 & -3.8 $\pm$ 6.2 & \textbf{4.6 $\pm$ 1.7} & 3.2 $\pm$ 1.3 \\
 & sc-zs & -1.5 $\pm$ 2.1 & -0.5 $\pm$ 3.5 & 2.5 $\pm$ 2.1 & -6.0 $\pm$ 2.8 & -6.5 $\pm$ 0.7 & \textbf{6.0 $\pm$ 0.0} & 2.5 $\pm$ 0.7 \\
 & sc-cot & -3.0 $\pm$ 1.4 & 1.0 $\pm$ 0.0 & 4.5 $\pm$ 3.5 & -2.5 $\pm$ 2.1 & -0.5 $\pm$ 4.9 & \textbf{9.0 $\pm$ 0.0} & 6.5 $\pm$ 0.7 \\
 & sc-spp & 2.5 $\pm$ 2.1 & 3.0 $\pm$ 4.2 & 1.0 $\pm$ 1.4 & 0.0 $\pm$ 2.8 & -8.5 $\pm$ 0.7 & \textbf{6.5 $\pm$ 0.7} & 6.0 $\pm$ 2.8 \\
\cline{1-9}
\multirow[t]{6}{*}{L3.3-70B} & zs & -1.4 $\pm$ 2.8 & -0.6 $\pm$ 1.7 & 0.6 $\pm$ 1.8 & 0.4 $\pm$ 1.7 & -7.4 $\pm$ 9.4 & \textbf{7.0 $\pm$ 1.6} & 4.8 $\pm$ 4.5 \\
 & cot & -0.6 $\pm$ 1.8 & 0.8 $\pm$ 3.2 & -0.2 $\pm$ 1.6 & 0.0 $\pm$ 1.6 & -8.2 $\pm$ 2.5 & \textbf{7.4 $\pm$ 1.7} & 3.2 $\pm$ 2.9 \\
 & spp & 0.0 $\pm$ 1.9 & 3.2 $\pm$ 4.5 & 2.6 $\pm$ 4.0 & 1.0 $\pm$ 2.0 & -10.4 $\pm$ 8.9 & 5.0 $\pm$ 4.0 & \textbf{5.4 $\pm$ 4.3} \\
 & sc-zs & 0.0 $\pm$ 0.0 & -2.5 $\pm$ 0.7 & 4.0 $\pm$ 2.8 & 3.5 $\pm$ 3.5 & -14.0 $\pm$ 14.1 & 9.5 $\pm$ 2.1 & \textbf{10.5 $\pm$ 2.1} \\
 & sc-cot & 0.0 $\pm$ 5.7 & 1.5 $\pm$ 6.4 & -1.5 $\pm$ 3.5 & -3.0 $\pm$ 1.4 & -14.0 $\pm$ 14.1 & \textbf{4.0 $\pm$ 1.4} & 1.0 $\pm$ 2.8 \\
 & sc-spp & 0.0 $\pm$ 0.0 & -1.5 $\pm$ 2.1 & 1.0 $\pm$ 2.8 & 1.0 $\pm$ 2.8 & -6.0 $\pm$ 2.8 & \textbf{8.5 $\pm$ 0.7} & 4.0 $\pm$ 1.4 \\
\cline{1-9}
\multirow[t]{6}{*}{Mistral} & zs & -0.2 $\pm$ 4.2 & -3.6 $\pm$ 5.2 & 2.4 $\pm$ 5.2 & -0.2 $\pm$ 3.0 & -12.4 $\pm$ 9.5 & 0.0 $\pm$ 2.4 & \textbf{8.4 $\pm$ 1.3} \\
 & cot & -3.2 $\pm$ 5.4 & 2.4 $\pm$ 5.5 & 0.4 $\pm$ 5.5 & -2.4 $\pm$ 4.8 & -6.2 $\pm$ 7.1 & 3.6 $\pm$ 5.9 & \textbf{5.0 $\pm$ 3.7} \\
 & spp & 1.4 $\pm$ 8.0 & -3.0 $\pm$ 5.0 & 0.8 $\pm$ 7.2 & 2.2 $\pm$ 4.3 & -4.8 $\pm$ 5.4 & \textbf{9.0 $\pm$ 10.7} & 5.4 $\pm$ 6.7 \\
 & sc-zs & 0.0 $\pm$ 0.0 & -0.5 $\pm$ 0.7 & -5.0 $\pm$ 12.7 & -0.5 $\pm$ 4.9 & -11.5 $\pm$ 17.7 & -0.5 $\pm$ 0.7 & \textbf{9.0 $\pm$ 4.2} \\
 & sc-cot & -3.0 $\pm$ 1.4 & 2.5 $\pm$ 3.5 & 8.0 $\pm$ 2.8 & 0.0 $\pm$ 1.4 & -14.5 $\pm$ 13.4 & \textbf{12.5 $\pm$ 13.4} & 11.0 $\pm$ 0.0 \\
 & sc-spp & -0.5 $\pm$ 0.7 & 0.5 $\pm$ 2.1 & 4.5 $\pm$ 4.9 & 3.0 $\pm$ 1.4 & -22.0 $\pm$ 2.8 & 7.5 $\pm$ 17.7 & \textbf{19.0 $\pm$ 4.2} \\
\bottomrule
\end{tabular}
\caption{Total Points Averaged Over All Iterations in the label-based RPS counterfactual.}
\label{tab:rps_total_points_avg_heatmap_eq1-alt}
\end{table*}

Label-based RPS counterfactuals are demonstrated in Table \ref{tab:rps_total_points_avg_heatmap_eq1-alt}. A brief takeaway from the label-based RPS counterfactual is that simply renaming / permuting the dominance structure hurts many models substantially, especially against PP, while performance against AP and sometimes TFT remains more positive. This is exactly the pattern expected if some models rely partly on the canonical “Rock beats Scissors, Scissors beats Paper, Paper beats Rock” template rather than fully recomputing the altered dominance relation.

Compared with default RPS, the most striking change is that several models now score negative points against PP, sometimes heavily so. This is especially visible for Claude 3.5, Llama 3.3, and parts of DeepSeek and Mistral, indicating that the cyclic patterned opponent becomes much harder to exploit once the familiar label semantics are disrupted. In contrast, Claude 3.7 and particularly Claude 3.7 Thinking remain among the strongest and most robust models: they still obtain strong positive scores against PP/TFT in several prompt settings, with some of the best results in the table, suggesting better adaptation to the remapped action semantics.

A second consistent pattern is that AP remains one of the most reliably exploitable opponents across models. For many rows, AP is the best or among the best-performing columns, often yielding moderate positive returns even when PP becomes negative. This suggests that models can still respond to short-horizon behavioral regularities even when they struggle with the deeper symbolic remapping of the whole RPS structure.

Model-wise, Claude 3.7 / Claude 3.7 Thinking look strongest overall under this counterfactual, with high peaks against PP, AP, and TFT. Claude 4 and Claude 4 Thinking are more mixed: they can still do well, especially with SC or against PP/AP, but are less uniformly robust. DeepSeek R1 is relatively stable but weaker, usually getting modest gains mainly against AP. Llama 3.3 degrades noticeably under the label shift, especially against PP, which supports the view that it is less robust in harder RPS settings. Mistral remains the most erratic: it sometimes achieves strong scores, but it also shows large negative drops and very high variability.

Finally, prompting effects are mixed but self-consistency often helps, especially for stronger Claude variants, by pushing scores upward on exploitable opponents such as TFT and PP. Still, SC is not a universal fix: under label remapping, some runs remain strongly negative, meaning that the counterfactual exposes a genuine robustness problem rather than just sampling noise. Overall, this table provides clear evidence that label-only perturbations in RPS are enough to break strategic transfer for many models.

\begin{table*}[h!]
\small
\centering
\begin{tabular}{p{1cm}p{0.9cm}p{1.6cm}p{1.6cm}p{1.6cm}p{1.4cm}p{1.5cm}p{1.5cm}p{1.5cm}}
\toprule
 &  & \multicolumn{7}{c}{\textbf{Payoff-based RPS counterfactual}} \\
 &  & zs & spp & cot & srep & pp & ap & tft \\
model & prompt &  &  &  &  &  &  &  \\
\midrule
\multirow[t]{6}{*}{C3.5Sv2} & zs & 3.8 $\pm$ 16.1 & -1.2 $\pm$ 25.5 & -5.0 $\pm$ 12.8 & 0.2 $\pm$ 5.6 & -4.0 $\pm$ 17.2 & 9.2 $\pm$ 6.2 & \textbf{29.6 $\pm$ 10.7} \\
 & cot & -3.0 $\pm$ 21.6 & 11.6 $\pm$ 23.1 & 7.0 $\pm$ 15.1 & -0.4 $\pm$ 6.3 & -7.0 $\pm$ 5.1 & 6.8 $\pm$ 2.8 & \textbf{24.2 $\pm$ 14.0} \\
 & spp & \textbf{20.2 $\pm$ 11.2} & 12.4 $\pm$ 11.8 & 7.4 $\pm$ 8.5 & 3.4 $\pm$ 3.9 & 1.6 $\pm$ 17.7 & 6.4 $\pm$ 2.1 & 13.2 $\pm$ 12.8 \\
 & sc-zs & -14.5 $\pm$ 21.9 & 3.5 $\pm$ 3.5 & 17.0 $\pm$ 5.7 & 2.0 $\pm$ 9.9 & 35.0 $\pm$ 0.0 & 18.0 $\pm$ 1.4 & \textbf{40.0 $\pm$ 0.0} \\
 & sc-cot & -0.5 $\pm$ 17.7 & 8.0 $\pm$ 15.6 & -18.5 $\pm$ 26.2 & 0.0 $\pm$ 0.0 & 16.0 $\pm$ 26.9 & 18.5 $\pm$ 9.2 & \textbf{32.5 $\pm$ 4.9} \\
 & sc-spp & 4.5 $\pm$ 9.2 & -24.5 $\pm$ 2.1 & 33.0 $\pm$ 8.5 & -3.0 $\pm$ 0.0 & 34.5 $\pm$ 0.7 & 16.0 $\pm$ 18.4 & \textbf{35.5 $\pm$ 0.7} \\
\cline{1-9}
\multirow[t]{6}{*}{C3.7S} & zs & 2.6 $\pm$ 20.1 & -3.0 $\pm$ 13.8 & -1.6 $\pm$ 11.1 & -3.8 $\pm$ 6.4 & \textbf{35.0 $\pm$ 5.3} & 13.6 $\pm$ 12.3 & 31.4 $\pm$ 7.3 \\
 & cot & 12.2 $\pm$ 8.6 & -9.2 $\pm$ 8.1 & -4.4 $\pm$ 11.5 & 0.0 $\pm$ 2.9 & \textbf{34.6 $\pm$ 0.9} & 16.6 $\pm$ 14.9 & 27.6 $\pm$ 11.9 \\
 & spp & -6.8 $\pm$ 18.5 & -4.4 $\pm$ 11.8 & -5.2 $\pm$ 12.7 & -3.8 $\pm$ 4.6 & 31.8 $\pm$ 4.1 & 10.6 $\pm$ 8.7 & \textbf{34.6 $\pm$ 2.9} \\
 & sc-zs & 10.5 $\pm$ 12.0 & -34.0 $\pm$ 7.1 & -6.5 $\pm$ 13.4 & 9.0 $\pm$ 8.5 & 11.0 $\pm$ 2.8 & 9.5 $\pm$ 7.8 & \textbf{37.0 $\pm$ 4.2} \\
 & sc-cot & 12.0 $\pm$ 2.8 & -12.0 $\pm$ 15.6 & 0.5 $\pm$ 6.4 & -1.0 $\pm$ 9.9 & 35.0 $\pm$ 0.0 & 9.0 $\pm$ 1.4 & \textbf{38.0 $\pm$ 2.8} \\
 & sc-spp & 7.5 $\pm$ 33.2 & -5.5 $\pm$ 6.4 & 1.5 $\pm$ 4.9 & 2.0 $\pm$ 7.1 & 34.0 $\pm$ 1.4 & 31.5 $\pm$ 0.7 & \textbf{36.5 $\pm$ 0.7} \\
\cline{1-9}
\multirow[t]{6}{*}{C3.7S(T)} & zs & 2.4 $\pm$ 7.6 & 7.2 $\pm$ 9.0 & -3.4 $\pm$ 8.3 & 4.6 $\pm$ 5.9 & \textbf{28.0 $\pm$ 4.6} & 10.6 $\pm$ 0.9 & 10.8 $\pm$ 3.0 \\
 & cot & 7.0 $\pm$ 7.8 & -3.2 $\pm$ 12.9 & -4.4 $\pm$ 3.4 & -0.6 $\pm$ 3.6 & \textbf{34.8 $\pm$ 0.4} & 17.2 $\pm$ 12.3 & 30.0 $\pm$ 7.4 \\
 & spp & -1.2 $\pm$ 14.7 & -5.6 $\pm$ 12.9 & -6.6 $\pm$ 14.1 & 2.8 $\pm$ 3.0 & \textbf{34.6 $\pm$ 0.5} & 28.8 $\pm$ 13.4 & 31.4 $\pm$ 5.2 \\
 & sc-zs & -8.5 $\pm$ 4.9 & -4.0 $\pm$ 7.1 & -6.5 $\pm$ 7.8 & 2.5 $\pm$ 4.9 & \textbf{35.0 $\pm$ 0.0} & 13.0 $\pm$ 2.8 & 33.0 $\pm$ 9.9 \\
 & sc-cot & 3.5 $\pm$ 0.7 & 7.5 $\pm$ 9.2 & 7.5 $\pm$ 0.7 & -1.5 $\pm$ 2.1 & \textbf{35.0 $\pm$ 0.0} & 9.5 $\pm$ 6.4 & 34.5 $\pm$ 2.1 \\
 & sc-spp & -7.0 $\pm$ 1.4 & -18.0 $\pm$ 1.4 & 5.0 $\pm$ 5.7 & 4.0 $\pm$ 7.1 & 34.5 $\pm$ 0.7 & 21.5 $\pm$ 14.8 & \textbf{37.0 $\pm$ 0.0} \\
\cline{1-9}
\multirow[t]{6}{*}{C4S} & zs & -3.0 $\pm$ 12.7 & -1.8 $\pm$ 10.5 & -4.0 $\pm$ 19.1 & -1.8 $\pm$ 7.4 & \textbf{34.4 $\pm$ 0.5} & 11.8 $\pm$ 4.0 & 23.0 $\pm$ 14.5 \\
 & cot & -2.8 $\pm$ 20.2 & -9.0 $\pm$ 17.4 & 6.8 $\pm$ 7.2 & 4.0 $\pm$ 5.5 & \textbf{31.8 $\pm$ 4.4} & 17.4 $\pm$ 11.1 & 23.2 $\pm$ 7.8 \\
 & spp & 13.8 $\pm$ 7.6 & 3.6 $\pm$ 10.5 & 4.8 $\pm$ 6.5 & 3.0 $\pm$ 8.0 & \textbf{28.2 $\pm$ 13.5} & 13.0 $\pm$ 5.4 & 20.4 $\pm$ 15.5 \\
 & sc-zs & -13.5 $\pm$ 27.6 & 1.0 $\pm$ 7.1 & -10.0 $\pm$ 2.8 & 1.0 $\pm$ 14.1 & \textbf{35.0 $\pm$ 0.0} & 16.5 $\pm$ 4.9 & 11.0 $\pm$ 5.7 \\
 & sc-cot & 7.0 $\pm$ 19.8 & -3.5 $\pm$ 0.7 & -7.0 $\pm$ 9.9 & -1.0 $\pm$ 2.8 & \textbf{36.0 $\pm$ 2.8} & 17.5 $\pm$ 14.8 & 20.0 $\pm$ 28.3 \\
 & sc-spp & -3.0 $\pm$ 22.6 & 0.0 $\pm$ 5.7 & 3.5 $\pm$ 44.5 & 4.5 $\pm$ 2.1 & \textbf{35.0 $\pm$ 0.0} & 12.5 $\pm$ 6.4 & 0.0 $\pm$ 0.0 \\
\cline{1-9}
\multirow[t]{6}{*}{C4S(T)} & zs & -8.8 $\pm$ 12.6 & -9.2 $\pm$ 4.5 & -6.0 $\pm$ 11.0 & 0.6 $\pm$ 2.2 & \textbf{30.6 $\pm$ 6.3} & 13.8 $\pm$ 3.2 & 12.0 $\pm$ 16.7 \\
 & cot & 4.0 $\pm$ 10.1 & 6.4 $\pm$ 7.6 & 1.6 $\pm$ 10.7 & -1.6 $\pm$ 5.3 & 33.6 $\pm$ 2.6 & 9.0 $\pm$ 7.2 & \textbf{33.8 $\pm$ 4.5} \\
 & spp & 2.8 $\pm$ 15.9 & 3.4 $\pm$ 12.0 & -0.6 $\pm$ 21.2 & -3.2 $\pm$ 4.0 & \textbf{34.0 $\pm$ 1.4} & 12.0 $\pm$ 1.6 & 28.8 $\pm$ 9.2 \\
 & sc-zs & -5.0 $\pm$ 9.9 & -9.5 $\pm$ 3.5 & -10.0 $\pm$ 29.7 & -5.5 $\pm$ 0.7 & \textbf{32.5 $\pm$ 3.5} & 17.0 $\pm$ 5.7 & 28.0 $\pm$ 12.7 \\
 & sc-cot & -5.5 $\pm$ 20.5 & 3.0 $\pm$ 1.4 & -26.5 $\pm$ 12.0 & -4.0 $\pm$ 14.1 & 27.5 $\pm$ 3.5 & 12.0 $\pm$ 4.2 & \textbf{37.0 $\pm$ 1.4} \\
 & sc-spp & 10.5 $\pm$ 4.9 & -4.5 $\pm$ 12.0 & 3.5 $\pm$ 2.1 & 2.0 $\pm$ 7.1 & \textbf{36.0 $\pm$ 1.4} & 10.5 $\pm$ 4.9 & 25.0 $\pm$ 21.2 \\
\cline{1-9}
\multirow[t]{6}{*}{DS-R1} & zs & -10.4 $\pm$ 15.5 & 1.6 $\pm$ 6.9 & 9.8 $\pm$ 10.1 & 1.0 $\pm$ 6.4 & 13.8 $\pm$ 12.8 & 8.4 $\pm$ 4.9 & \textbf{29.4 $\pm$ 4.2} \\
 & cot & 9.0 $\pm$ 8.4 & -8.6 $\pm$ 14.6 & -0.4 $\pm$ 10.6 & 1.2 $\pm$ 7.8 & 14.8 $\pm$ 5.4 & 11.4 $\pm$ 6.3 & \textbf{20.2 $\pm$ 3.8} \\
 & spp & -7.8 $\pm$ 9.4 & 3.0 $\pm$ 12.8 & 2.0 $\pm$ 6.8 & -1.2 $\pm$ 9.2 & 12.4 $\pm$ 10.6 & 9.8 $\pm$ 3.8 & \textbf{17.0 $\pm$ 12.9} \\
 & sc-zs & 4.5 $\pm$ 10.6 & 11.5 $\pm$ 9.2 & 5.5 $\pm$ 4.9 & -2.0 $\pm$ 8.5 & 22.0 $\pm$ 11.3 & 18.5 $\pm$ 6.4 & \textbf{36.5 $\pm$ 0.7} \\
 & sc-cot & 8.5 $\pm$ 0.7 & -6.5 $\pm$ 19.1 & -5.5 $\pm$ 4.9 & 9.0 $\pm$ 8.5 & 18.0 $\pm$ 1.4 & 11.5 $\pm$ 0.7 & \textbf{24.5 $\pm$ 3.5} \\
 & sc-spp & -14.5 $\pm$ 12.0 & -8.5 $\pm$ 4.9 & -2.0 $\pm$ 8.5 & 4.5 $\pm$ 3.5 & \textbf{25.0 $\pm$ 8.5} & 19.0 $\pm$ 4.2 & 23.5 $\pm$ 2.1 \\
\cline{1-9}
\multirow[t]{6}{*}{L3.3-70B} & zs & 0.4 $\pm$ 17.6 & -3.8 $\pm$ 16.1 & 12.2 $\pm$ 15.0 & 2.0 $\pm$ 5.3 & \textbf{15.2 $\pm$ 14.5} & 11.6 $\pm$ 8.2 & 1.6 $\pm$ 3.4 \\
 & cot & 0.8 $\pm$ 7.3 & 0.4 $\pm$ 14.2 & 2.6 $\pm$ 8.5 & 0.2 $\pm$ 1.9 & \textbf{37.0 $\pm$ 6.7} & 17.0 $\pm$ 0.7 & 1.6 $\pm$ 1.1 \\
 & spp & -0.2 $\pm$ 1.8 & -0.4 $\pm$ 10.3 & 2.6 $\pm$ 4.7 & 0.6 $\pm$ 7.1 & 9.2 $\pm$ 5.6 & 10.4 $\pm$ 5.9 & \textbf{13.2 $\pm$ 15.7} \\
 & sc-zs & -10.0 $\pm$ 24.0 & 13.0 $\pm$ 18.4 & -1.0 $\pm$ 1.4 & 3.0 $\pm$ 7.1 & \textbf{24.5 $\pm$ 21.9} & 17.5 $\pm$ 12.0 & -1.0 $\pm$ 1.4 \\
 & sc-cot & -17.0 $\pm$ 25.5 & -0.5 $\pm$ 3.5 & 1.5 $\pm$ 2.1 & 2.5 $\pm$ 6.4 & \textbf{20.5 $\pm$ 27.6} & 13.5 $\pm$ 0.7 & 1.0 $\pm$ 1.4 \\
 & sc-spp & \textbf{27.0 $\pm$ 2.8} & 3.0 $\pm$ 2.8 & 2.5 $\pm$ 0.7 & -4.5 $\pm$ 9.2 & 2.5 $\pm$ 2.1 & 13.5 $\pm$ 16.3 & 16.5 $\pm$ 26.2 \\
\cline{1-9}
\multirow[t]{6}{*}{Mistral} & zs & -0.2 $\pm$ 2.8 & -7.8 $\pm$ 10.2 & -5.4 $\pm$ 4.8 & -1.0 $\pm$ 2.7 & \textbf{34.8 $\pm$ 11.6} & 1.8 $\pm$ 13.0 & 1.8 $\pm$ 1.6 \\
 & cot & \textbf{15.6 $\pm$ 11.5} & -3.8 $\pm$ 13.7 & -1.6 $\pm$ 20.9 & -0.6 $\pm$ 4.2 & 8.6 $\pm$ 7.1 & 6.8 $\pm$ 7.4 & 14.0 $\pm$ 16.5 \\
 & spp & 9.0 $\pm$ 19.0 & -13.0 $\pm$ 19.1 & 2.6 $\pm$ 13.7 & 2.2 $\pm$ 6.4 & -4.8 $\pm$ 7.7 & -3.0 $\pm$ 5.1 & \textbf{27.8 $\pm$ 10.4} \\
 & sc-zs & 0.0 $\pm$ 0.0 & -18.5 $\pm$ 17.7 & -9.0 $\pm$ 5.7 & 4.5 $\pm$ 4.9 & \textbf{20.5 $\pm$ 20.5} & 2.5 $\pm$ 0.7 & 0.5 $\pm$ 0.7 \\
 & sc-cot & 11.0 $\pm$ 15.6 & 17.5 $\pm$ 31.8 & 3.0 $\pm$ 15.6 & 0.5 $\pm$ 2.1 & 15.5 $\pm$ 21.9 & 13.0 $\pm$ 2.8 & \textbf{22.0 $\pm$ 25.5} \\
 & sc-spp & 5.0 $\pm$ 11.3 & -7.5 $\pm$ 23.3 & -8.5 $\pm$ 30.4 & 3.5 $\pm$ 7.8 & 11.0 $\pm$ 17.0 & -1.0 $\pm$ 21.2 & \textbf{35.5 $\pm$ 3.5} \\
\bottomrule
\end{tabular}
\caption{Total Points Averaged Over All Iterations for the payoff-based RPS counterfactual.}
\label{tab:rps_total_points_avg_heatmap_ba3}
\end{table*}

Table \ref{tab:rps_total_points_avg_heatmap_ba3} exhibits results regarding payoff-based RPS counterfactuals. Changing rewards proves to be more disruptive than the label-only shift, because strong play now requires abandoning uniform randomization and biasing toward the amplified interaction. This creates larger spreads in total points, more variance across prompts, and clearer separation between models that adapt to incentives and those that remain closer to default RPS behavior.

The most consistent pattern is that PP and TFT become the most exploitable opponents. Many of the best scores in the table appear in these two columns, often reaching the mid-30s or even around 40, which is much higher than in default RPS. This suggests that once an LLM both detects the opponent pattern and aligns it with the altered payoff asymmetry, it can extract substantially more reward than in the canonical game. In contrast, SREP remains near zero for most rows, as expected, since an equilibrium-style opponent leaves little room for systematic gain even under modified payoffs. AP usually yields moderate positive returns, but generally not as high as PP/TFT.

Model-wise, Claude 3.7, Claude 3.7 Thinking, and the Claude 4 variants remain among the strongest overall under this counterfactual, especially against PP and TFT. They repeatedly achieve the highest or near-highest totals, indicating relatively good adaptation to the new payoff landscape. Claude 3.5 is more uneven but still capable of strong gains, especially under SC, where TFT scores become very large. DeepSeek R1 is mixed: it sometimes performs strongly, particularly against TFT and under SC, but remains less stable across prompts. Llama 3.3 is the least consistent among the stronger models: it has some high peaks, but also several near-zero or highly variable results, suggesting weaker payoff sensitivity. Mistral remains the most erratic, with occasional strong positive runs but also many unstable or negative ones.

Prompting effects are again mixed, but here self-consistency often helps more clearly than in the default game. For several models, SC substantially boosts scores against PP/TFT, which fits the intuition that majority voting can stabilize a biased strategy once the model has partially inferred the new incentives. At the same time, some prompt/model combinations still collapse into strongly negative totals, showing that payoff changes expose not just variance but genuine failures to recompute the optimal mixed strategy.

All in all, presented results support the main claim of the paper: payoff-based counterfactuals are a stronger test of strategic reasoning than default play. Many models that look competent in standard RPS do not transfer cleanly when the reward structure changes, while the strongest Claude-family models show the best, though still imperfect, adaptation

\begin{table*}[h!]
\small
\centering
\begin{tabular}{p{1cm}p{0.9cm}p{1.5cm}p{1.5cm}p{1.6cm}p{1.4cm}p{1.6cm}p{1.6cm}p{1.5cm}}
\toprule
 &  & \multicolumn{7}{c}{\textbf{Joint RPS counterfactual}} \\
 &  & zs & spp & cot & srep & pp & ap & tft \\
model & prompt &  &  &  &  &  &  &  \\
\midrule
\multirow[t]{6}{*}{C3.5Sv2} & zs & 0.6 $\pm$ 6.0 & 2.2 $\pm$ 10.3 & -1.0 $\pm$ 10.9 & -3.4 $\pm$ 4.7 & -8.6 $\pm$ 22.2 & 10.6 $\pm$ 5.0 & \textbf{11.8 $\pm$ 12.1} \\
 & cot & 2.0 $\pm$ 6.8 & -5.8 $\pm$ 15.9 & -4.2 $\pm$ 13.0 & -0.2 $\pm$ 5.4 & -8.8 $\pm$ 16.5 & 6.2 $\pm$ 11.4 & \textbf{12.2 $\pm$ 7.4} \\
 & spp & -1.4 $\pm$ 11.0 & 5.0 $\pm$ 10.2 & 2.6 $\pm$ 9.8 & 3.6 $\pm$ 3.9 & -25.8 $\pm$ 10.0 & \textbf{14.0 $\pm$ 4.8} & 11.8 $\pm$ 7.7 \\
 & sc-zs & -7.0 $\pm$ 8.5 & -7.5 $\pm$ 4.9 & 4.5 $\pm$ 0.7 & 0.5 $\pm$ 6.4 & -17.0 $\pm$ 19.8 & \textbf{8.5 $\pm$ 4.9} & 7.0 $\pm$ 5.7 \\
 & sc-cot & 8.0 $\pm$ 5.7 & 0.5 $\pm$ 9.2 & 2.0 $\pm$ 9.9 & -2.5 $\pm$ 7.8 & -20.5 $\pm$ 20.5 & \textbf{14.0 $\pm$ 7.1} & 6.5 $\pm$ 2.1 \\
 & sc-spp & 5.0 $\pm$ 4.2 & -0.5 $\pm$ 7.8 & 2.5 $\pm$ 2.1 & -4.5 $\pm$ 9.2 & -14.0 $\pm$ 5.7 & -11.0 $\pm$ 7.1 & \textbf{18.0 $\pm$ 5.7} \\
\cline{1-9}
\multirow[t]{6}{*}{C3.7S} & zs & 1.6 $\pm$ 11.9 & -2.6 $\pm$ 6.9 & -10.2 $\pm$ 11.0 & -1.2 $\pm$ 6.5 & -13.4 $\pm$ 15.8 & 12.0 $\pm$ 7.0 & \textbf{16.6 $\pm$ 16.6} \\
 & cot & 1.2 $\pm$ 9.9 & 2.4 $\pm$ 9.4 & -1.6 $\pm$ 5.3 & 4.2 $\pm$ 6.4 & \textbf{33.4 $\pm$ 10.0} & 12.4 $\pm$ 1.1 & 20.0 $\pm$ 15.4 \\
 & spp & 6.4 $\pm$ 14.7 & 1.6 $\pm$ 11.7 & 3.0 $\pm$ 8.9 & -2.6 $\pm$ 6.1 & 27.4 $\pm$ 18.1 & 18.0 $\pm$ 10.1 & \textbf{29.0 $\pm$ 11.9} \\
 & sc-zs & 10.0 $\pm$ 28.3 & -3.0 $\pm$ 0.0 & -19.5 $\pm$ 4.9 & 0.5 $\pm$ 4.9 & 16.0 $\pm$ 2.8 & 24.0 $\pm$ 21.2 & \textbf{31.0 $\pm$ 8.5} \\
 & sc-cot & -2.0 $\pm$ 2.8 & -12.0 $\pm$ 9.9 & -13.0 $\pm$ 2.8 & 1.5 $\pm$ 7.8 & \textbf{31.0 $\pm$ 2.8} & 19.5 $\pm$ 0.7 & 22.5 $\pm$ 24.7 \\
 & sc-spp & 5.5 $\pm$ 14.8 & 3.5 $\pm$ 7.8 & 2.5 $\pm$ 3.5 & 2.0 $\pm$ 8.5 & \textbf{33.0 $\pm$ 2.8} & 11.0 $\pm$ 4.2 & 24.5 $\pm$ 14.8 \\
\cline{1-9}
\multirow[t]{6}{*}{C3.7S(T)} & zs & 4.0 $\pm$ 7.0 & -6.2 $\pm$ 12.2 & 0.8 $\pm$ 14.7 & 7.6 $\pm$ 6.1 & -8.6 $\pm$ 13.7 & 13.6 $\pm$ 4.8 & \textbf{15.6 $\pm$ 12.0} \\
 & cot & 6.8 $\pm$ 7.3 & -1.2 $\pm$ 11.5 & -1.0 $\pm$ 6.0 & -0.2 $\pm$ 4.3 & \textbf{18.4 $\pm$ 20.2} & 15.0 $\pm$ 4.4 & 16.6 $\pm$ 18.5 \\
 & spp & 6.8 $\pm$ 5.6 & -0.8 $\pm$ 10.8 & 7.8 $\pm$ 13.6 & -1.6 $\pm$ 4.6 & \textbf{25.8 $\pm$ 15.1} & 14.0 $\pm$ 4.6 & 19.6 $\pm$ 13.6 \\
 & sc-zs & -1.5 $\pm$ 7.8 & -1.0 $\pm$ 12.7 & -14.0 $\pm$ 11.3 & 4.0 $\pm$ 1.4 & -14.0 $\pm$ 8.5 & 8.0 $\pm$ 0.0 & \textbf{9.0 $\pm$ 5.7} \\
 & sc-cot & 2.0 $\pm$ 7.1 & -13.0 $\pm$ 0.0 & 3.5 $\pm$ 6.4 & -2.0 $\pm$ 1.4 & \textbf{35.0 $\pm$ 7.1} & 16.5 $\pm$ 2.1 & 30.5 $\pm$ 13.4 \\
 & sc-spp & 10.0 $\pm$ 14.1 & 5.0 $\pm$ 26.9 & -3.0 $\pm$ 1.4 & 2.5 $\pm$ 6.4 & \textbf{33.0 $\pm$ 9.9} & 16.5 $\pm$ 0.7 & 28.5 $\pm$ 0.7 \\
\cline{1-9}
\multirow[t]{6}{*}{C4S} & zs & -2.6 $\pm$ 12.7 & -11.2 $\pm$ 8.9 & -0.6 $\pm$ 4.9 & 2.2 $\pm$ 7.9 & 7.4 $\pm$ 27.3 & \textbf{15.6 $\pm$ 2.3} & 10.2 $\pm$ 21.1 \\
 & cot & 7.0 $\pm$ 2.5 & -2.8 $\pm$ 13.1 & -4.8 $\pm$ 12.6 & 4.4 $\pm$ 9.2 & \textbf{35.4 $\pm$ 4.4} & 8.0 $\pm$ 11.9 & 6.8 $\pm$ 4.4 \\
 & spp & 2.8 $\pm$ 10.0 & -6.8 $\pm$ 8.2 & 2.4 $\pm$ 11.9 & -3.4 $\pm$ 12.8 & \textbf{24.4 $\pm$ 13.8} & 15.6 $\pm$ 2.1 & 7.6 $\pm$ 12.9 \\
 & sc-zs & 5.0 $\pm$ 7.1 & -6.0 $\pm$ 5.7 & -7.0 $\pm$ 4.2 & 4.0 $\pm$ 9.9 & \textbf{15.0 $\pm$ 14.1} & 13.5 $\pm$ 4.9 & -1.5 $\pm$ 6.4 \\
 & sc-cot & 8.0 $\pm$ 19.8 & -8.5 $\pm$ 7.8 & 0.0 $\pm$ 5.7 & 4.5 $\pm$ 12.0 & \textbf{36.0 $\pm$ 5.7} & 8.5 $\pm$ 0.7 & 3.0 $\pm$ 1.4 \\
 & sc-spp & -6.5 $\pm$ 10.6 & 5.5 $\pm$ 4.9 & -0.5 $\pm$ 10.6 & -8.0 $\pm$ 14.1 & 27.5 $\pm$ 6.4 & 11.0 $\pm$ 1.4 & \textbf{33.5 $\pm$ 9.2} \\
\cline{1-9}
\multirow[t]{6}{*}{C4S(T)} & zs & -3.2 $\pm$ 6.1 & -8.6 $\pm$ 17.8 & 4.4 $\pm$ 15.1 & 5.8 $\pm$ 4.6 & -16.2 $\pm$ 11.1 & \textbf{12.0 $\pm$ 2.0} & 8.8 $\pm$ 16.5 \\
 & cot & 1.6 $\pm$ 5.1 & 2.0 $\pm$ 13.8 & -1.2 $\pm$ 14.9 & 5.4 $\pm$ 8.6 & 9.8 $\pm$ 27.8 & 12.0 $\pm$ 4.5 & \textbf{21.6 $\pm$ 11.3} \\
 & spp & 0.4 $\pm$ 8.7 & 9.2 $\pm$ 9.7 & 2.0 $\pm$ 4.1 & -2.4 $\pm$ 7.8 & \textbf{32.0 $\pm$ 4.9} & 7.6 $\pm$ 7.7 & 8.8 $\pm$ 10.7 \\
 & sc-zs & -3.0 $\pm$ 4.2 & -5.5 $\pm$ 3.5 & 13.0 $\pm$ 4.2 & -3.5 $\pm$ 12.0 & -0.5 $\pm$ 29.0 & \textbf{16.0 $\pm$ 5.7} & -12.5 $\pm$ 2.1 \\
 & sc-cot & -3.5 $\pm$ 16.3 & -9.5 $\pm$ 9.2 & -16.0 $\pm$ 7.1 & -1.0 $\pm$ 9.9 & -2.5 $\pm$ 0.7 & \textbf{15.0 $\pm$ 1.4} & 8.0 $\pm$ 7.1 \\
 & sc-spp & 10.5 $\pm$ 0.7 & -1.5 $\pm$ 10.6 & -3.0 $\pm$ 1.4 & -0.5 $\pm$ 7.8 & \textbf{23.5 $\pm$ 0.7} & 12.5 $\pm$ 3.5 & 14.5 $\pm$ 10.6 \\
\cline{1-9}
\multirow[t]{6}{*}{DS-R1} & zs & 2.4 $\pm$ 7.9 & -3.4 $\pm$ 8.7 & 0.2 $\pm$ 8.8 & -4.0 $\pm$ 8.9 & -12.6 $\pm$ 5.8 & \textbf{12.4 $\pm$ 3.8} & -1.4 $\pm$ 7.3 \\
 & cot & -4.8 $\pm$ 11.6 & -0.2 $\pm$ 0.8 & -4.2 $\pm$ 4.3 & -3.2 $\pm$ 2.8 & -7.2 $\pm$ 6.9 & 8.0 $\pm$ 5.7 & \textbf{8.2 $\pm$ 4.3} \\
 & spp & -4.8 $\pm$ 6.9 & -2.8 $\pm$ 4.8 & -1.8 $\pm$ 7.1 & 1.4 $\pm$ 4.5 & -14.8 $\pm$ 5.3 & 9.4 $\pm$ 7.7 & \textbf{10.6 $\pm$ 9.2} \\
 & sc-zs & -0.5 $\pm$ 10.6 & 2.5 $\pm$ 3.5 & -1.5 $\pm$ 3.5 & 5.5 $\pm$ 2.1 & -14.0 $\pm$ 0.0 & \textbf{13.5 $\pm$ 0.7} & -0.5 $\pm$ 0.7 \\
 & sc-cot & 6.5 $\pm$ 2.1 & 4.0 $\pm$ 4.2 & -6.5 $\pm$ 2.1 & 6.5 $\pm$ 7.8 & -27.5 $\pm$ 0.7 & \textbf{15.5 $\pm$ 0.7} & 2.0 $\pm$ 5.7 \\
 & sc-spp & 3.5 $\pm$ 2.1 & 2.5 $\pm$ 3.5 & 0.0 $\pm$ 1.4 & 0.0 $\pm$ 8.5 & -16.5 $\pm$ 9.2 & \textbf{10.5 $\pm$ 3.5} & 4.0 $\pm$ 4.2 \\
\cline{1-9}
\multirow[t]{6}{*}{L3.3-70B} & zs & 0.0 $\pm$ 3.9 & 0.0 $\pm$ 5.5 & -1.2 $\pm$ 2.9 & 3.2 $\pm$ 3.3 & -16.2 $\pm$ 16.5 & 5.4 $\pm$ 5.9 & \textbf{7.8 $\pm$ 2.9} \\
 & cot & -4.8 $\pm$ 2.9 & 2.2 $\pm$ 2.7 & -0.6 $\pm$ 5.6 & -0.4 $\pm$ 3.2 & -8.8 $\pm$ 8.6 & \textbf{7.2 $\pm$ 4.4} & 5.2 $\pm$ 5.5 \\
 & spp & -6.6 $\pm$ 10.4 & 1.0 $\pm$ 3.3 & 4.4 $\pm$ 3.4 & -4.2 $\pm$ 7.2 & -11.8 $\pm$ 9.7 & \textbf{11.0 $\pm$ 3.7} & 3.8 $\pm$ 2.8 \\
 & sc-zs & 0.5 $\pm$ 0.7 & 5.0 $\pm$ 1.4 & 2.5 $\pm$ 0.7 & -1.5 $\pm$ 4.9 & -20.5 $\pm$ 27.6 & \textbf{6.5 $\pm$ 9.2} & \textbf{6.5 $\pm$ 9.2} \\
 & sc-cot & -4.0 $\pm$ 0.0 & 4.0 $\pm$ 2.8 & -0.5 $\pm$ 3.5 & 2.5 $\pm$ 2.1 & -4.5 $\pm$ 4.9 & \textbf{5.0 $\pm$ 5.7} & 3.5 $\pm$ 2.1 \\
 & sc-spp & -6.0 $\pm$ 1.4 & -0.5 $\pm$ 3.5 & 1.5 $\pm$ 7.8 & -10.0 $\pm$ 1.4 & -30.5 $\pm$ 13.4 & \textbf{11.5 $\pm$ 0.7} & 0.5 $\pm$ 0.7 \\
\cline{1-9}
\multirow[t]{6}{*}{Mistral} & zs & 0.0 $\pm$ 6.4 & 1.2 $\pm$ 8.0 & -0.4 $\pm$ 10.5 & 0.0 $\pm$ 6.5 & -11.6 $\pm$ 17.6 & -16.4 $\pm$ 13.8 & \textbf{10.2 $\pm$ 9.7} \\
 & cot & 2.0 $\pm$ 12.9 & -0.2 $\pm$ 6.9 & 6.8 $\pm$ 8.5 & -0.6 $\pm$ 8.1 & -14.6 $\pm$ 12.5 & -11.2 $\pm$ 14.9 & \textbf{21.8 $\pm$ 8.9} \\
 & spp & 0.4 $\pm$ 10.7 & 2.8 $\pm$ 8.9 & 0.0 $\pm$ 15.0 & 4.2 $\pm$ 5.3 & \textbf{17.4 $\pm$ 18.4} & 13.8 $\pm$ 13.5 & 14.6 $\pm$ 8.4 \\
 & sc-zs & 1.0 $\pm$ 26.9 & 2.0 $\pm$ 1.4 & 1.0 $\pm$ 5.7 & 2.5 $\pm$ 4.9 & -40.0 $\pm$ 0.0 & -4.5 $\pm$ 20.5 & \textbf{8.5 $\pm$ 14.8} \\
 & sc-cot & -3.5 $\pm$ 3.5 & 4.5 $\pm$ 4.9 & 9.5 $\pm$ 14.8 & 6.5 $\pm$ 3.5 & -16.0 $\pm$ 14.1 & -6.0 $\pm$ 21.2 & \textbf{19.5 $\pm$ 21.9} \\
 & sc-spp & 9.5 $\pm$ 33.2 & -1.0 $\pm$ 11.3 & -4.5 $\pm$ 21.9 & 1.0 $\pm$ 2.8 & -32.0 $\pm$ 11.3 & 8.5 $\pm$ 16.3 & \textbf{29.0 $\pm$ 8.5} \\
\bottomrule
\end{tabular}
\caption{Total Points Averaged Over All Iterations for the joint RPS counterfactual.}
\label{tab:rps_total_points_avg_heatmap_ba3-alt}
\end{table*}
Finally, joint counterfactual results are presented in Table \ref{tab:rps_total_points_avg_heatmap_ba3-alt}. This proves to be the hardest RPS setting overall, since models must handle both the label permutation and the payoff asymmetry at the same time. As a result, the table shows the strongest instability: many model–prompt combinations that were competitive in default RPS or in the single-intervention counterfactuals now collapse to near-zero or strongly negative scores, while only a subset of stronger models still manage to exploit structured opponents reliably.

The clearest pattern is that LLM–LLM and SREP columns remain mostly near zero or highly variable, which is expected because the joint counterfactual makes stable equilibrium adaptation harder. The real separation appears against structured algorithmic opponents, especially PP and TFT, and to a lesser extent AP. For the strongest models, these are still the main sources of positive reward, but performance is much less uniform than in the payoff-only setup. In particular, Claude 3.7 Sonnet, Claude 3.7 Thinking, and parts of the Claude 4 family remain the strongest overall: they still achieve high totals against PP and TFT, often in the 20s or low 30s, showing partial adaptation to the altered dominance and incentive structure. Among them, Claude 3.7 appears the most robust on average, with repeated strong scores across several prompts and opponents.

By contrast, DeepSeek R1, Llama 3.3, and especially Mistral are much more brittle here. DeepSeek tends to obtain its best results mainly against AP, while frequently performing poorly or negatively against PP, suggesting that it does not consistently internalize the jointly modified structure. Llama 3.3 is even more telling: compared to the payoff-only counterfactual, it drops substantially, with many negative scores against PP and only modest gains against AP/TFT, indicating weak robustness to the combined intervention. Mistral remains the most unstable model overall, with very large swings and some extreme failures, such as highly negative PP scores under SC, confirming that the joint counterfactual is particularly damaging for weaker or less consistent models.

Prompting remains mixed, but the joint setting again suggests that self-consistency can help only when the base reasoning is already somewhat aligned with the new structure. For stronger Claude models, SC sometimes lifts PP/TFT performance further, but for weaker models it often amplifies bad policies instead of correcting them. This is especially visible in rows with consistently negative PP scores despite SC.

Overall, this table provides some of the strongest evidence for your paper’s core claim: strategic competence in default RPS does not reliably transfer when both labels and payoffs are changed simultaneously. The joint counterfactual exposes the sharpest reasoning brittleness, with only the strongest Claude-family models retaining substantial exploitation ability, while weaker models and some prompt variants fail to recompute the altered game structure.

\subsection{RPS Opponent comprehension}
\begin{table*}[h!]
\small
\centering
\begin{tabular}{p{1cm}p{0.9cm}lllllll}
\toprule
 &  & \multicolumn{7}{c}{\textbf{RPS}} \\
 &  & zs & spp & cot & srep & pp & ap & tft \\
model & prompt &  &  &  &  &  &  &  \\
\midrule
\multirow[t]{6}{*}{C3.5Sv2} & zs & 10.6 $\pm$ 13.1 & 21.4 $\pm$ 4.6 & 19.6 $\pm$ 5.6 & 21.0 $\pm$ 3.5 & 14.6 $\pm$ 12.4 & 10.4 $\pm$ 11.2 & \textbf{1.0 $\pm$ 0.0} \\
 & cot & 17.2 $\pm$ 7.3 & 19.6 $\pm$ 7.5 & 20.6 $\pm$ 5.6 & 19.6 $\pm$ 7.1 & 16.0 $\pm$ 11.9 & 22.6 $\pm$ 2.1 & \textbf{5.8 $\pm$ 10.7} \\
 & spp & 11.2 $\pm$ 10.6 & 11.8 $\pm$ 10.9 & 20.4 $\pm$ 5.9 & 23.0 $\pm$ 1.9 & 11.4 $\pm$ 11.3 & 18.4 $\pm$ 9.3 & \textbf{1.6 $\pm$ 0.9} \\
 & sc-zs & 25.0 $\pm$ 0.0 & 14.0 $\pm$ 15.6 & 21.5 $\pm$ 4.9 & 13.5 $\pm$ 16.3 & 12.0 $\pm$ 15.6 & 19.5 $\pm$ 6.4 & \textbf{1.0 $\pm$ 0.0} \\
 & sc-cot & 15.0 $\pm$ 14.1 & 20.5 $\pm$ 0.7 & 22.5 $\pm$ 3.5 & 25.0 $\pm$ 0.0 & 12.5 $\pm$ 16.3 & 7.0 $\pm$ 8.5 & \textbf{1.0 $\pm$ 0.0} \\
 & sc-spp & 14.5 $\pm$ 14.8 & 15.0 $\pm$ 14.1 & 14.5 $\pm$ 14.8 & 24.0 $\pm$ 1.4 & 16.0 $\pm$ 8.5 & 21.0 $\pm$ 0.0 & \textbf{1.0 $\pm$ 0.0} \\
\cline{1-9}
\multirow[t]{6}{*}{C3.7S} & zs & 21.8 $\pm$ 5.5 & 22.2 $\pm$ 4.2 & 20.2 $\pm$ 9.7 & 23.4 $\pm$ 2.5 & \textbf{10.0 $\pm$ 12.4} & 17.6 $\pm$ 9.3 & 15.0 $\pm$ 12.8 \\
 & cot & 15.4 $\pm$ 7.1 & 16.2 $\pm$ 10.0 & 18.8 $\pm$ 4.3 & 18.0 $\pm$ 7.4 & \textbf{1.0 $\pm$ 0.0} & 15.6 $\pm$ 10.1 & \textbf{1.0 $\pm$ 0.0} \\
 & spp & 24.4 $\pm$ 1.3 & 20.0 $\pm$ 5.6 & 22.4 $\pm$ 4.3 & 23.6 $\pm$ 1.5 & \textbf{1.2 $\pm$ 0.4} & 12.0 $\pm$ 10.6 & \textbf{1.2 $\pm$ 0.4} \\
 & sc-zs & 13.0 $\pm$ 17.0 & 23.0 $\pm$ 0.0 & 24.5 $\pm$ 0.7 & 22.0 $\pm$ 2.8 & \textbf{11.5 $\pm$ 14.8} & 15.0 $\pm$ 14.1 & 12.5 $\pm$ 16.3 \\
 & sc-cot & 4.5 $\pm$ 4.9 & 24.5 $\pm$ 0.7 & 11.5 $\pm$ 14.8 & 24.0 $\pm$ 1.4 & \textbf{1.0 $\pm$ 0.0} & 24.5 $\pm$ 0.7 & \textbf{1.0 $\pm$ 0.0} \\
 & sc-spp & 9.5 $\pm$ 12.0 & 21.5 $\pm$ 3.5 & 22.0 $\pm$ 1.4 & \textbf{12.5 $\pm$ 16.3} & \textbf{1.0 $\pm$ 0.0} & 22.5 $\pm$ 2.1 & \textbf{1.0 $\pm$ 0.0} \\
\cline{1-9}
\multirow[t]{6}{*}{C3.7S(T)} & zs & 24.2 $\pm$ 1.3 & 19.4 $\pm$ 8.3 & 22.6 $\pm$ 4.8 & 22.4 $\pm$ 2.2 & \textbf{1.4 $\pm$ 0.9} & 22.6 $\pm$ 1.7 & 1.6 $\pm$ 0.5 \\
 & cot & 19.2 $\pm$ 5.1 & 21.8 $\pm$ 4.9 & 22.0 $\pm$ 2.7 & 23.8 $\pm$ 2.2 & \textbf{1.0 $\pm$ 0.0} & 15.6 $\pm$ 12.0 & 4.4 $\pm$ 7.6 \\
 & spp & 23.6 $\pm$ 1.5 & 20.6 $\pm$ 4.3 & 21.2 $\pm$ 5.3 & 15.2 $\pm$ 11.3 & \textbf{1.0 $\pm$ 0.0} & 17.8 $\pm$ 9.7 & 1.2 $\pm$ 0.4 \\
 & sc-zs & 24.5 $\pm$ 0.7 & 22.5 $\pm$ 3.5 & 13.0 $\pm$ 17.0 & 22.5 $\pm$ 3.5 & \textbf{1.0 $\pm$ 0.0} & 12.0 $\pm$ 15.6 & \textbf{1.0 $\pm$ 0.0} \\
 & sc-cot & 24.5 $\pm$ 0.7 & 25.0 $\pm$ 0.0 & 25.0 $\pm$ 0.0 & 19.5 $\pm$ 7.8 & \textbf{1.0 $\pm$ 0.0} & 10.0 $\pm$ 12.7 & \textbf{1.0 $\pm$ 0.0} \\
 & sc-spp & 24.0 $\pm$ 1.4 & 25.0 $\pm$ 0.0 & 25.0 $\pm$ 0.0 & 25.0 $\pm$ 0.0 & \textbf{1.0 $\pm$ 0.0} & 23.0 $\pm$ 0.0 & \textbf{1.0 $\pm$ 0.0} \\
\cline{1-9}
\multirow[t]{6}{*}{C4S} & zs & 15.8 $\pm$ 8.5 & 20.0 $\pm$ 10.6 & 23.8 $\pm$ 2.2 & 22.2 $\pm$ 0.8 & \textbf{1.0 $\pm$ 0.0} & 23.2 $\pm$ 1.1 & 11.8 $\pm$ 11.3 \\
 & cot & 15.2 $\pm$ 13.0 & 15.6 $\pm$ 1.9 & 10.8 $\pm$ 11.8 & 19.4 $\pm$ 8.6 & \textbf{1.0 $\pm$ 0.0} & 19.0 $\pm$ 10.1 & 3.2 $\pm$ 4.9 \\
 & spp & 16.0 $\pm$ 10.5 & 20.8 $\pm$ 9.4 & 18.6 $\pm$ 10.0 & 19.2 $\pm$ 5.5 & \textbf{3.4 $\pm$ 4.8} & 14.4 $\pm$ 12.3 & 9.0 $\pm$ 8.6 \\
 & sc-zs & 7.5 $\pm$ 9.2 & 25.0 $\pm$ 0.0 & 25.0 $\pm$ 0.0 & 22.0 $\pm$ 4.2 & \textbf{1.0 $\pm$ 0.0} & 23.0 $\pm$ 1.4 & \textbf{1.0 $\pm$ 0.0} \\
 & sc-cot & 21.5 $\pm$ 4.9 & 15.0 $\pm$ 14.1 & 14.5 $\pm$ 14.8 & 21.0 $\pm$ 1.4 & \textbf{1.0 $\pm$ 0.0} & 24.0 $\pm$ 1.4 & \textbf{1.0 $\pm$ 0.0} \\
 & sc-spp & 10.0 $\pm$ 7.1 & 11.0 $\pm$ 8.5 & 23.0 $\pm$ 2.8 & 21.5 $\pm$ 0.7 & \textbf{1.0 $\pm$ 0.0} & 13.0 $\pm$ 17.0 & \textbf{1.0 $\pm$ 0.0} \\
\cline{1-9}
\multirow[t]{6}{*}{C4S(T)} & zs & 13.4 $\pm$ 12.0 & 21.0 $\pm$ 5.3 & 17.0 $\pm$ 10.0 & 17.6 $\pm$ 8.5 & \textbf{3.4 $\pm$ 5.4} & 22.6 $\pm$ 1.8 & 10.6 $\pm$ 13.1 \\
 & cot & 14.4 $\pm$ 12.3 & 21.8 $\pm$ 2.7 & 20.2 $\pm$ 10.7 & 21.6 $\pm$ 4.2 & \textbf{1.0 $\pm$ 0.0} & 19.6 $\pm$ 10.4 & 3.2 $\pm$ 4.9 \\
 & spp & 24.2 $\pm$ 1.1 & 15.2 $\pm$ 9.9 & 17.8 $\pm$ 8.6 & 18.8 $\pm$ 10.1 & \textbf{1.0 $\pm$ 0.0} & 19.4 $\pm$ 10.4 & 5.6 $\pm$ 6.5 \\
 & sc-zs & 13.0 $\pm$ 17.0 & 22.0 $\pm$ 4.2 & 24.0 $\pm$ 1.4 & 18.0 $\pm$ 5.7 & \textbf{1.0 $\pm$ 0.0} & 24.5 $\pm$ 0.7 & \textbf{1.0 $\pm$ 0.0} \\
 & sc-cot & \textbf{1.0 $\pm$ 0.0} & 7.0 $\pm$ 4.2 & 10.5 $\pm$ 13.4 & 23.5 $\pm$ 0.7 & \textbf{1.0 $\pm$ 0.0} & 11.0 $\pm$ 14.1 & \textbf{1.0 $\pm$ 0.0} \\
 & sc-spp & 24.0 $\pm$ 1.4 & 23.0 $\pm$ 2.8 & 24.5 $\pm$ 0.7 & 20.5 $\pm$ 2.1 & \textbf{1.0 $\pm$ 0.0} & 23.5 $\pm$ 0.7 & \textbf{1.0 $\pm$ 0.0} \\
\cline{1-9}
\multirow[t]{6}{*}{DS-R1} & zs & 21.2 $\pm$ 5.5 & 21.2 $\pm$ 5.3 & 19.4 $\pm$ 10.3 & 23.4 $\pm$ 1.3 & 23.2 $\pm$ 0.4 & 16.8 $\pm$ 11.1 & \textbf{7.8 $\pm$ 9.7} \\
 & cot & 21.2 $\pm$ 4.3 & 20.0 $\pm$ 5.5 & 24.4 $\pm$ 0.9 & 13.6 $\pm$ 10.9 & 12.8 $\pm$ 10.3 & 15.4 $\pm$ 12.7 & \textbf{1.0 $\pm$ 0.0} \\
 & spp & 18.0 $\pm$ 10.4 & 22.6 $\pm$ 2.3 & 20.6 $\pm$ 8.8 & 23.0 $\pm$ 1.4 & 20.2 $\pm$ 4.1 & 16.8 $\pm$ 8.1 & \textbf{6.2 $\pm$ 9.0} \\
 & sc-zs & 12.0 $\pm$ 15.6 & 21.0 $\pm$ 2.8 & 24.0 $\pm$ 1.4 & 22.5 $\pm$ 3.5 & 13.5 $\pm$ 16.3 & 22.5 $\pm$ 2.1 & \textbf{1.5 $\pm$ 0.7} \\
 & sc-cot & 12.0 $\pm$ 15.6 & 23.0 $\pm$ 1.4 & 19.0 $\pm$ 7.1 & 21.5 $\pm$ 4.9 & 3.5 $\pm$ 2.1 & \textbf{1.0 $\pm$ 0.0} & \textbf{1.0 $\pm$ 0.0} \\
 & sc-spp & 16.0 $\pm$ 8.5 & 24.0 $\pm$ 1.4 & 20.5 $\pm$ 3.5 & 23.0 $\pm$ 0.0 & 13.5 $\pm$ 16.3 & 24.5 $\pm$ 0.7 & \textbf{1.0 $\pm$ 0.0} \\
\cline{1-9}
\multirow[t]{6}{*}{L3.3-70B} & zs & 5.8 $\pm$ 10.7 & 16.2 $\pm$ 10.4 & 12.4 $\pm$ 11.8 & 21.8 $\pm$ 2.3 & \textbf{1.0 $\pm$ 0.0} & 12.8 $\pm$ 7.9 & \textbf{1.0 $\pm$ 0.0} \\
 & cot & 5.6 $\pm$ 10.3 & \textbf{5.8 $\pm$ 10.7} & 10.0 $\pm$ 12.4 & 23.0 $\pm$ 1.9 & 2.8 $\pm$ 4.0 & 21.2 $\pm$ 4.1 & \textbf{1.0 $\pm$ 0.0} \\
 & spp & \textbf{1.0 $\pm$ 0.0} & 7.8 $\pm$ 10.0 & 21.2 $\pm$ 5.8 & 21.8 $\pm$ 5.5 & 8.6 $\pm$ 11.0 & 15.2 $\pm$ 9.4 & \textbf{1.0 $\pm$ 0.0} \\
 & sc-zs & \textbf{1.0 $\pm$ 0.0} & 13.0 $\pm$ 17.0 & 20.0 $\pm$ 7.1 & 25.0 $\pm$ 0.0 & \textbf{1.0 $\pm$ 0.0} & 5.5 $\pm$ 4.9 & \textbf{1.0 $\pm$ 0.0} \\
 & sc-cot & 10.5 $\pm$ 13.4 & 23.5 $\pm$ 2.1 & \textbf{1.0 $\pm$ 0.0} & 25.0 $\pm$ 0.0 & \textbf{1.0 $\pm$ 0.0} & 12.5 $\pm$ 16.3 & \textbf{1.0 $\pm$ 0.0} \\
 & sc-spp & \textbf{1.0 $\pm$ 0.0} & 13.0 $\pm$ 17.0 & 24.5 $\pm$ 0.7 & 16.0 $\pm$ 8.5 & 2.5 $\pm$ 2.1 & 12.0 $\pm$ 15.6 & \textbf{1.0 $\pm$ 0.0} \\
\cline{1-9}
\multirow[t]{6}{*}{Mistral} & zs & 1.2 $\pm$ 0.4 & 19.2 $\pm$ 10.3 & 23.6 $\pm$ 1.1 & 22.6 $\pm$ 4.8 & \textbf{1.0 $\pm$ 0.0} & 6.0 $\pm$ 10.6 & \textbf{1.0 $\pm$ 0.0} \\
 & cot & 18.8 $\pm$ 10.4 & 20.2 $\pm$ 6.1 & 15.8 $\pm$ 10.7 & 21.2 $\pm$ 4.3 & 12.4 $\pm$ 12.1 & 17.0 $\pm$ 9.4 & \textbf{1.0 $\pm$ 0.0} \\
 & spp & 6.6 $\pm$ 10.4 & 10.6 $\pm$ 13.1 & 17.8 $\pm$ 9.6 & 23.0 $\pm$ 0.7 & 17.2 $\pm$ 10.3 & 19.8 $\pm$ 10.5 & \textbf{1.0 $\pm$ 0.0} \\
 & sc-zs & \textbf{1.0 $\pm$ 0.0} & 13.0 $\pm$ 17.0 & 24.5 $\pm$ 0.7 & 23.0 $\pm$ 0.0 & \textbf{1.0 $\pm$ 0.0} & 24.0 $\pm$ 0.0 & \textbf{1.0 $\pm$ 0.0} \\
 & sc-cot & \textbf{1.0 $\pm$ 0.0} & 24.0 $\pm$ 0.0 & 13.0 $\pm$ 17.0 & 24.5 $\pm$ 0.7 & 12.5 $\pm$ 16.3 & 24.5 $\pm$ 0.7 & \textbf{1.0 $\pm$ 0.0} \\
 & sc-spp & 12.5 $\pm$ 16.3 & 11.5 $\pm$ 14.8 & 12.5 $\pm$ 16.3 & 25.0 $\pm$ 0.0 & 11.5 $\pm$ 14.8 & 24.5 $\pm$ 0.7 & \textbf{6.5 $\pm$ 7.8} \\
\bottomrule
\end{tabular}
\caption{Round $m$ \# where the LLM player understood the opponent's Strategy in the default RPS setup. Lower is better; $m=25$ indicates no stable comprehension within the 24-round game.}
\label{tab:rps_round_heatmap_eq1}
\end{table*}

Table \ref{tab:rps_round_heatmap_eq1} presents opponent comprehension results for the default RPS game. opponent comprehension is markedly weaker and more variable than in PD, which is expected given the absence of a dominant response and the need to reason over three actions rather than two. In LLM--LLM interactions (first three columns), most values remain relatively high and often close to the horizon, indicating that models generally do not infer and exploit other LLMs’ play quickly; for instance, Claude 3.5 Sonnet v2 under zero-shot reaches only 
$m=10.6\pm 13.1$ against its ZS counterpart and rises to 
$21.4\pm 4.6$ and 
$19.6\pm 5.6$ against SPP and CoT counterparts, respectively. By contrast, algorithmic opponents are more informative: \textbf{PP} and especially \textbf{TFT} are the most exploitable, with Claude 3.7, Claude 3.7(T), Claude 4, Claude 4(T), and Llama 3.3 often reaching very early comprehension ($m \approx 1$) in several prompting settings, showing successful recognition of cyclic or reactive behavior. \textbf{AP} is noticeably harder, with comprehension typically delayed across models, while \textbf{SREP} also remains difficult for many rows, consistent with the fact that an equilibrium-style opponent offers little exploitable structure. Overall, the strongest and most consistent RPS comprehension is shown by the advanced Claude models, Llama 3.3 acts as a competitive but less stable runner-up, DeepSeek R1 is mixed, and Mistral Large remains the most inconsistent despite a few isolated low-$m$ outcomes. This confirms that, unlike PD, default RPS rarely leads to early and stable mutual understanding; instead, successful play mainly comes from detecting exploitable algorithmic regularities rather than quickly countering another LLM.

\begin{table*}[h!]
\small
\centering
\begin{tabular}{p{1cm}p{0.9cm}lllllll}
\toprule
 &  & \multicolumn{7}{c}{\textbf{Label-based RPS counterfactual}} \\
 &  & zs & spp & cot & srep & pp & ap & tft \\
model & prompt &  &  &  &  &  &  &  \\
\midrule
\multirow[t]{6}{*}{C3.5Sv2} & zs & 19.0 $\pm$ 10.3 & 22.4 $\pm$ 4.2 & 24.4 $\pm$ 1.3 & 20.4 $\pm$ 4.2 & 21.6 $\pm$ 3.8 & \textbf{11.4 $\pm$ 12.5} & 14.8 $\pm$ 11.4 \\
 & cot & 21.8 $\pm$ 6.1 & \textbf{19.6 $\pm$ 7.3} & 19.8 $\pm$ 6.8 & 20.8 $\pm$ 5.1 & 21.6 $\pm$ 6.5 & 24.4 $\pm$ 0.9 & 23.6 $\pm$ 1.7 \\
 & spp & 21.8 $\pm$ 2.6 & 24.0 $\pm$ 0.7 & \textbf{14.6 $\pm$ 8.7} & 22.6 $\pm$ 3.0 & 18.4 $\pm$ 9.0 & 22.8 $\pm$ 2.7 & 16.4 $\pm$ 11.3 \\
 & sc-zs & \textbf{7.5 $\pm$ 9.2} & 22.5 $\pm$ 0.7 & 24.0 $\pm$ 1.4 & \textbf{12.0 $\pm$ 4.2} & 25.0 $\pm$ 0.0 & 19.0 $\pm$ 5.7 & 13.0 $\pm$ 17.0 \\
 & sc-cot & 23.0 $\pm$ 1.4 & \textbf{11.5 $\pm$ 3.5} & 25.0 $\pm$ 0.0 & 22.5 $\pm$ 0.7 & 25.0 $\pm$ 0.0 & 25.0 $\pm$ 0.0 & 24.0 $\pm$ 1.4 \\
 & sc-spp & 19.0 $\pm$ 8.5 & 12.5 $\pm$ 16.3 & 18.5 $\pm$ 9.2 & 17.0 $\pm$ 8.5 & 25.0 $\pm$ 0.0 & 25.0 $\pm$ 0.0 & \textbf{12.0 $\pm$ 15.6} \\
\cline{1-9}
\multirow[t]{6}{*}{C3.7S} & zs & 23.4 $\pm$ 1.8 & 21.0 $\pm$ 5.3 & 22.6 $\pm$ 2.3 & 19.2 $\pm$ 7.2 & 21.2 $\pm$ 5.8 & 22.8 $\pm$ 0.8 & \textbf{7.8 $\pm$ 10.0} \\
 & cot & 19.8 $\pm$ 10.0 & 19.2 $\pm$ 10.2 & 24.0 $\pm$ 1.0 & 23.2 $\pm$ 2.0 & 3.2 $\pm$ 3.9 & 18.4 $\pm$ 9.8 & \textbf{1.6 $\pm$ 1.3} \\
 & spp & 18.4 $\pm$ 9.5 & 19.4 $\pm$ 8.3 & 21.6 $\pm$ 6.5 & 22.2 $\pm$ 5.2 & 2.4 $\pm$ 1.9 & 23.6 $\pm$ 2.1 & \textbf{1.0 $\pm$ 0.0} \\
 & sc-zs & 25.0 $\pm$ 0.0 & 22.0 $\pm$ 2.8 & 20.0 $\pm$ 7.1 & 19.0 $\pm$ 5.7 & \textbf{16.5 $\pm$ 12.0} & 25.0 $\pm$ 0.0 & 23.0 $\pm$ 2.8 \\
 & sc-cot & 25.0 $\pm$ 0.0 & 21.5 $\pm$ 0.7 & 18.5 $\pm$ 4.9 & 25.0 $\pm$ 0.0 & \textbf{4.0 $\pm$ 4.2} & 21.5 $\pm$ 2.1 & 13.0 $\pm$ 17.0 \\
 & sc-spp & 12.0 $\pm$ 2.8 & 13.0 $\pm$ 17.0 & 24.0 $\pm$ 0.0 & 25.0 $\pm$ 0.0 & 5.0 $\pm$ 0.0 & 22.5 $\pm$ 2.1 & \textbf{1.0 $\pm$ 0.0} \\
\cline{1-9}
\multirow[t]{6}{*}{C3.7S(T)} & zs & 22.6 $\pm$ 4.3 & 24.6 $\pm$ 0.9 & 23.6 $\pm$ 1.3 & 22.4 $\pm$ 2.2 & 15.0 $\pm$ 9.7 & 17.6 $\pm$ 10.0 & \textbf{10.4 $\pm$ 12.9} \\
 & cot & 19.2 $\pm$ 4.4 & 22.4 $\pm$ 2.6 & 22.0 $\pm$ 2.6 & 23.2 $\pm$ 1.9 & 10.6 $\pm$ 13.1 & 21.8 $\pm$ 1.8 & \textbf{7.2 $\pm$ 9.7} \\
 & spp & 17.4 $\pm$ 10.4 & 21.6 $\pm$ 4.9 & 23.6 $\pm$ 0.5 & 21.6 $\pm$ 3.8 & 3.4 $\pm$ 3.9 & 24.0 $\pm$ 1.0 & \textbf{1.2 $\pm$ 0.4} \\
 & sc-zs & 24.5 $\pm$ 0.7 & 22.0 $\pm$ 4.2 & 12.5 $\pm$ 16.3 & 22.5 $\pm$ 3.5 & 3.0 $\pm$ 2.8 & 23.0 $\pm$ 0.0 & \textbf{1.0 $\pm$ 0.0} \\
 & sc-cot & 19.0 $\pm$ 8.5 & 13.0 $\pm$ 17.0 & 25.0 $\pm$ 0.0 & 22.5 $\pm$ 0.7 & 11.5 $\pm$ 14.8 & 19.0 $\pm$ 7.1 & \textbf{1.0 $\pm$ 0.0} \\
 & sc-spp & 3.0 $\pm$ 1.4 & 16.0 $\pm$ 7.1 & 24.5 $\pm$ 0.7 & 25.0 $\pm$ 0.0 & \textbf{1.0 $\pm$ 0.0} & 24.5 $\pm$ 0.7 & \textbf{1.0 $\pm$ 0.0} \\
\cline{1-9}
\multirow[t]{6}{*}{C4S} & zs & 19.4 $\pm$ 10.3 & 21.6 $\pm$ 7.1 & 24.0 $\pm$ 2.2 & 22.8 $\pm$ 3.0 & 16.2 $\pm$ 10.3 & 21.6 $\pm$ 6.5 & \textbf{9.4 $\pm$ 9.2} \\
 & cot & 13.2 $\pm$ 10.2 & 24.2 $\pm$ 1.8 & 23.8 $\pm$ 1.3 & 24.0 $\pm$ 1.2 & \textbf{4.8 $\pm$ 2.8} & 18.0 $\pm$ 9.6 & 13.6 $\pm$ 6.8 \\
 & spp & 19.6 $\pm$ 10.4 & 17.0 $\pm$ 10.1 & 17.6 $\pm$ 10.9 & 19.6 $\pm$ 4.3 & \textbf{4.2 $\pm$ 2.2} & 21.2 $\pm$ 4.0 & 21.2 $\pm$ 5.4 \\
 & sc-zs & 25.0 $\pm$ 0.0 & 22.5 $\pm$ 3.5 & 24.5 $\pm$ 0.7 & 24.5 $\pm$ 0.7 & \textbf{2.5 $\pm$ 0.7} & 25.0 $\pm$ 0.0 & 13.0 $\pm$ 17.0 \\
 & sc-cot & 24.0 $\pm$ 0.0 & 23.5 $\pm$ 2.1 & 24.0 $\pm$ 1.4 & 12.5 $\pm$ 16.3 & \textbf{2.5 $\pm$ 2.1} & 25.0 $\pm$ 0.0 & 20.5 $\pm$ 3.5 \\
 & sc-spp & 23.5 $\pm$ 0.7 & 24.0 $\pm$ 1.4 & 23.5 $\pm$ 0.7 & 24.0 $\pm$ 1.4 & \textbf{3.0 $\pm$ 2.8} & 15.0 $\pm$ 8.5 & 24.5 $\pm$ 0.7 \\
\cline{1-9}
\multirow[t]{6}{*}{C4S(T)} & zs & 22.2 $\pm$ 4.7 & 23.8 $\pm$ 0.8 & 24.2 $\pm$ 0.8 & 22.4 $\pm$ 4.2 & 20.6 $\pm$ 9.3 & 21.4 $\pm$ 4.5 & \textbf{16.4 $\pm$ 9.7} \\
 & cot & 24.8 $\pm$ 0.4 & 24.6 $\pm$ 0.5 & 22.6 $\pm$ 4.3 & 21.8 $\pm$ 4.4 & \textbf{12.2 $\pm$ 10.7} & 21.0 $\pm$ 5.7 & 17.6 $\pm$ 10.6 \\
 & spp & 23.6 $\pm$ 1.1 & 21.8 $\pm$ 4.9 & 22.4 $\pm$ 2.1 & 23.4 $\pm$ 0.5 & \textbf{7.8 $\pm$ 10.0} & 22.0 $\pm$ 1.6 & 14.8 $\pm$ 12.7 \\
 & sc-zs & 25.0 $\pm$ 0.0 & 24.5 $\pm$ 0.7 & 25.0 $\pm$ 0.0 & 23.0 $\pm$ 1.4 & 24.5 $\pm$ 0.7 & \textbf{22.0 $\pm$ 1.4} & 24.5 $\pm$ 0.7 \\
 & sc-cot & 21.5 $\pm$ 4.9 & 9.0 $\pm$ 11.3 & 24.0 $\pm$ 1.4 & 25.0 $\pm$ 0.0 & \textbf{1.0 $\pm$ 0.0} & 22.0 $\pm$ 1.4 & 12.0 $\pm$ 15.6 \\
 & sc-spp & 18.5 $\pm$ 7.8 & 24.0 $\pm$ 1.4 & 13.0 $\pm$ 17.0 & 21.5 $\pm$ 3.5 & \textbf{6.5 $\pm$ 3.5} & 23.0 $\pm$ 1.4 & 13.0 $\pm$ 17.0 \\
\cline{1-9}
\multirow[t]{6}{*}{DS-R1} & zs & \textbf{14.4 $\pm$ 12.3} & 23.6 $\pm$ 2.1 & 17.0 $\pm$ 6.5 & 20.6 $\pm$ 4.6 & 23.8 $\pm$ 0.8 & 23.0 $\pm$ 1.0 & 19.8 $\pm$ 4.9 \\
 & cot & \textbf{15.8 $\pm$ 10.8} & 23.2 $\pm$ 1.3 & 23.6 $\pm$ 2.6 & 24.0 $\pm$ 1.7 & 24.6 $\pm$ 0.9 & 21.2 $\pm$ 3.5 & 16.8 $\pm$ 10.5 \\
 & spp & 23.2 $\pm$ 2.2 & 21.0 $\pm$ 3.7 & \textbf{18.6 $\pm$ 8.8} & 23.2 $\pm$ 0.8 & 24.6 $\pm$ 0.9 & 22.6 $\pm$ 1.1 & 23.4 $\pm$ 0.5 \\
 & sc-zs & 22.5 $\pm$ 3.5 & 16.5 $\pm$ 12.0 & \textbf{8.0 $\pm$ 9.9} & 23.0 $\pm$ 2.8 & 25.0 $\pm$ 0.0 & 24.0 $\pm$ 0.0 & 16.5 $\pm$ 3.5 \\
 & sc-cot & 24.0 $\pm$ 1.4 & 22.0 $\pm$ 2.8 & 18.0 $\pm$ 9.9 & 24.5 $\pm$ 0.7 & 25.0 $\pm$ 0.0 & \textbf{13.0 $\pm$ 0.0} & 15.0 $\pm$ 14.1 \\
 & sc-spp & 21.5 $\pm$ 4.9 & 19.0 $\pm$ 5.7 & 24.0 $\pm$ 1.4 & 24.0 $\pm$ 0.0 & 24.0 $\pm$ 0.0 & 25.0 $\pm$ 0.0 & \textbf{13.0 $\pm$ 17.0} \\
\cline{1-9}
\multirow[t]{6}{*}{L3.3-70B} & zs & 16.2 $\pm$ 9.8 & 7.4 $\pm$ 9.0 & 16.4 $\pm$ 9.6 & 23.8 $\pm$ 1.6 & 10.4 $\pm$ 9.5 & 18.6 $\pm$ 9.9 & \textbf{6.8 $\pm$ 9.5} \\
 & cot & 16.2 $\pm$ 9.1 & 10.2 $\pm$ 12.6 & 12.2 $\pm$ 11.1 & 20.2 $\pm$ 5.0 & 21.0 $\pm$ 5.2 & 19.8 $\pm$ 10.0 & \textbf{1.0 $\pm$ 0.0} \\
 & spp & 13.0 $\pm$ 11.4 & \textbf{1.4 $\pm$ 0.9} & 10.2 $\pm$ 12.6 & 23.0 $\pm$ 1.2 & 12.2 $\pm$ 9.8 & 18.8 $\pm$ 10.1 & 11.2 $\pm$ 10.7 \\
 & sc-zs & 5.0 $\pm$ 5.7 & 6.5 $\pm$ 3.5 & 12.5 $\pm$ 13.4 & 21.0 $\pm$ 1.4 & 14.0 $\pm$ 15.6 & 23.5 $\pm$ 2.1 & \textbf{1.0 $\pm$ 0.0} \\
 & sc-cot & 11.5 $\pm$ 14.8 & \textbf{8.5 $\pm$ 6.4} & 13.0 $\pm$ 17.0 & 20.5 $\pm$ 4.9 & 14.0 $\pm$ 15.6 & 24.0 $\pm$ 1.4 & 13.0 $\pm$ 17.0 \\
 & sc-spp & 6.0 $\pm$ 7.1 & 6.0 $\pm$ 7.1 & 9.0 $\pm$ 11.3 & 25.0 $\pm$ 0.0 & 7.0 $\pm$ 5.7 & \textbf{1.0 $\pm$ 0.0} & \textbf{1.0 $\pm$ 0.0} \\
\cline{1-9}
\multirow[t]{6}{*}{Mistral} & zs & 14.2 $\pm$ 12.2 & 19.0 $\pm$ 10.2 & 16.0 $\pm$ 12.0 & 23.8 $\pm$ 1.3 & 19.4 $\pm$ 9.1 & 24.2 $\pm$ 0.8 & \textbf{1.0 $\pm$ 0.0} \\
 & cot & 20.8 $\pm$ 4.2 & 13.8 $\pm$ 9.0 & 9.0 $\pm$ 7.6 & 23.4 $\pm$ 2.1 & 24.2 $\pm$ 0.8 & 17.2 $\pm$ 10.5 & \textbf{3.2 $\pm$ 4.9} \\
 & spp & 18.0 $\pm$ 7.0 & 22.2 $\pm$ 4.6 & 20.4 $\pm$ 6.4 & 22.2 $\pm$ 2.9 & 23.8 $\pm$ 1.6 & 12.8 $\pm$ 11.3 & \textbf{5.6 $\pm$ 10.3} \\
 & sc-zs & \textbf{1.0 $\pm$ 0.0} & 1.5 $\pm$ 0.7 & 12.0 $\pm$ 15.6 & 21.5 $\pm$ 4.9 & 13.0 $\pm$ 17.0 & 24.5 $\pm$ 0.7 & 7.5 $\pm$ 9.2 \\
 & sc-cot & 17.5 $\pm$ 10.6 & 14.5 $\pm$ 13.4 & \textbf{10.0 $\pm$ 7.1} & 18.5 $\pm$ 4.9 & 14.5 $\pm$ 14.8 & 13.0 $\pm$ 17.0 & 12.5 $\pm$ 16.3 \\
 & sc-spp & 12.5 $\pm$ 16.3 & 19.0 $\pm$ 7.1 & 23.5 $\pm$ 0.7 & 19.0 $\pm$ 7.1 & 25.0 $\pm$ 0.0 & 12.5 $\pm$ 16.3 & \textbf{1.0 $\pm$ 0.0} \\
\bottomrule
\end{tabular}
\caption{Round $m$ \# where the LLM player understood the opponent's Strategy in the label-based RPS counterfactual}
\label{tab:rps_round_heatmap_eq1-alt}
\end{table*}

In Table \ref{tab:rps_round_heatmap_eq1-alt} we demonstrate results for the label-based RPS counterfactual. The label-based RPS counterfactual makes opponent comprehension noticeably harder and more uneven than in default RPS, especially in LLM--LLM interactions and against the cyclic \textbf{PP} opponent. Since only the dominance induced by action labels is changed, these delays suggest that many models rely partly on canonical RPS label associations rather than fully recomputing the altered dominance relation. This is most visible for Claude 3.5 Sonnet v2, whose comprehension often moves close to the horizon across almost all opponent types (e.g., $m=21.6\pm3.8$
 against \textbf{PP} under ZS, and repeated 
$m=25.0$ under several SC settings), indicating substantial difficulty adapting to the relabeled game.

A clearer separation emerges among stronger models. Claude 3.7, Claude 3.7(T), Claude 4, and to a lesser extent Claude 4(T), still achieve early comprehension against structured algorithmic opponents in several prompting settings, particularly against \textbf{PP} and \textbf{TFT}. For example, Claude 3.7 reaches 
$3.2\pm 3.9$ against \textbf{PP} and $1.6
\pm 1.3$ against \textbf{TFT}) under CoT, while Claude 3.7(T) reaches $
m=3.4\pm 3.9$ and 
$m=1.2\pm 0.4$ under SPP. Claude 4 is similarly robust against \textbf{PP}, often obtaining 
$m \approx 2.5-4.8$, although its comprehension against \textbf{TFT} is less consistent. These results suggest that the advanced Claude models retain some ability to infer reactive or cyclic structure even when familiar action semantics are permuted.

By contrast, \textbf{AP} and \textbf{SREP} remain difficult for most models under this counterfactual, with many 
$m$-values close to 20--25. This indicates that the label remapping does not merely slow exploitation of cyclic opponents, but more broadly disrupts stable identification of opponent policies. DeepSeek R1 is relatively weak in this setting: most of its comprehension values remain late across all opponent types, with only isolated improvements such as 
$m=8.0\pm 9.9$ against CoT under SC-ZS or 
$m=13.0\pm 0.0$ against \textbf{AP} under SC-CoT. Llama 3.3 is more competitive, showing several very early values---for example $m=1.0\pm 0.0$ against \textbf{TFT} in multiple prompting settings and $m=1.0\pm 0.0$ against \textbf{AP} under SC-SPP---but it remains less uniformly robust than the strongest Claude variants. Mistral Large is again highly unstable: although it occasionally achieves immediate comprehension (e.g., 
$m=1.0\pm 0.0$ against \textbf{TFT} or in some LLM--LLM settings), many other entries stay near the horizon, confirming brittle and inconsistent adaptation.

Overall, the label-based RPS counterfactual experiment shows that label-only perturbations in RPS significantly impair opponent comprehension. The most robust models are Claude 3.7, Claude 3.7(T), and Claude 4, which still exploit \textbf{PP}/\textbf{TFT} in several settings, while Claude 3.5, DeepSeek R1, and especially Mistral exhibit much greater degradation. In contrast to default RPS, where some models could reliably infer cyclic or reactive opponents, the relabeled game often pushes comprehension toward the horizon, supporting the claim that many LLMs depend in part on memorized action semantics rather than abstract reasoning over the modified dominance structure.

\begin{table*}[h!]
\small
\centering
\begin{tabular}{p{1cm}p{0.9cm}lllllll}
\toprule
 &  & \multicolumn{7}{c}{\textbf{Payoff-based RPS counterfactual}} \\
 &  & zs & spp & cot & srep & pp & ap & tft \\
model & prompt &  &  &  &  &  &  &  \\
\midrule
\multirow[t]{6}{*}{C3.5Sv2} & zs & 16.2 $\pm$ 11.0 & 17.6 $\pm$ 10.3 & 24.4 $\pm$ 0.5 & 23.4 $\pm$ 1.9 & 19.0 $\pm$ 10.4 & 19.4 $\pm$ 10.4 & \textbf{5.6 $\pm$ 10.3} \\
 & cot & 20.2 $\pm$ 9.1 & 12.6 $\pm$ 11.0 & 21.0 $\pm$ 5.3 & 17.6 $\pm$ 9.8 & 24.2 $\pm$ 1.3 & 18.4 $\pm$ 9.8 & \textbf{10.0 $\pm$ 12.4} \\
 & spp & \textbf{10.6 $\pm$ 10.1} & 13.4 $\pm$ 11.5 & 22.2 $\pm$ 4.1 & 18.4 $\pm$ 6.9 & 15.0 $\pm$ 9.1 & 24.2 $\pm$ 0.8 & 14.2 $\pm$ 12.3 \\
 & sc-zs & 23.5 $\pm$ 2.1 & 24.0 $\pm$ 1.4 & 13.0 $\pm$ 14.1 & 18.0 $\pm$ 9.9 & \textbf{1.0 $\pm$ 0.0} & \textbf{1.0 $\pm$ 0.0} & \textbf{1.0 $\pm$ 0.0} \\
 & sc-cot & 25.0 $\pm$ 0.0 & 18.5 $\pm$ 6.4 & 25.0 $\pm$ 0.0 & 19.5 $\pm$ 2.1 & 12.0 $\pm$ 15.6 & 11.5 $\pm$ 14.8 & \textbf{1.0 $\pm$ 0.0} \\
 & sc-spp & 23.0 $\pm$ 1.4 & 25.0 $\pm$ 0.0 & \textbf{2.0 $\pm$ 1.4} & 24.5 $\pm$ 0.7 & \textbf{1.0 $\pm$ 0.0} & 12.5 $\pm$ 16.3 & \textbf{1.0 $\pm$ 0.0} \\
\cline{1-9}
\multirow[t]{6}{*}{C3.7S} & zs & 20.4 $\pm$ 8.6 & 17.6 $\pm$ 10.6 & 23.2 $\pm$ 2.0 & 22.4 $\pm$ 2.5 & \textbf{1.4 $\pm$ 0.9} & 18.8 $\pm$ 10.1 & 3.2 $\pm$ 3.5 \\
 & cot & 16.6 $\pm$ 9.5 & 23.0 $\pm$ 2.9 & 23.2 $\pm$ 1.6 & 18.8 $\pm$ 6.2 & \textbf{1.0 $\pm$ 0.0} & 7.4 $\pm$ 8.9 & 1.2 $\pm$ 0.4 \\
 & spp & 21.0 $\pm$ 4.6 & 24.4 $\pm$ 1.3 & 24.0 $\pm$ 1.0 & 20.8 $\pm$ 6.0 & 3.0 $\pm$ 4.5 & 18.4 $\pm$ 9.8 & \textbf{1.0 $\pm$ 0.0} \\
 & sc-zs & 16.5 $\pm$ 12.0 & 25.0 $\pm$ 0.0 & 24.5 $\pm$ 0.7 & 17.0 $\pm$ 11.3 & 12.0 $\pm$ 15.6 & 12.0 $\pm$ 15.6 & \textbf{1.0 $\pm$ 0.0} \\
 & sc-cot & 19.0 $\pm$ 7.1 & 21.0 $\pm$ 5.7 & 21.5 $\pm$ 0.7 & 24.5 $\pm$ 0.7 & \textbf{1.0 $\pm$ 0.0} & 22.0 $\pm$ 1.4 & \textbf{1.0 $\pm$ 0.0} \\
 & sc-spp & 13.0 $\pm$ 17.0 & 23.5 $\pm$ 2.1 & 23.5 $\pm$ 2.1 & 24.0 $\pm$ 1.4 & \textbf{1.0 $\pm$ 0.0} & \textbf{1.0 $\pm$ 0.0} & \textbf{1.0 $\pm$ 0.0} \\
\cline{1-9}
\multirow[t]{6}{*}{C3.7S(T)} & zs & 19.2 $\pm$ 6.8 & 20.8 $\pm$ 3.5 & 23.6 $\pm$ 2.1 & 17.8 $\pm$ 8.6 & 5.8 $\pm$ 9.1 & 23.6 $\pm$ 1.5 & \textbf{4.8 $\pm$ 4.8} \\
 & cot & 18.8 $\pm$ 6.6 & 21.4 $\pm$ 4.9 & 23.6 $\pm$ 1.1 & 21.2 $\pm$ 5.8 & \textbf{1.0 $\pm$ 0.0} & 13.8 $\pm$ 11.9 & 3.4 $\pm$ 5.4 \\
 & spp & 21.4 $\pm$ 5.0 & 22.4 $\pm$ 1.9 & 24.8 $\pm$ 0.4 & 19.6 $\pm$ 8.8 & \textbf{1.0 $\pm$ 0.0} & 5.8 $\pm$ 10.7 & \textbf{1.0 $\pm$ 0.0} \\
 & sc-zs & 25.0 $\pm$ 0.0 & 20.0 $\pm$ 7.1 & 13.0 $\pm$ 17.0 & 24.0 $\pm$ 0.0 & \textbf{1.0 $\pm$ 0.0} & 19.5 $\pm$ 7.8 & \textbf{1.0 $\pm$ 0.0} \\
 & sc-cot & 22.5 $\pm$ 2.1 & 24.5 $\pm$ 0.7 & 25.0 $\pm$ 0.0 & 18.5 $\pm$ 6.4 & \textbf{1.0 $\pm$ 0.0} & 23.5 $\pm$ 2.1 & \textbf{1.0 $\pm$ 0.0} \\
 & sc-spp & 24.0 $\pm$ 1.4 & 25.0 $\pm$ 0.0 & 24.0 $\pm$ 1.4 & 20.5 $\pm$ 0.7 & \textbf{1.0 $\pm$ 0.0} & 12.0 $\pm$ 15.6 & \textbf{1.0 $\pm$ 0.0} \\
\cline{1-9}
\multirow[t]{6}{*}{C4S} & zs & 23.8 $\pm$ 1.6 & 23.2 $\pm$ 2.0 & 18.0 $\pm$ 7.6 & 24.2 $\pm$ 0.8 & \textbf{1.0 $\pm$ 0.0} & 23.0 $\pm$ 1.9 & 7.2 $\pm$ 8.5 \\
 & cot & 19.8 $\pm$ 10.5 & 21.8 $\pm$ 6.1 & 19.6 $\pm$ 5.6 & 20.0 $\pm$ 4.2 & \textbf{1.0 $\pm$ 0.0} & 18.8 $\pm$ 10.0 & 5.6 $\pm$ 10.3 \\
 & spp & 10.2 $\pm$ 9.7 & 19.2 $\pm$ 10.3 & 23.0 $\pm$ 1.4 & 24.0 $\pm$ 1.7 & \textbf{5.8 $\pm$ 10.7} & 18.6 $\pm$ 5.8 & 15.6 $\pm$ 11.6 \\
 & sc-zs & 20.0 $\pm$ 7.1 & 25.0 $\pm$ 0.0 & 18.0 $\pm$ 9.9 & 25.0 $\pm$ 0.0 & \textbf{1.0 $\pm$ 0.0} & 23.0 $\pm$ 0.0 & 20.5 $\pm$ 3.5 \\
 & sc-cot & 12.0 $\pm$ 15.6 & 19.0 $\pm$ 8.5 & 13.0 $\pm$ 17.0 & 16.5 $\pm$ 9.2 & \textbf{1.0 $\pm$ 0.0} & 13.0 $\pm$ 17.0 & \textbf{1.0 $\pm$ 0.0} \\
 & sc-spp & 25.0 $\pm$ 0.0 & 23.5 $\pm$ 2.1 & 13.0 $\pm$ 17.0 & 13.5 $\pm$ 0.7 & \textbf{1.0 $\pm$ 0.0} & 24.0 $\pm$ 1.4 & \textbf{1.0 $\pm$ 0.0} \\
\cline{1-9}
\multirow[t]{6}{*}{C4S(T)} & zs & 24.0 $\pm$ 0.7 & 24.8 $\pm$ 0.4 & 21.0 $\pm$ 4.3 & 21.4 $\pm$ 4.3 & \textbf{4.2 $\pm$ 7.2} & 23.4 $\pm$ 1.5 & 15.0 $\pm$ 12.3 \\
 & cot & 22.8 $\pm$ 2.3 & 10.6 $\pm$ 10.3 & 20.2 $\pm$ 10.7 & 17.4 $\pm$ 10.1 & 1.2 $\pm$ 0.4 & 18.2 $\pm$ 9.8 & \textbf{1.0 $\pm$ 0.0} \\
 & spp & 19.2 $\pm$ 6.2 & 19.4 $\pm$ 4.6 & 12.2 $\pm$ 11.5 & 21.2 $\pm$ 4.2 & \textbf{1.0 $\pm$ 0.0} & 20.4 $\pm$ 9.2 & 6.0 $\pm$ 10.1 \\
 & sc-zs & 25.0 $\pm$ 0.0 & 24.0 $\pm$ 1.4 & 19.5 $\pm$ 7.8 & \textbf{11.5 $\pm$ 0.7} & \textbf{1.0 $\pm$ 0.0} & 23.0 $\pm$ 1.4 & \textbf{1.0 $\pm$ 0.0} \\
 & sc-cot & 23.5 $\pm$ 0.7 & 23.0 $\pm$ 1.4 & 25.0 $\pm$ 0.0 & 23.0 $\pm$ 0.0 & 3.5 $\pm$ 2.1 & 18.5 $\pm$ 7.8 & \textbf{1.0 $\pm$ 0.0} \\
 & sc-spp & 22.5 $\pm$ 2.1 & 24.0 $\pm$ 1.4 & 21.5 $\pm$ 0.7 & 24.5 $\pm$ 0.7 & \textbf{1.0 $\pm$ 0.0} & 23.0 $\pm$ 2.8 & 12.5 $\pm$ 16.3 \\
\cline{1-9}
\multirow[t]{6}{*}{DS-R1} & zs & 24.8 $\pm$ 0.4 & 19.0 $\pm$ 6.7 & 20.0 $\pm$ 10.6 & 23.0 $\pm$ 2.0 & 18.6 $\pm$ 9.3 & 18.8 $\pm$ 8.3 & \textbf{5.0 $\pm$ 8.9} \\
 & cot & 19.6 $\pm$ 7.4 & 19.0 $\pm$ 8.0 & 20.8 $\pm$ 5.4 & 17.8 $\pm$ 9.1 & 24.4 $\pm$ 0.9 & 15.2 $\pm$ 11.7 & \textbf{7.0 $\pm$ 7.0} \\
 & spp & 22.2 $\pm$ 3.3 & 21.2 $\pm$ 4.7 & 19.4 $\pm$ 3.0 & 24.4 $\pm$ 1.3 & 17.8 $\pm$ 9.7 & 14.6 $\pm$ 11.6 & \textbf{9.4 $\pm$ 11.5} \\
 & sc-zs & 20.0 $\pm$ 7.1 & 20.0 $\pm$ 0.0 & 23.0 $\pm$ 1.4 & 24.0 $\pm$ 0.0 & 11.5 $\pm$ 14.8 & 1.5 $\pm$ 0.7 & \textbf{1.0 $\pm$ 0.0} \\
 & sc-cot & 17.0 $\pm$ 5.7 & 24.5 $\pm$ 0.7 & 25.0 $\pm$ 0.0 & 18.5 $\pm$ 6.4 & 25.0 $\pm$ 0.0 & 18.5 $\pm$ 6.4 & \textbf{5.5 $\pm$ 6.4} \\
 & sc-spp & 25.0 $\pm$ 0.0 & 18.0 $\pm$ 5.7 & 24.5 $\pm$ 0.7 & 24.0 $\pm$ 0.0 & 13.0 $\pm$ 17.0 & \textbf{1.5 $\pm$ 0.7} & \textbf{1.5 $\pm$ 0.7} \\
\cline{1-9}
\multirow[t]{6}{*}{L3.3-70B} & zs & 5.8 $\pm$ 10.7 & 15.0 $\pm$ 9.7 & 8.4 $\pm$ 10.9 & 23.8 $\pm$ 0.8 & 7.2 $\pm$ 10.4 & 6.8 $\pm$ 9.7 & \textbf{1.0 $\pm$ 0.0} \\
 & cot & 9.6 $\pm$ 11.9 & 16.8 $\pm$ 10.8 & 13.4 $\pm$ 11.3 & 21.6 $\pm$ 4.9 & \textbf{1.0 $\pm$ 0.0} & 16.4 $\pm$ 10.5 & \textbf{1.0 $\pm$ 0.0} \\
 & spp & \textbf{1.0 $\pm$ 0.0} & 5.0 $\pm$ 4.1 & 7.6 $\pm$ 10.5 & 24.2 $\pm$ 0.8 & 11.6 $\pm$ 11.1 & 20.2 $\pm$ 7.3 & 1.2 $\pm$ 0.4 \\
 & sc-zs & 13.0 $\pm$ 17.0 & \textbf{1.0 $\pm$ 0.0} & 11.5 $\pm$ 14.8 & 23.0 $\pm$ 0.0 & \textbf{1.0 $\pm$ 0.0} & \textbf{1.0 $\pm$ 0.0} & \textbf{1.0 $\pm$ 0.0} \\
 & sc-cot & 13.0 $\pm$ 17.0 & 7.0 $\pm$ 8.5 & 10.5 $\pm$ 13.4 & 16.0 $\pm$ 7.1 & \textbf{1.0 $\pm$ 0.0} & 22.0 $\pm$ 0.0 & \textbf{1.0 $\pm$ 0.0} \\
 & sc-spp & \textbf{1.0 $\pm$ 0.0} & 12.0 $\pm$ 15.6 & 19.0 $\pm$ 5.7 & 18.5 $\pm$ 9.2 & 6.5 $\pm$ 7.8 & 6.0 $\pm$ 7.1 & \textbf{1.0 $\pm$ 0.0} \\
\cline{1-9}
\multirow[t]{6}{*}{Mistral} & zs & 7.6 $\pm$ 10.1 & 15.2 $\pm$ 13.0 & 12.4 $\pm$ 10.5 & 24.8 $\pm$ 0.4 & \textbf{1.0 $\pm$ 0.0} & 19.8 $\pm$ 10.5 & \textbf{1.0 $\pm$ 0.0} \\
 & cot & 9.8 $\pm$ 11.2 & 17.0 $\pm$ 6.8 & 19.2 $\pm$ 10.2 & 24.0 $\pm$ 1.7 & 13.8 $\pm$ 11.9 & 15.4 $\pm$ 13.1 & \textbf{4.8 $\pm$ 7.9} \\
 & spp & 10.0 $\pm$ 12.3 & 20.6 $\pm$ 4.6 & 11.6 $\pm$ 9.0 & 22.0 $\pm$ 5.6 & 20.6 $\pm$ 6.6 & 21.0 $\pm$ 6.4 & \textbf{8.4 $\pm$ 9.7} \\
 & sc-zs & \textbf{1.0 $\pm$ 0.0} & 24.5 $\pm$ 0.7 & 11.0 $\pm$ 0.0 & 22.0 $\pm$ 2.8 & \textbf{1.0 $\pm$ 0.0} & \textbf{1.0 $\pm$ 0.0} & \textbf{1.0 $\pm$ 0.0} \\
 & sc-cot & \textbf{1.0 $\pm$ 0.0} & 11.5 $\pm$ 14.8 & 12.0 $\pm$ 15.6 & 23.5 $\pm$ 0.7 & 13.0 $\pm$ 17.0 & \textbf{1.0 $\pm$ 0.0} & 1.5 $\pm$ 0.7 \\
 & sc-spp & 24.0 $\pm$ 0.0 & 12.5 $\pm$ 16.3 & 16.5 $\pm$ 10.6 & 17.0 $\pm$ 8.5 & 13.0 $\pm$ 17.0 & 12.5 $\pm$ 16.3 & \textbf{1.0 $\pm$ 0.0} \\
\bottomrule
\end{tabular}
\caption{Round $m$ \# where the LLM player understood the opponent's Strategy in the payoff-based RPS counterfactual.}
\label{tab:rps_round_heatmap_ba3}
\end{table*}
Table \ref{tab:rps_round_heatmap_ba3} illustrated results for the payoff-based RPS counterfactual. In this case, opponent comprehension becomes more polarized than in the default or label-only setting. The clearest pattern is that \textbf{PP} and \textbf{TFT} are often understood very early, especially by the stronger Claude models, whereas \textbf{SREP} and many LLM--LLM matchups still remain close to the horizon. This suggests that once exploitable cyclic or reactive structure is present, several models can adapt quickly to the altered payoff asymmetry, but they still struggle to infer behavior in near-equilibrium or symmetric play. In particular, Claude 3.7, Claude 3.7(T), and Claude 4 show the strongest results, frequently reaching 
$m \approx 1$ against \textbf{PP} and/or \textbf{TFT} across several prompting settings; for example, Claude 3.7 obtains $m=1.0\pm 0.0$ against \textbf{PP}) and 
$m=1.2\pm 0.4$ against \textbf{TFT}) under CoT, while Claude 4 repeatedly reaches $m=1.0\pm 0.0$ against \textbf{PP}) under multiple prompts. Claude 3.5 also improves substantially under self-consistency, with several $m=1.0\pm 0.0$ outcomes against \textbf{PP}), \textbf{AP}, and \textbf{TFT}, although its non-SC variants remain much less stable.

A second notable trend is that \textbf{AP} becomes easier than in the default and label-based setups for some models, but this improvement is uneven. Claude 3.5 under SC, Claude 3.7 under SC-SPP, DeepSeek R1 under SC-ZS/SC-SPP, and Llama 3.3 under SC-ZS all achieve very early or immediate comprehension on \textbf{AP}, yet many other entries remain delayed, often with $m$ near 15--24. This suggests that the payoff asymmetry can make adaptive behavior more exploitable once recognized, but models do not consistently detect this advantage.

DeepSeek R1 remains mixed overall: it shows some good outcomes under SC, especially on \textbf{AP} and \textbf{TFT}, but many non-SC values remain relatively late. Llama 3.3 is more competitive in this counterfactual than in the label-only one, with many early values across \textbf{PP}, \textbf{TFT}, and occasionally \textbf{AP}, indicating relatively good payoff sensitivity. Mistral Large remains the most unstable model, although it also exhibits several isolated $m=1.0\pm 0.0$ results under SC, particularly against \textbf{PP}), \textbf{AP}, and \textbf{TFT}. Overall, the table suggests that payoff-based changes are highly diagnostic: the strongest models can recompute the new incentives and quickly exploit structured opponents, but comprehension remains weak in LLM--LLM play and against \textbf{SREP}, confirming that altered payoffs help mainly when exploitable regularities are present rather than by generally improving strategic inference.

\begin{table*}[h!]
\small
\centering
\begin{tabular}{p{1cm}p{0.9cm}lllllll}
\toprule
 &  & \multicolumn{7}{c}{Joint RPS counterfactual} \\
 &  & zs & spp & cot & srep & pp & ap & tft \\
model & prompt &  &  &  &  &  &  &  \\
\midrule
\multirow[t]{6}{*}{C3.5Sv2} & zs & 23.8 $\pm$ 1.3 & 18.8 $\pm$ 7.6 & 14.6 $\pm$ 10.7 & 24.2 $\pm$ 0.8 & 21.2 $\pm$ 7.4 & 15.0 $\pm$ 12.8 & \textbf{10.2 $\pm$ 9.3} \\
 & cot & 16.2 $\pm$ 9.5 & 17.6 $\pm$ 9.6 & 19.0 $\pm$ 7.4 & 17.4 $\pm$ 10.3 & 24.4 $\pm$ 0.9 & 18.6 $\pm$ 5.6 & \textbf{14.2 $\pm$ 12.1} \\
 & spp & 23.4 $\pm$ 0.9 & 24.6 $\pm$ 0.5 & 15.8 $\pm$ 9.6 & 20.0 $\pm$ 3.7 & 25.0 $\pm$ 0.0 & 19.0 $\pm$ 8.0 & \textbf{7.8 $\pm$ 10.2} \\
 & sc-zs & 24.0 $\pm$ 1.4 & 24.0 $\pm$ 0.0 & \textbf{15.0 $\pm$ 0.0} & 19.0 $\pm$ 8.5 & 21.0 $\pm$ 5.7 & 23.5 $\pm$ 2.1 & 24.5 $\pm$ 0.7 \\
 & sc-cot & 22.5 $\pm$ 2.1 & 14.0 $\pm$ 15.6 & 12.5 $\pm$ 13.4 & 24.0 $\pm$ 1.4 & 22.5 $\pm$ 3.5 & 24.5 $\pm$ 0.7 & \textbf{10.5 $\pm$ 13.4} \\
 & sc-spp & 10.5 $\pm$ 12.0 & 12.5 $\pm$ 16.3 & 17.5 $\pm$ 10.6 & 22.0 $\pm$ 2.8 & 24.0 $\pm$ 1.4 & 23.5 $\pm$ 2.1 & \textbf{10.0 $\pm$ 12.7} \\
\cline{1-9}
\multirow[t]{6}{*}{C3.7S} & zs & 21.0 $\pm$ 4.2 & 23.2 $\pm$ 2.2 & 23.4 $\pm$ 1.8 & 22.8 $\pm$ 2.9 & 16.6 $\pm$ 9.0 & 19.0 $\pm$ 10.2 & \textbf{12.2 $\pm$ 11.8} \\
 & cot & 20.0 $\pm$ 4.5 & 20.4 $\pm$ 4.6 & 21.4 $\pm$ 4.8 & 18.0 $\pm$ 7.5 & \textbf{2.6 $\pm$ 3.6} & 23.2 $\pm$ 1.6 & 9.4 $\pm$ 11.7 \\
 & spp & 17.0 $\pm$ 9.7 & 22.8 $\pm$ 2.9 & 24.6 $\pm$ 0.5 & 20.8 $\pm$ 8.8 & \textbf{4.6 $\pm$ 8.0} & 17.4 $\pm$ 9.9 & 7.0 $\pm$ 10.4 \\
 & sc-zs & 12.5 $\pm$ 16.3 & 23.5 $\pm$ 0.7 & 25.0 $\pm$ 0.0 & 18.5 $\pm$ 9.2 & 7.5 $\pm$ 0.7 & \textbf{1.0 $\pm$ 0.0} & 3.0 $\pm$ 2.8 \\
 & sc-cot & 13.0 $\pm$ 17.0 & 23.0 $\pm$ 2.8 & 24.5 $\pm$ 0.7 & 23.0 $\pm$ 1.4 & 2.0 $\pm$ 1.4 & 22.5 $\pm$ 2.1 & \textbf{1.0 $\pm$ 0.0} \\
 & sc-spp & 19.0 $\pm$ 5.7 & 23.0 $\pm$ 1.4 & 24.5 $\pm$ 0.7 & 23.0 $\pm$ 1.4 & \textbf{1.0 $\pm$ 0.0} & 25.0 $\pm$ 0.0 & \textbf{1.0 $\pm$ 0.0} \\
\cline{1-9}
\multirow[t]{6}{*}{C3.7S(T)} & zs & 22.6 $\pm$ 4.3 & 20.0 $\pm$ 4.7 & 21.6 $\pm$ 4.6 & 24.2 $\pm$ 0.4 & 24.8 $\pm$ 0.4 & 19.0 $\pm$ 10.1 & \textbf{9.4 $\pm$ 8.8} \\
 & cot & 17.4 $\pm$ 6.7 & 21.6 $\pm$ 4.1 & 22.6 $\pm$ 3.2 & 21.6 $\pm$ 5.1 & 11.8 $\pm$ 12.1 & 21.4 $\pm$ 3.9 & \textbf{2.4 $\pm$ 3.1} \\
 & spp & 17.0 $\pm$ 7.2 & 19.8 $\pm$ 8.4 & 22.8 $\pm$ 1.9 & 17.4 $\pm$ 7.4 & \textbf{2.4 $\pm$ 1.3} & 22.0 $\pm$ 1.6 & 6.2 $\pm$ 10.0 \\
 & sc-zs & 24.5 $\pm$ 0.7 & 18.5 $\pm$ 0.7 & 24.5 $\pm$ 0.7 & 21.0 $\pm$ 1.4 & 25.0 $\pm$ 0.0 & 23.0 $\pm$ 1.4 & \textbf{8.0 $\pm$ 9.9} \\
 & sc-cot & 17.0 $\pm$ 5.7 & 23.5 $\pm$ 0.7 & 25.0 $\pm$ 0.0 & 24.0 $\pm$ 0.0 & 1.5 $\pm$ 0.7 & 23.5 $\pm$ 0.7 & \textbf{1.0 $\pm$ 0.0} \\
 & sc-spp & 14.0 $\pm$ 12.7 & 18.0 $\pm$ 9.9 & 23.5 $\pm$ 2.1 & \textbf{13.0 $\pm$ 8.5} & 3.0 $\pm$ 2.8 & 25.0 $\pm$ 0.0 & \textbf{1.0 $\pm$ 0.0} \\
\cline{1-9}
\multirow[t]{6}{*}{C4S} & zs & 20.4 $\pm$ 6.3 & 23.4 $\pm$ 1.7 & 19.8 $\pm$ 10.5 & 21.6 $\pm$ 3.8 & \textbf{13.6 $\pm$ 9.9} & 23.4 $\pm$ 0.5 & 17.4 $\pm$ 9.9 \\
 & cot & 24.0 $\pm$ 1.0 & 20.4 $\pm$ 7.6 & 23.2 $\pm$ 2.2 & 19.2 $\pm$ 5.9 & \textbf{1.4 $\pm$ 0.9} & 24.2 $\pm$ 0.8 & 18.6 $\pm$ 9.9 \\
 & spp & 23.4 $\pm$ 1.3 & 24.2 $\pm$ 1.3 & 21.4 $\pm$ 4.8 & 19.0 $\pm$ 5.6 & \textbf{6.4 $\pm$ 8.8} & 22.0 $\pm$ 4.0 & 14.0 $\pm$ 9.7 \\
 & sc-zs & 22.5 $\pm$ 2.1 & 23.0 $\pm$ 1.4 & 24.0 $\pm$ 1.4 & 18.5 $\pm$ 6.4 & \textbf{8.5 $\pm$ 4.9} & 24.0 $\pm$ 1.4 & 22.0 $\pm$ 1.4 \\
 & sc-cot & 15.0 $\pm$ 14.1 & 25.0 $\pm$ 0.0 & 22.5 $\pm$ 3.5 & 21.0 $\pm$ 1.4 & \textbf{1.0 $\pm$ 0.0} & 24.0 $\pm$ 1.4 & 24.0 $\pm$ 0.0 \\
 & sc-spp & 23.5 $\pm$ 2.1 & 11.5 $\pm$ 2.1 & 25.0 $\pm$ 0.0 & 24.0 $\pm$ 1.4 & 3.0 $\pm$ 2.8 & 24.5 $\pm$ 0.7 & \textbf{1.0 $\pm$ 0.0} \\
\cline{1-9}
\multirow[t]{6}{*}{C4S(T)} & zs & 17.4 $\pm$ 10.0 & 24.8 $\pm$ 0.4 & 19.4 $\pm$ 10.3 & 18.2 $\pm$ 9.1 & 23.4 $\pm$ 3.0 & 24.0 $\pm$ 2.2 & \textbf{14.6 $\pm$ 12.4} \\
 & cot & 21.0 $\pm$ 6.2 & 20.4 $\pm$ 8.6 & 23.2 $\pm$ 1.6 & 15.2 $\pm$ 11.3 & \textbf{12.2 $\pm$ 11.5} & 24.6 $\pm$ 0.9 & 13.6 $\pm$ 11.8 \\
 & spp & 23.6 $\pm$ 1.3 & 19.4 $\pm$ 7.2 & 23.6 $\pm$ 0.9 & 16.8 $\pm$ 7.0 & \textbf{2.6 $\pm$ 3.0} & 22.0 $\pm$ 1.9 & 19.8 $\pm$ 9.4 \\
 & sc-zs & 25.0 $\pm$ 0.0 & 24.0 $\pm$ 1.4 & 23.5 $\pm$ 2.1 & 21.0 $\pm$ 2.8 & \textbf{14.0 $\pm$ 12.7} & 24.0 $\pm$ 1.4 & 25.0 $\pm$ 0.0 \\
 & sc-cot & 23.5 $\pm$ 0.7 & 24.5 $\pm$ 0.7 & 24.0 $\pm$ 1.4 & \textbf{18.5 $\pm$ 9.2} & 24.5 $\pm$ 0.7 & 24.5 $\pm$ 0.7 & 24.0 $\pm$ 0.0 \\
 & sc-spp & 14.5 $\pm$ 13.4 & \textbf{8.0 $\pm$ 9.9} & 23.5 $\pm$ 0.7 & 24.5 $\pm$ 0.7 & 13.5 $\pm$ 13.4 & 23.5 $\pm$ 0.7 & 22.5 $\pm$ 2.1 \\
\cline{1-9}
\multirow[t]{6}{*}{DS-R1} & zs & \textbf{19.2 $\pm$ 7.1} & 24.0 $\pm$ 1.2 & 22.6 $\pm$ 2.5 & 21.8 $\pm$ 6.6 & 24.6 $\pm$ 0.5 & 20.8 $\pm$ 5.0 & 22.4 $\pm$ 2.1 \\
 & cot & 23.6 $\pm$ 1.1 & 21.0 $\pm$ 5.7 & \textbf{20.4 $\pm$ 8.6} & 24.6 $\pm$ 0.5 & 24.4 $\pm$ 0.5 & 23.4 $\pm$ 0.9 & 22.0 $\pm$ 2.2 \\
 & spp & 22.2 $\pm$ 4.1 & 24.2 $\pm$ 0.4 & 19.6 $\pm$ 4.9 & 21.0 $\pm$ 5.9 & 24.8 $\pm$ 0.4 & 20.8 $\pm$ 5.3 & \textbf{16.6 $\pm$ 9.8} \\
 & sc-zs & 25.0 $\pm$ 0.0 & \textbf{10.5 $\pm$ 13.4} & 19.0 $\pm$ 7.1 & 22.0 $\pm$ 4.2 & 24.0 $\pm$ 1.4 & 18.5 $\pm$ 4.9 & 12.0 $\pm$ 2.8 \\
 & sc-cot & \textbf{18.0 $\pm$ 4.2} & 22.0 $\pm$ 2.8 & 25.0 $\pm$ 0.0 & 24.0 $\pm$ 1.4 & 24.5 $\pm$ 0.7 & 19.5 $\pm$ 7.8 & 25.0 $\pm$ 0.0 \\
 & sc-spp & 24.5 $\pm$ 0.7 & 20.0 $\pm$ 7.1 & \textbf{6.0 $\pm$ 2.8} & 25.0 $\pm$ 0.0 & 25.0 $\pm$ 0.0 & 23.0 $\pm$ 0.0 & 13.0 $\pm$ 15.6 \\
\cline{1-9}
\multirow[t]{6}{*}{L3.3-70B} & zs & \textbf{10.4 $\pm$ 11.6} & 14.4 $\pm$ 12.3 & 14.4 $\pm$ 10.7 & 24.2 $\pm$ 1.3 & 11.4 $\pm$ 12.4 & 16.8 $\pm$ 11.1 & 10.6 $\pm$ 13.1 \\
 & cot & 18.2 $\pm$ 5.4 & 6.8 $\pm$ 6.3 & \textbf{5.6 $\pm$ 10.3} & 22.4 $\pm$ 4.2 & 16.2 $\pm$ 8.9 & 22.2 $\pm$ 1.8 & 10.4 $\pm$ 12.9 \\
 & spp & 20.0 $\pm$ 10.6 & 6.2 $\pm$ 10.5 & \textbf{4.6 $\pm$ 5.1} & 24.2 $\pm$ 1.1 & 16.0 $\pm$ 9.9 & 14.8 $\pm$ 12.7 & \textbf{1.0 $\pm$ 0.0} \\
 & sc-zs & 10.0 $\pm$ 12.7 & \textbf{1.0 $\pm$ 0.0} & 7.5 $\pm$ 9.2 & 23.5 $\pm$ 2.1 & 13.0 $\pm$ 17.0 & 24.0 $\pm$ 1.4 & \textbf{1.0 $\pm$ 0.0} \\
 & sc-cot & 12.5 $\pm$ 10.6 & 12.0 $\pm$ 15.6 & 20.5 $\pm$ 4.9 & 25.0 $\pm$ 0.0 & 10.5 $\pm$ 13.4 & 24.5 $\pm$ 0.7 & \textbf{1.0 $\pm$ 0.0} \\
 & sc-spp & 21.0 $\pm$ 1.4 & 13.0 $\pm$ 14.1 & 14.0 $\pm$ 15.6 & 20.0 $\pm$ 1.4 & 19.0 $\pm$ 8.5 & \textbf{1.0 $\pm$ 0.0} & \textbf{1.0 $\pm$ 0.0} \\
\cline{1-9}
\multirow[t]{6}{*}{Mistral} & zs & \textbf{6.0 $\pm$ 6.2} & 17.6 $\pm$ 10.3 & 14.6 $\pm$ 12.4 & 23.8 $\pm$ 0.8 & 13.4 $\pm$ 11.2 & 19.2 $\pm$ 10.2 & 7.8 $\pm$ 10.5 \\
 & cot & 11.2 $\pm$ 12.2 & 16.8 $\pm$ 5.8 & 14.0 $\pm$ 12.1 & 22.6 $\pm$ 2.1 & 25.0 $\pm$ 0.0 & 22.0 $\pm$ 4.6 & \textbf{1.6 $\pm$ 1.3} \\
 & spp & 17.2 $\pm$ 10.0 & 19.6 $\pm$ 6.1 & 17.6 $\pm$ 9.7 & 21.4 $\pm$ 4.5 & 7.0 $\pm$ 10.0 & 12.0 $\pm$ 10.4 & \textbf{5.8 $\pm$ 10.7} \\
 & sc-zs & 13.0 $\pm$ 17.0 & 20.0 $\pm$ 7.1 & 19.0 $\pm$ 8.5 & 19.0 $\pm$ 8.5 & 25.0 $\pm$ 0.0 & 24.5 $\pm$ 0.7 & \textbf{7.5 $\pm$ 9.2} \\
 & sc-cot & 24.5 $\pm$ 0.7 & 23.5 $\pm$ 0.7 & 23.5 $\pm$ 2.1 & 24.5 $\pm$ 0.7 & 17.5 $\pm$ 10.6 & 12.0 $\pm$ 15.6 & \textbf{1.5 $\pm$ 0.7} \\
 & sc-spp & 13.0 $\pm$ 17.0 & 20.5 $\pm$ 0.7 & 23.5 $\pm$ 0.7 & 24.0 $\pm$ 1.4 & 25.0 $\pm$ 0.0 & 24.5 $\pm$ 0.7 & \textbf{1.0 $\pm$ 0.0} \\
\bottomrule
\end{tabular}
\caption{Round $m$ \# where the LLM player understood the opponent's Strategy in the joint RPS counterfactual.}
\label{tab:rps_round_heatmap_ba3-alt}
\end{table*}
We finally present results for the joint RPS counterfactual in Table \ref{tab:rps_round_heatmap_ba3-alt}. The joint RPS counterfactual is the hardest comprehension setting overall, since it simultaneously perturbs both action semantics and payoff structure. Relative to the label-only and payoff-only variants, many 
$m$-values move closer to the horizon, showing that stable opponent inference becomes substantially harder. The most robust models are Claude 3.7 and Claude 3.7(T), which still achieve very early comprehension against structured algorithmic opponents in several prompting settings, especially on \textbf{PP}, \textbf{AP}, and \textbf{TFT}. Claude 4 is somewhat competitive but less stable, while Claude 3.5 is much more disrupted, with many late values across opponent types. DeepSeek R1 also degrades markedly in this joint setting, whereas Llama 3.3 remains moderately competitive with a few strong low-$m$ cases. Mistral is again the most brittle and inconsistent. Overall, the table shows that combining label and payoff perturbations exposes stronger reasoning limitations than either intervention alone, with only the strongest models retaining reliable comprehension of exploitable opponents.

\subsection{RPS Efficiency}
\begin{table*}[h!]
\centering \small
\begin{tabular}{llllll}
\toprule
 &  & RPS & RPS label-based & RPS payoff-based & RPS joint \\
model & prompt &  &  &  &  \\
\midrule
\multirow[t]{6}{*}{C3.5v2} & zs & \textbf{1.09 $\pm$ 2.37} & 0.09 $\pm$ 1.34 & 0.47 $\pm$ 2.65 & 0.25 $\pm$ 1.61 \\
 & cot & 0.44 $\pm$ 0.74 & -0.23 $\pm$ 0.84 & \textbf{0.49 $\pm$ 1.64} & -0.03 $\pm$ 1.28 \\
 & spp & 0.49 $\pm$ 0.71 & 0.01 $\pm$ 0.48 & \textbf{0.66 $\pm$ 0.83} & 0.05 $\pm$ 1.10 \\
 & sc-zs & 0.20 $\pm$ 0.60 & 0.10 $\pm$ 0.48 & \textbf{0.46 $\pm$ 0.56} & -0.03 $\pm$ 0.35 \\
 & sc-cot & 0.14 $\pm$ 0.18 & -0.08 $\pm$ 0.25 & \textbf{0.15 $\pm$ 0.41} & -0.01 $\pm$ 0.36 \\
 & sc-spp & 0.01 $\pm$ 0.20 & -0.03 $\pm$ 0.17 & \textbf{0.19 $\pm$ 0.34} & -0.01 $\pm$ 0.17 \\
\cline{1-6}
\multirow[t]{6}{*}{C3.7S} & zs & 0.58 $\pm$ 2.39 & 0.20 $\pm$ 2.12 & \textbf{2.02 $\pm$ 4.02} & 0.07 $\pm$ 2.02 \\
 & cot & 0.65 $\pm$ 0.75 & 0.51 $\pm$ 0.80 & 0.63 $\pm$ 1.07 & \textbf{0.71 $\pm$ 1.10} \\
 & spp & 0.51 $\pm$ 0.70 & 0.41 $\pm$ 0.61 & 0.43 $\pm$ 1.13 & \textbf{0.56 $\pm$ 0.76} \\
 & sc-zs & -0.08 $\pm$ 0.72 & 0.00 $\pm$ 0.36 & 0.00 $\pm$ 0.93 & \textbf{0.32 $\pm$ 0.71} \\
 & sc-cot & \textbf{0.21 $\pm$ 0.25} & 0.14 $\pm$ 0.17 & 0.17 $\pm$ 0.26 & 0.11 $\pm$ 0.28 \\
 & sc-spp & 0.14 $\pm$ 0.18 & 0.09 $\pm$ 0.12 & \textbf{0.16 $\pm$ 0.21} & 0.13 $\pm$ 0.16 \\
\cline{1-6}
\multirow[t]{6}{*}{C3.7S(T)} & zs & 1.92 $\pm$ 3.30 & 0.60 $\pm$ 2.55 & \textbf{3.26 $\pm$ 4.15} & 1.19 $\pm$ 4.43 \\
 & cot & 0.61 $\pm$ 1.04 & 0.25 $\pm$ 1.02 & \textbf{0.87 $\pm$ 1.27} & 0.60 $\pm$ 1.15 \\
 & spp & 0.57 $\pm$ 0.89 & 0.60 $\pm$ 0.80 & \textbf{1.06 $\pm$ 1.53} & 0.70 $\pm$ 0.97 \\
 & sc-zs & 0.23 $\pm$ 0.77 & 0.31 $\pm$ 0.98 & \textbf{0.70 $\pm$ 1.44} & -0.08 $\pm$ 0.81 \\
 & sc-cot & 0.11 $\pm$ 0.27 & 0.15 $\pm$ 0.21 & \textbf{0.24 $\pm$ 0.28} & 0.18 $\pm$ 0.31 \\
 & sc-spp & 0.04 $\pm$ 0.26 & 0.11 $\pm$ 0.18 & 0.17 $\pm$ 0.35 & \textbf{0.18 $\pm$ 0.24} \\
\cline{1-6}
\multirow[t]{6}{*}{C4S} & zs & 0.80 $\pm$ 1.82 & -0.12 $\pm$ 2.25 & \textbf{0.92 $\pm$ 2.96} & 0.23 $\pm$ 2.49 \\
 & cot & 0.62 $\pm$ 0.89 & 0.31 $\pm$ 0.71 & \textbf{0.75 $\pm$ 1.41} & 0.57 $\pm$ 1.10 \\
 & spp & 0.37 $\pm$ 0.70 & 0.19 $\pm$ 0.56 & \textbf{0.73 $\pm$ 0.83} & 0.30 $\pm$ 0.74 \\
 & sc-zs & \textbf{0.13 $\pm$ 0.50} & 0.07 $\pm$ 0.29 & 0.12 $\pm$ 0.43 & 0.07 $\pm$ 0.21 \\
 & sc-cot & 0.10 $\pm$ 0.18 & 0.07 $\pm$ 0.13 & \textbf{0.15 $\pm$ 0.29} & 0.11 $\pm$ 0.25 \\
 & sc-spp & 0.10 $\pm$ 0.14 & 0.05 $\pm$ 0.09 & 0.09 $\pm$ 0.20 & \textbf{0.12 $\pm$ 0.21} \\
\cline{1-6}
\multirow[t]{6}{*}{C4S(T)} & zs & \textbf{2.07 $\pm$ 3.74} & -0.16 $\pm$ 2.00 & 1.51 $\pm$ 4.69 & 0.04 $\pm$ 3.76 \\
 & cot & 1.11 $\pm$ 1.68 & 0.60 $\pm$ 0.92 & \textbf{1.50 $\pm$ 2.05} & 0.81 $\pm$ 1.93 \\
 & spp & 0.31 $\pm$ 0.72 & 0.28 $\pm$ 0.68 & \textbf{0.69 $\pm$ 1.20} & 0.57 $\pm$ 0.91 \\
 & sc-zs & \textbf{0.44 $\pm$ 0.67} & -0.21 $\pm$ 0.39 & 0.37 $\pm$ 1.14 & 0.03 $\pm$ 0.75 \\
 & sc-cot & \textbf{0.35 $\pm$ 0.26} & 0.20 $\pm$ 0.28 & 0.16 $\pm$ 0.61 & -0.04 $\pm$ 0.27 \\
 & sc-spp & 0.13 $\pm$ 0.13 & 0.07 $\pm$ 0.12 & \textbf{0.14 $\pm$ 0.18} & 0.10 $\pm$ 0.13 \\
\cline{1-6}
\multirow[t]{6}{*}{DS-R1} & zs & 0.47 $\pm$ 0.73 & 0.17 $\pm$ 0.44 & \textbf{0.89 $\pm$ 1.49} & 0.03 $\pm$ 0.86 \\
 & cot & 0.80 $\pm$ 0.84 & 0.23 $\pm$ 0.49 & \textbf{1.04 $\pm$ 1.77} & 0.02 $\pm$ 0.76 \\
 & spp & 0.41 $\pm$ 0.74 & 0.03 $\pm$ 0.31 & \textbf{0.49 $\pm$ 0.96} & 0.09 $\pm$ 0.85 \\
 & sc-zs & 0.13 $\pm$ 0.22 & -0.01 $\pm$ 0.11 & \textbf{0.42 $\pm$ 0.48} & 0.03 $\pm$ 0.20 \\
 & sc-cot & 0.19 $\pm$ 0.28 & 0.07 $\pm$ 0.14 & \textbf{0.29 $\pm$ 0.32} & 0.08 $\pm$ 0.27 \\
 & sc-spp & 0.09 $\pm$ 0.20 & 0.05 $\pm$ 0.10 & \textbf{0.22 $\pm$ 0.38} & 0.05 $\pm$ 0.18 \\
\cline{1-6}
\multirow[t]{6}{*}{L3.3-70B} & zs & 1.16 $\pm$ 3.16 & 0.14 $\pm$ 2.06 & \textbf{2.38 $\pm$ 4.85} & -0.07 $\pm$ 3.64 \\
 & cot & 1.18 $\pm$ 2.20 & 0.12 $\pm$ 1.26 & \textbf{1.59 $\pm$ 2.91} & -0.01 $\pm$ 1.46 \\
 & spp & 0.57 $\pm$ 1.41 & 0.26 $\pm$ 1.05 & \textbf{1.28 $\pm$ 2.29} & -0.13 $\pm$ 2.19 \\
 & sc-zs & 0.13 $\pm$ 0.63 & 0.11 $\pm$ 0.56 & \textbf{0.53 $\pm$ 1.31} & 0.02 $\pm$ 0.85 \\
 & sc-cot & \textbf{0.22 $\pm$ 0.28} & -0.10 $\pm$ 0.45 & 0.12 $\pm$ 0.73 & 0.04 $\pm$ 0.21 \\
 & sc-spp & 0.20 $\pm$ 0.53 & 0.04 $\pm$ 0.24 & \textbf{0.46 $\pm$ 0.72} & -0.22 $\pm$ 0.69 \\
\cline{1-6}
\multirow[t]{6}{*}{Mistral} & zs & 0.78 $\pm$ 4.58 & -0.35 $\pm$ 3.42 & \textbf{1.44 $\pm$ 6.57} & -1.05 $\pm$ 5.91 \\
 & cot & 0.36 $\pm$ 1.04 & 0.02 $\pm$ 0.59 & \textbf{0.51 $\pm$ 1.16} & 0.13 $\pm$ 1.21 \\
 & spp & 0.50 $\pm$ 0.75 & 0.15 $\pm$ 0.79 & 0.24 $\pm$ 1.68 & \textbf{0.62 $\pm$ 1.14} \\
 & sc-zs & \textbf{0.11 $\pm$ 0.96} & -0.10 $\pm$ 0.74 & 0.01 $\pm$ 1.21 & -0.37 $\pm$ 1.70 \\
 & sc-cot & 0.15 $\pm$ 0.29 & 0.03 $\pm$ 0.22 & \textbf{0.22 $\pm$ 0.31} & 0.03 $\pm$ 0.24 \\
 & sc-spp & 0.03 $\pm$ 0.18 & 0.02 $\pm$ 0.26 & \textbf{0.13 $\pm$ 0.35} & 0.02 $\pm$ 0.39 \\
\bottomrule
\end{tabular}
\caption{Average Efficiency (Points per kilo-token). Bold values denote higher efficiency per model-prompting setup.}
\label{tab:rps_efficiency_avg_heatmap}
\end{table*}

In RPS, efficiency (Table \ref{tab:rps_efficiency_avg_heatmap}) is generally lower and more variable than in PD, reflecting the greater difficulty of converting reasoning into reliable payoff gains in a three-action adversarial game. A consistent pattern across almost all models is that the payoff-based counterfactual yields the highest efficiency values, often by a clear margin. This is visible for Claude 3.7 ($2.02\pm 4.02$ under ZS), Claude 3.7(T) (
$3.26\pm 4.15$ under ZS), Claude 4(T) (
$1.50\pm 2.05$ under CoT), DeepSeek R1 (
$1.04\pm 1.77$ under CoT), and Llama 3.3 (
$2.38\pm 4.85$ under ZS). The reason is intuitive: once the altered payoff asymmetry is recognized, exploitable opponents such as PP or TFT can produce larger score gains per token than in the default game, so the extra reasoning is more often rewarded.

By contrast, the label-based and especially the joint counterfactuals are usually less efficient. In the label-based setup, efficiency often drops close to zero or even becomes negative, suggesting that relabeling disrupts strategic adaptation without producing compensating payoff gains. This effect is particularly strong for Claude 3.5, Claude 4, Claude 4(T), Llama 3.3, and Mistral in several prompt settings. The joint counterfactual is often even harsher: many values are near zero or negative, especially for weaker or less stable models, which indicates that simultaneously changing labels and payoffs increases token cost without reliably improving performance.

Across models, Claude 3.7(T) is the strongest overall efficiency performer in RPS, especially in the payoff-based setting, where it achieves the highest values in the table. Llama 3.3 is also highly competitive in payoff-based RPS, although its efficiency drops sharply in the joint setup, showing weaker robustness when both semantics and incentives change together. Claude 4(T) can also be very efficient in default and payoff-based RPS, but becomes much less efficient in the label-based and joint settings. Claude 3.7 and Claude 4 are more moderate but stable performers, usually remaining positive and competitive without reaching the peaks of Claude 3.7(T). DeepSeek R1 is fairly competitive in efficiency, especially in payoff-based RPS, even though its strategic robustness is less consistent. Mistral is the least reliable: while it occasionally attains strong efficiency, especially in payoff-based or joint SPP settings, it also produces some of the most negative and variable results, showing that low token cost can coincide with brittle behavior rather than genuinely better adaptation.

Prompting effects broadly mirror the PD findings. ZS and standard prompted variants often achieve the highest raw efficiency when the model already exploits the opponent effectively, because they avoid the heavy token overhead of aggregation. Self-consistency usually lowers efficiency, even when it stabilizes behavior, since majority voting adds substantial token cost and only sometimes yields proportional score gains. This is especially clear in the stronger models, where SC frequently reduces the large positive efficiencies seen in non-SC payoff-based settings. Overall, the table reinforces three main points: (i) payoff-based RPS is the most efficiency-favorable setting because altered incentives can yield larger returns per token, (ii) label-based and joint counterfactuals are substantially less efficient because they disrupt strategic transfer, and (iii) the best efficiency-capability trade-off in RPS appears to come from Claude 3.7(T), with Llama 3.3 and Claude 4(T) also competitive in the payoff-based setup.

\subsection{RPS validity rate}
\begin{table}[h!]
\centering \small
\begin{tabular}{ll}
\toprule
model & avg validity rate\\
\midrule
C3.5Sv2 & 100.0 $\pm$ 0.0 \\
C3.7S & 100.0 $\pm$ 0.0 \\
C3.7S(T) & 100.0 $\pm$ 0.2 \\
C4S & 99.9 $\pm$ 0.9 \\
C4S(T) & 100.0 $\pm$ 0.0 \\
DS-R1 & 100.0 $\pm$ 0.2 \\
L3.3-70B Instruct & 100.0 $\pm$ 0.0 \\
Mistral & 99.2 $\pm$ 6.8 \\
\bottomrule
\end{tabular}
\caption{Average Valid Rate (\% of Valid Outcomes)}
\label{tab:rps_valid_rates}
\end{table}

In Table \ref{tab:rps_valid_rates} we present validity rates for the various LLMs used in our experimentation. Validity in RPS is effectively perfect across almost all models, showing that formatting and action-space compliance are not a meaningful confound for the strategic results. Claude 3.5 Sonnet v2, Claude 3.7, Claude 4(T), and Llama 3.3 all achieve a strict $100.0\%\pm 0.0$ valid rate, while Claude 3.7(T) and DeepSeek R1 remain essentially perfect with only negligible variation ($100.0\%\pm 0.2$). Claude 4 is similarly reliable at 
$99.9\%\pm 0.9$, indicating only extremely rare invalid outputs. The only clear outlier is Mistral, with a lower and much more variable validity rate of 
$99.2\%\pm 6.8$. Although this still indicates high average compliance, the larger variance suggests occasional formatting or parsing failures that are substantially more frequent than for the other models. Overall, these results confirm that the differences observed in RPS total points, opponent comprehension, and efficiency are not driven by invalid-action artifacts, but by genuine differences in strategic behavior; validity issues are too rare to materially affect the main conclusions, except that Mistral’s instability is also reflected at the output-format level.

\subsection{Cross-model analysis}

\paragraph{Overall model comparison.} Across settings, the strongest models differ by game and by robustness criterion. In the default PD, Claude 3.5/3.7 Sonnet and Llama 3.3-70B are the strongest overall, as they reliably achieve perfect or near-perfect mutual cooperation in LLM--LLM play while also adapting well to algorithmic opponents. In the default RPS setting, the strongest performance is shown by the advanced Claude models, especially Claude 3.7 and Claude 4, which most consistently exploit cyclic algorithmic opponents and maintain near-equilibrium behavior against stronger opposition. Under counterfactual interventions, Claude 3.7 and the stronger Claude variants remain the most robust overall, showing the smallest degradation under label-only changes and the best adaptation when payoff structure changes require recomputing incentives; by contrast, Mistral Large is the most brittle, with DeepSeek R1 also showing noticeable instability. In terms of prompting stability, Claude 3.5/3.7 and Llama 3.3 are generally the most consistent across ZS, CoT, and SPP, whereas Mistral Large exhibits the highest variability and Claude 4 as well as DeepSeek R1 are more sensitive to prompting, particularly in settings where additional reasoning appears to induce overthinking rather than better strategic adaptation.

\begin{table*}[h!]
\centering
\small
\begin{tabular}{lccccc}
\toprule
\textbf{Model} & \textbf{PD} & \textbf{RPS} & \textbf{Counterfactuals} & \textbf{Efficiency} & \textbf{Main weakness} \\
\midrule
Claude 3.7 & strong & strong & relatively robust & moderate & overreasoning \\
Claude 4 & mixed & strong & mixed & lower & instability \\
DeepSeek R1 & mixed & moderate & brittle & lower & coordination \\
Llama 3.3 & strong & moderate & moderate & good & RPS adaptation \\
Mistral Large & weak & weak & weakest & low & comprehension \\
\bottomrule
\end{tabular}
\caption{Cross-model qualitative comparison across games, counterfactual robustness, and efficiency.}
\label{tab:cross-model}
\end{table*}

\paragraph{Robustness across prompting strategies.} Across prompting strategies, the benefits of explicit reasoning are uneven. CoT is most helpful for models that already display strong strategic competence but occasionally need more structured adaptation, with Claude 3.5/3.7 and, in some PD settings, Mistral Large showing improvements in stability or recovery relative to zero-shot behavior. By contrast, several stronger reasoning-oriented models degrade under thinking-enabled variants: Claude 4 and DeepSeek R1 often become more distrustful or unstable, particularly in PD LLM--LLM interactions and in the harder counterfactual settings, suggesting that additional deliberation can induce overthinking rather than better strategic play. In terms of prompting stability, Claude 3.5/3.7 and Llama 3.3 are the most consistent across ZS, CoT, SPP, and SC, maintaining similar qualitative behavior and relatively low variance, whereas Mistral Large is the most prompt-sensitive overall and Claude 4/DeepSeek R1 exhibit greater fluctuations across prompting conditions. Self-consistency generally improves robustness by reducing run-to-run variance, but it does not fundamentally change the strategic tendencies of a model; rather, it tends to reinforce whichever behavior—cooperative, exploitative, or unstable—the underlying prompting setup already induces.

\paragraph{Capability vs Efficiency.} A clear trade-off emerges between strategic capability and efficiency. Models that achieve the strongest total-point outcomes and earliest opponent comprehension are not always the most efficient, since superior performance often comes at the cost of longer reasoning traces and higher token usage. This pattern is especially visible in RPS, where advanced Claude models often outperform weaker baselines in total points and adaptation, but do so with greater computational cost. Conversely, some lighter or less deliberative models can appear more efficient simply because they generate fewer tokens, even when their strategic performance is weaker or less stable. The same trade-off appears within model families: thinking-enabled variants do not consistently improve total points or comprehension in proportion to their additional token usage, and in several cases—particularly for Claude 4 and DeepSeek R1—the extra reasoning is associated with lower efficiency and occasionally worse strategic outcomes. Overall, efficiency should therefore be interpreted jointly with performance metrics: higher token expenditure can support stronger play in harder settings, but it does not guarantee proportionate gains and may instead reflect overthinking or unstable adaptation.

\paragraph{Robustness across opponent types.} Robustness also varies markedly across opponent types. Against algorithmic opponents, the strongest models are those that can quickly identify deterministic or cyclic structure and exploit it reliably: advanced Claude models perform best overall in this regime, with Llama 3.3 also showing strong and consistent adaptation, especially in PD. In contrast, Mistral Large is the least reliable exploiter of deterministic opponents, often adapting late or inconsistently, while DeepSeek R1 tends to be more variable across settings. Against LLM opponents, the challenge shifts from exploitation to coordination or equilibrium maintenance. In PD, Claude 3.5/3.7 and Llama 3.3 coordinate best with other LLMs, consistently converging to mutual cooperation, whereas Claude 4 and DeepSeek R1 remain more brittle, often defaulting to defection or unstable exploration. In RPS, where cooperative convergence is impossible, robustness in LLM--LLM play is better reflected by maintaining near-equilibrium behavior without introducing exploitable biases; here again, stronger Claude models are the most stable, while Mistral Large and DeepSeek R1 are more prone to deviations and instability. Overall, some models are strong exploiters of deterministic opponents, while others are better coordinators or stabilizers against adaptive LLM opposition, and the most robust models are those that perform well in both regimes.

\paragraph{Counterfactual behaviors.} Counterfactual sensitivity differs substantially by model. Claude 3.7 and Claude 4 generally deliver the strongest performance in the default games, especially in exploiting structured opponents in RPS and achieving high payoffs in PD, but their robustness under counterfactual changes is uneven: Claude 3.7 remains comparatively stable, whereas Claude 4 is more prone to degradation, particularly when altered incentives or additional reasoning induce overthinking. Llama 3.3 is often one of the most stable models in default PD, especially in LLM--LLM cooperation, but it is weaker in the more difficult RPS and payoff-shifted counterfactual settings, where adaptation requires finer-grained strategic recalibration. DeepSeek R1 is strong in some regimes, including early adaptation against certain algorithmic opponents, yet remains unstable overall and often over-defective in PD, which hurts both coordination and counterfactual robustness. Mistral Large is the most brittle model across the board: it shows the greatest sensitivity to label shifts, the highest instability in opponent comprehension, and the weakest ability to preserve performance when familiar game structure is altered. Overall, counterfactual sensitivity reveals that strong default-game performance does not necessarily imply robust strategic generalization, and models differ sharply in how well they recompute behavior when labels or incentives are changed.

\paragraph{Final takeaways.}
Table~\ref{tab:cross-model} summarizes the main qualitative differences across models, aggregating performance across games, prompting strategies, and counterfactual settings. 

Claude 3.7 emerges as the most consistently strong model, combining high performance in both PD and RPS with relatively robust behavior under counterfactual changes, albeit with moderate efficiency due to additional reasoning. Claude 4 achieves strong results in RPS but is less stable overall, showing mixed counterfactual robustness and a tendency toward distrust or unstable strategies. Llama 3.3 is particularly strong and stable in PD, especially in cooperative settings, while remaining moderately effective in RPS with good efficiency. DeepSeek R1 demonstrates competitive performance in some scenarios but is overall brittle, especially in coordination-heavy settings, often defaulting to unstable or overly defensive strategies. Finally, Mistral Large is consistently the weakest across all dimensions, with delayed opponent comprehension, high variability, and the strongest degradation under counterfactual interventions.

Overall, the table highlights that strong default-game performance does not necessarily translate to robustness or efficiency, reinforcing the importance of multi-dimensional evaluation.

\end{document}